%% file: acl_latex.tex
\pdfoutput=1

\documentclass[11pt]{article}

\usepackage[]{acl}

\usepackage{times}
\usepackage{latexsym}

\usepackage[T1]{fontenc}

\usepackage[utf8]{inputenc}

\usepackage{microtype}

\usepackage{inconsolata}

\usepackage{graphicx}
\usepackage{rotating}
\usepackage{booktabs}
\usepackage{multirow}
\usepackage{subcaption}
\usepackage{amsmath,amssymb}
\usepackage{algorithm2e}
\usepackage{colortbl}
\usepackage{siunitx}

\title{As easy as PIE: understanding when pruning causes language models to disagree}

\author{Pietro Tropeano \\
  University of Copenhagen \\ Copenhagen, Denmark \\
  \texttt{pitr@di.ku.dk} \\\And
  Maria Maistro \\
  University of Copenhagen \\ Copenhagen, Denmark \\
  \texttt{mm@di.ku.dk} \\\AND
  Tuukka Ruotsalo \\
  University of Copenhagen \\ Copenhagen, Denmark \\
  LUT University \\ Lahti, Finland \\
  \texttt{tr@di.ku.dk} \\\And
  Christina Lioma \\
  University of Copenhagen \\ Copenhagen, Denmark \\
  \texttt{c.lioma@di.ku.dk} \\}

\begin{document}
\maketitle
\begin{abstract}
Language Model (LM) pruning compresses the model by removing weights, nodes, or other parts of its architecture. 
Typically, pruning focuses on the resulting efficiency gains at the cost of effectiveness.
However, when looking at how individual data points
are affected by pruning, it turns out that a particular subset of data points always bears most of the brunt (in terms of reduced accuracy) when pruning,
but this effect goes unnoticed when reporting 
the mean accuracy of all data points. These data points are called PIEs and have been studied in image processing, but not in NLP.
In a study of various NLP datasets, pruning methods, and levels of compression, we find that PIEs impact inference quality considerably, regardless of class frequency, and
that BERT is more prone to this than BiLSTM. We also find that PIEs contain a high amount of data points that have the largest influence on how well the model generalises to unseen data. This means that when pruning, with seemingly moderate loss to accuracy across all data points, we in fact hurt tremendously those data points that matter the most. We trace what makes PIEs both hard and impactful to inference to their overall longer and more semantically complex text. 
These findings are novel and contribute to understanding how LMs are affected by pruning. 
The code is available at: \url{https://github.com/pietrotrope/AsEasyAsPIE}
\end{abstract}

\section{Introduction}

Neural networks (NNs) are becoming increasingly larger, with remarkable improvements to their inference capabilities, but also very high computational demands. 
The latter has motivated research in the area of NN pruning, whose goal is to reduce a model (in terms of its parameters, nodes, layers, or any other aspect of its architecture) to a smaller version, without significant loss of inference quality. Pruning has been shown to produce smaller, hence more efficient NNs, with small loss to their effectiveness \cite{pmlr-v119-li20m,hooker2019compressed}. Similar findings are also reported when pruning Language Models (LMs) \cite{gupta2021compression, wang-etal-2020-structured, sun2023simple, NEURIPS2020_eae15aab, NEURIPS2019_2c601ad9} in NLP. 

\begin{table}
\centering
\input{Tables/PIEs/Examples/PIEs_examples}
\caption{Examples where pruned and unpruned models disagree (from the SNLI dataset).}
\label{tab:PIEs_preds}
\end{table}

When pruning NNs, typically the focus is on the high efficiency gains achieved at the cost of  effectiveness, commonly measured in terms of test set accuracy. However, when zooming in on precisely how individual data points are affected by pruning, it turns out that models of similar accuracy scores can have notably different weights and therefore make wildly different inferences on a subset of data points. In other words, the similar accuracy scores between pruned and unpruned models do not mean that pruning affects all data points in a uniform way, but rather that some parts of the data distribution are much more sensitive to pruning than others. This effect can go unnoticed when one measures pruning effectiveness in terms of mean accuracy, because taking the mean can hide such important score variations in the data. 
However, this does not change the fact that certain types of data are disproportionately impacted by pruning, which begs the question: what are the characteristics of these data points and how important is their detection?

In response to this, \textit{Pruned Identified Exemplars} (PIEs) are defined as the subset of data points where pruned and unpruned models disagree \citep{hooker2019compressed} (see example in Table \ref{tab:PIEs_preds}). Studies in image processing reveal that PIEs are harder to classify, not only for NNs, but also for humans, because they a) tend to be mislabeled (ground truth noise), b) may have overall lower quality (inherently noisy signal), or c) may depict multiple objects (more challenging task) \citep{hooker2019compressed}. Hence, this subset of data points where pruned and unpruned models tend to disagree are also some of the most difficult data points for the model to handle. PIEs are those critical data points on which we would suffer the most damage, if the model were to be deployed out in the wild. Despite this, to our knowledge, PIEs have not been studied in NLP.

Motivated by this gap in understanding how LMs are actually affected by pruning, we study whether PIEs exist in text, what are their textual characteristics, and what this practically means for inference. Using eight pruning methods on two different LM architectures (BiLSTM and BERT) and four common NLP datasets for sentiment classification, document categorisation and natural language inference, we contribute the first study of PIEs in LM pruning for NLP. Our empirical analysis shows that there is always a subset of data points where pruned and unpruned models disagree, and that this subset is larger for BERT than BiLSTM. We also find that these data points, namely PIEs, are overall semantically more complex, contain on average more difficult words, and have generally longer text than the rest of the data. Furthermore, we find that PIEs contain a high amount of \textit{influential examples}, i.e., data points that have the largest influence on how well the model generalises to unseen data \citep{jin2022pruning}.
These novel findings mean that when pruning LMs for efficiency, with seemingly small drops in overall accuracy, we are impacting notably the accuracy on subsets of data that are critical for model generalisation.
More simply put, we impact how well the model learns. This effect is more pronounced for BERT than for BiLSTM.

\section{Pruned Identified Exemplars (PIEs)} \label{sec:design_PIEs}

 \noindent \textbf{Formal definition of PIEs.} \label{sec:def_PIEs}
Pruned Identified Exemplars (PIEs) are data points where the predictions of pruned and unpruned models 
differ~\citep{hooker2019compressed}.
Assume a single-label classification task, where each data point $x$ belongs to a single class.
Let $P=\{p_1,...,p_N\}$ be the set of $N$ different initializations of the pruned model, and $U=\{u_1,...,u_N\}$ the set of $N$ different initializations of the unpruned model.\footnote{$N$ must be the same for pruned and unpruned models.}
Let $m(P,x)$ be the majority class assigned to $x$ over all the initializations of the pruned model after training. 
This is computed as the most frequently predicted class for the data point $x$ across all $N$ initializations in $P$, i.e., the mode of the $N$ predicted classes.\footnote{In case of ties, classes are sorted ascendingly by their associated number, and the first class is assigned.}
Similarly, $m(U,x)$ is the most frequent class predicted by the unpruned model initializations.
Then,
$x\text{ is a PIE if } m(P,x) \neq m(U,x)$, 
i.e., the majority class assigned to $x$ by the pruned and unpruned model is different.

\noindent \textbf{PIEs in multi-label classification.} \label{sec:multi-label_PIEs}
We extend the above definition of PIEs
to multi-label classification, where a data point $x$ can belong to more than one class. 
We treat multi-label classification as multiple single-label classifications: a data point $x$ is a PIE, if there exists a class such that the pruned and unpruned models disagree.
Let $\tilde{m}(P,x)$ be the set of majority classes assigned to $x$ over all the initializations of the pruned models.
A class is assigned to the set of majority classes if $>N/2$ initializations of the pruned model predict that $x$ belongs to that class.
Similarly, $\tilde{m}(U,x)$ is the set of majority classes assigned by the unpruned model.
Then,
$x\text{ is a PIE if }  \tilde{m}(P,x)\neq \tilde{m}(U,x)$, 
i.e., the sets of majority classes predicted for $x$ by the pruned and unpruned models differ.
The inequality between $\tilde{m}(P,x)\nsubseteq \tilde{m}(U,x)$ and $\tilde{m}(P,x)\nsupseteq \tilde{m}(U,x)$ means that $x$ is a PIE even if the pruned and unpruned model disagree only on a single class.

Unlike PIEs, which consider qualitative data aspects, \citet{dutta2024accuracy} introduce \textit{flips}, a distance metric that counts the number of predictions that switch from correct to incorrect (and vice-versa) after compression. 
\citet{holste2023does} propose the following alternative way of selecting PIEs in a multi-label setting. 
For each data point, they compute the average prediction over all initializations. Then, the data points are ranked by the average prediction, and agreement is measured as the Spearman rank correlation between the rankings for the pruned and unpruned models.
The $5^{th}$ percentile of data points with highest disagreement (lowest Spearman rank correlation) are considered PIEs. 
This approach does not quantify the amount of PIEs for the pruned and unpruned models. 
In addition, in~\citet{holste2023does}, a data point can be considered as non PIE even if there is disagreement between the pruned and unpruned models, simply because that data point is outside the $5^{th}$ percentile.
Our definition of PIEs is stricter than~\citeposs{holste2023does}, since disagreement even on a single class determines the data point to be a PIE.

\section{Study design} 
\label{sec:study_design}
Our aim is to study whether PIEs exist in text data, what are their textual characteristics, and what this practically means for inference.
We present the datasets, LMs, and pruning methods of our study.

\noindent \textbf{Datasets.}\label{sec:study_design_datasets}
We use two single-label datasets: IMDB \citep{maas-EtAl:2011:ACL-HLT2011} for sentiment analysis, and SNLI \citep{bowman2015large} for natural language inference. We also use two multi-label datasets for document categorisation: Reuters-21578\footnote{\url{https://www.daviddlewis.com/resources/}.}, and AAPD \citep{yang2018sgm}.
Statistics are in Table \ref{tab:dataset_stats} (see Appendix \ref{app:datasets_and_preprocessing} for preprocessing details).

\begin{table}
\centering

\input{Tables/Setup/Dataset/stats}
\caption{Dataset statistics after preprocessing.}
\label{tab:dataset_stats}
\end{table}
\begin{table}
\input{Tables/Setup/Pruning_algorithms/algorithms}
\caption{Our 8 pruning methods. \textit{Random} cannot be combined with \textit{Weight Rewinding} because weights that are rewinded to their initial values are not random.}
\label{tab:pruningAlgoTable}
\end{table}

\noindent \textbf{Language Model Architectures.}\label{sec:study_design_architectures}
We select two common types of LMs to represent both transformers and Recurrent Neural Networks (RNNs): BERT~\citep{devlin2018bert}, and bidirectional LSTM (BiLSTM)~\citep{hochreiter1997long}.
We train BiLSTM from scratch, but we finetune a pretrained version of BERT\textsubscript{BASE}. 
See Table \ref{tab:architectures_stats} in Appendix \ref{app:language_model_architectures} for details on the LMs, and Appendix \ref{app:language_model_architectures} for our tuning methodology.

\noindent \textbf{Pruning methods.}\label{sec:study_design_pruning_methods}
 We use eight common pruning methods, shown in Table \ref{tab:pruningAlgoTable}. Each of them is a combination of  \textit{scheduling}  and \textit{scoring}.
 
 Scheduling controls 
the moment and frequency of the pruning iterations during training. We use two scheduling variations: (i) pruning the model before training (\textit{at initialization}), and (ii) pruning in multiple iterations during training (\textit{iterative}). Only for iterative pruning, we use two tuning strategies: \textit{finetuning} and \textit{weight rewinding}.
In finetuning, we retrain the model after pruning and update its weights. In weight rewinding, we rewind weights to their initial state \citep{frankle2018lottery}. 

Scoring refers to selecting which weights to prune. A score is given to each LM weight, and the weights with the lowest score according to a threshold are pruned.
We use three scoring variations: 
(i) \textit{Magnitude}, where the score is the absolute value of a weight \citep{frankle2018lottery}; 
(ii) \textit{Impact}, where the score is the weight multiplied by its accumulated gradient on 100 randomly sampled data points of the training set \cite{lee2018snip}; 
(iii) \textit{Random}, where the score is randomly assigned a value between 0 and 1.

We chose the above pruning methods as they are architecture-agnostic (i.e., applicable to any NN architecture). Recent methods build on these, are architecture-specific, or optimized for one architecture \citep{sun2023simple, yu2019playing, NEURIPS2019_2c601ad9, frantar2023sparsegpt}.
We prune 20\%, 50\%, 70\%, 90\%, and 99\% of the parameters of each LM (see Table \ref{tab:architectures_stats} in Appendix \ref{app:language_model_architectures} for details). For each configuration, we train 30 initializations. This results in 9840 runs (= 2 LMs x 4 datasets x 8 pruning methods x 5 pruning thresholds x 30 initializations + 2 LMs x 4 datasets x 30 unpruned model initializations), that require ca. 28000 AMD MI250X GPU hours. 
Our tuning methodology for pruning is detailed in Appendix \ref{app:hyperparameter_tuning}.

\section{Experimental findings}
We show how pruning impacts inference, the role of PIEs, and the textual characteristics of PIEs. 
\subsection{Pruning and occurrence of PIEs} \label{ss:pruning-occurrence}
\begin{figure*}
\centering
\begin{tabular}{cccc}
 \includegraphics[width=0.221\textwidth]{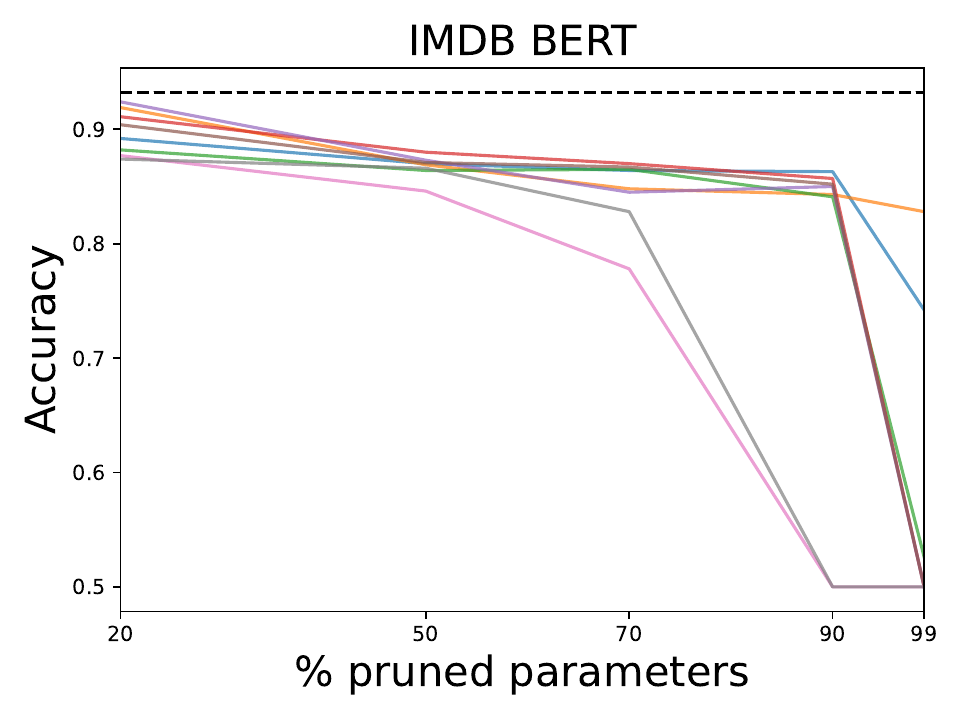} &
 \includegraphics[width=0.221\textwidth]{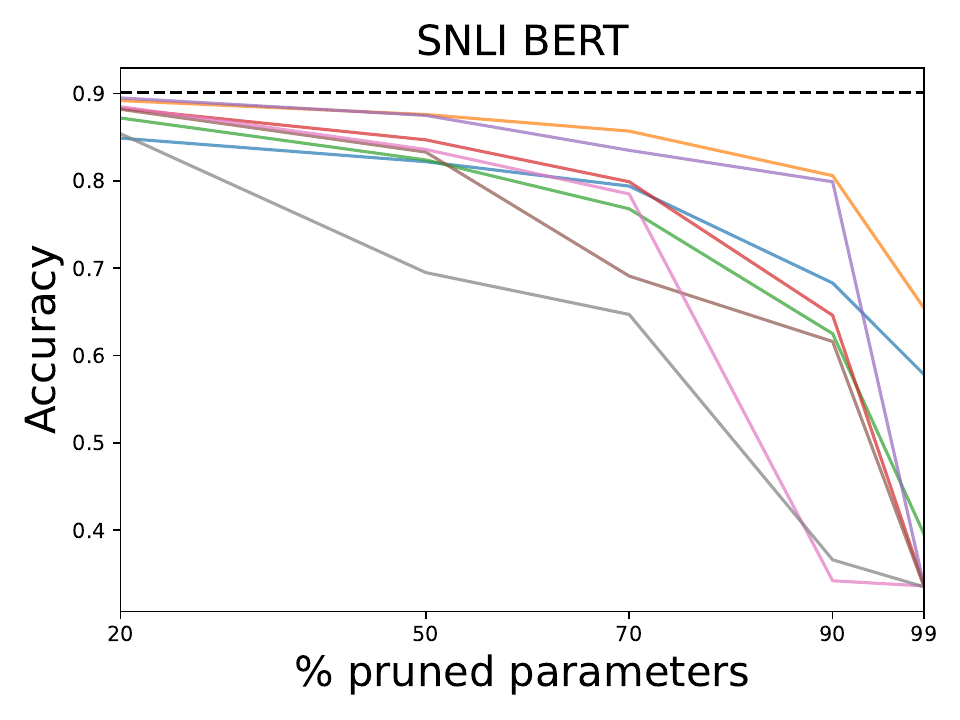} &
 \includegraphics[width=0.221\textwidth]{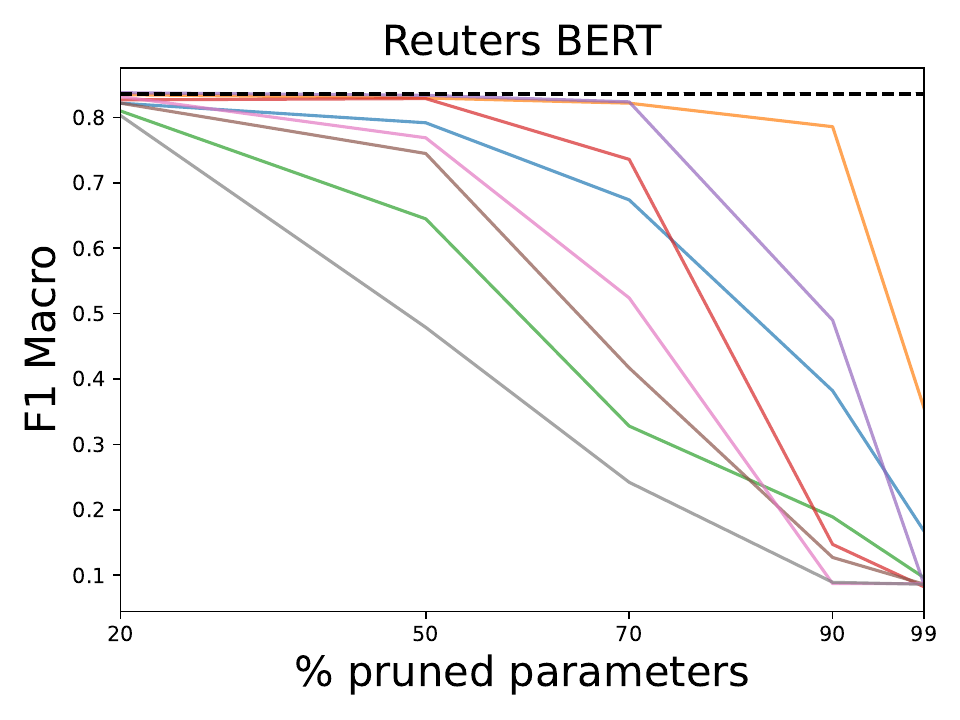} &
 \includegraphics[width=0.221\textwidth]{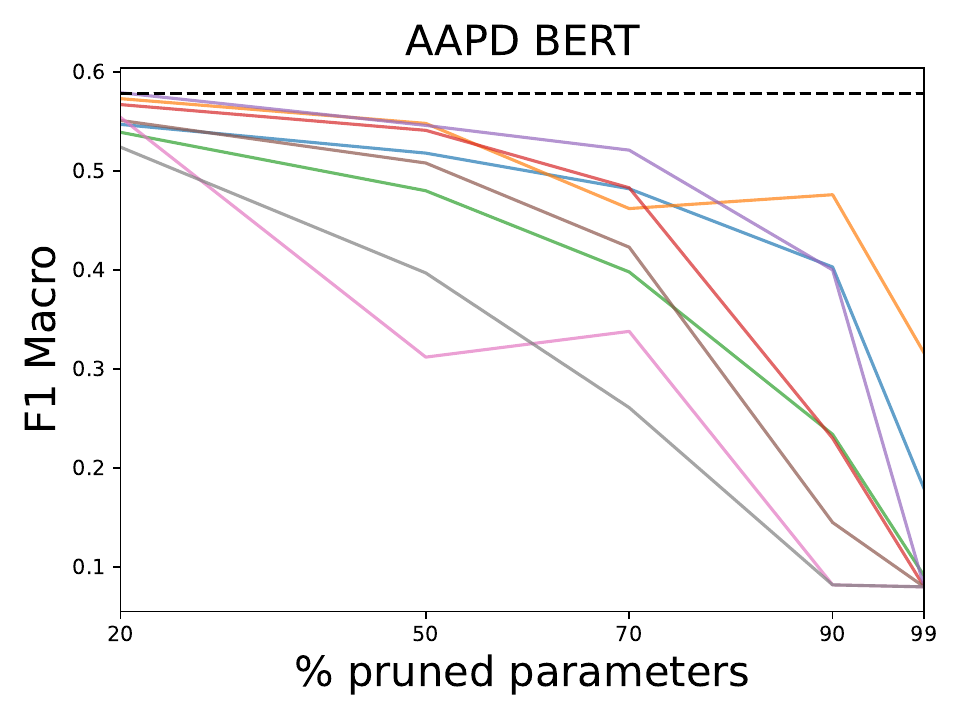} \\

 \includegraphics[width=0.221\textwidth]{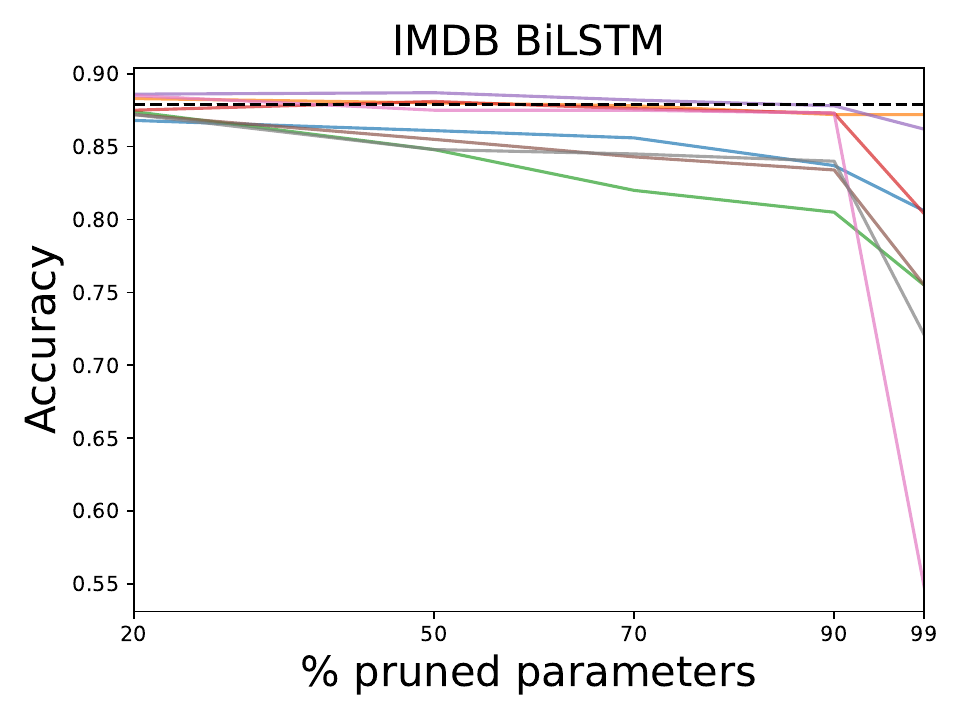} &
 \includegraphics[width=0.221\textwidth]{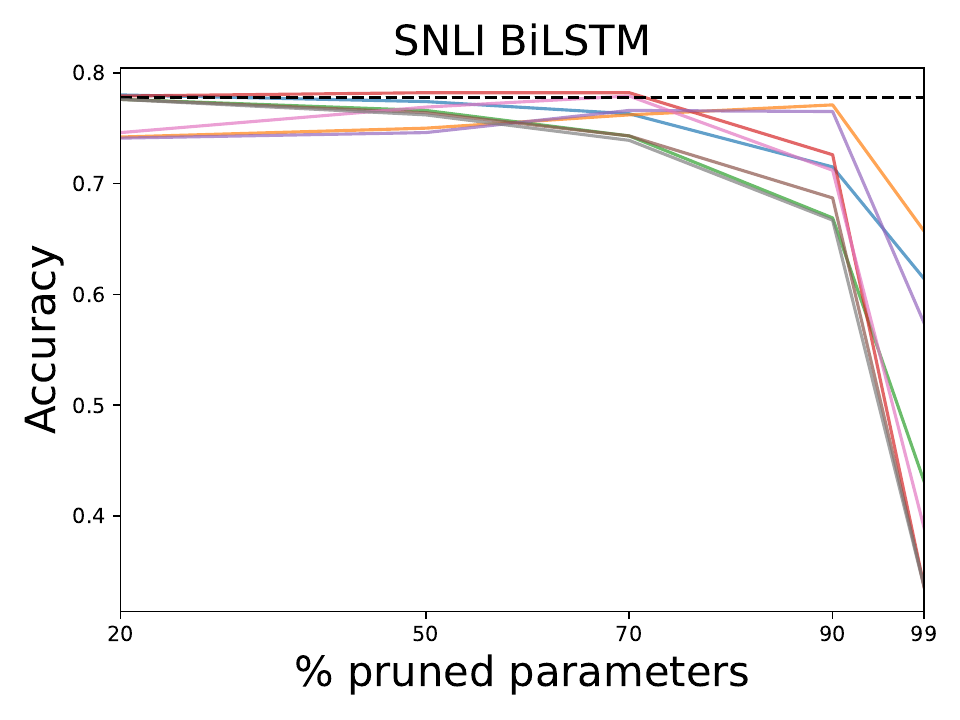} &
 \includegraphics[width=0.221\textwidth]{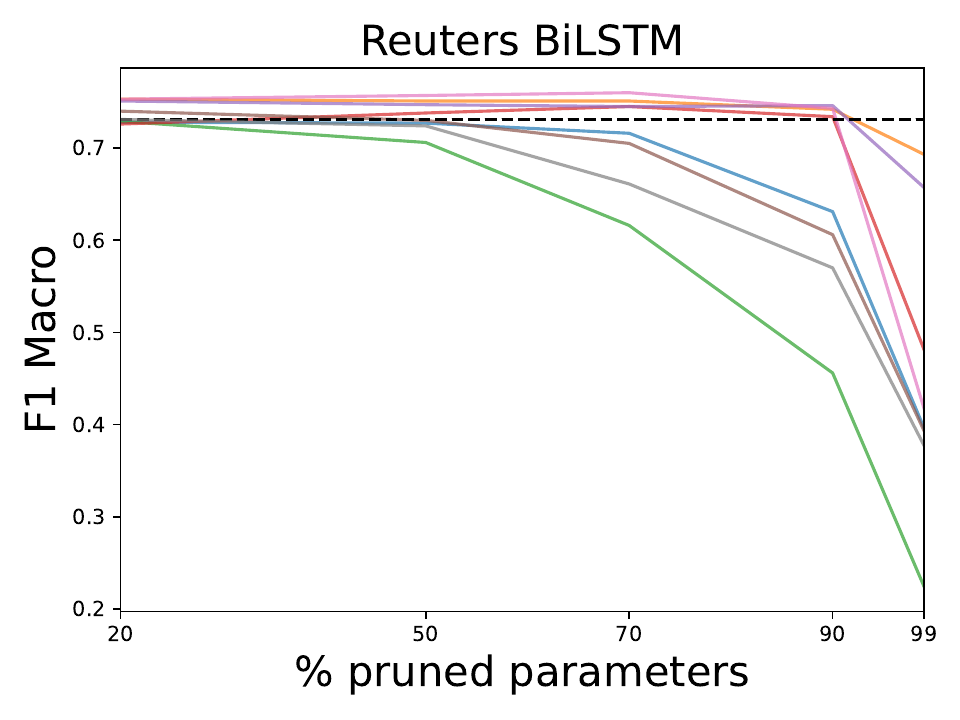} &
 \includegraphics[width=0.221\textwidth]{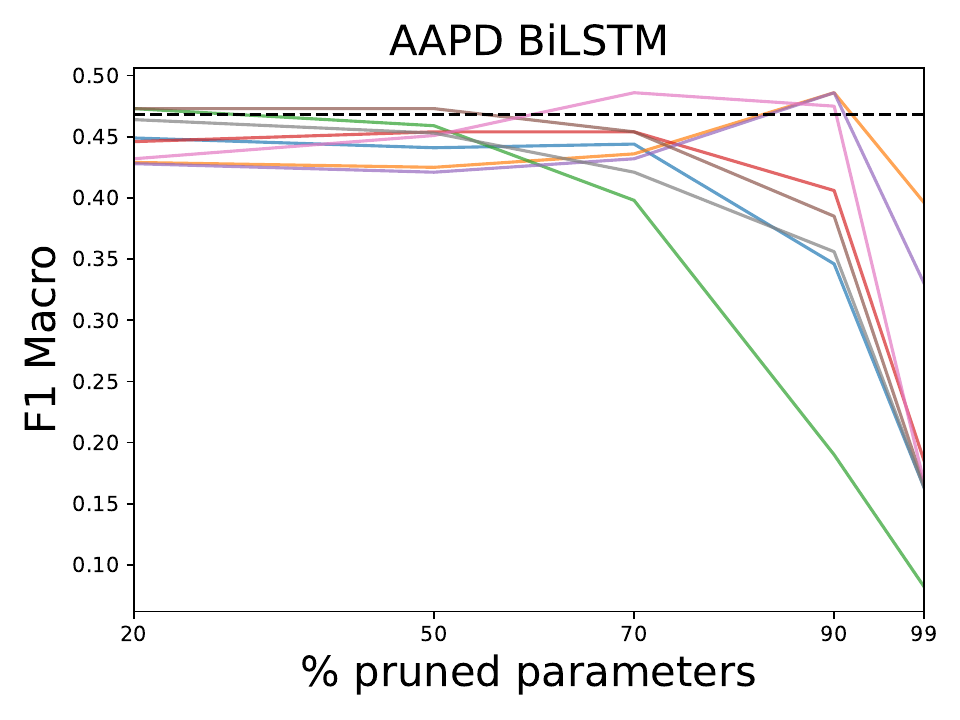}  \\

 \multicolumn{4}{c}{\includegraphics[width=0.75\textwidth]{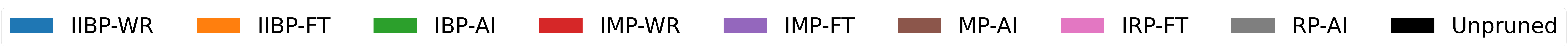}}         
\end{tabular}
\caption{Accuracy/F1 (y axis) of unpruned and pruned LMs per pruning threshold (x axis), over 30 initializations.
}
\label{fig:effectiveness_all}

\centering

\begin{tabular}{cccc}
 \includegraphics[width=0.223\textwidth]{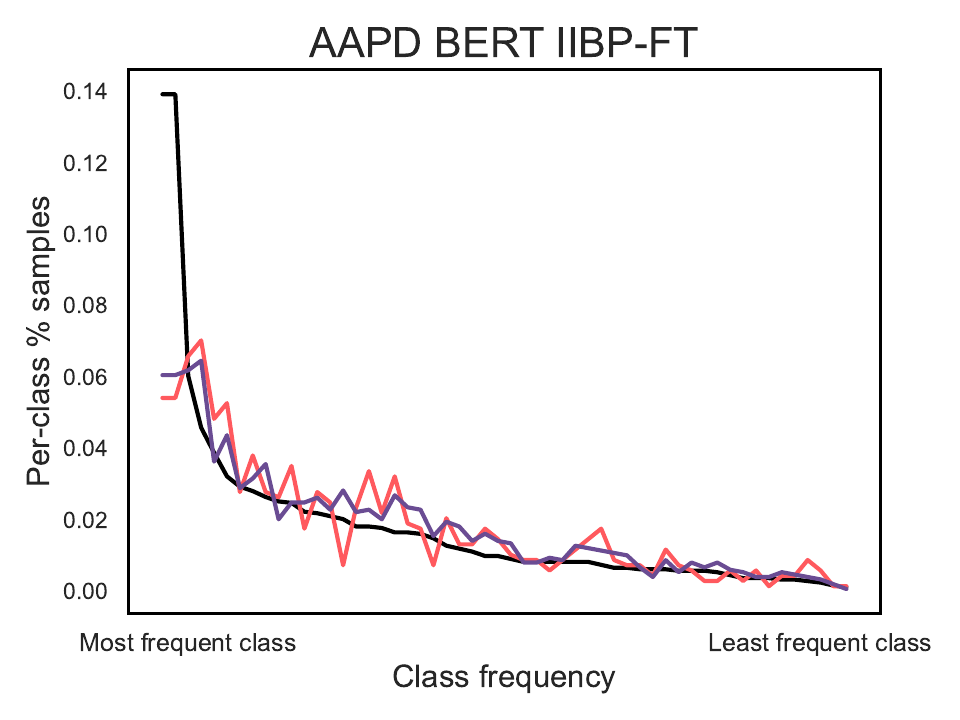} &
 \includegraphics[width=0.223\textwidth]{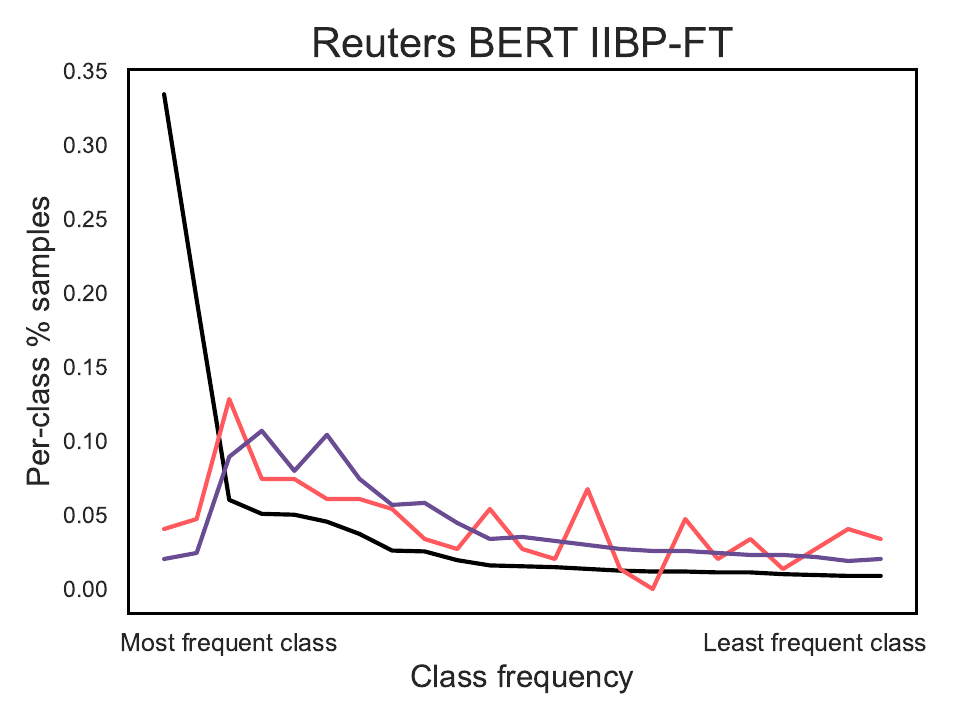} &
 \includegraphics[width=0.223\textwidth]{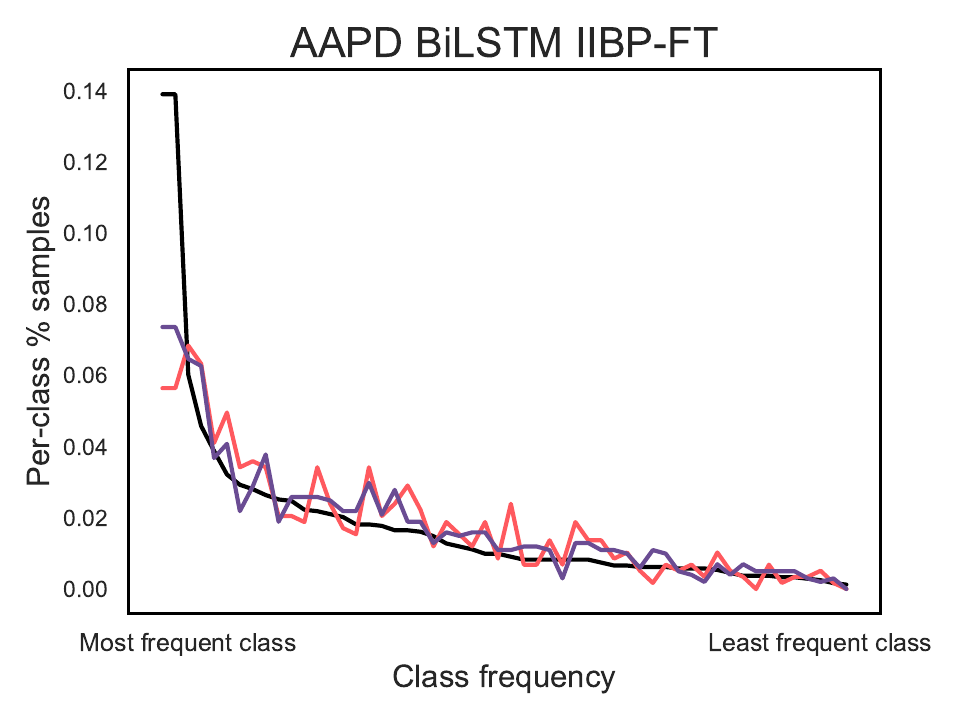} &
 \includegraphics[width=0.223\textwidth]{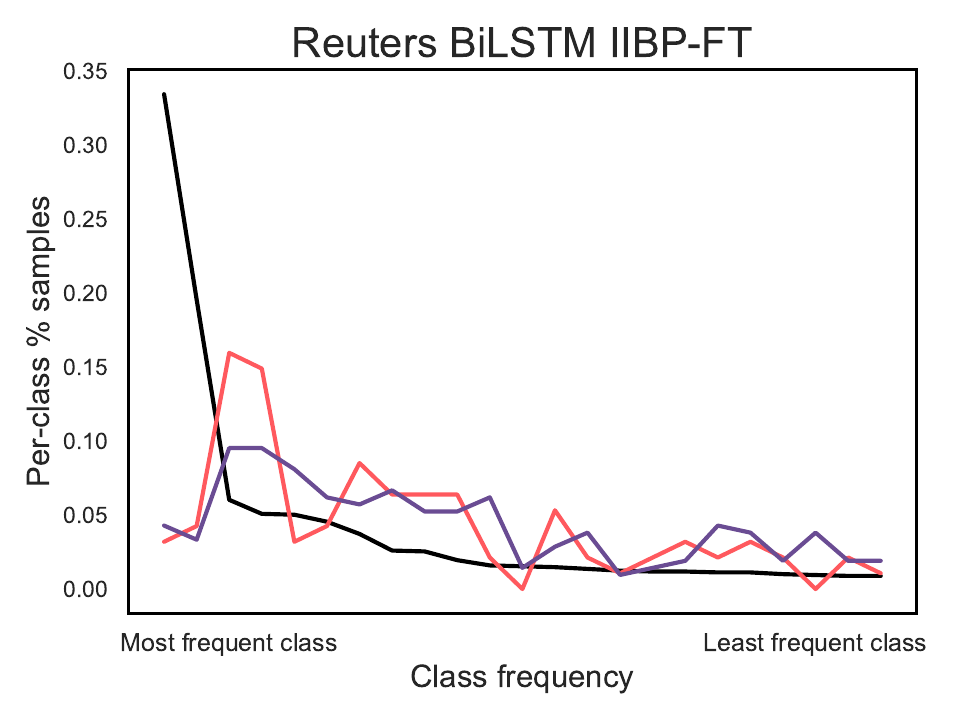} \\  
 \multicolumn{4}{c}{\includegraphics[width=0.21\textwidth]{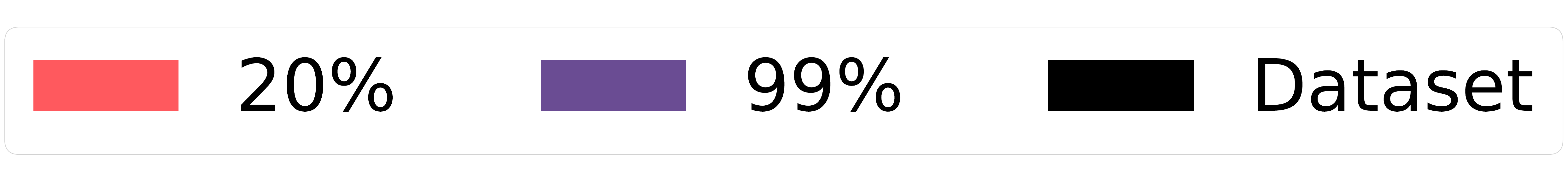}} 
\end{tabular}

\caption{Distribution of all data points and of PIEs at 20\% and 99\% pruning, across classes sorted by frequency (x axis), for the multi-label datasets (test set) and IIBP-FT pruner.}
\label{fig:pies_class_dist_pic}%
\end{figure*}

Figure \ref{fig:effectiveness_all} shows the accuracy/F1 of pruned versus unpruned models (see Table \ref{tab:architecture_remaining_params} in Appendix \ref{app:language_model_architectures} for details on the number of parameters pruned). We see that pruning BERT/BiLSTM up to 50\% gives overall tolerable drops to accuracy/F1 for most pruning methods. IIBP-FT is the pruning method with the overall smallest drop in accuracy/F1 compared to the unpruned model, and even outperforms the unpruned BiLSTM at times. We also see that, while unpruned BERT outperforms unpruned BiLSTM, pruning BERT hurts accuracy/F1 more than pruning BiLSTM, especially for pruning at 70\%-99\%. Hence \textbf{BERT is more sensitive to pruning than BiLSTM}, indicating that
parameters in BERT are not as easily disposable as in BiLSTM. Otherwise put, BERT seems to make better use of its parameters than BiLSTM, because their removal has a bigger impact on it than on BiLSTM.

\begin{table}
\centering
\input{Tables/PIEs/merged_PIEs_percentage_orginal}
\caption{Percentage of datapoints that are PIEs per configuration. Green and gray mark the percentages of datapoints that are PIEs for the best (green) and worst (gray) pruner per dataset and pruning threshold.
}
\label{tab:percentage_pies_app}
\end{table}
Table \ref{tab:percentage_pies_app} shows the \% of all data points\footnote{From now on, whenever we refer to all data points, we mean all data points in the test set, unless otherwise specified.} that are PIEs per model, dataset, pruning method and pruning threshold. We see that, as the pruning threshold increases, so does the proportion of PIEs, with very few marginal exceptions. This means that the particular subset of data points where unpruned and pruned models disagree becomes larger, the more we prune. 
In Table \ref{tab:percentage_pies_app} we shade the PIEs of the best and worst pruned model (according to their accuracy/F1 in Figure \ref{fig:effectiveness_all}) as green and gray respectively. We see that the best pruned model (green) has almost always a smaller percentage of PIEs than the worst pruned model (gray), per pruning threshold. In other words, \textbf{as the amount of PIEs increases, overall accuracy/F1 lowers, meaning that PIEs clearly impact inference quality}.

For the multi-label datasets, it is important to know, not only the proportion of data points that are PIEs, but also their distribution across classes. So, Figure \ref{fig:pies_class_dist_pic} plots the distribution of all data points versus PIEs, across classes, for IIBP-FT, which is the pruner with the best overall F1 in Figure \ref{fig:effectiveness_all}. The plots of the other configurations are in Appendix \ref{app:pruning_and_occurence_of_PIEs}.
We show PIEs when pruning 20\% and 99\% of the parameters which captures the lowest and highest \% of PIEs according to Table \ref{tab:percentage_pies_app}. Figure \ref{fig:pies_class_dist_pic} shows that PIEs are found across all classes of the dataset, from the least frequent to the most frequent class, and roughly follow the distribution of all data points across classes, except for the most frequent ones.
This observation, combined with the findings of Table \ref{tab:percentage_pies_app}, means that \textbf{besides the most frequent class, pruned and unpruned models disagree on all classes, regardless of their frequency.}

\begin{figure*}
\centering
\begin{tabular}{cccc}
 \includegraphics[width=0.22\textwidth]{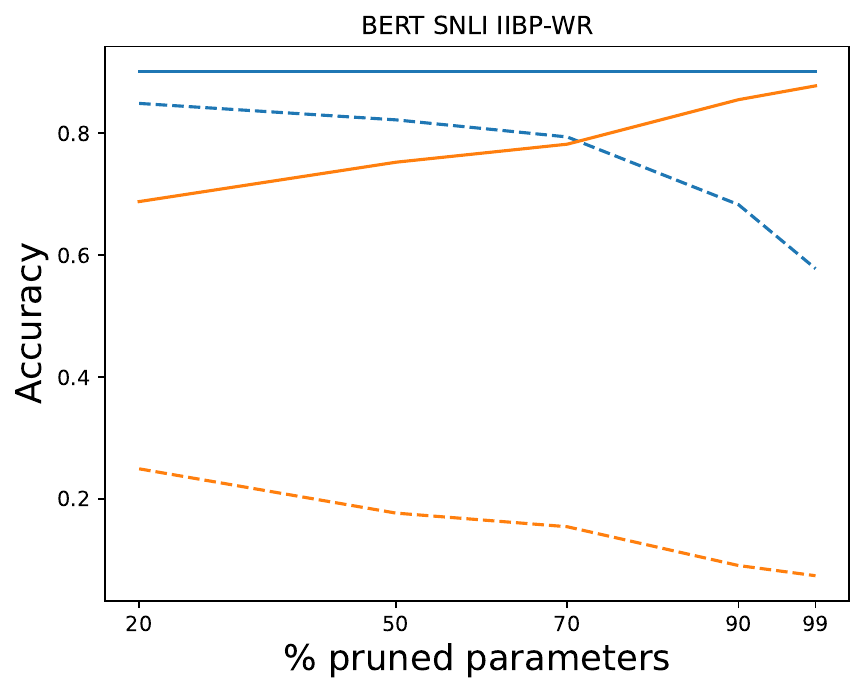} &
 \includegraphics[width=0.22\textwidth]{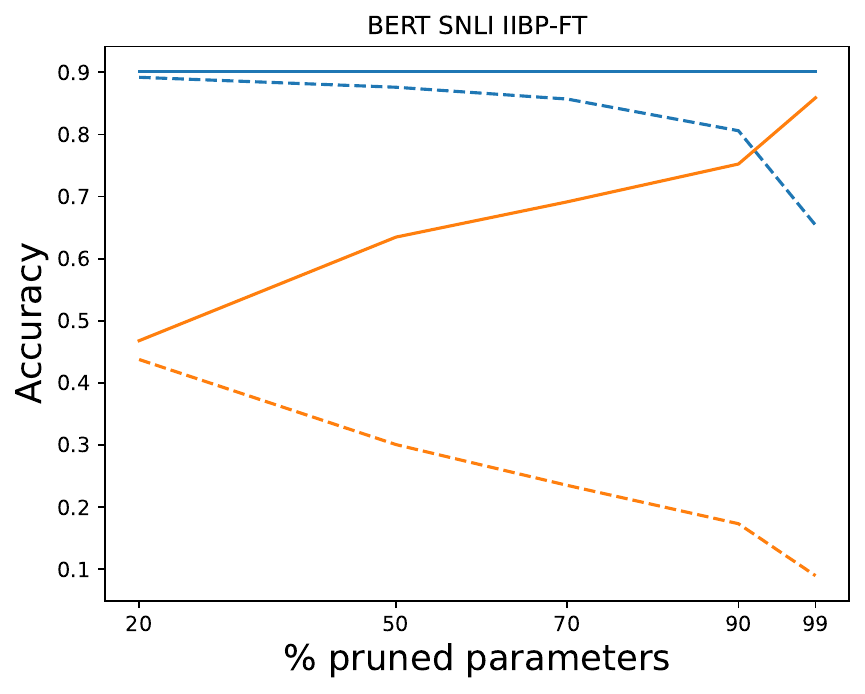} &
 \includegraphics[width=0.22\textwidth]{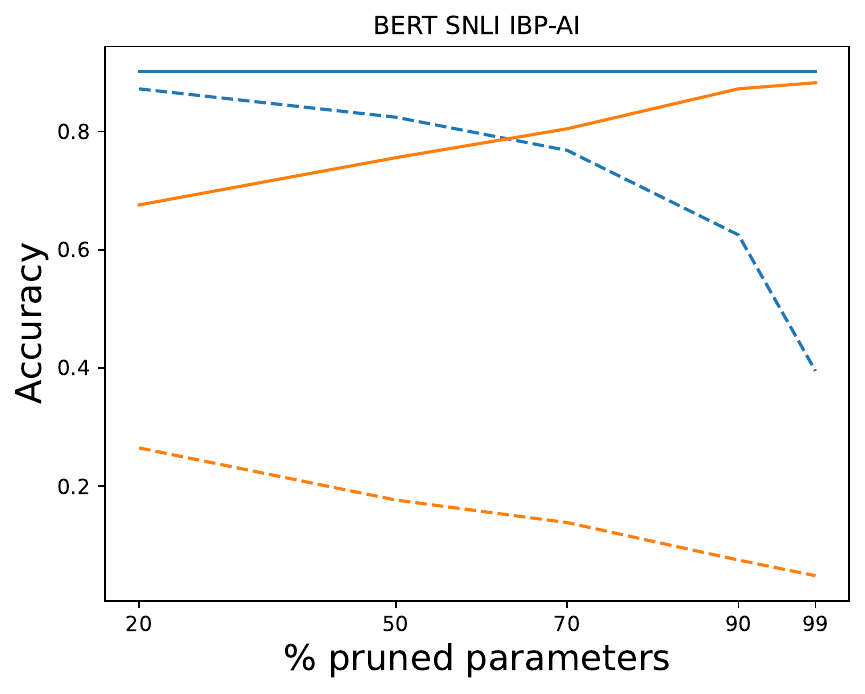} &
 \includegraphics[width=0.22\textwidth]{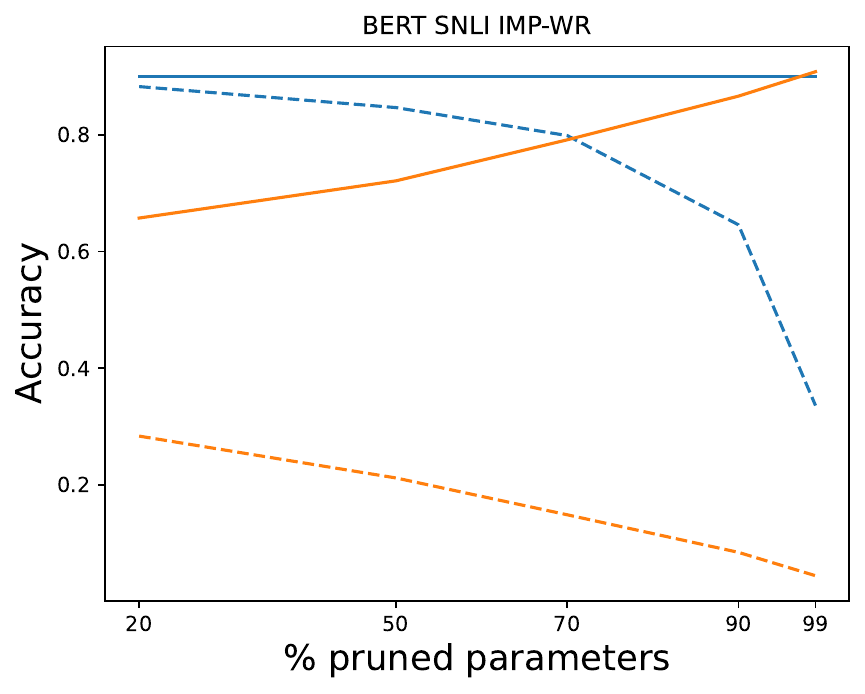} \\

 \includegraphics[width=0.22\textwidth]{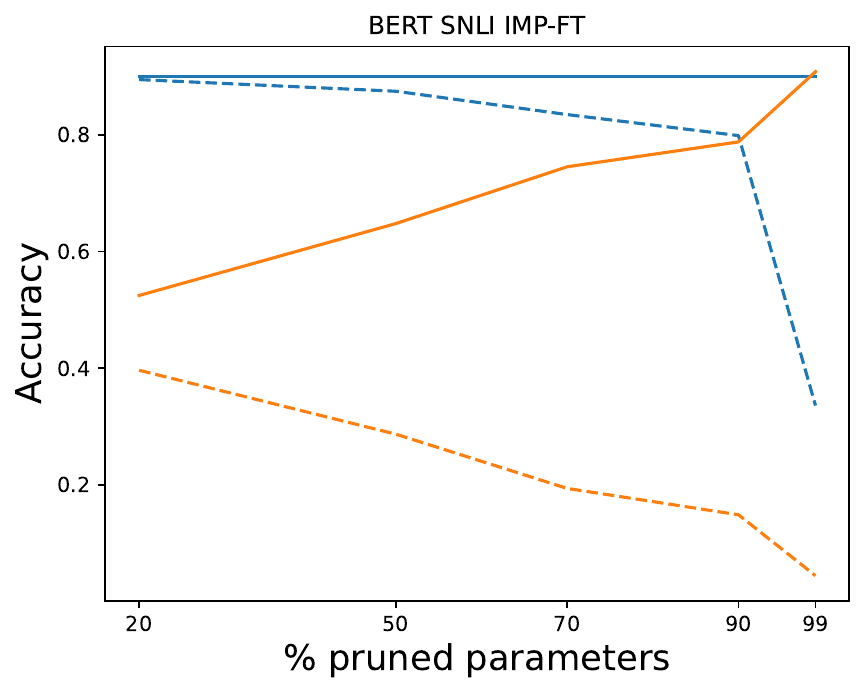} &
 \includegraphics[width=0.22\textwidth]{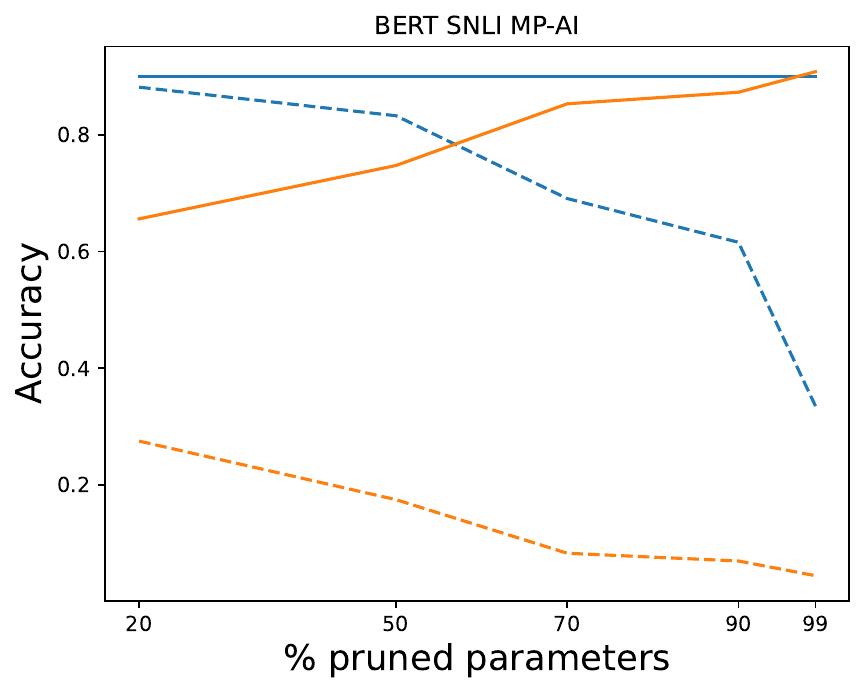} &
 \includegraphics[width=0.22\textwidth]{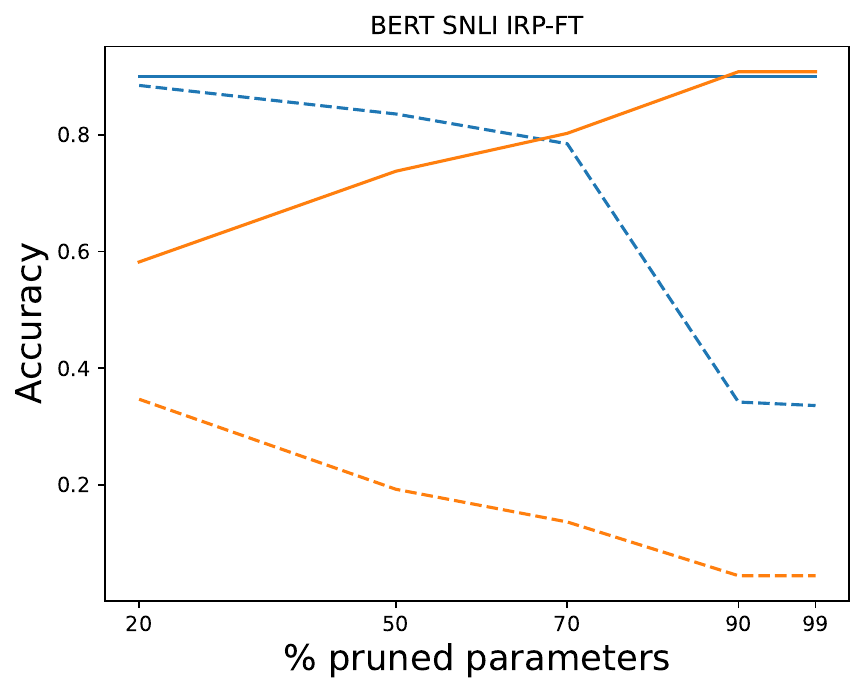} &
 \includegraphics[width=0.22\textwidth]{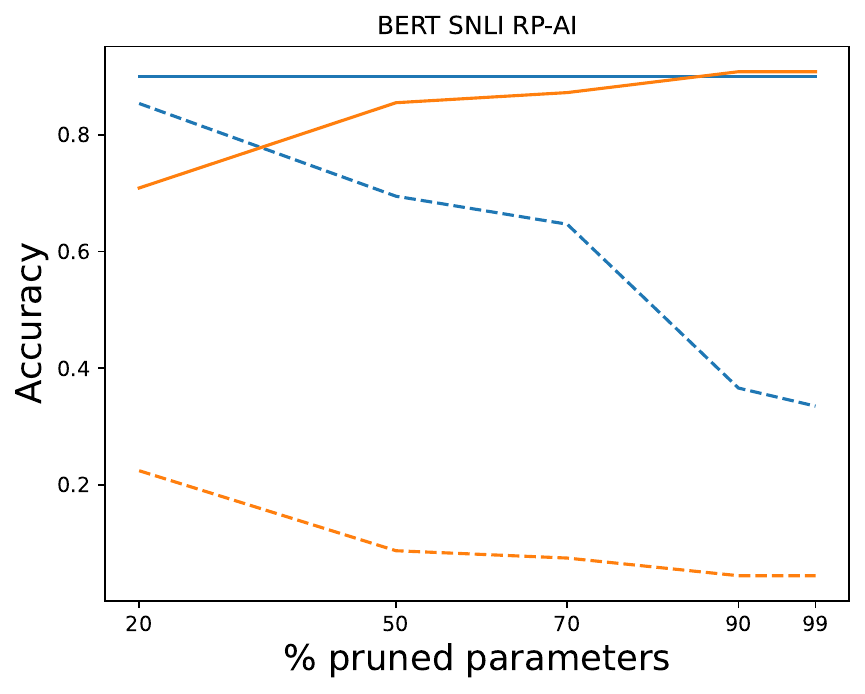}  \\

 \multicolumn{4}{c}{\includegraphics[width=0.5\textwidth]{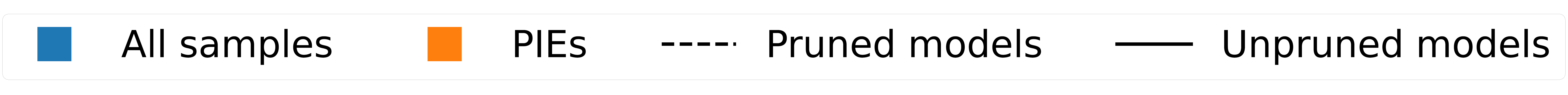}}         
\end{tabular}
\caption{Accuracy (y axis) of unpruned (solid line) \& pruned (dotted line) BERT on SNLI, for all data points (blue) or only for PIEs (orange), per pruning threshold (x axis), over 30 initializations. Each plot is a different pruner.
}
\label{fig:PIEs_against_all}%
\end{figure*}

To probe further into the extent of this impact, Figure \ref{fig:PIEs_against_all} shows accuracy only on PIEs versus accuracy on all data points, for BERT and SNLI. The plots of the other configurations are in Appendix \ref{app:pruning_and_occurence_of_PIEs} and have overall similar trends. We see that accuracy is overall lower on PIEs (orange) than on all data points (blue), for both pruned and unpruned models, with few marginal exceptions for 99\% pruning and BILSTM, where the scores are almost the same. The fact that accuracy is lower for PIEs than for all data points confirms the findings reported above. However, interestingly, Figure \ref{fig:PIEs_against_all} also shows that the impact of pruning upon accuracy is much larger on the subset of PIEs than on all data points: the gap between the two orange lines (PIEs) in Figure \ref{fig:PIEs_against_all} is notably larger than the gap between the two blue lines (all data points). Even when pruning 20\%-50\%, which according to Figure \ref{fig:effectiveness_all} has overall small drops to the mean accuracy of all data points for most pruning methods, still, the drop in accuracy to the data points of the dataset that are PIEs is much larger. 
When pruning highly accurate models, accuracy decreases, and PIEs are more likely to be misclassified. Thus, \textbf{PIEs bear most of the brunt and more focus is needed on them. This effect goes unnoticed when reporting the mean accuracy over all data points}. 

\subsection{Influential examples in PIEs}\label{sec:exp4}
The above findings suggest that PIEs are hard for inference. Next, we try to quantify this hardness, by studying how many of the PIEs are in fact \textit{influential examples}, i.e., data points that have the largest influence on how well the model generalises to unseen data, irrespective of whether this influence is positive or negative. We do this using the EL2N score  \cite{paul2021deep} as per \citet{jin2022pruning}.

Given a model with weights $w_t$ during training iteration $t$, and given an example $(x,y)$ where $x$ is the input and $y$ is its label, $EL2N(x,y)$ is the L2 distance between the predicted probabilities $p(w_t,x)$ during  $t$\footnote{As the EL2N score is not reliable until at least one epoch of fine-tuning has been computed \cite{fayyaz2022bert}, we only monitor the scores after the model has undergone training for at least one epoch (the first epoch that exceeds 30\% of the total training epochs).} and the one-hot label:
\begin{equation}
   EL2N(x,y) = \mathop{\mathbb{E}}{\left[||p(w_t,x)-y||_2\right]}
\end{equation}
Examples are grouped into 20 bins based on their EL2N score percentiles.  
Higher EL2N scores mean that the model undergoes larger weight updates when the example is presented early in training. So, the bigger the weight changes, the higher the EL2N score, and the higher the influence of an example. 
Note that the above takes place during training, so we obtain PIEs on the training set.

\begin{figure*}
\setlength\tabcolsep{-1pt}
\centering
\begin{tabular}{cccc}

 \includegraphics[width=0.23\textwidth]{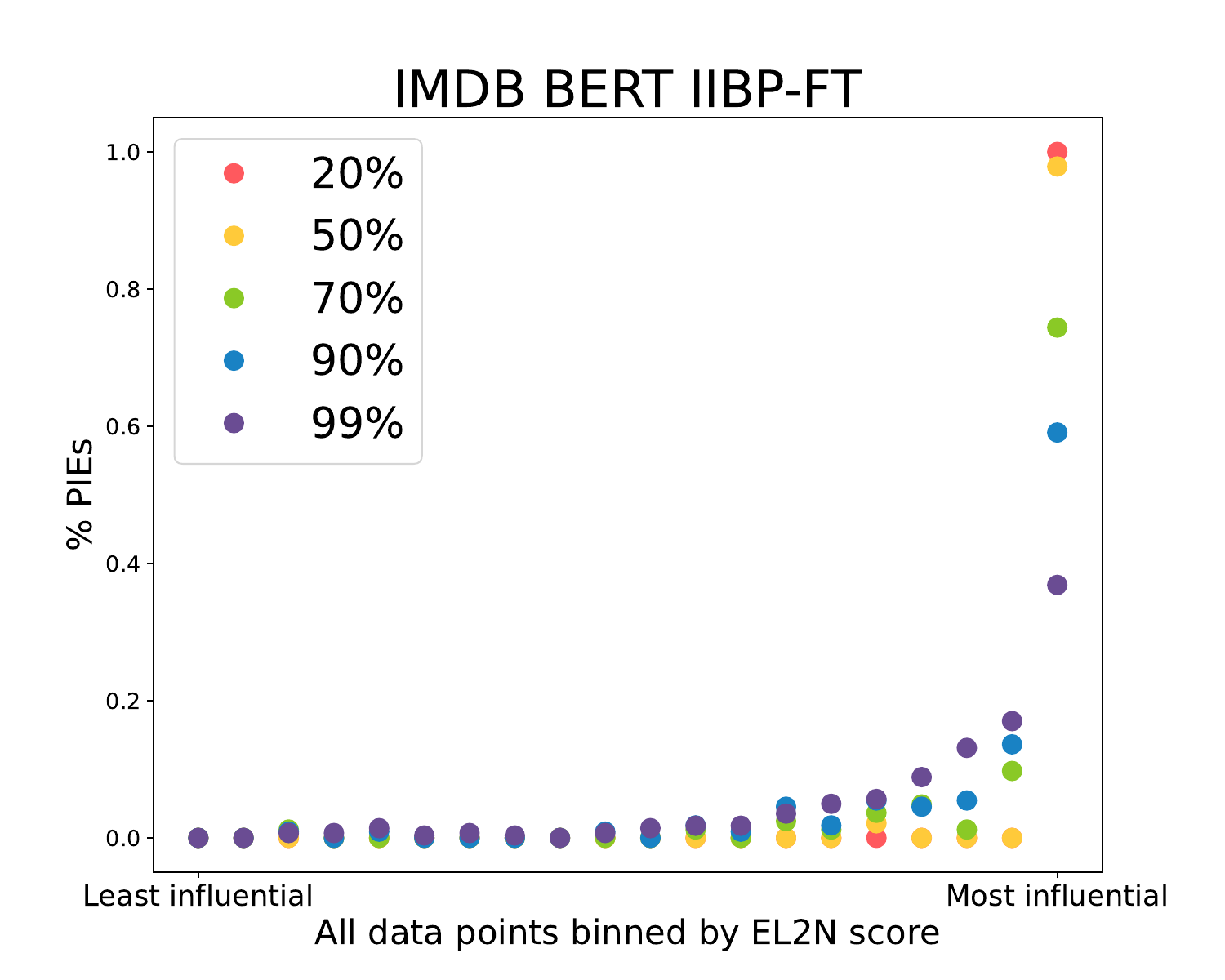} &
 \includegraphics[width=0.23\textwidth]{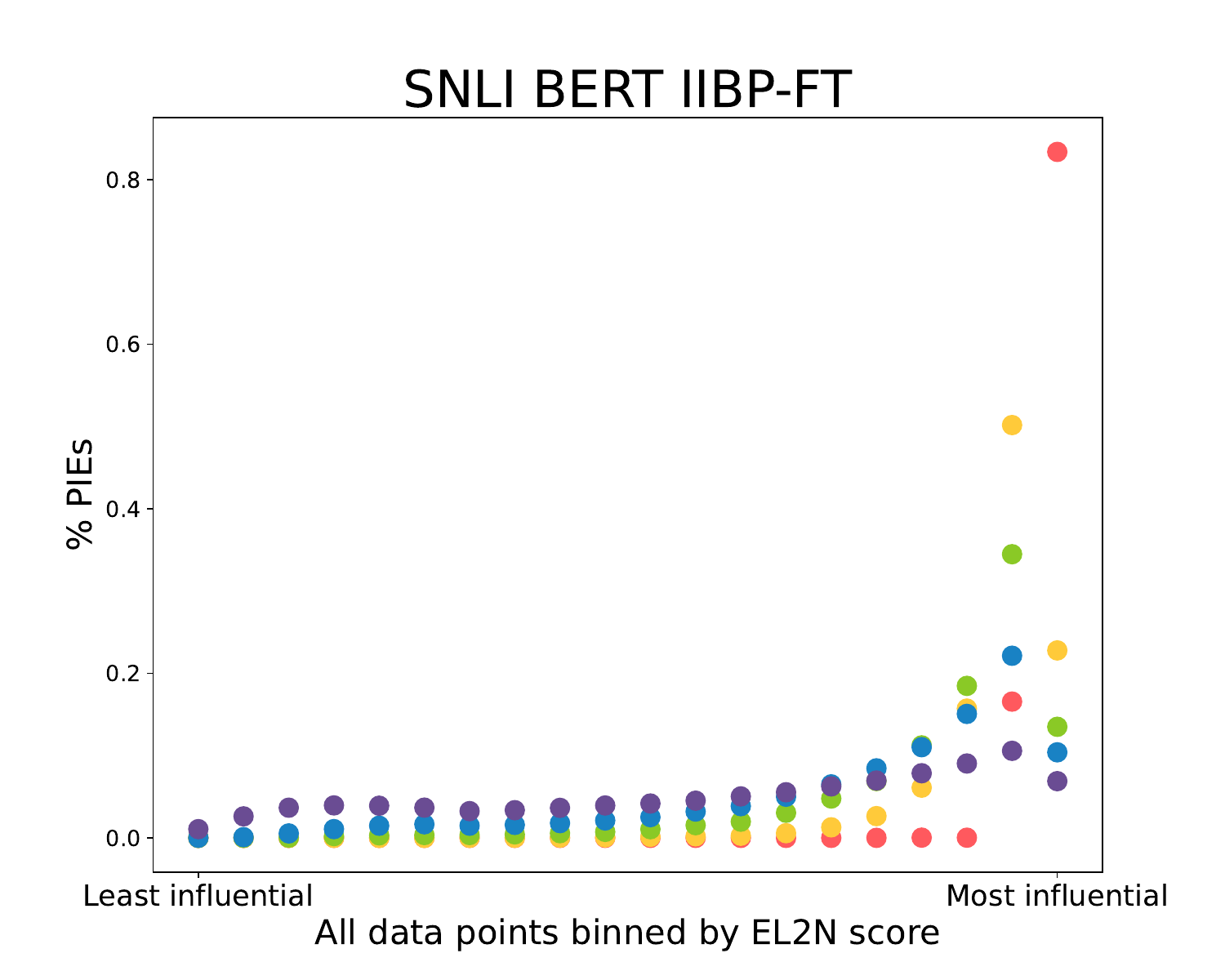} &
 \includegraphics[width=0.23\textwidth]{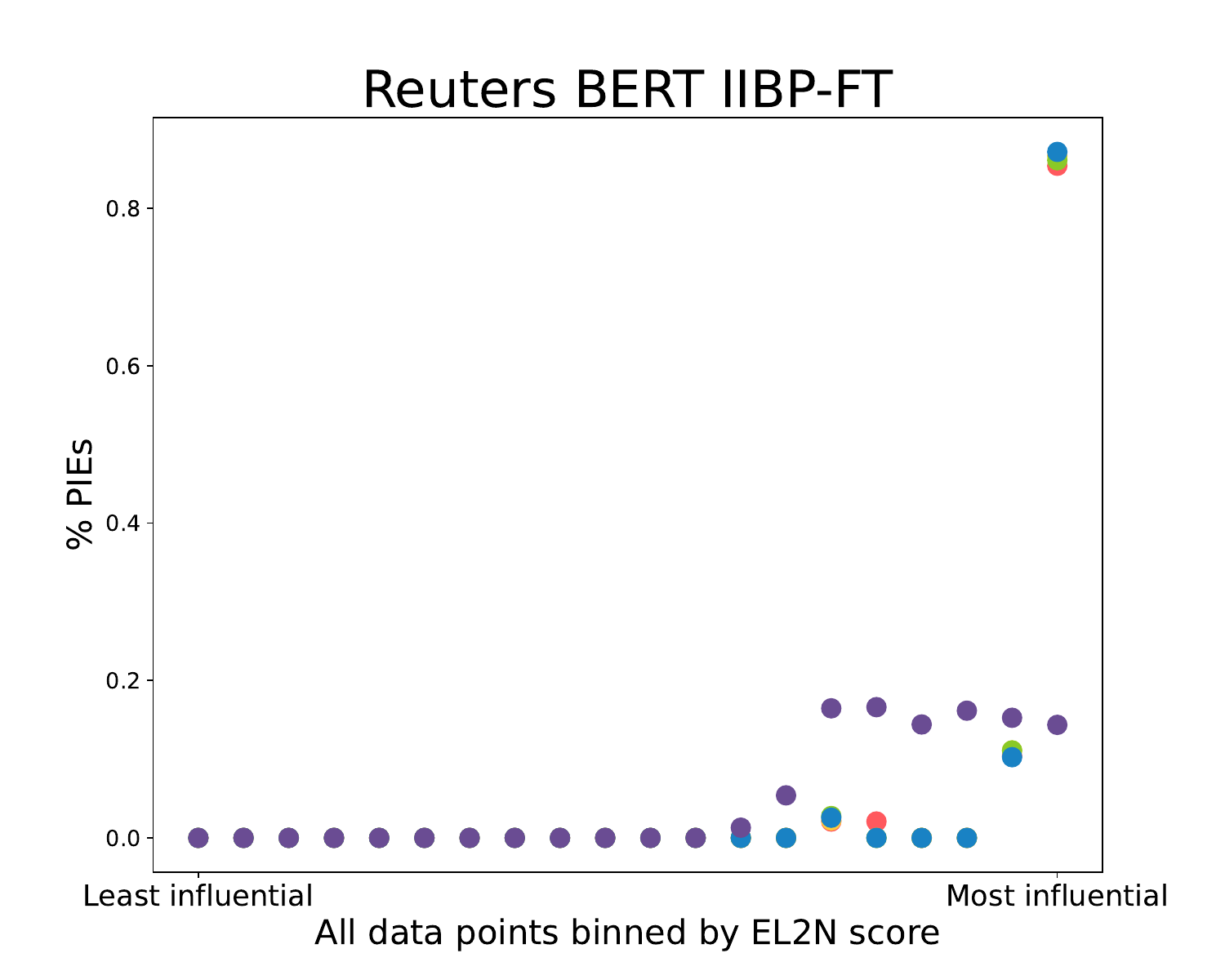} &
 \includegraphics[width=0.23\textwidth]{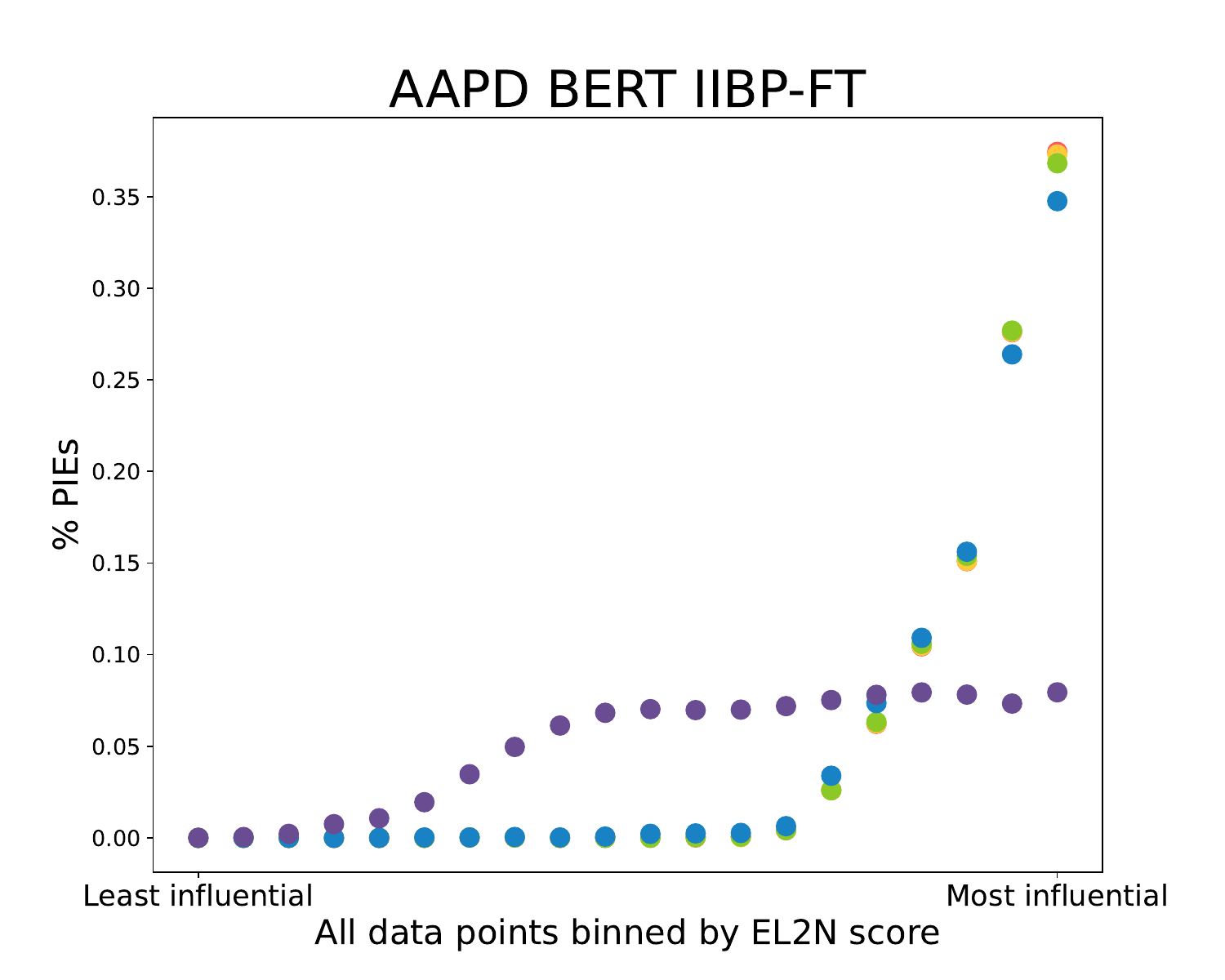} 
  \\

 \includegraphics[width=0.23\textwidth]{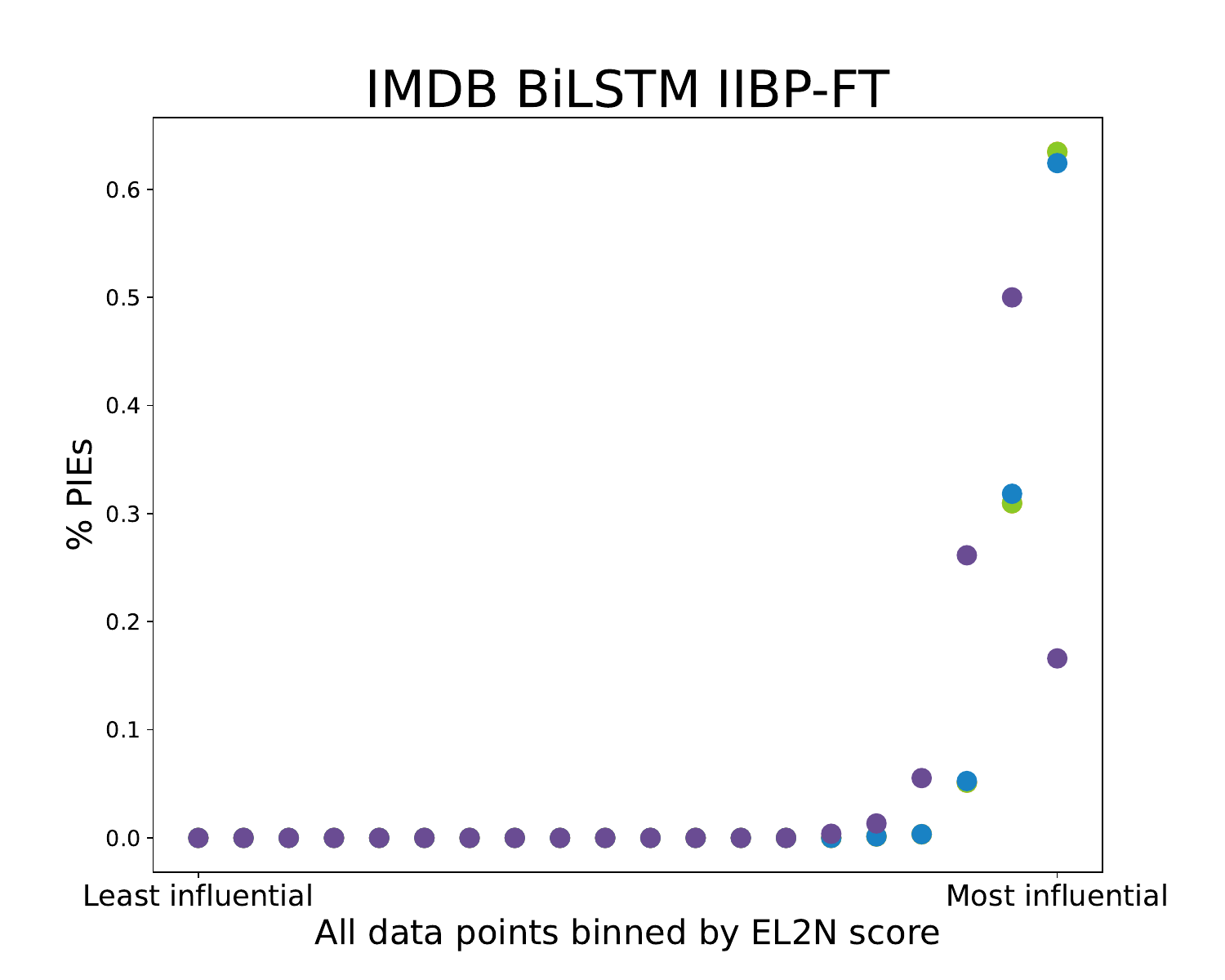} &
 \includegraphics[width=0.23\textwidth]{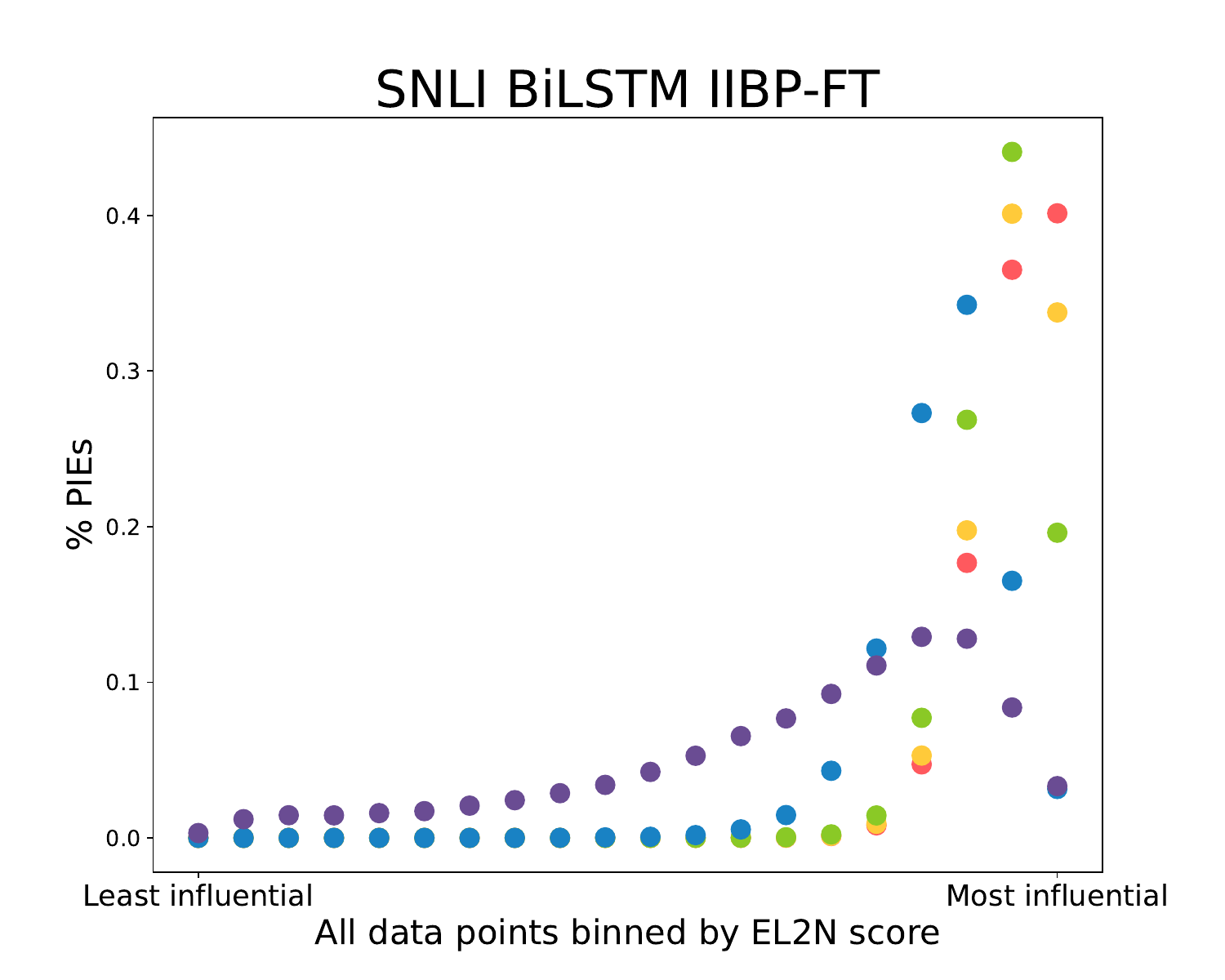} &
 \includegraphics[width=0.23\textwidth]{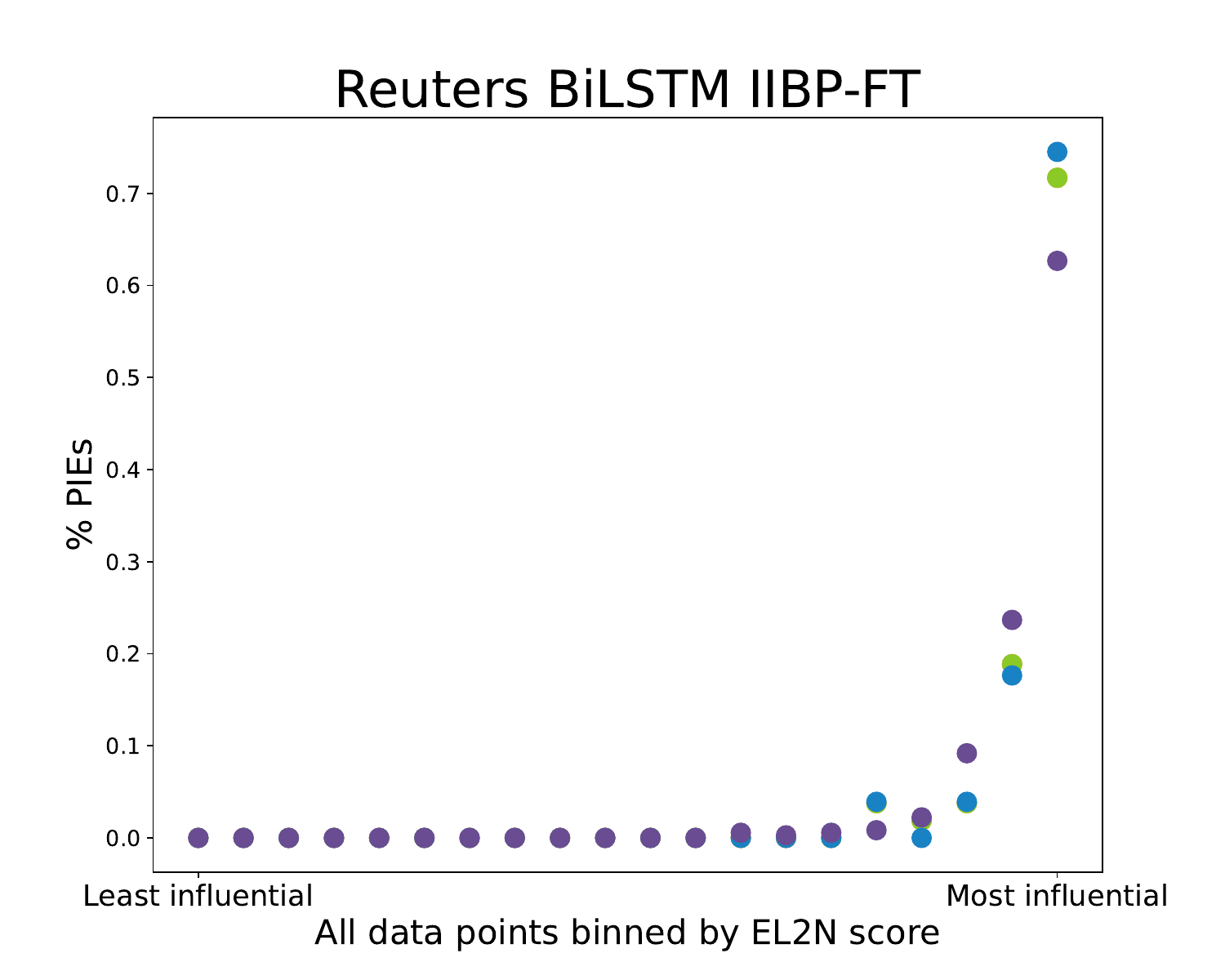} &
 \includegraphics[width=0.23\textwidth]{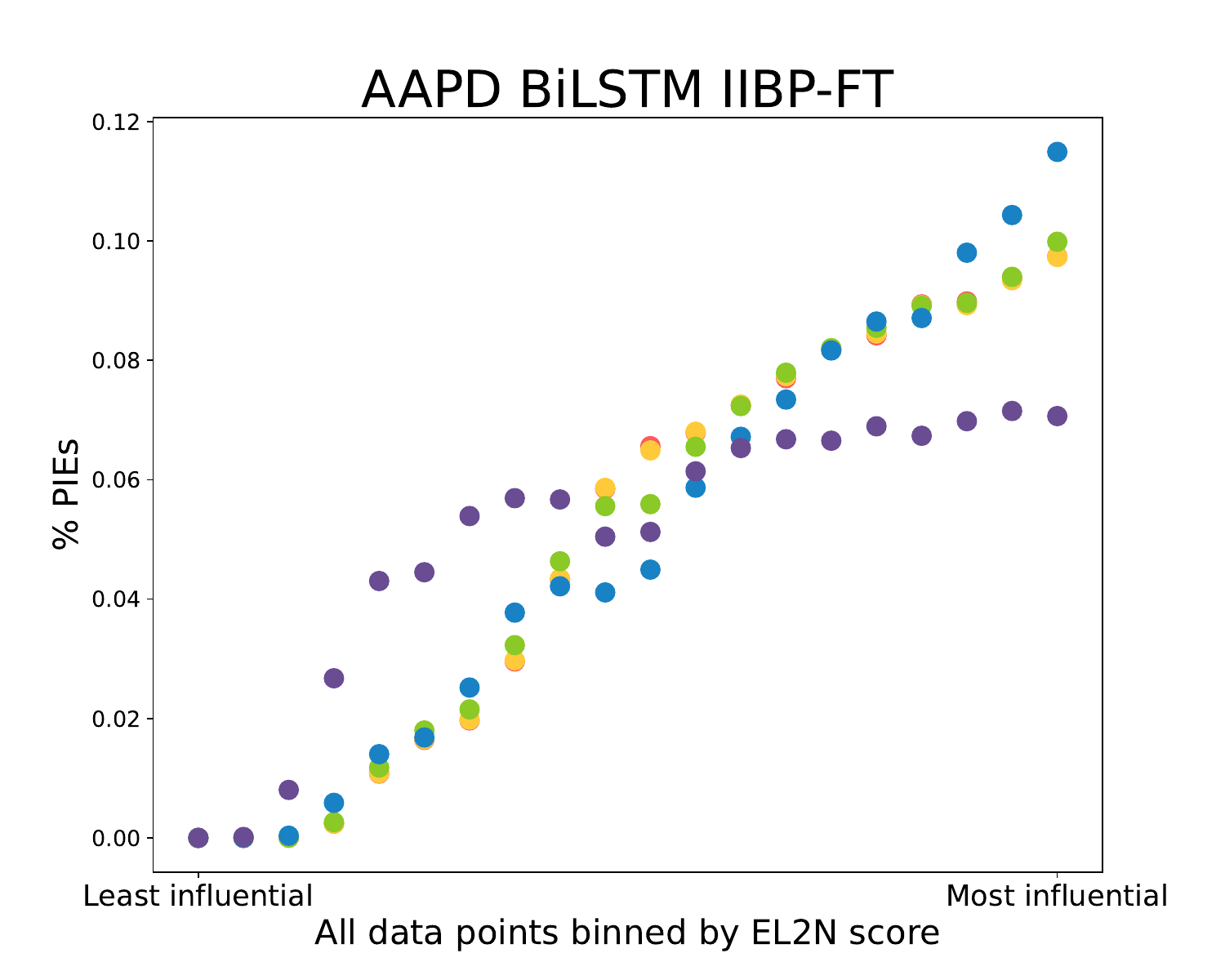} \\

\end{tabular}
\caption{
Percentage of data points that are PIEs (y axis) versus degree of influence (EL2N score) of all data points in the training set (x axis) for IIBP-FT across pruning thresholds (different colours).
}
\label{fig:IMDB_BERT_anal4_grid}%
\end{figure*}

Figure \ref{fig:IMDB_BERT_anal4_grid} shows the distribution of PIEs across the degree of influence of all data points in the training set for IIBP-FT (the rest of the plots are in Appendix \ref{app:influential_examples_in_PIEs}). We see that PIEs are concentrated among the most influential data points (right hand side of the plots). This is even more so for BERT, where up to 80\% - 100\% of its most influential data points are in fact PIEs, compared to up to 70\% for BiLSTM. This explains the finding of Section \ref{ss:pruning-occurrence} that BERT is more affected by pruning than BiLSTM, because (a) more influential examples are PIEs in BERT than in BiLSTM, and (b) accuracy/F1 is lower among PIEs than among all data points, as we saw in Figure \ref{fig:PIEs_against_all}. We conclude that \textbf{a considerable amount of those data points that have the largest influence on how
well the model generalises to unseen data are PIEs}.

\subsection{Textual characteristics of PIEs}\label{sec:Textual_characteristics_of_PIEs}
The above findings motivate the need to understand what the text of PIEs actually looks like. We do this using the following eight scores of text readability and length: (1) Automated readability index \cite{senter1967automated}; (2) Coleman–Liau index \cite{coleman1975computer}; (3) Flesch–Kincaid grade level \cite{kincaid1975derivation}; (4) Linsear Write \cite{o1966gobbledygook}; (5) Gunning Fog index \cite{gunning1969fog}; (6) Dale–Chall readability \cite{dale1948formula}; (7) Number of difficult words; and (8) Text length, counted as the number of tokens per text. (1)-(6) are different approximators of text readability in terms of what formal education level would be needed in order to understand the text. (6) approximates comprehension difficulty based on a list of 3000 easily understandable words. (7) is a count of the number of words that are not in the Dale-Chall list of understandable words. 

\begin{figure*}
\centering

\resizebox{2\columnwidth}{!}{%
\begin{tabular}{cccc}

 \includegraphics[width=0.25\textwidth]{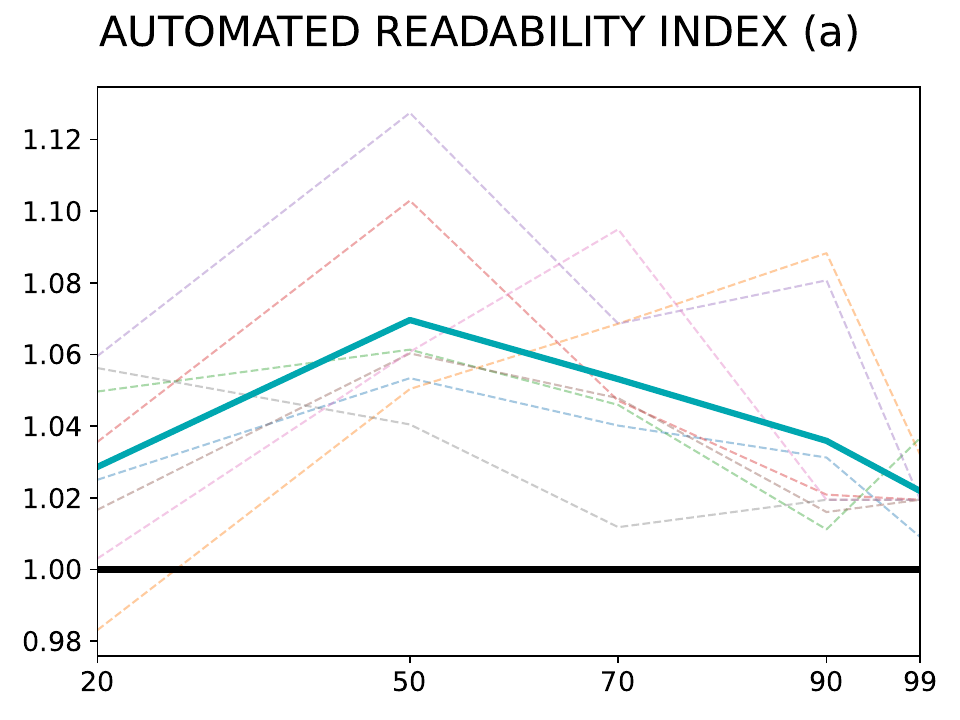} &
 \includegraphics[width=0.25\textwidth]{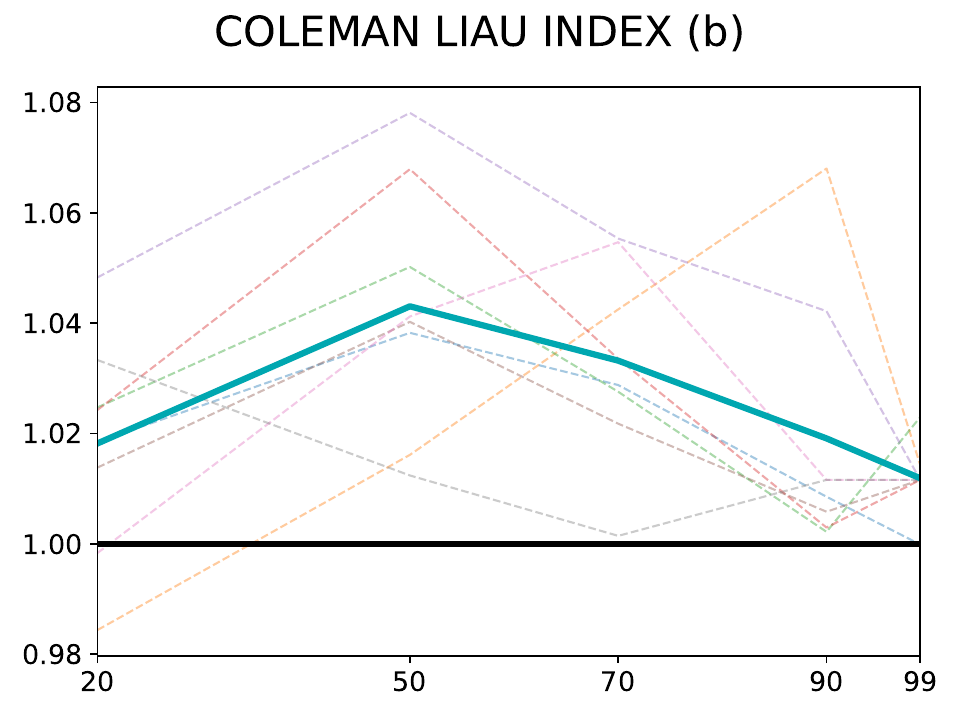} &
 \includegraphics[width=0.25\textwidth]{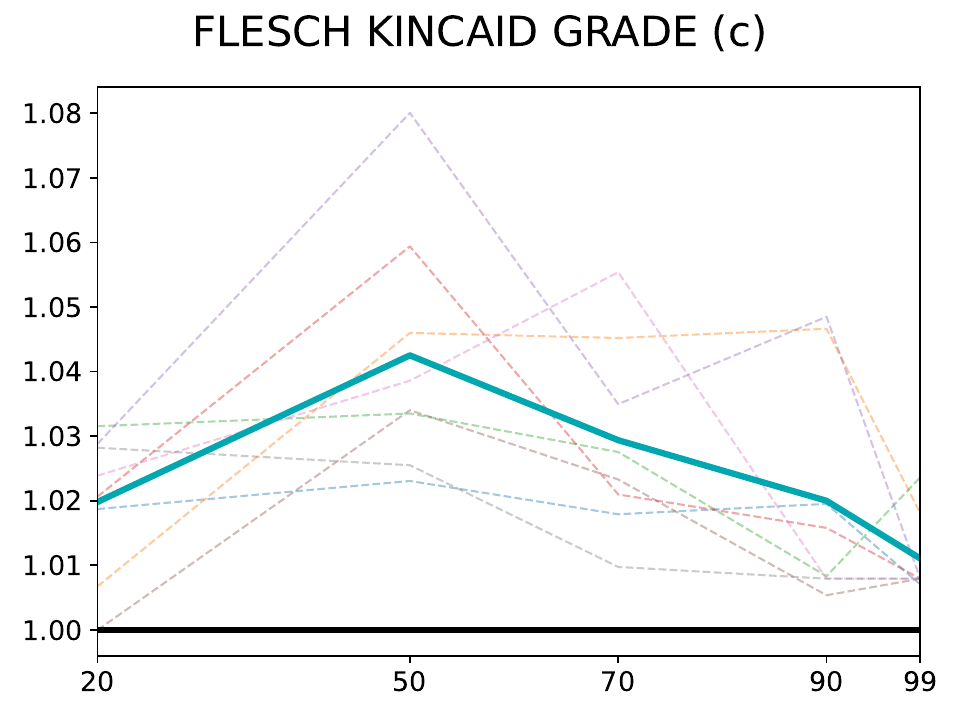} &
 \includegraphics[width=0.250\textwidth]{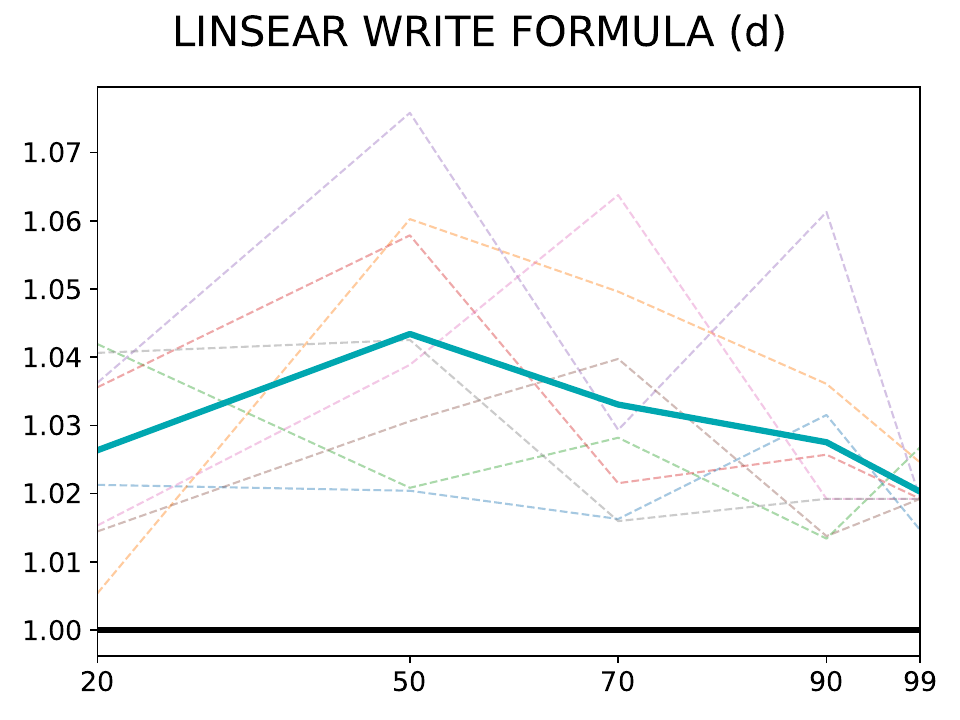} \\
 \includegraphics[width=0.250\textwidth]{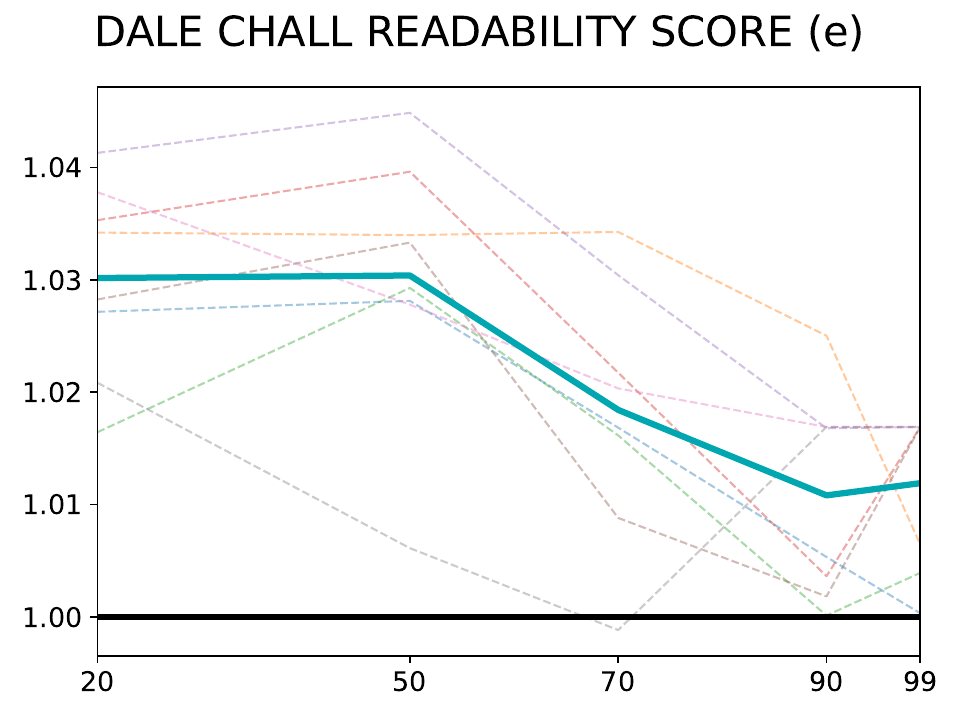} &
 \includegraphics[width=0.250\textwidth]{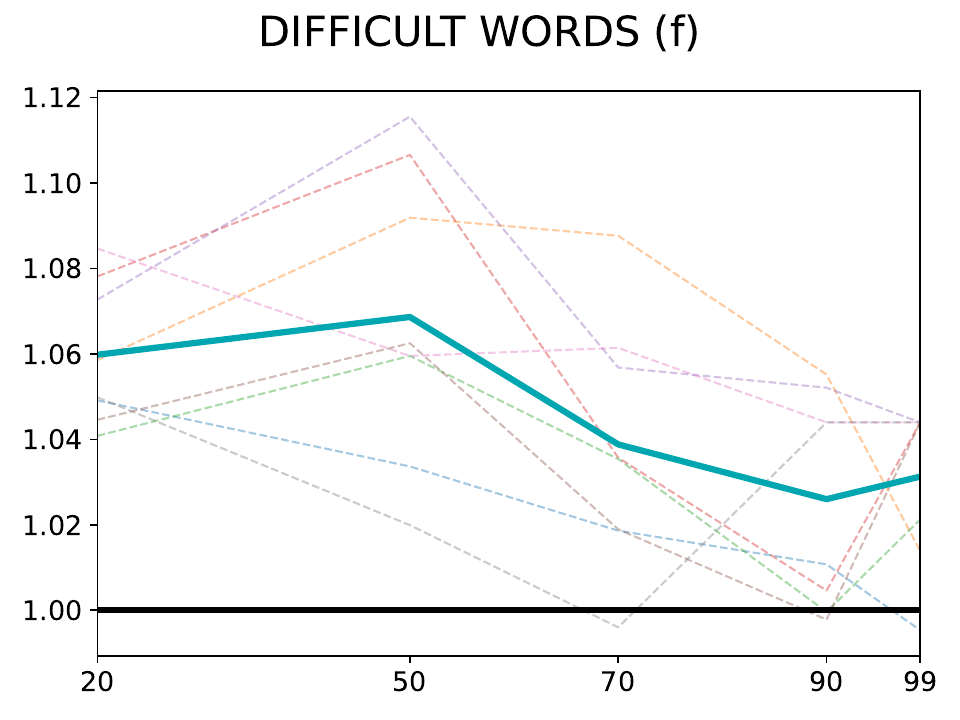} &
 \includegraphics[width=0.250\textwidth]{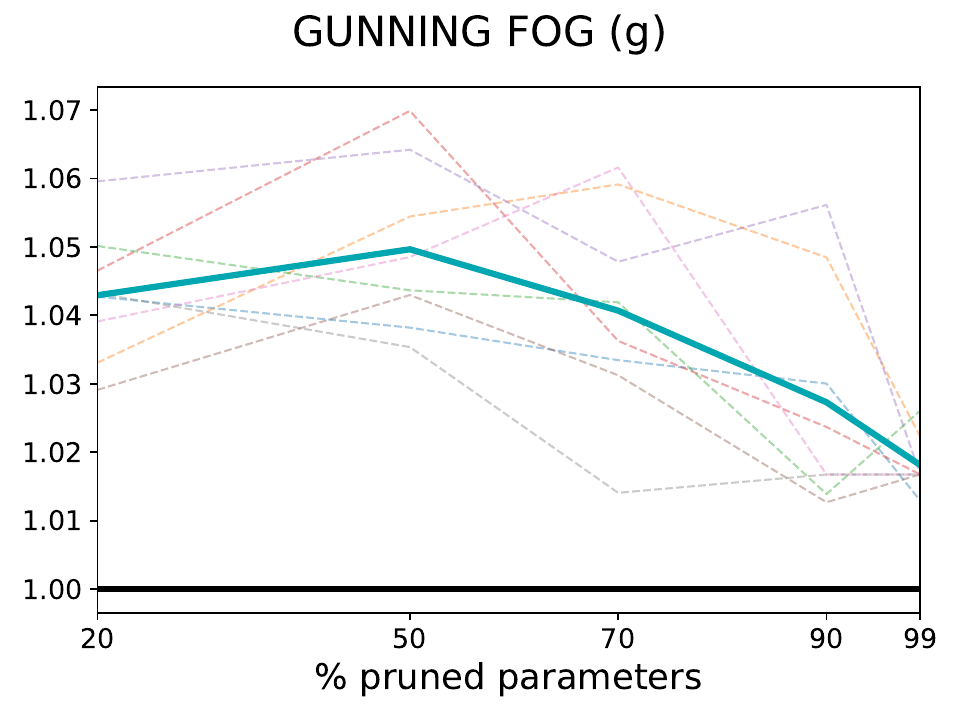} &
 \includegraphics[width=0.250\textwidth]{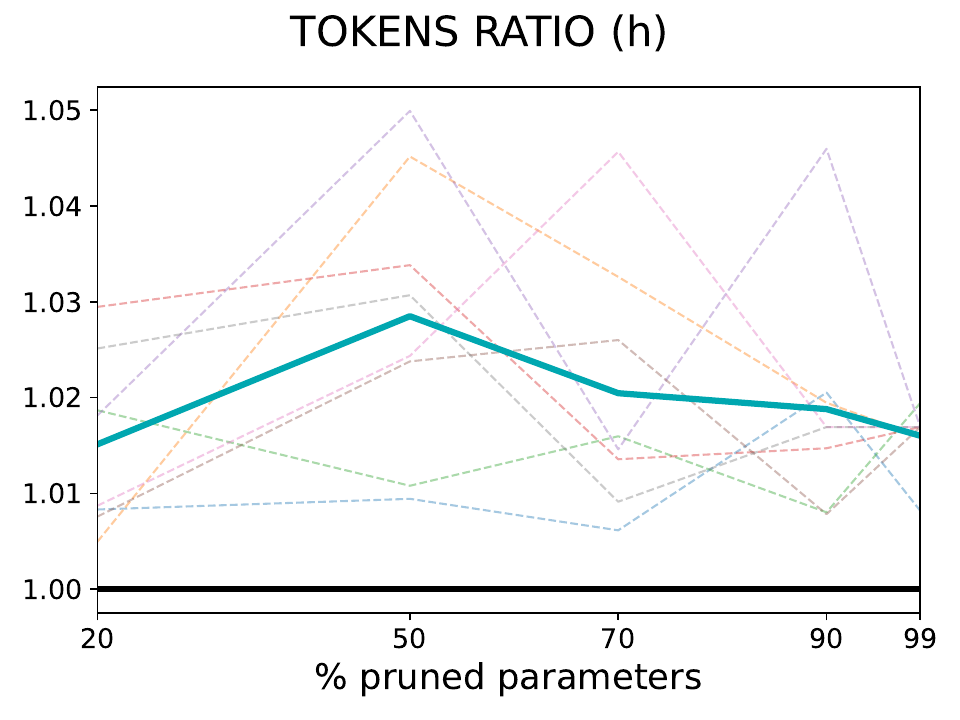} \\
 
 \multicolumn{4}{c}{\includegraphics[width=\textwidth]{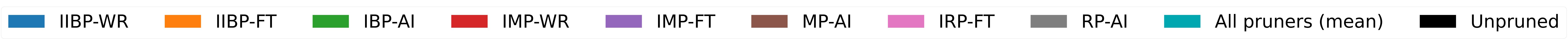}}    
\end{tabular}
}
\caption{How the text of PIEs differs from the text of all data points, according to 7 readability scores (plots (a)-(g)) and text length (plot (h)). 
Ratio between the scores of PIEs and the scores of all data points (y axis), across pruning thresholds (x axis), for BERT and SNLI.
The solid black horizontal line represents equal scores in PIEs and all data points. The solid turquoise line is the mean score of all pruners. Any line above the solid black line means that PIEs are harder to understand (plots (a)-(g)) or have longer text (plot (h)), on average, than all data points.
}
\label{fig:readability_SNLI_BERT}%
\end{figure*}

We compute the above scores first on all data points and then only on PIEs.  Figure \ref{fig:readability_SNLI_BERT} shows the resulting plots for SNLI and BERT (the plots of the other configurations are in Appendix \ref{app:textual_characteristics_of_PIEs}). The black horizontal line represents all data points and PIEs having the same scores. Any divergence from this line reflects how much the scores of PIEs differ from those of all data points. E.g., the point 1.05 on the y axis of the Gunning Fog index plot means that the text of PIEs is approximately 1.05 times harder to understand than the text of all data points.  

In Figure \ref{fig:readability_SNLI_BERT} we see that the formal education level needed for text understanding is overall higher for PIEs than for all data points (plots (a)-(e) and (g)). We also see that the text of PIEs has overall a larger amount of difficult words (plot (f)), and is on average longer than the text of all data points (plot (h)). Overall, according to the average scores of all pruning methods (turquoise line), PIE text is up to 1.03 times harder to understand than the text of all data points (plots (a)-(e) and (g)), with words that are up to 1.06 times more difficult (plot (f)), and text length that is up to 1.02 times longer (plot (h)). This means that \textbf{PIEs tend to be semantically more complex than the average text}. Note that the scores presented in plots (a)-(g) are designed to approximate human (as opposed to computational) difficulty in understanding text. This implies that \textbf{PIEs are more difficult than the average text, not only for LMs} (as shown in Figure \ref{fig:PIEs_against_all}), \textbf{but also for humans} (as shown in Figure \ref{fig:readability_SNLI_BERT}).

\section{Related work}

\textbf{Pruning LMs.} LM pruning has typically been successful when models are first trained and then pruned \cite{large-then-compress:2020}.
Pruning works either by removing structures simultaneously (entire neurons, layers, or even larger sections of NNs) or parameters \cite{zhu2023survey, sun2023simple, frantar2023sparsegpt}.
Examples include pruning entire attention heads in transformer models like BERT without severe inference degradation \cite{NEURIPS2019_2c601ad9} and pruning entire blocks of layers with substantial efficiency gains and minimal effectiveness loss \cite{lagunas-etal-2021-block,llmpruner:2024}. 
Such pruning methods can lead to more interpretable and manageable models but have the disadvantage that they tend to be architecture-specific. Unlike these approaches, in unstructured pruning, LM parameters/weights are pruned individually. This makes unstructured pruning agnostic to particular model architectures \cite{lecun1989optimal}, making it possible to compare the effect of pruning on different types of LMs. As a result, unstructured pruning has been successfully applied in NLP \cite{zhu2023survey, sun2023simple, frantar2023sparsegpt,lottery-nlp:2020}.
Moreover, pruning can operate on the entire network at once \citep{janowsky1989pruning}, i.e., global pruning, or on a fraction of the network \citep{han2015learning}.
The latter is called local pruning and can result in models that are more effective than global pruning \citep{mishra2021does}. 
In our study, we use only local unstructured pruning methods, allowing us to study PIEs in both transformers and RNNs.

For BERT in particular, it has been shown that a substantial amount of pruning can be applied during pre-training without significant loss in inference \cite{Sanh2020CompressingBERT}. It has also been shown that specific parameters that are redundant to such transformer architectures can be accurately identified by dedicated second-order pruning methods, such as Optimal BERT Surgeon \cite{Frantar2022OptimalBERTSurgeon}. However, another body of recent work also shows that complex LM pruning methods do not always work better than simpler, more straightforward pruning
\cite{sun2023simple,frantar2023sparsegpt}.  

Finally, researchers have also assessed, not only the accuracy, but also the loyalty (preservation of individual predictions) and robustness (resilience to adversarial attacks) of pruned BERT models \cite{Xu2021BeyondPreservedAccuracy}. The findings reveal that traditional pruning methods that seem to maintain overall accuracy,  may in fact affect the loyalty and robustness of the model. This line of work, similarly to ours, suggests that more nuanced analyses and evaluation approaches are needed to understand how pruning affects LMs beyond simple average accuracy.

\noindent \textbf{Impact of pruning on subsets of data.} While conventional pruned model evaluation has focused on inference time, number of pruned parameters, and effectiveness of the pruned models \cite{surveyPruning, gupta2021compression, paganini2020iterative, renda2020comparing}, an understudied aspect has been the impact of model pruning on subsets of data. As language data is often power distributed, pruning can have a more severe effect on the performance of the least frequent, tail classes \citep{holste2023does}. This can make models less robust and more prone to overfit shortcuts \citep{du-etal-2023-robustness}, result in disparate accuracy across subgroups of data \citep{tran2022pruning, hooker2020characterising}, and affect prediction quality based on sample frequency \citep{ogueji-etal-2022-intriguing}. 
\citet{kuzmin2024pruning} show that in most cases quantization outperforms pruning.
Close to ours is the study of \citet{hooker2019compressed}, who defined PIEs, and found them harder for both NNs and humans to classify. This study was limited to image processing. To our knowledge, our study is the first in-depth examination of PIEs for NLP, with novel findings about where and how often PIEs occur in text data, how they impact inference, and why. 

\section{Conclusions \& Discussion}

We empirically studied how LMs are affected by pruning in the text domain. Unlike most work in this area which looks at overall gains in efficiency and costs to inference effectiveness, we zoomed in on precisely how pruning affects a particular subset of data points where pruned and unpruned models systematically disagree (\textit{Pruning Identified Exemplars} (PIEs)). Using two LM architectures, 
four datasets, eight pruning methods, and five pruning thresholds, we found that PIEs impact inference quality considerably, but this effect goes undetected when reporting the mean accuracy across all data points. This effect is invariable to class frequency and increases the more we prune. BERT is overall more susceptible to this effect than BiLSTM. We also found that PIEs tend to contain a high amount of influential examples (data points that have the largest influence on how well the model generalises to unseen data). Probing into what it is about PIEs that makes them both hard and impactful to inference, we found that their text is overall longer and more semantically complex, and harder to process not only for LMs but also for humans, based on human text readability approximations.

Overall, our findings suggest that, the more influential and complex a data point is, the higher the chance that pruned and unpruned models will disagree on its prediction, impacting disproportionately a subset of the dataset, yet going generally unnoticed when reporting mean accuracy on the whole test set.
This can pose significant risks to LMs, such as focusing on easier examples and sacrificing inference quality on more difficult examples that are however linked to better generalisation. 

Given the increased call for compressing LMs, pruning them without considering the effect on PIEs can make models vulnerable in high-stakes applications, where relying solely on good top-line performance is inadequate to guarantee the model's reliability and trustworthiness across data points and independently of class distribution. For instance, in fact-checking a pruned LM may misclassify long and complex documents more frequently, potentially labelling them as factually correct or incorrect based on their length and complexity rather than their content. 
In medicine, domain expertise can lead to higher text complexity and lower readability. This underscores the need for accurate models on complex and longer documents, as pruned models may significantly affect prediction quality. Similar considerations apply to the legal domain, where documents often contain terminology and complex sentences that are challenging even for humans.

PIEs are not only complex for the model but also include complex language according to different readability scores used to assess human readability. This makes the models more vulnerable to texts that are less readable for humans, limiting the usefulness of pruned models even in applications with a human in the loop.
Identifying PIEs in these fields can help determine if the pruned model can be trusted, while also making the model's behaviour more interpretable, enabling the study of the shared textual characteristics of the data points that the pruned model processes differently from the unpruned one.
More broadly, PIEs focus on cases whose impact goes unnoticed with averaged scores, becoming a valuable tool for understanding pruned models' generalization beyond top-line scores, which is relevant across all areas.
Lastly, identifying PIEs in the previously mentioned scenarios is crucial for understanding the extent to which the model is affected by pruning. 
Leveraging PIEs to produce new, more reliable pruned models can help reduce the impact on longer and more complex documents, making the pruned models more trustworthy.

Future work includes studying PIEs when pruning LLMs, extending PIEs to tasks such as text generation, and researching how to balance the impact of pruning across PIEs and all data points. 
Moreover, PIEs can be used to study in isolation the data points where pruning aids or hinders inference. 

\section*{Limitations}

We evaluated the effects of pruning across eight pruning methods, two LM architectures, and four datasets. While these are representative, we cannot rule out the possibility that other pruning methods or model architectures might yield different results. Moreover, while we train BiLSTM from scratch, BERT utilizes an existing backbone model. This may affect some specific findings. Nonetheless, our findings across all tested experimental conditions, datasets, and models consistently point in the same direction and unanimously support our conclusions.

More work is needed to understand the effect of pruning through PIEs on different architectures (e.g., decoder-only models) and models, varying unpruned model sizes, and pruning methods. While we utilized extensive resources from the LUMI supercomputing infrastructure (over 28000 AMD MI250X GPU hours), it was not practically feasible to experiment with the latest large language models or include decoder-only models in our setting where we aimed at varying many pruning thresholds, methods, and datasets, while experimenting on both attention based and RNN based models. To make investigations of PIEs computationally more accessible, future studies could investigate individual architectures and pruning methods in isolation and benchmark their results against our findings.

We also did not explore the design of new pruning algorithms that take into account properties of the data, such as the link between the influence of the examples and pruned and unpruned models' disagreement. These could help to mitigate both general effectiveness drops as well as improved handling of examples that are important for training and downstream usage of the models, which we leave for future work.
We show that PIEs are useful for understanding how pruning affects model predictions on harder-to-read, longer texts that are more influential for model generalisation. Future work should use these insights to guide pruning and mitigate its impact on models' ability to process harder and longer texts, as well as other text properties. For instance, \citet{williams-aletras-2024-impact} observed how calibration data directly affects pruned models. Based on this, one mitigation strategy could enhance gradient-based pruning algorithms by selecting calibration data that reflect complex textual characteristics or by sampling calibration samples according to the most influential training examples. This approach could lead to a text-complexity-aware pruning strategy.

Alternative mitigation strategies might involve selecting a subset of data to fine-tune the model after pruning, ensuring that examples with complex or long text are presented more frequently to the model. This would improve the model's handling of such data points.
Additionally, future work could develop new pruning methods that consider the importance of PIEs and explore rebalancing techniques to mitigate pruning's adverse effects.

\section*{Ethics Statement}

We adhere to the ACM Code of Ethics and Professional Conduct to ensure our work’s integrity, fairness, and transparency.

Our study aims to enhance the understanding of natural language model pruning. Our results reveal the trade-offs between performance and the impact of pruning for examples that are potentially lower frequency and minority class, but may be highly important for downstream usage of the models. This can be particularly the case for high stakes domains, such as fact checking, medical informatics, and conversational and retrieval models that can impact decisions and opinions of individuals. By investigating the nuances of model pruning, we aim to inform modeling practices that consider both technical performance and potential weaknesses of compressed models. This can be critical in many specific application domains, but that is not always accounted for in standard performance analysis focusing on average effectiveness. To this end, our research identifies cases and settings where pruned models may underperform, providing valuable insights to avoid potential harm.

We have conducted our research fully transparently, documenting our methodologies and choices. While our study did not involve human subjects directly, it utilized publicly available datasets that include human annotations. We ensured that the use of these datasets complied with their respective terms of use.

We have respected all intellectual property rights in our research, and to our best knowledge properly citing all sources and datasets used. Our work builds on existing literature while providing new contributions to the field. We have also appropriately acknowledged the contributions of other researchers and sources that have informed our work. 

We acknowledge that access to computing resources can be a barrier for some researchers aiming to reproduce our results. Our code to run the models was trained with a LUMI supercomputer,\footnote{https://www.lumi-supercomputer.eu/} available for academic use to reproduce the results.

\section*{Acknowledgements}
This work was funded by the Algorithms, Data \& Democracy project (Villum \& Velux funds).
We acknowledge the Danish e-Infrastructure Consortium (DeiC), Denmark, for awarding this project access to the LUMI supercomputer, owned by the EuroHPC Joint Undertaking, hosted by CSC (Finland) and the LUMI consortium through DeiC, Denmark.
We also thank Theresia Veronika Rampisela for her invaluable assistance in revising the draft of this paper.

\bibliography{anthology,custom}

\appendix

\section{Implementation Details}\label{app:implementation_details}

\subsection{Language Model Architectures}\label{app:language_model_architectures}

We use the pretrained uncased version of BERT-base from HuggingFace as is, which has 12 encoders with 12 self-attention heads \citep{wolf2019huggingface}.
BERT takes as input the tokenized text. We set the output layer size to match the number of classes of the dataset the model is trained on.
During training, we tune all of BERT's parameters

Our BiLSTM models receive as input a vector representation of the words in the text. To build such a vector we use Glove embeddings of size 300 \citep{pennington2014glove}.
We input the embeddings to a multilayer BiLSTM.
We set the output layer size of the BiLSTM models to match the number of classes of the data set the model is trained on. On BiLSTM, we always use rectified linear units (ReLu) as activation functions.

We present the “percentage of pruned parameters” based on the total number of parameters that can be pruned in the model, instead of all of the parameters of the model \citep{chen2020lottery}. 
In Table \ref{tab:architectures_stats} and Table \ref{tab:architecture_remaining_params} we report information about the number of remaining parameters in the architectures at different pruning amounts.

\begin{table}[h]
\input{Tables/Setup/Architectures/stats}
\caption{Number of LM parameters and \% of parameters that are removed when pruning at 20\%--99\%. Numbers differ per dataset because the different size of the classification layer leads LMs to a different final amount of parameters.}
\label{tab:architectures_stats}
\end{table}

\begin{table}
\resizebox{\columnwidth}{!}{%
\begin{tabular}{lllllll}
\textbf{Architecture} & \textbf{Unpruned} & \textbf{20} & \textbf{50} & \textbf{70} & \textbf{90} & \textbf{99} \\ \hline
BERT & $1.1x10^8$ & $9.2x10^7$ & $6.7x10^7$ & $5.0x10^7$ & $3.2x10^7$ & $2.5x10^7$ \\
BiLSTM & $6.5x10^5$ & $5.2x10^5$ & $3.3x10^5$ & $2.0x10^5$ & $6.8x10^4$ & $1.0x10^4$
\end{tabular}
}
\caption{Number of parameters for the unpruned models, and remaining parameters when pruning at 20\%-99\%.}
\label{tab:architecture_remaining_params}
\end{table}

\subsection{Datasets and Preprocessing}\label{app:datasets_and_preprocessing}

\begin{table*}
\centering
\input{Tables/Setup/Dataset/stats_appendix}
\caption{Datasets' statistics after preprocessing. \# train, \# test, and \# val are respectively the number of instances in train, test, and validation sets. Mean/median, and Min/max are respectively the mean, median, minimum, and maximum number of tokens in the dataset's instances. Tokens 85\% represent a value such that 85\% of the datasets' texts have fewer or equal tokens than such value. Max tokens are the number of tokens, starting from the beginning of the text, after which we truncate texts. \# classes is the number of classes. Task is the task solved using the dataset.}
\label{tab:dataset_stats_appendix}
\end{table*}

In table \ref{tab:dataset_stats_appendix} we report dataset statistics after pre-processing.
IMDB \cite{maas-EtAl:2011:ACL-HLT2011} is a single-label sentiment analysis dataset, made of reviews of movies. Each review is either positive or negative.
IMDB has the longest sentences and the fewest classes across all our datasets on average. 
SNLI is a single-label natural language inference dataset. Each sample contains two sentences, and the task is to determine if the relationship between them is entailment, contradiction, or neutral.
The dataset is available under a CC BY-SA license.
SNLI has the most training samples and the shortest sentences among all our datasets on average.
Reuters-21578 is a multi-label document categorization dataset, made of Reuters news belonging to 120 topics. Each news item is categorized and can belong to multiple topics. After preprocessing, the dataset has 23 classes. 
The dataset is available under CC BY license.
Reuters has the fewest training samples among our datasets.
AAPD is a multi-label document categorization dataset of article abstracts in computer science. Each arrticle can belong to multiple subjects, and the task is to identify the subjects given the abstract.
The dataset is available under CC BY license. AAPD has the most classes across our datasets.

\noindent\textbf{Dataset preprocessing.}
IMDB has 25000 training examples and 25000 test examples. To perform hyperparameter tuning of our models, we apply stratified sampling from the original training set to create a validation set of 5000 samples.
On SNLI we use the original data set splits.
On Reuters-21578 we remove all of the topics that do not appear in at least 100 documents and all of the documents that do not belong to at least one of the remaining topics. We perform stratified sampling and create three partitions by allocating 30\% of the samples to the training set, 15\% to the validation set, and 15\% to the test set.
For computational efficiency, before computing the statistics shown in Table \ref{tab:dataset_stats_appendix}, we convert texts in the Reuters dataset to lowercase and remove punctuation and numbers.
Lastly, we use the original splits for the AAPD data set.

We further pad and truncate texts to submit training examples in batches, and we select a strategy to handle terms that are not present in the model's vocabulary (OOV). We explain these two steps next.

To fully take advantage of the available hardware, we submit training examples to the models in batches. When multiple texts with a different amount of tokens are present in a batch, our models require padding on the shorter texts in such a way that each input has the same amount of tokens.
To have batches where each text is of equal size, we truncate long texts and pad short ones. Note that we do not remove documents based on a minimum amount of tokens in the text.
To truncate the texts, we find a threshold after which we perform truncation. 
We define this threshold as the first power of two after which, by selecting the value as a threshold, at least 85\% of the texts in the dataset do not need to be truncated. The resulting thresholds are reported as ``Max tokens'' in Table \ref{tab:dataset_stats_appendix}.
An exception is made for SNLI. The SNLI dataset is made of short texts, and even the longest text is under 128 tokens. Hence we consider 128 tokens, representing the whole text for each sample in the data set.
We then proceed to pad short texts in each batch to always exactly match the number of tokens specified in Table \ref{tab:dataset_stats_appendix}.
For BERT we use the huggingface's tokenizer padding and pad all of the texts in each dataset to the respective ``Max tokens'' value in Table \ref{tab:dataset_stats_appendix}. BERT will mask and ignore the padding.
For the BiLSTM model, we represent padding as a randomly generated embedding according to the mean and std distribution in Glove.

On BERT, OOV terms are assigned the default UNK token.
On BiLSTM, we represent OOV terms with a vector defined as the average over all of the present word embeddings.
The result of our preprocessing will be texts with exactly ``Max tokens'' tokens in which OOV terms are represented by the UNK token on BERT and as the average embedding vector on BiLSTM.

\subsection{Pruning Methods}\label{app:pruning_methods}

Model parameters are pruned one layer at a time. We prune uniformly across layers, i.e., we remove the same percentage of parameters in each layer. 
Following \citet{chen2020lottery} and \citet{lottery-nlp:2020,prasanna-etal-2020-bert}, we do not prune embedding layers and biases of the LMs \citep{gupta2021compression}.
We also do not prune the final classification layer, because its weights are likely disproportionately important to reach high effectiveness \cite{frankle2020pruning}.

With iterative pruning, we select a pruning percentage and keep it fixed for each pruning iteration to reach our pruning goal in exactly three iterations across all datasets, LMs, and pruning percentages. We train the model (BERT or BiLSTM) fully for N epochs, prune according to the selected percentage, and then retrain for N epochs. This process repeats until we achieve our pruning target as per \citep{jin2022pruning}. In total, this procedure requires four times the training iterations when compared to pruning at initialization.

\subsection{Hyperparameter Tuning}\label{app:hyperparameter_tuning}

We tune the unpruned model's hyperparameters for each combination of architecture and dataset.
The resulting hyperparameters are then used to train both unpruned and pruned models.
We do not tune hyperparameters of the pruning algorithms. The only tunable aspect when pruning at initialization is the percentage of parameters to prune. However, in our experiments, we fix five different values for this hyperparameter and we test such values on all pruning algorithms, hence, we do not optimize the percentage of pruned parameters.
When pruning iteratively (with or without weight rewinding) we also need to select the number of pruning iterations and the amount of parameters to prune at each pruning iteration.
To allow for comparison between pruning algorithms, we select a fixed percentage of parameters to remove during each iteration, such that in exactly 3 iterations the desired amount of parameters will be pruned. Hence those hyperparameters are inferred and fixed in each setting, leaving no hyperparameters to be optimized when pruning iteratively.

\input{Tables/Setup/Hyperparameters_tuning/stats}

The hyperparameter tuning is performed separately on architectures and separately for each data set. We tune the hyperparameters using the random optimization from the weights and biases (WandB) platform with a budget of 100 objective function evaluations \citep{wandb}.
Hyperparameter tuning is set to maximize accuracy and macro F1 in the validation set for the single-label and multi-label tasks respectively.
The search spaces optimal hyperparameter values are summarized in Table \ref{tab:hyperparam_tuning}.

\section{Results}\label{app:additional_results}
In Table \ref{tab:pruningAlgoTable} we report accuracy and F1 score with their standard deviation, obtained by unpruned models and pruned models at different amounts of pruned parameters.

\input{Tables/Effectiveness/effectiveness_merged}

In Table \ref{tab:exp3_mc_app} we report accuracy and F1 score on PIEs obtained by unpruned models and pruned models at different amounts of pruned parameters. We highlight in blue the cases where the pruned models are on average more effective than the unpruned models on PIEs.
    \input{Tables/PIEs/merged_original}

\subsection{Pruning and occurrence of PIEs}\label{app:pruning_and_occurence_of_PIEs}

We report here the additional results of Section \ref{ss:pruning-occurrence}.

\begin{figure*}
\centering
\resizebox{2\columnwidth}{!}{%
\begin{tabular}{cccc}
 \includegraphics[width=0.25\textwidth]{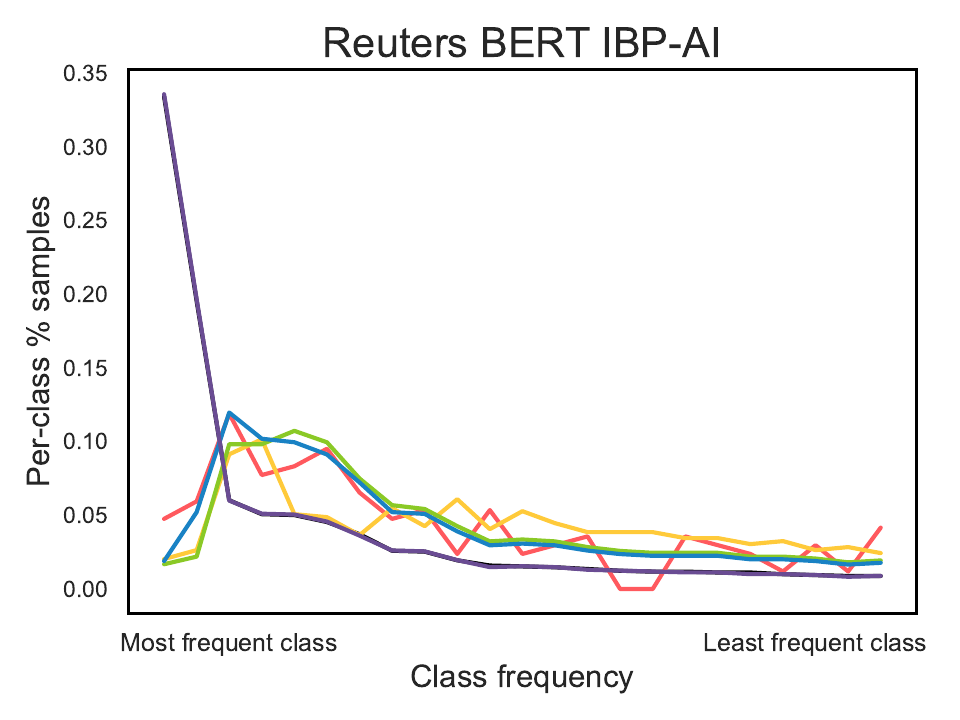} &
 \includegraphics[width=0.25\textwidth]{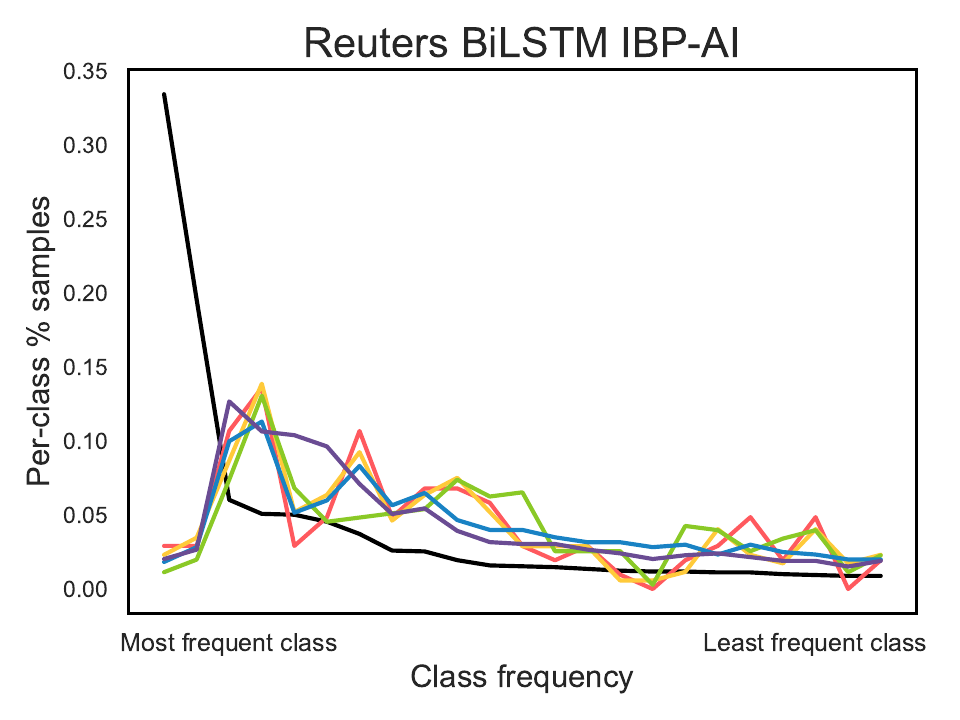} &
 \includegraphics[width=0.25\textwidth]{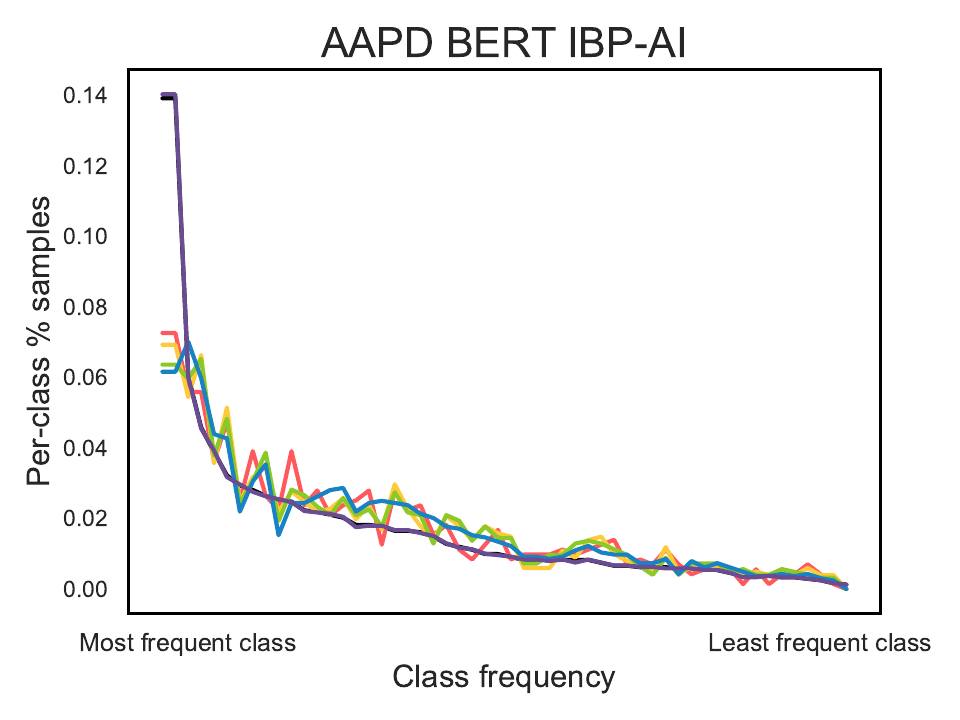} &
 \includegraphics[width=0.25\textwidth]{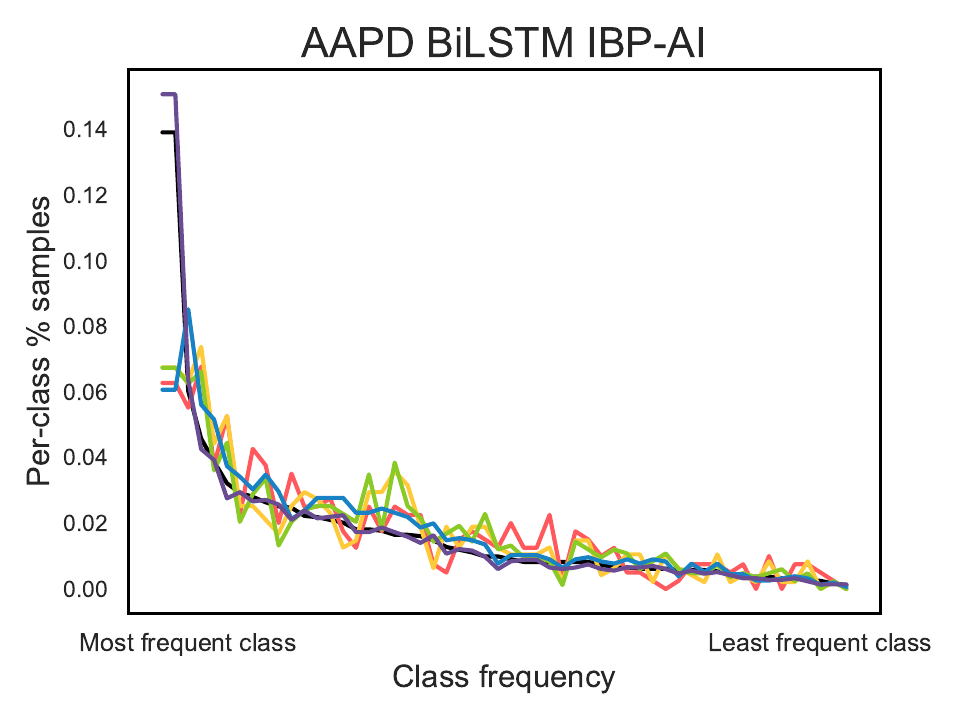} \\  
 
 \includegraphics[width=0.25\textwidth]{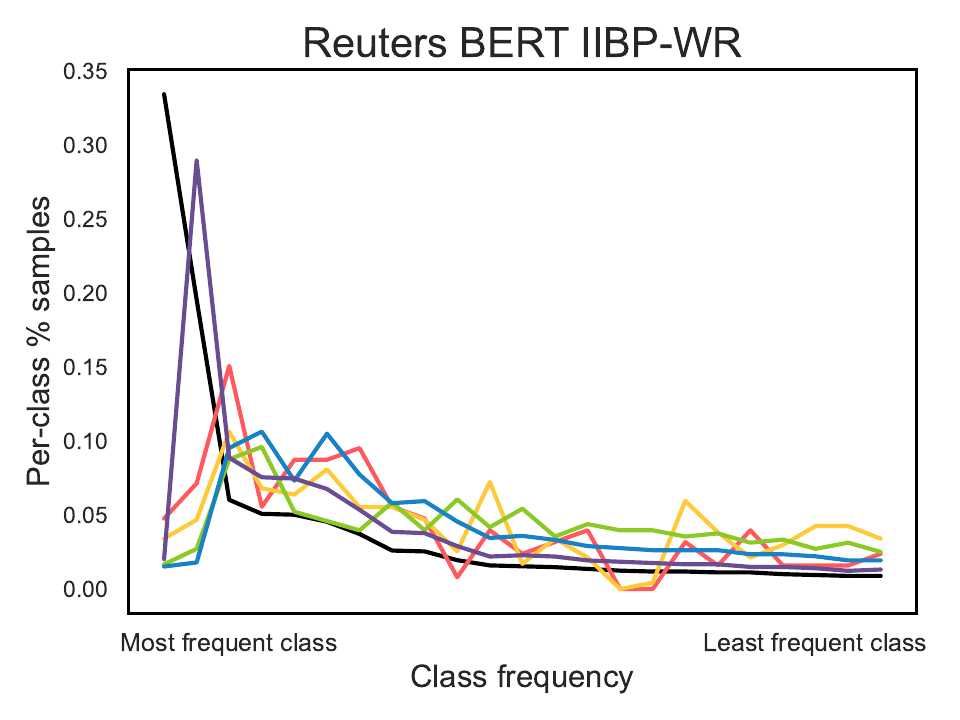} &
 \includegraphics[width=0.25\textwidth]{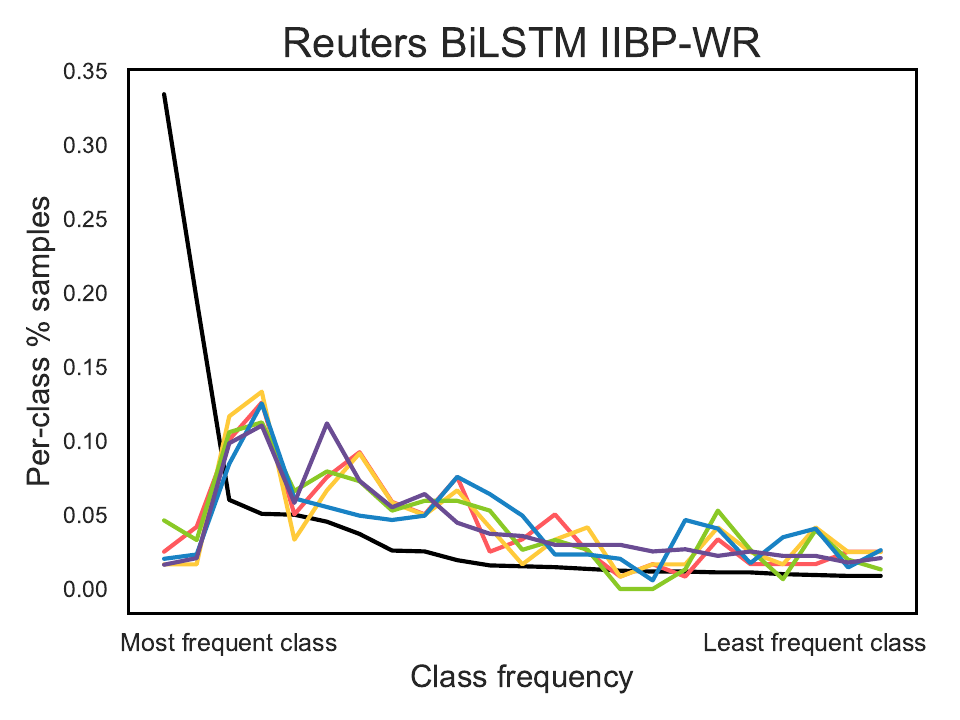} &
 \includegraphics[width=0.25\textwidth]{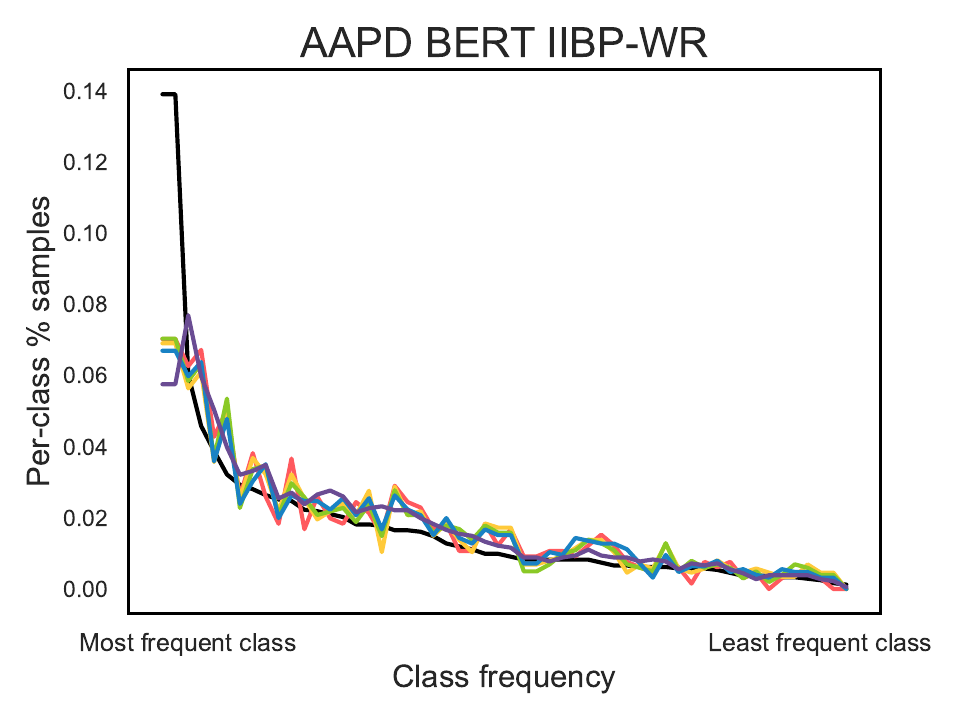} &
 \includegraphics[width=0.25\textwidth]{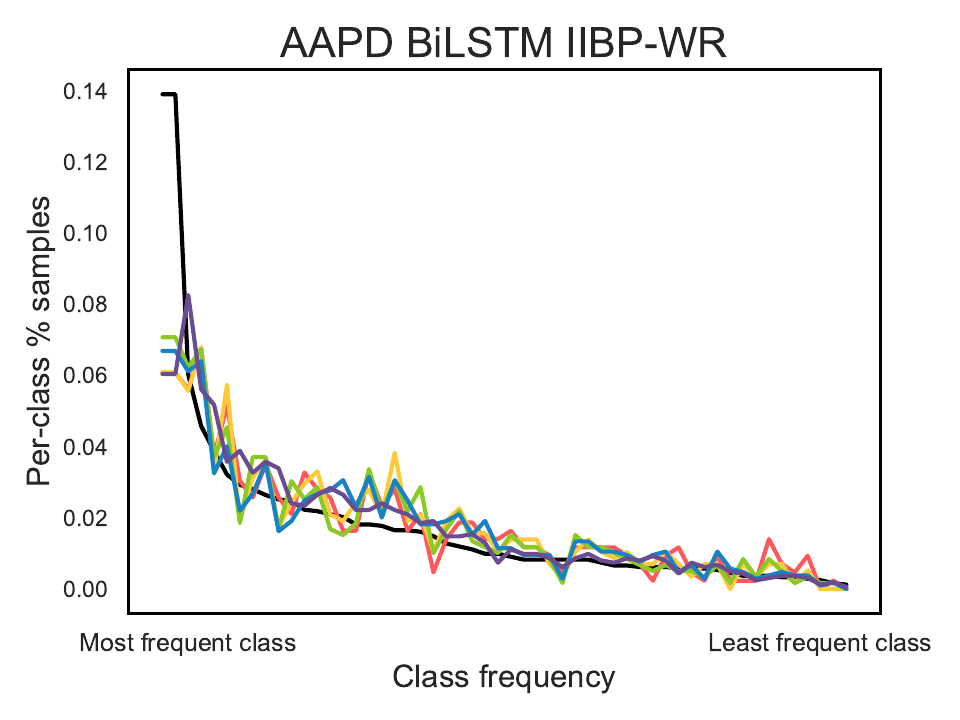} \\  
 
 \includegraphics[width=0.25\textwidth]{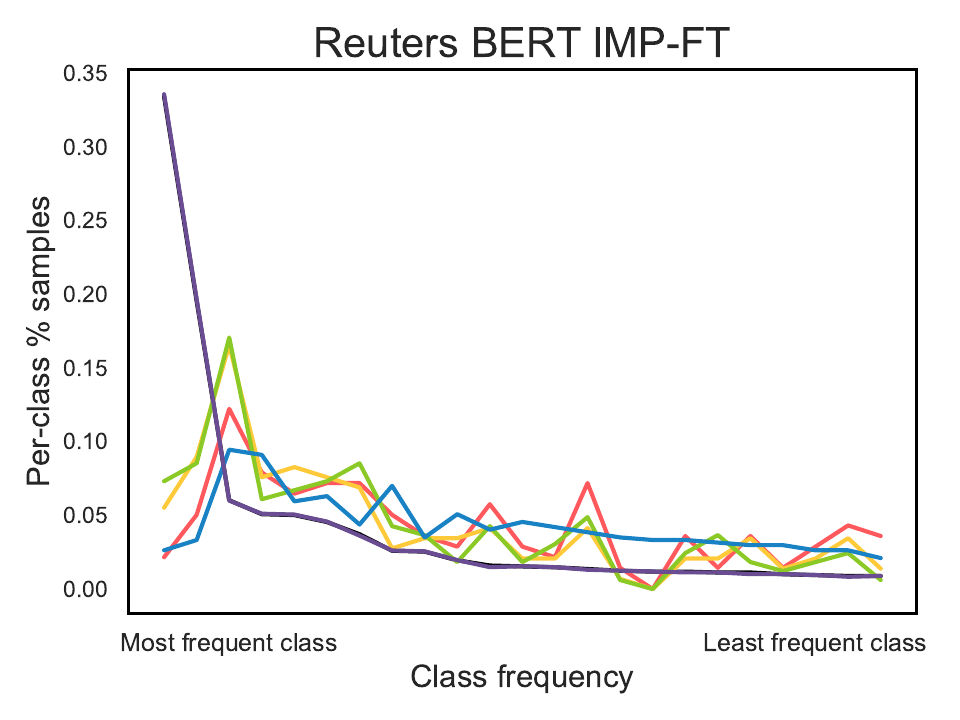} &
 \includegraphics[width=0.25\textwidth]{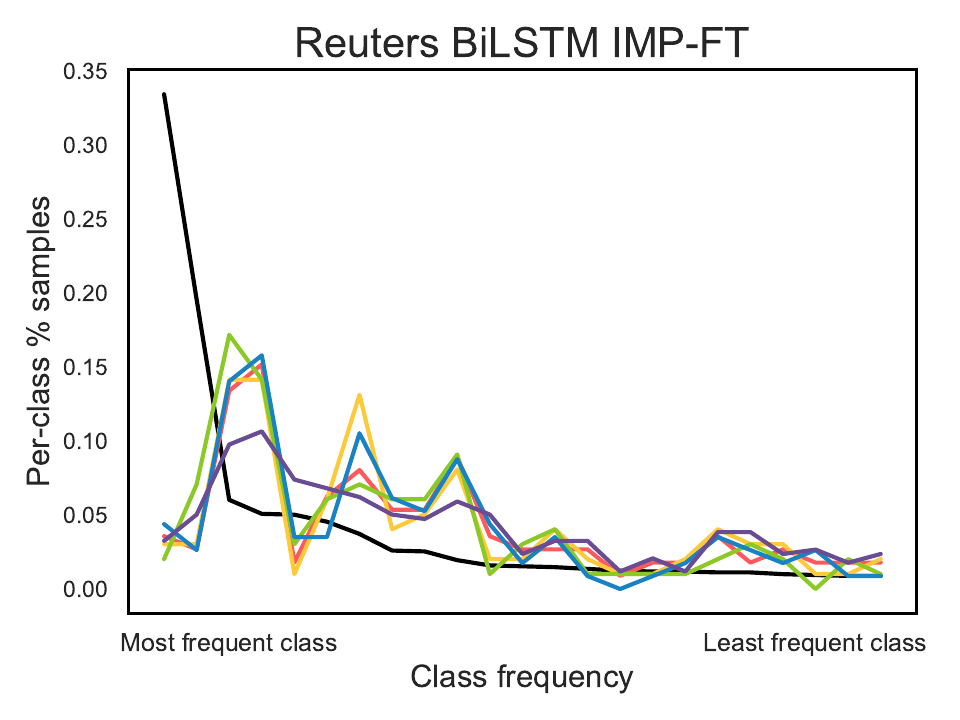} &
 \includegraphics[width=0.25\textwidth]{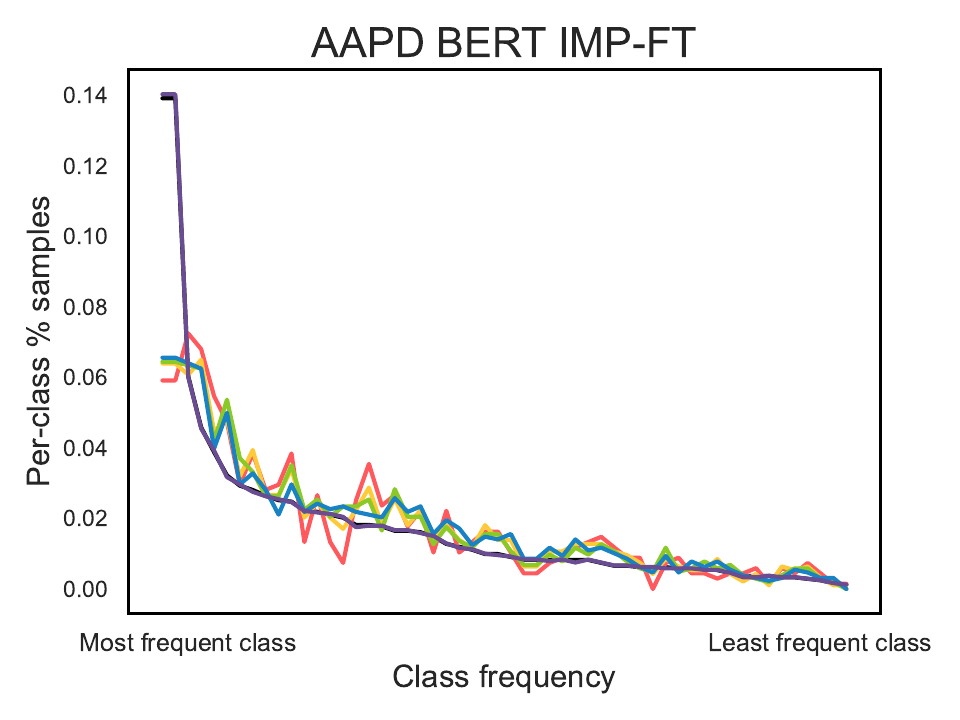} &
 \includegraphics[width=0.25\textwidth]{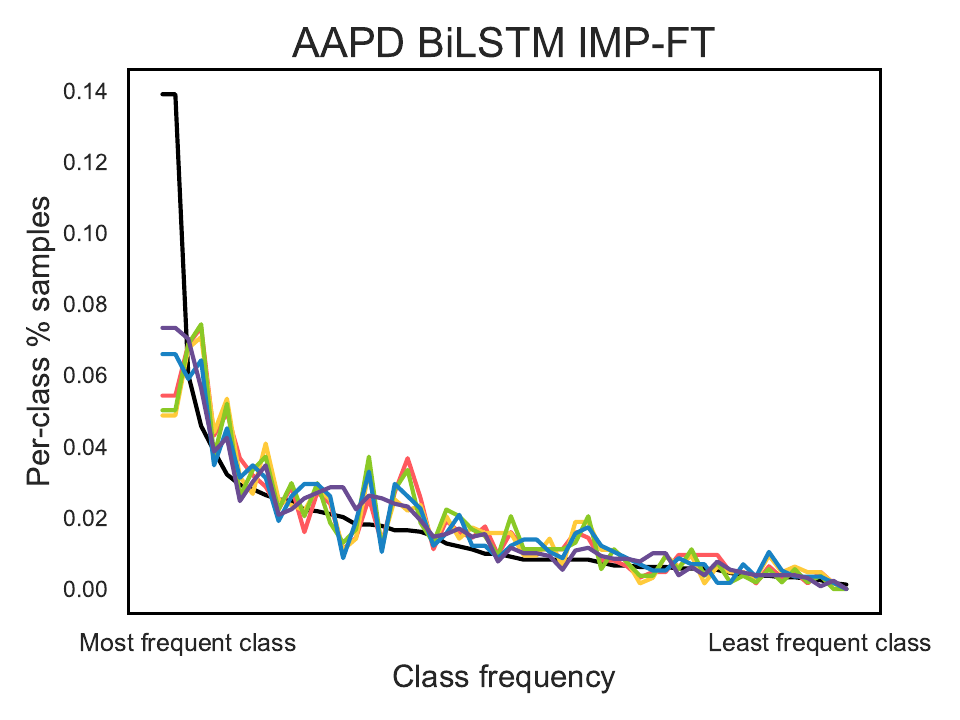} \\  

 \includegraphics[width=0.25\textwidth]{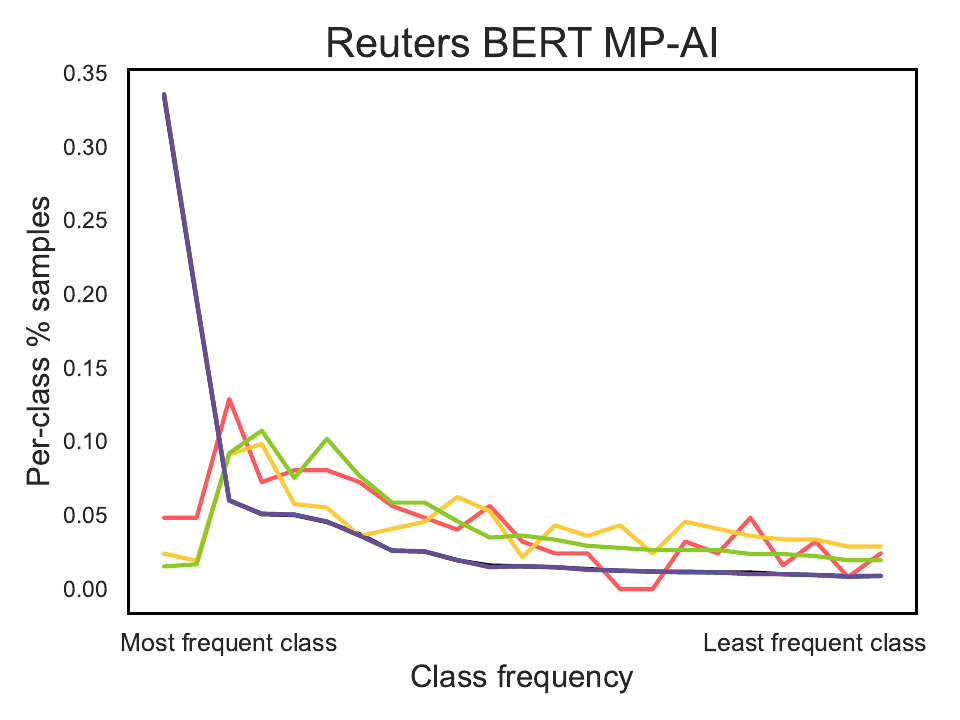} &
 \includegraphics[width=0.25\textwidth]{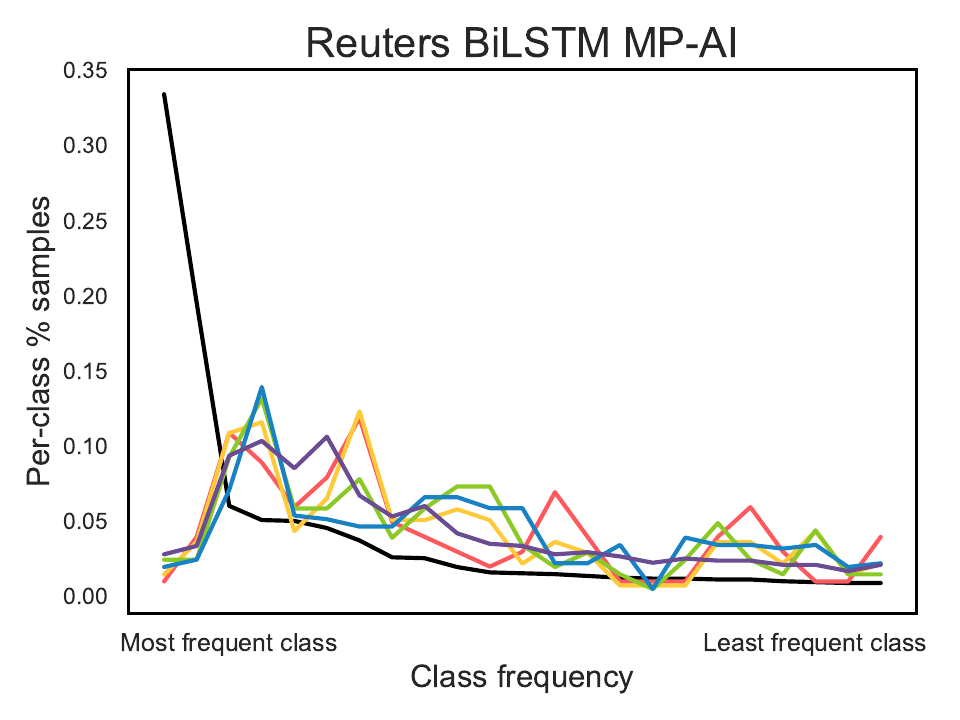} &
 \includegraphics[width=0.25\textwidth]{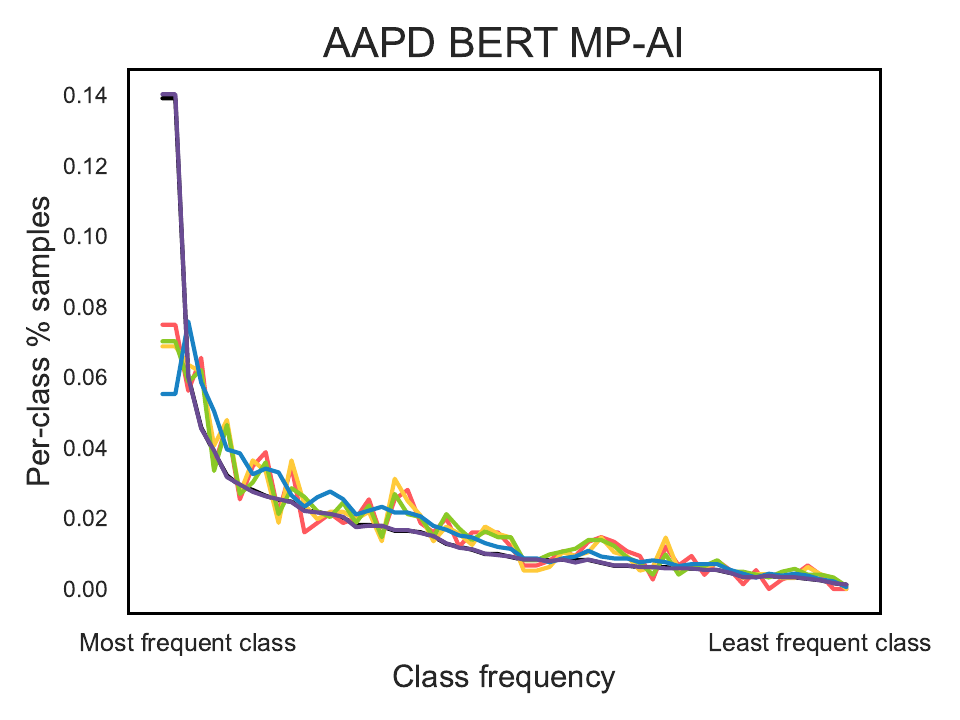} &
 \includegraphics[width=0.25\textwidth]{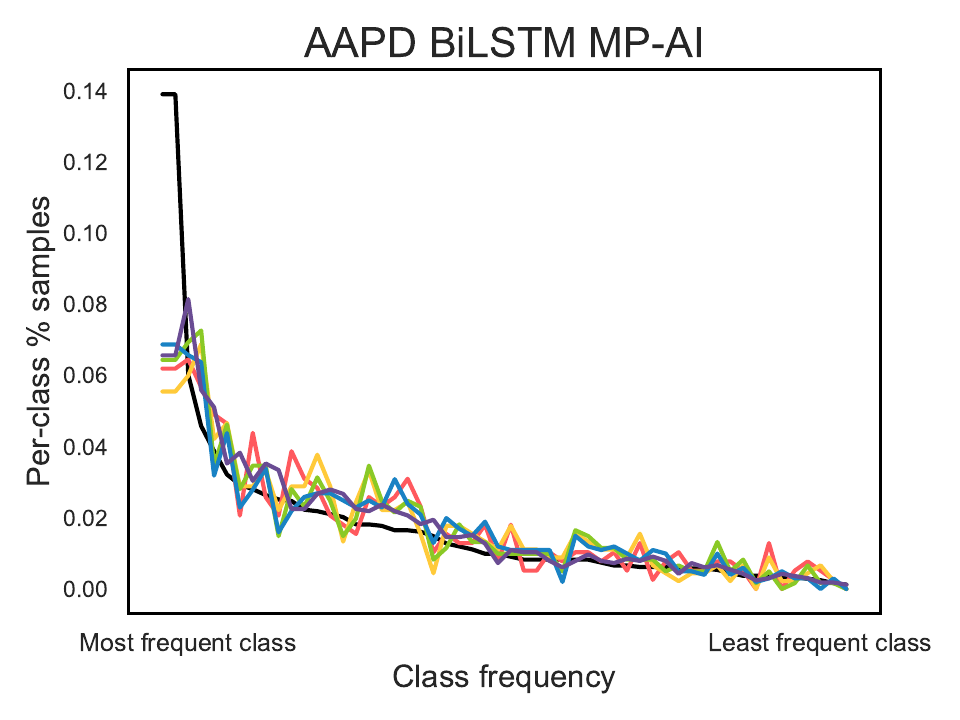} \\  
 
 \includegraphics[width=0.25\textwidth]{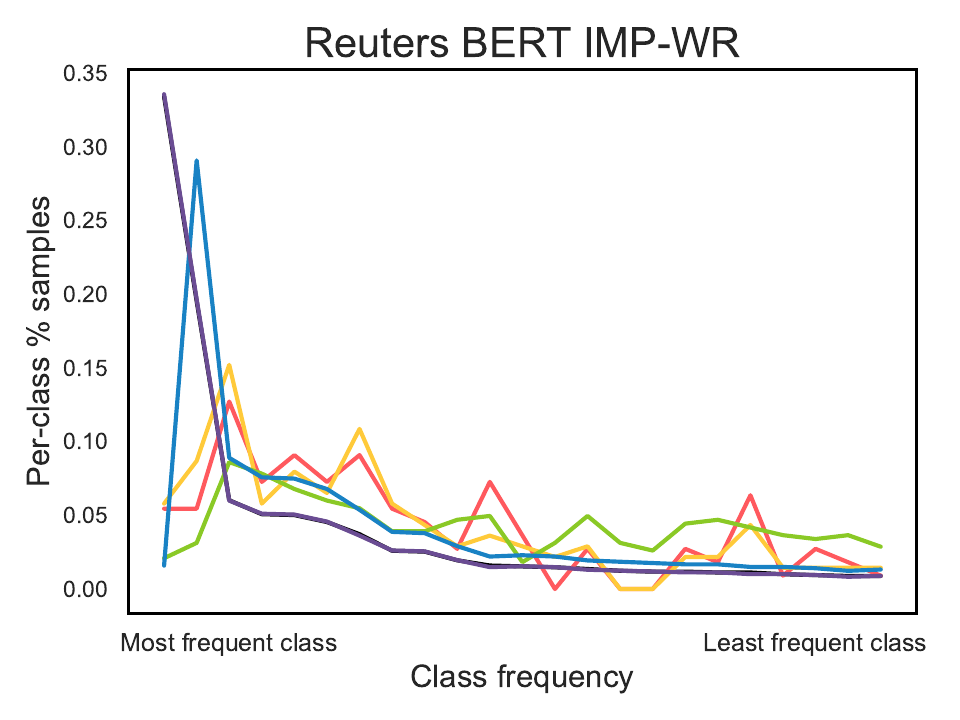} &
 \includegraphics[width=0.25\textwidth]{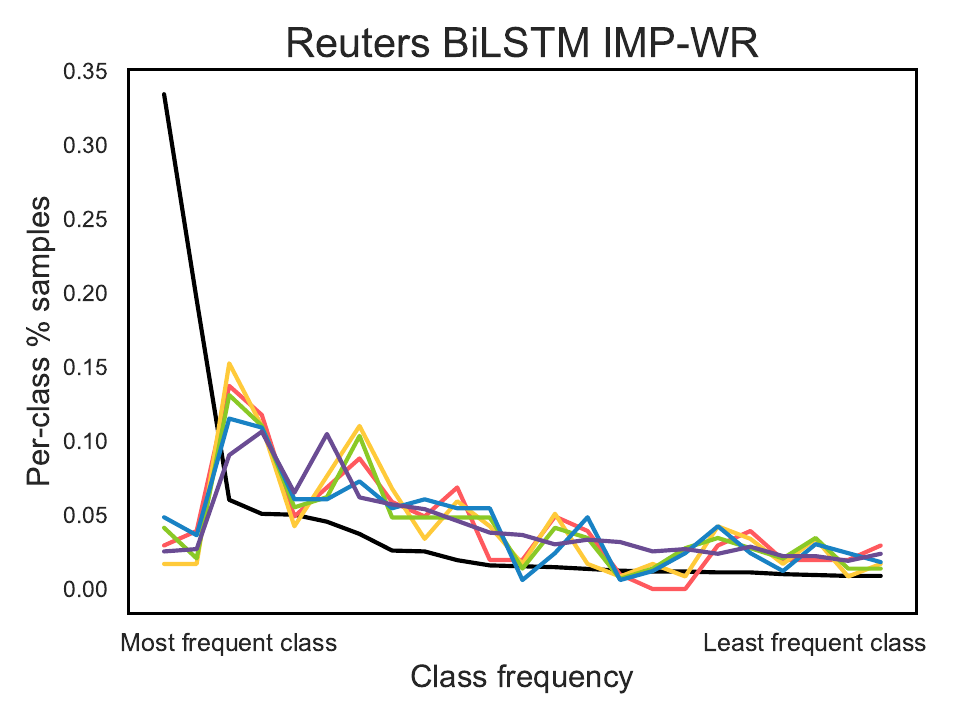} &
 \includegraphics[width=0.25\textwidth]{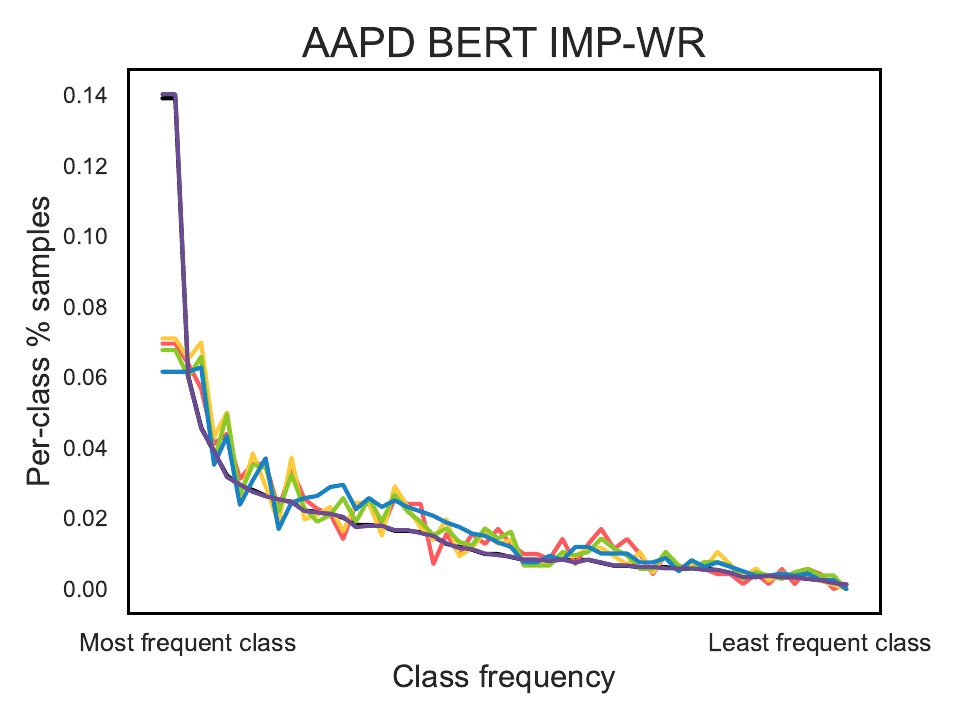} &
 \includegraphics[width=0.25\textwidth]{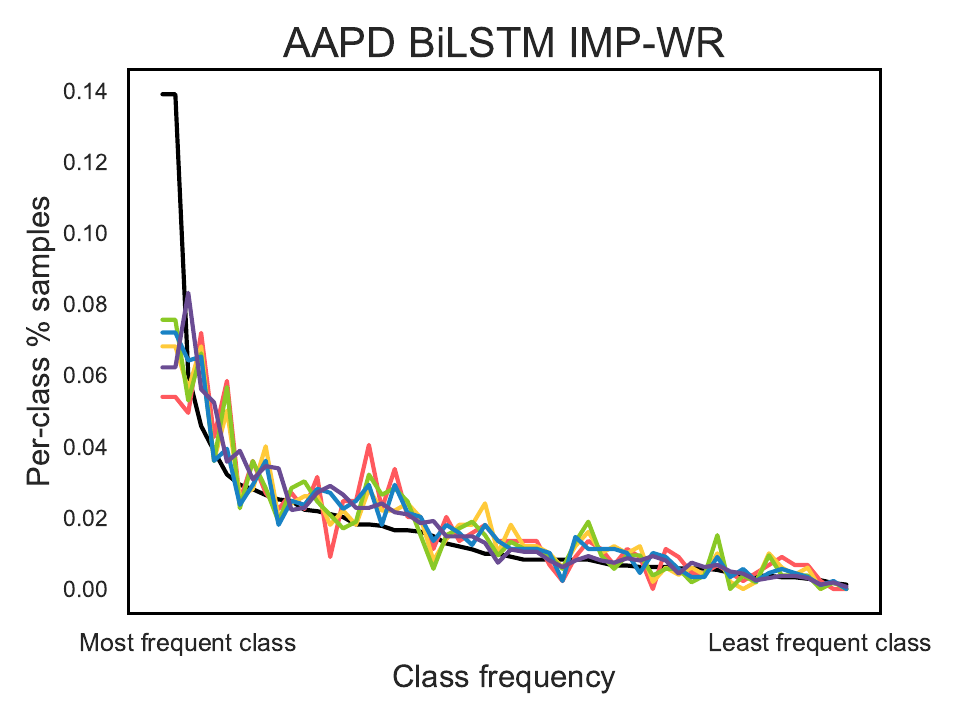} \\  
 
 \includegraphics[width=0.25\textwidth]{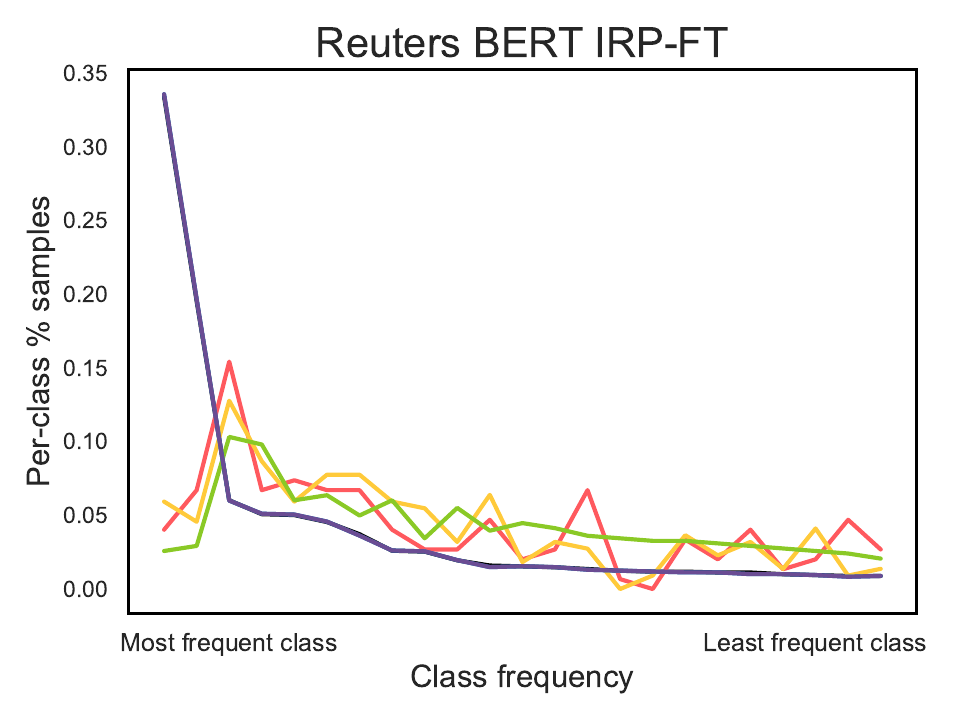} &
 \includegraphics[width=0.25\textwidth]{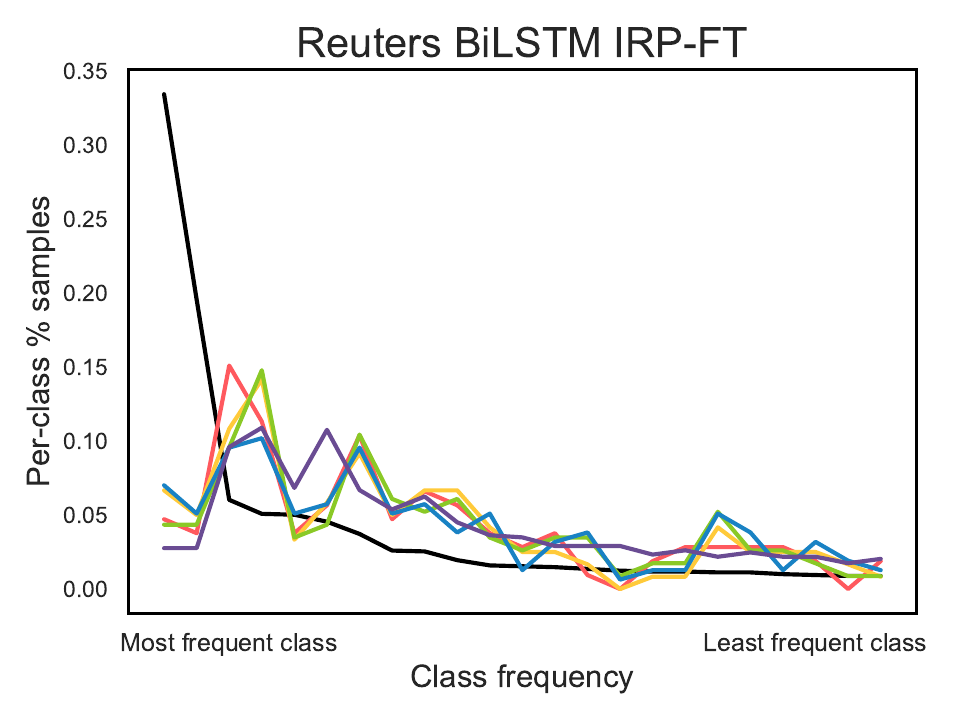} &
 \includegraphics[width=0.25\textwidth]{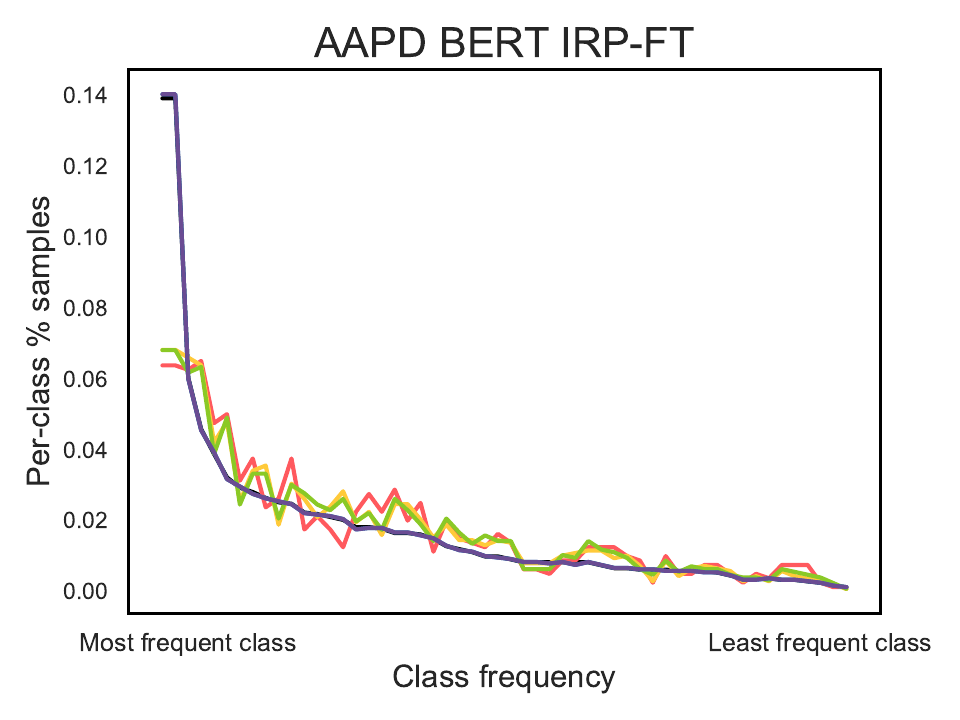} &
 \includegraphics[width=0.25\textwidth]{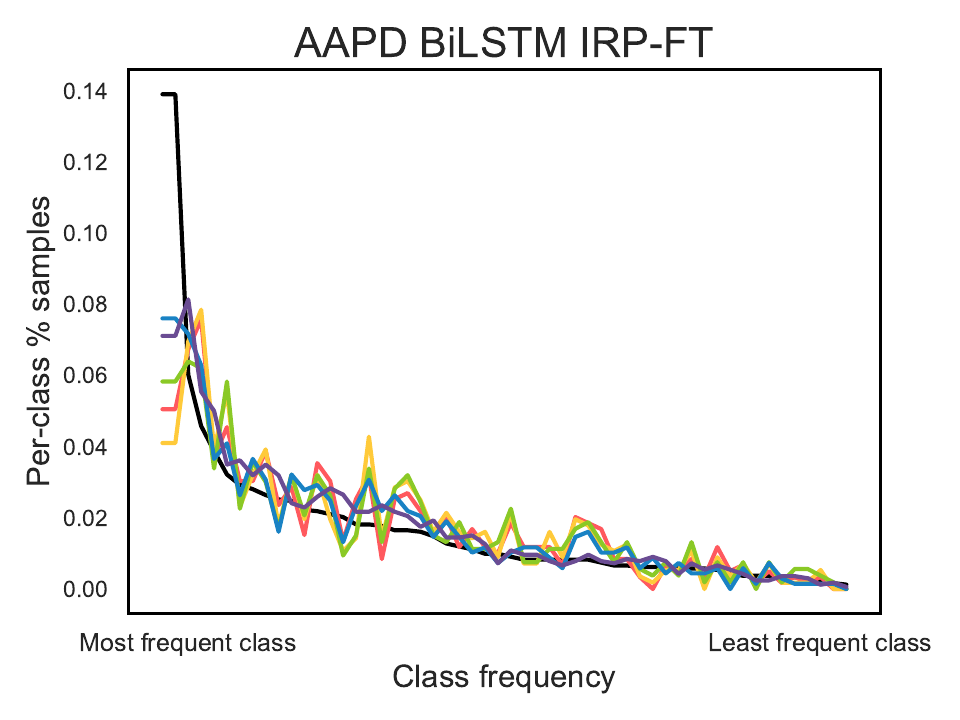} \\  
 
 \includegraphics[width=0.25\textwidth]{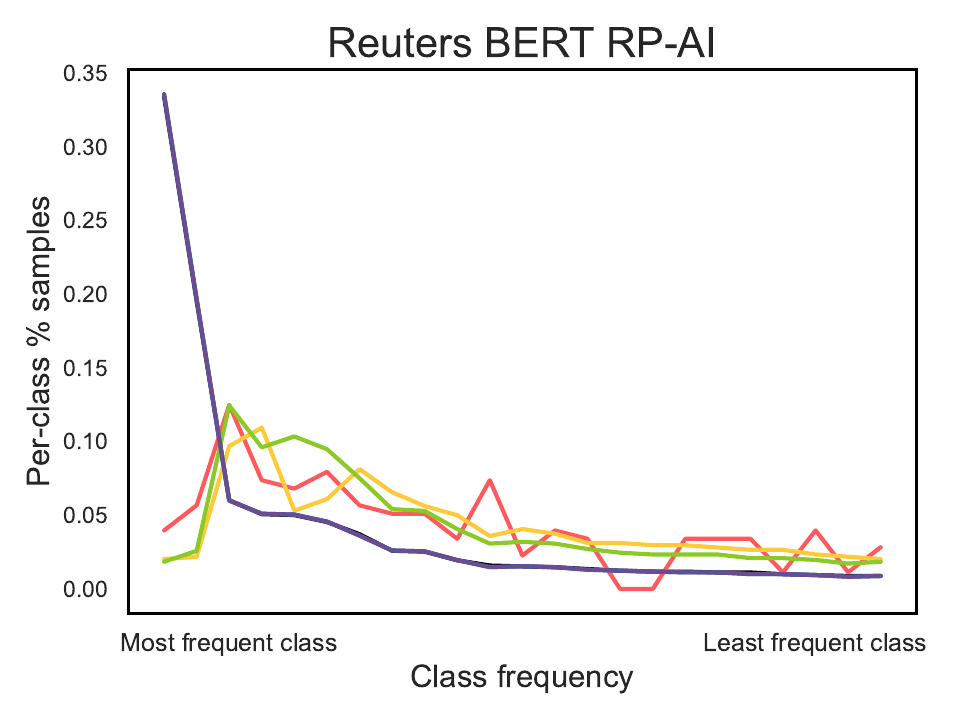} &
 \includegraphics[width=0.25\textwidth]{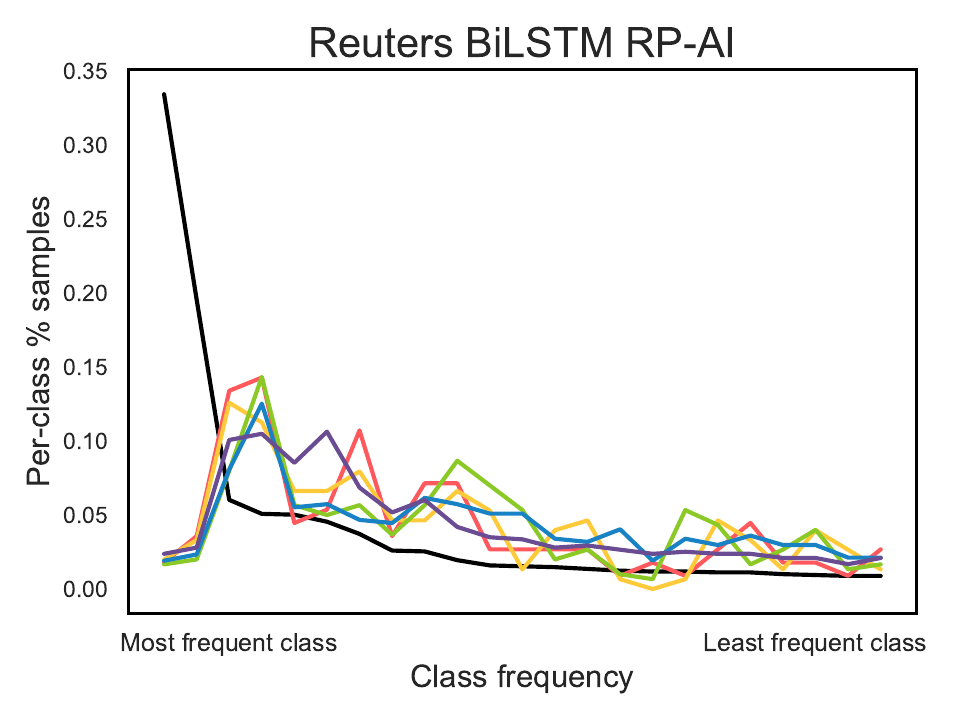} &
 \includegraphics[width=0.25\textwidth]{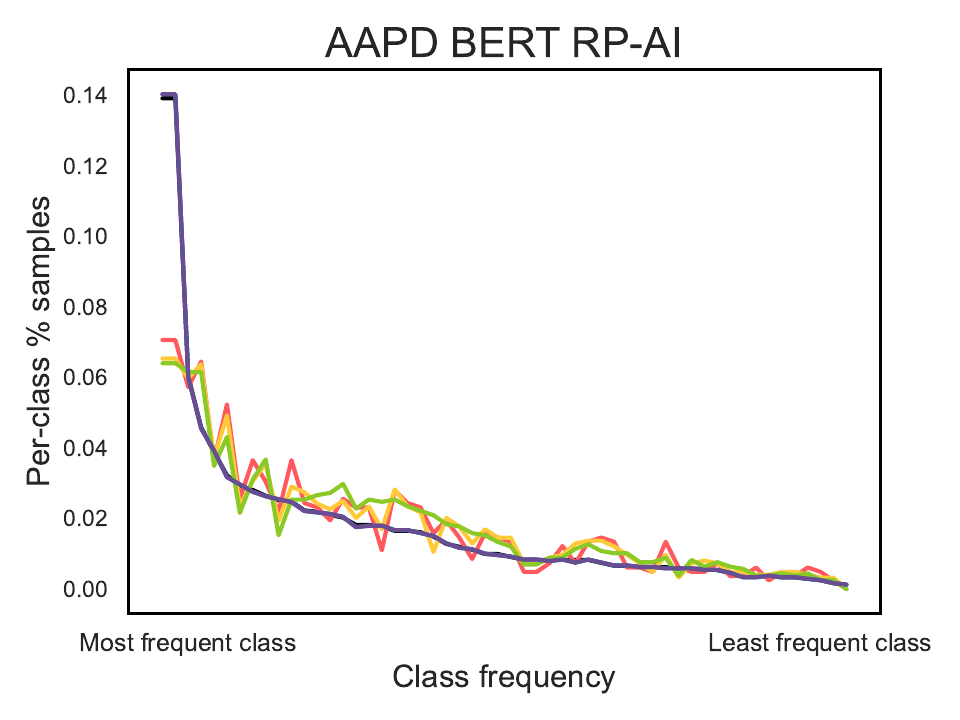} &
 \includegraphics[width=0.25\textwidth]{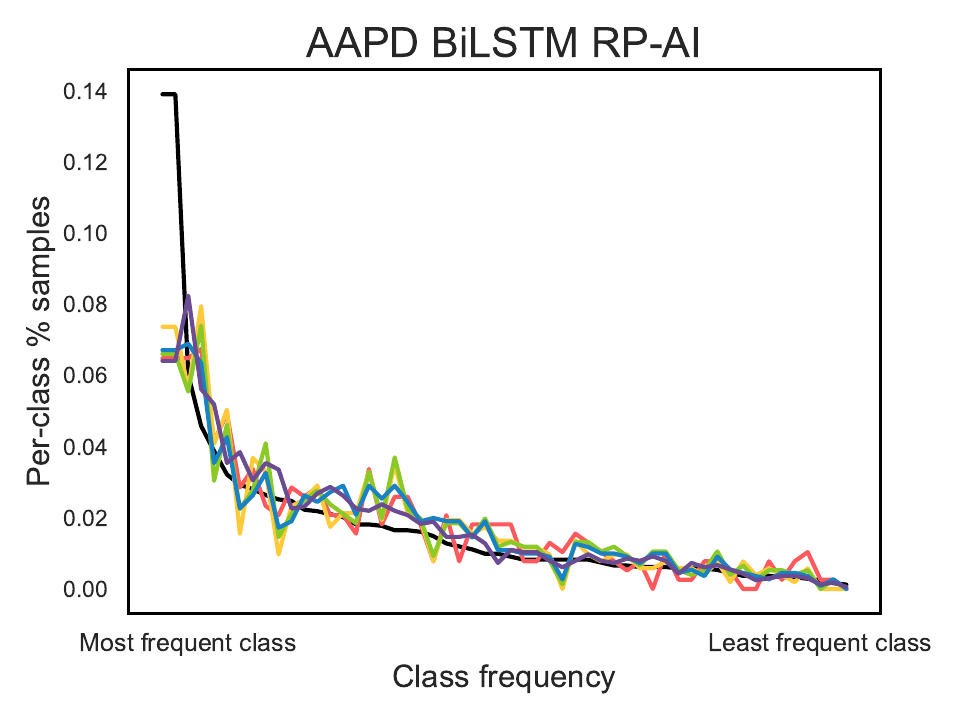} \\  
 \multicolumn{4}{c}{\includegraphics[width=0.4\textwidth]{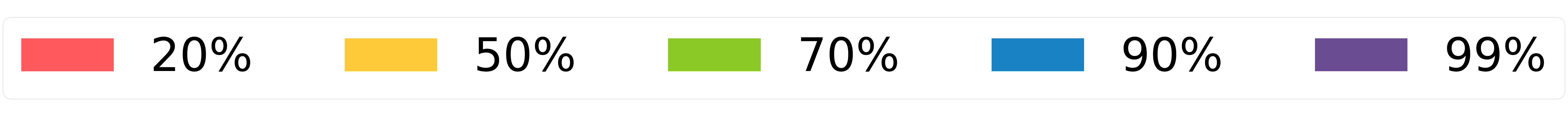}} 
\end{tabular}
}
\caption{Distribution of all data points and of PIEs at 20\% to 99\% pruning, across classes sorted by frequency (x axis), for the multi-label datasets (test set).}
\label{fig:pies_class_dist_pic_appendix}%
\end{figure*}

In Figure \ref{fig:pies_class_dist_pic_appendix} we show the distribution of all data points and of PIEs at 20\% to 99\% pruning, across classes sorted by frequency for the multi-label datasets. We observe the same overall trend in all settings. Regardless of the language model architecture, the percentage of PIEs in the most frequent class for Reuters is much lower than the percentage of examples belonging to the same class in all data points. This means that the disagreement between pruned and unpruned models is not focused on the most frequent class of Reuters. The disagreement is skewed instead towards the less frequent classes.
On AAPD we observe a similar behaviour, however, the percentage of PIEs belonging to the most frequent class is higher, hence the disagreement is slightly more balanced across all classes.

\begin{figure*}
\centering
\begin{tabular}{cccc}
 \includegraphics[width=0.223\textwidth]{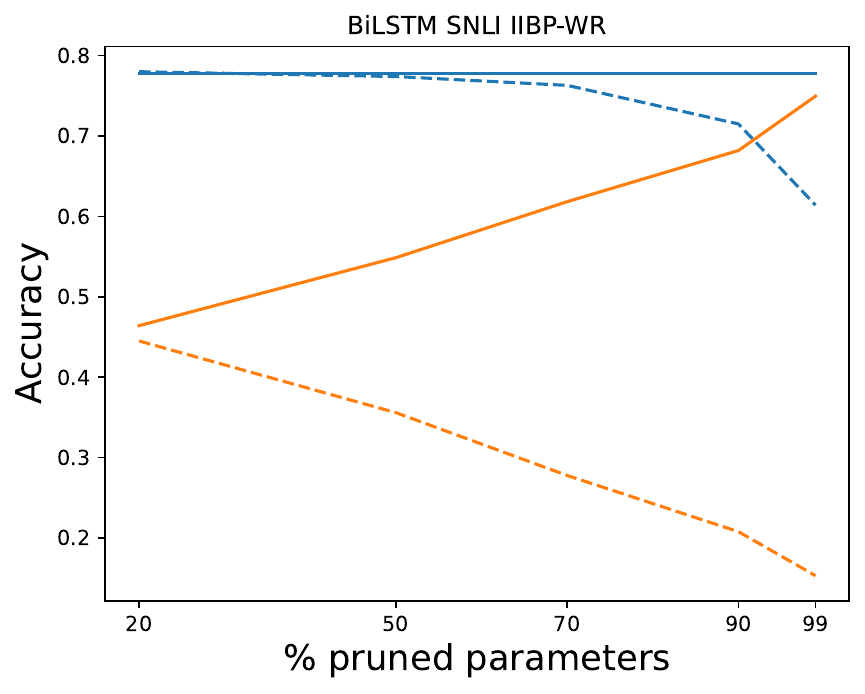} &
 \includegraphics[width=0.223\textwidth]{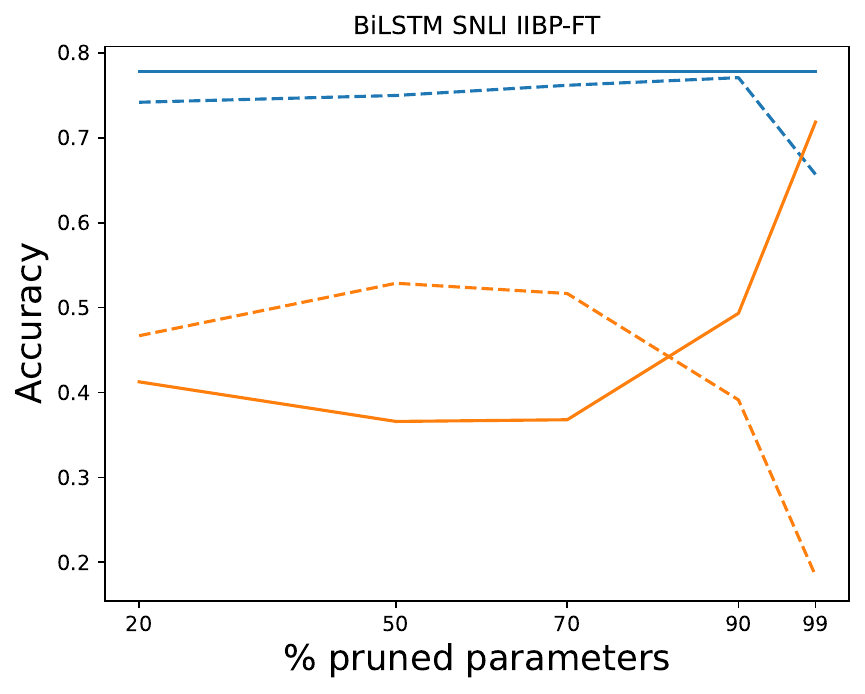} &
 \includegraphics[width=0.223\textwidth]{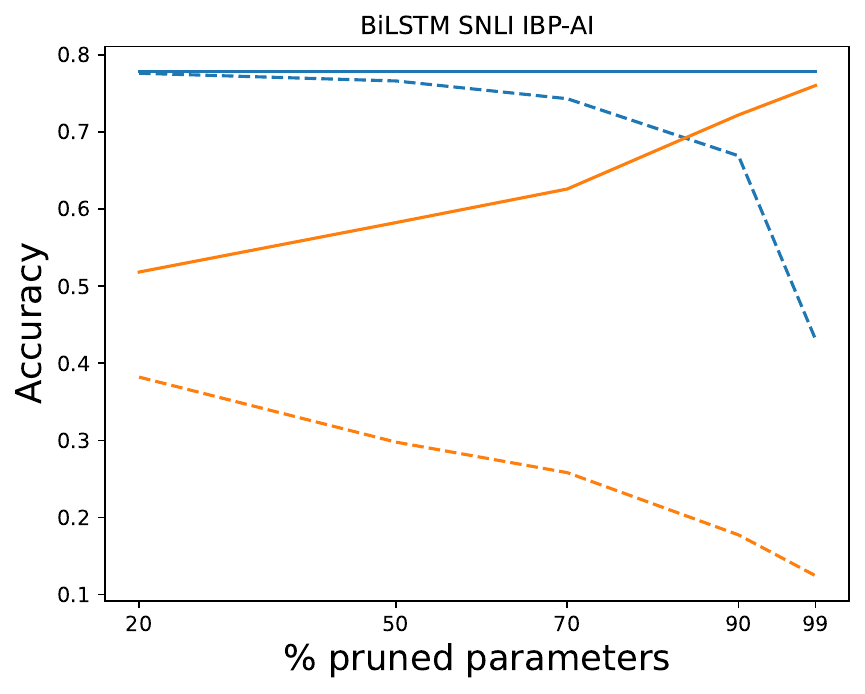} &
 \includegraphics[width=0.223\textwidth]{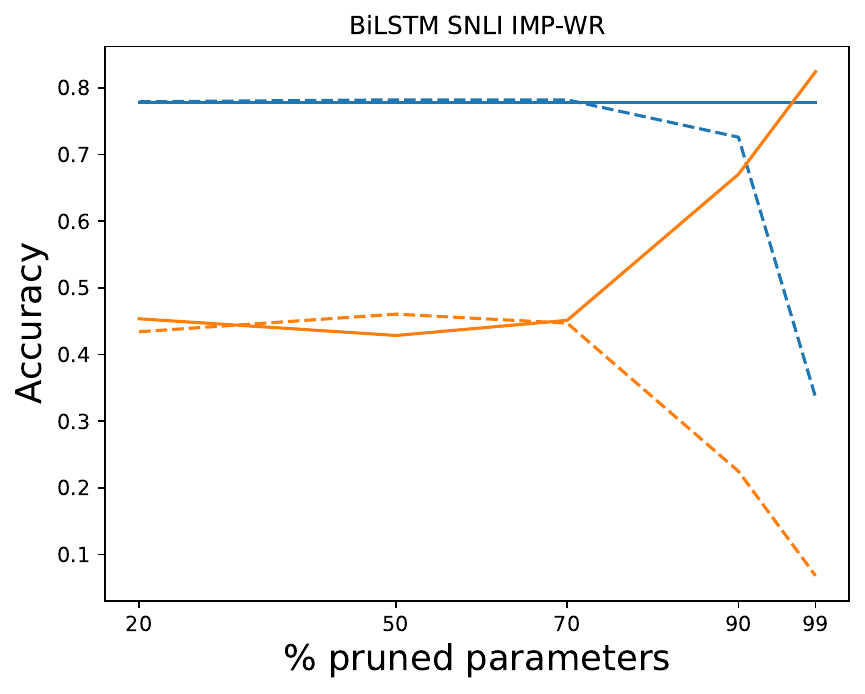} \\

 \includegraphics[width=0.223\textwidth]{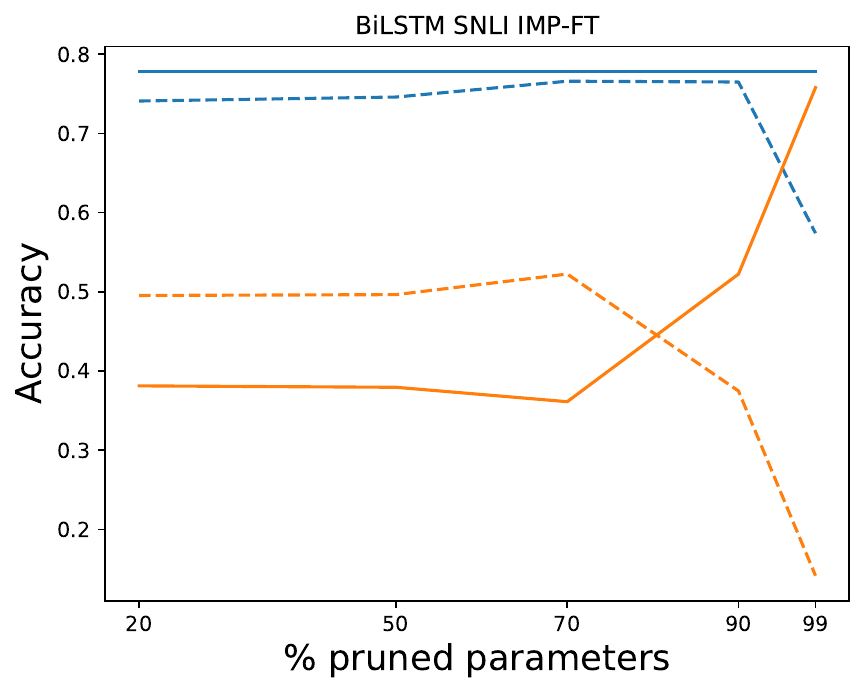} &
 \includegraphics[width=0.223\textwidth]{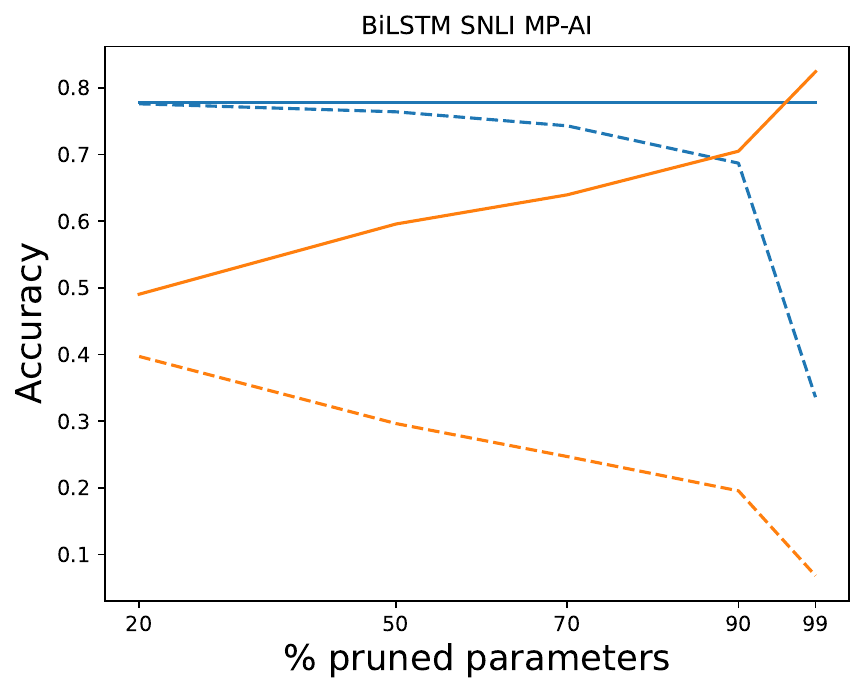} &
 \includegraphics[width=0.223\textwidth]{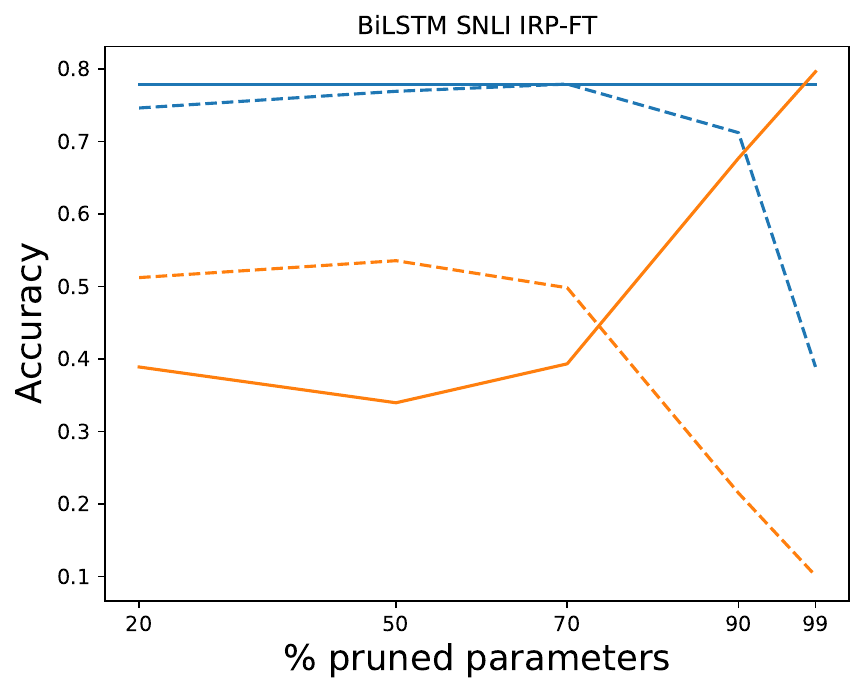} &
 \includegraphics[width=0.223\textwidth]{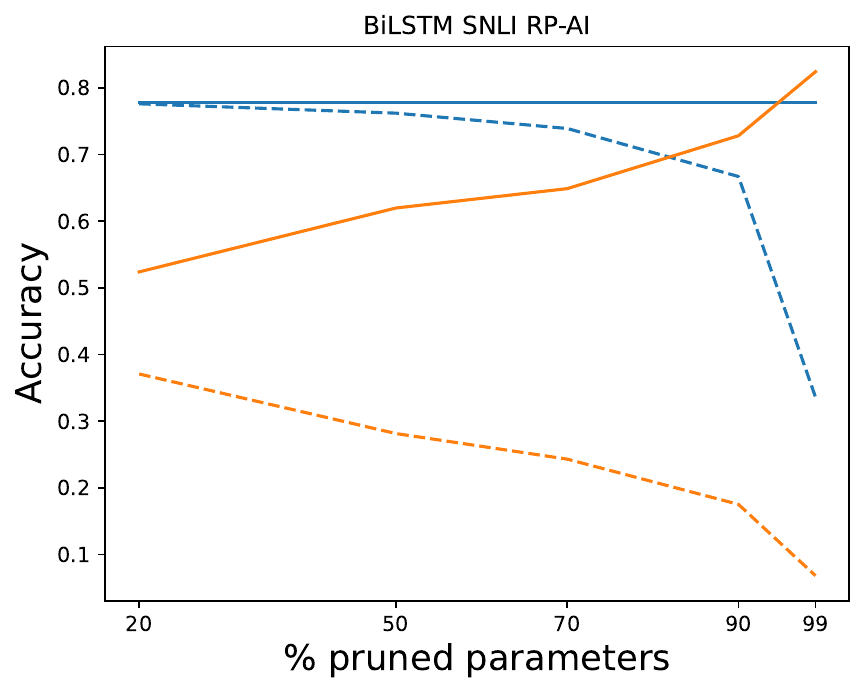}  \\

 \multicolumn{4}{c}{\includegraphics[width=0.7\textwidth]{Images/Effectiveness_all_against_PIEs/legend.pdf}}         
\end{tabular}
\caption{Accuracy of unpruned (black line) and pruned models on PIEs and all samples in the dataset per pruning method, across pruning thresholds (x-axis), over 30 initializations.
}
\label{fig:PIEs_against_all_BiLSTM_SNLI}%
\end{figure*}

\begin{figure*}
\centering
\begin{tabular}{cccc}
 \includegraphics[width=0.223\textwidth]{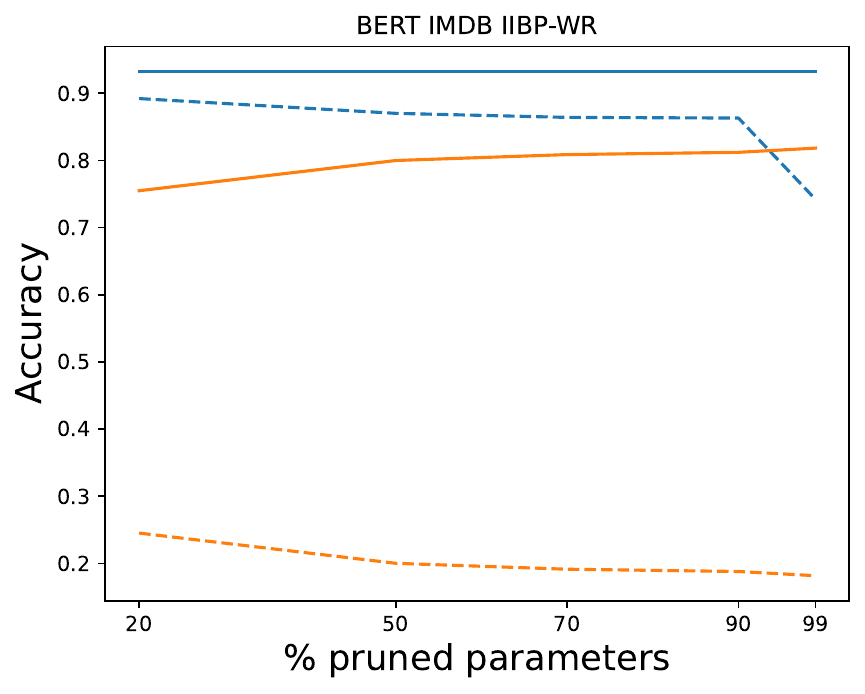} &
 \includegraphics[width=0.223\textwidth]{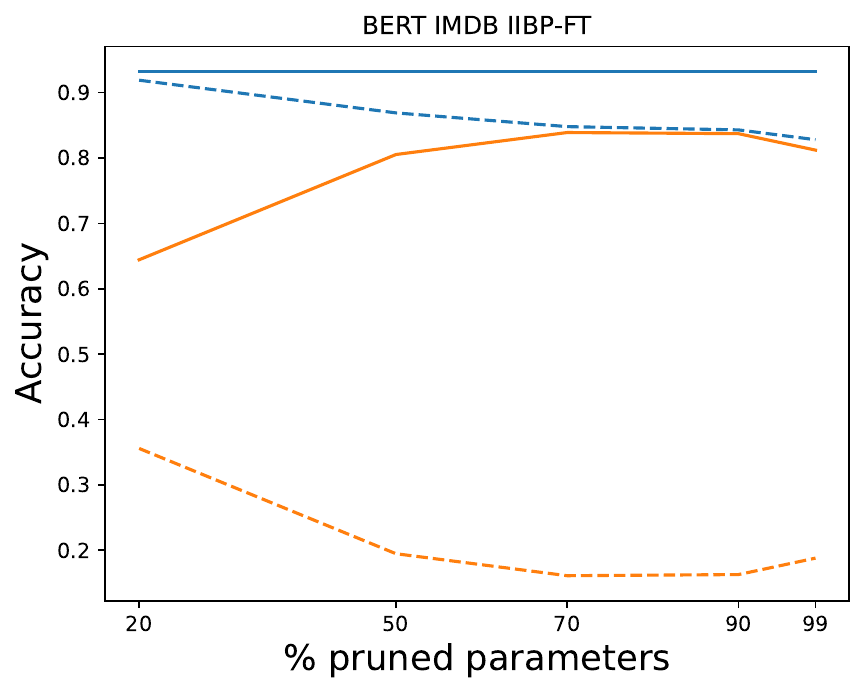} &
 \includegraphics[width=0.223\textwidth]{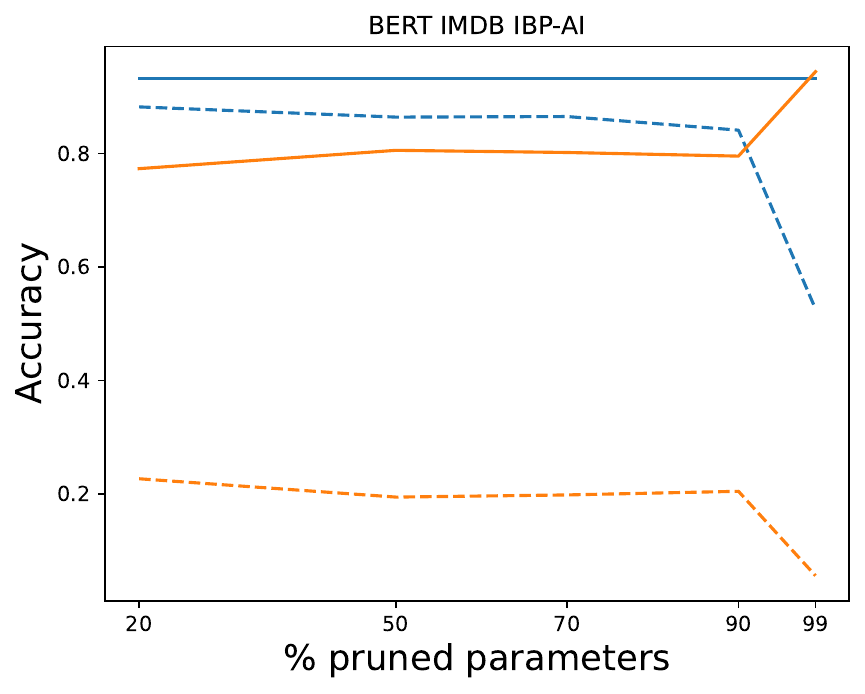} &
 \includegraphics[width=0.223\textwidth]{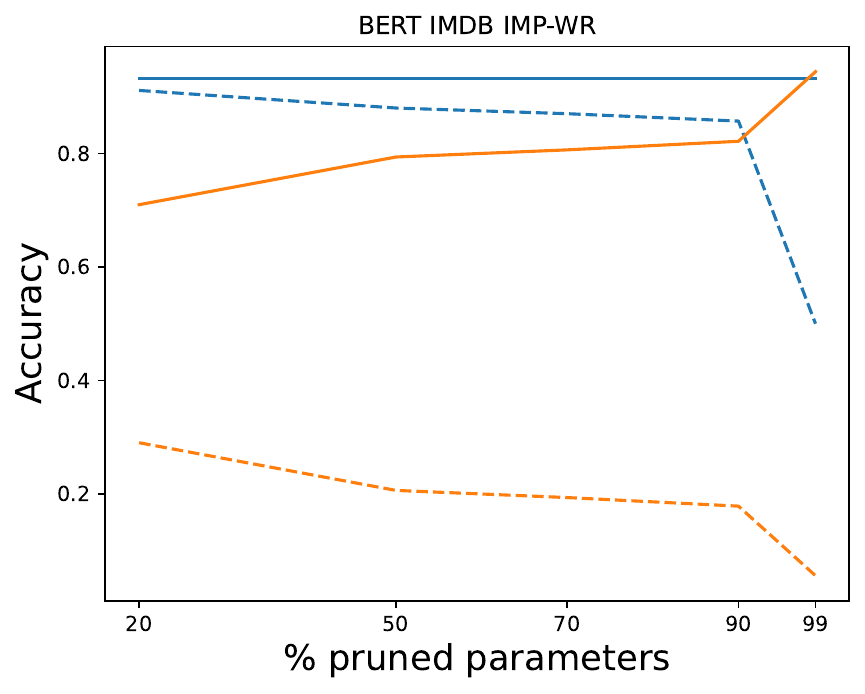} \\

 \includegraphics[width=0.223\textwidth]{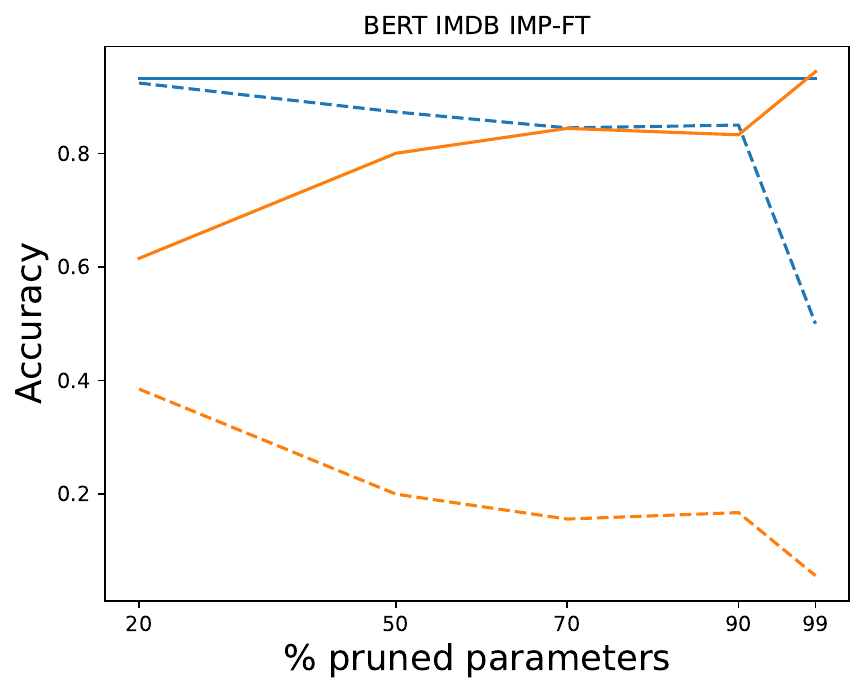} &
 \includegraphics[width=0.223\textwidth]{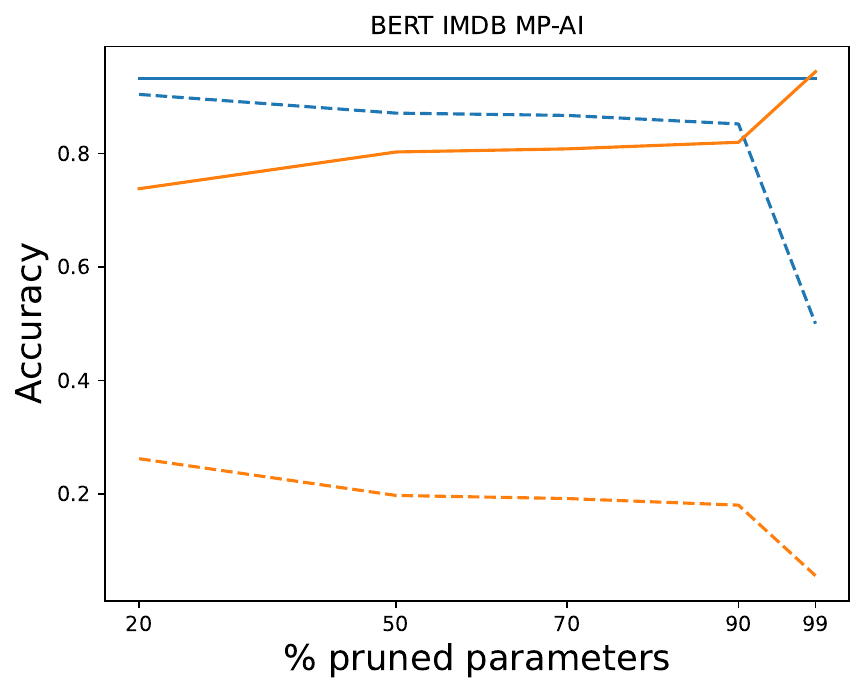} &
 \includegraphics[width=0.223\textwidth]{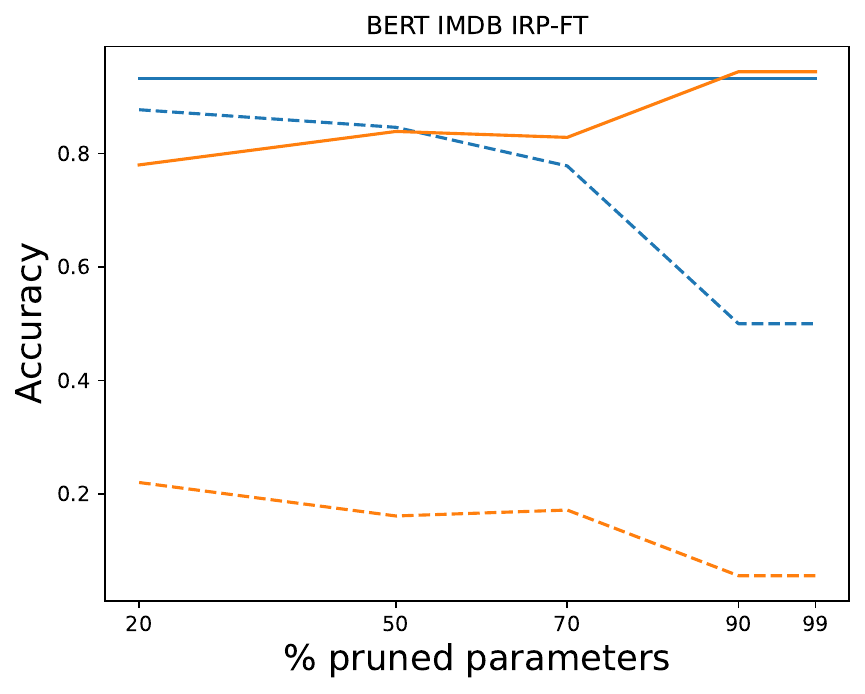} &
 \includegraphics[width=0.223\textwidth]{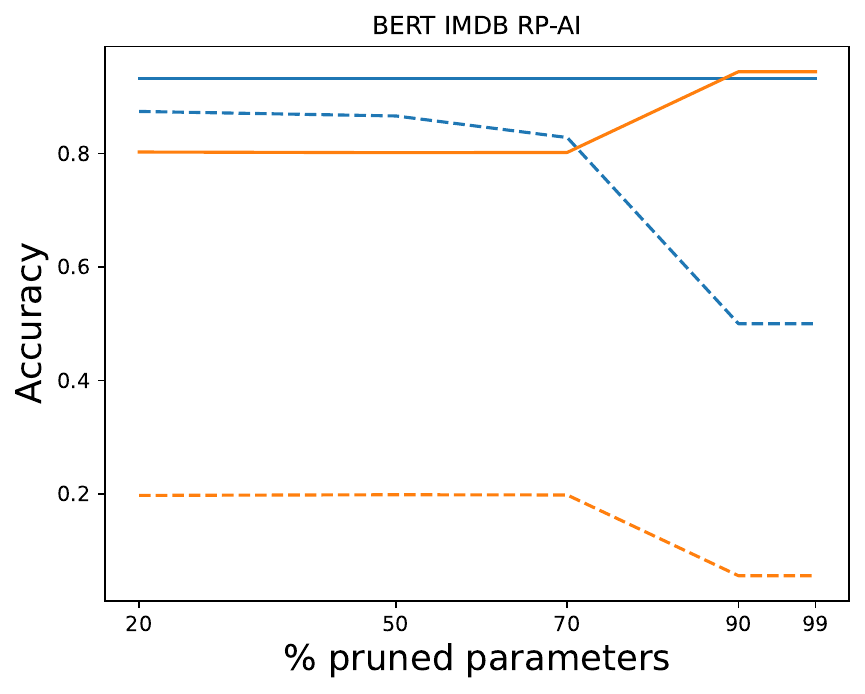}  \\

 \multicolumn{4}{c}{\includegraphics[width=0.7\textwidth]{Images/Effectiveness_all_against_PIEs/legend.pdf}}         
\end{tabular}
\caption{Accuracy of unpruned (black line) and pruned models on PIEs and all samples in the dataset per pruning method, across pruning thresholds (x-axis), over 30 initializations.
}
\label{fig:PIEs_against_all_BERT_IMDB}%
\end{figure*}

\begin{figure*}
\centering
\begin{tabular}{cccc}
 \includegraphics[width=0.223\textwidth]{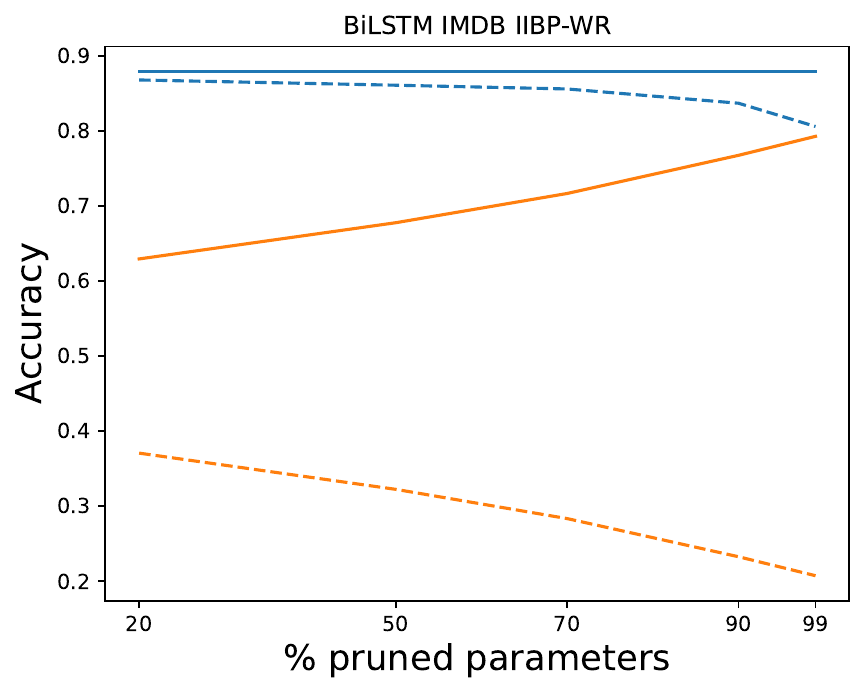} &
 \includegraphics[width=0.223\textwidth]{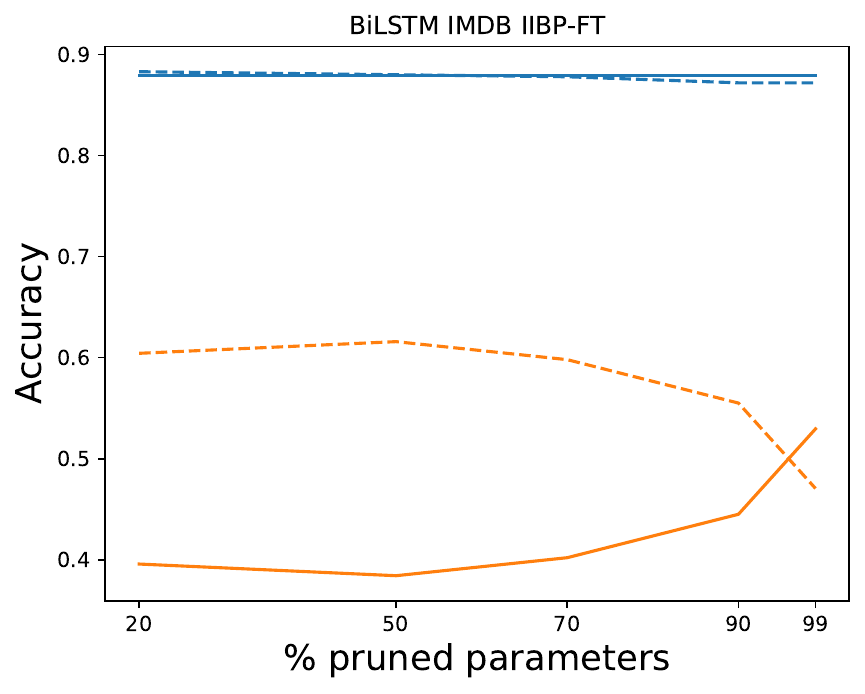} &
 \includegraphics[width=0.223\textwidth]{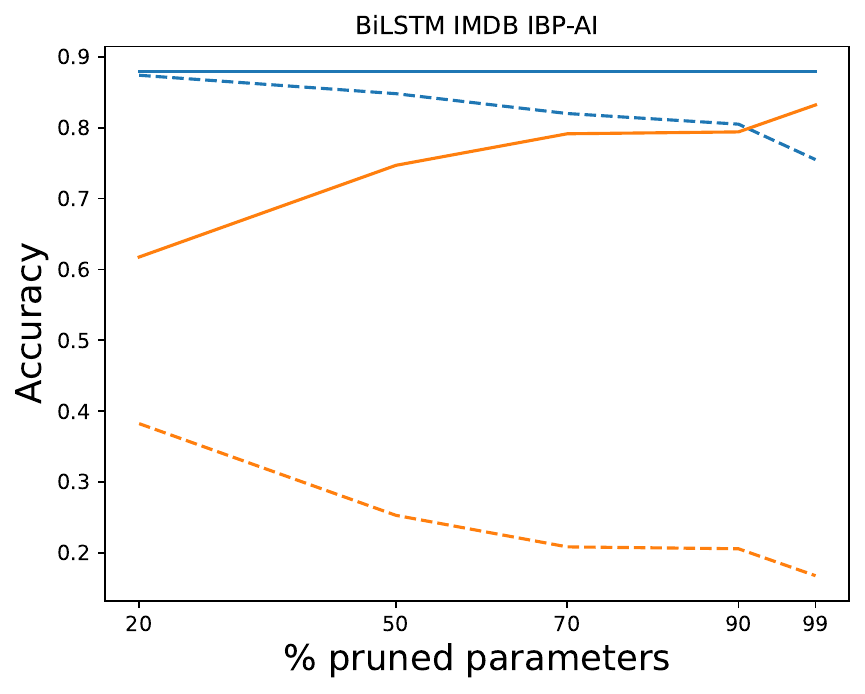} &
 \includegraphics[width=0.223\textwidth]{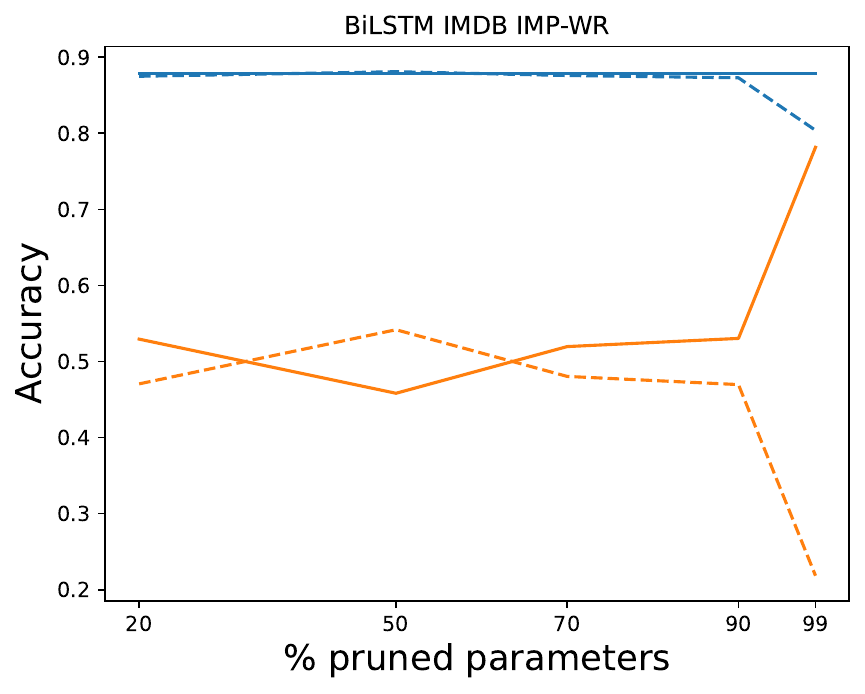} \\

 \includegraphics[width=0.223\textwidth]{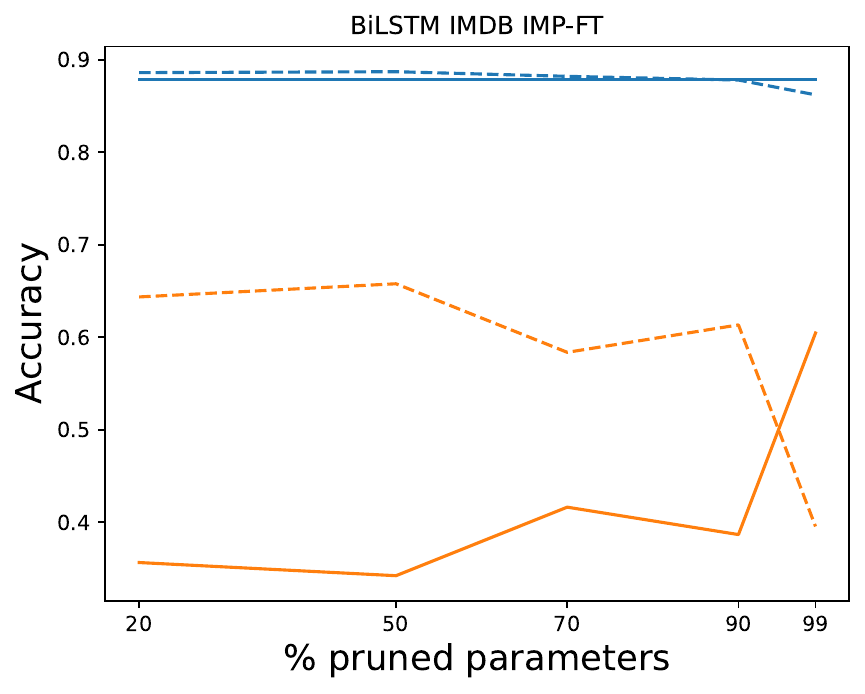} &
 \includegraphics[width=0.223\textwidth]{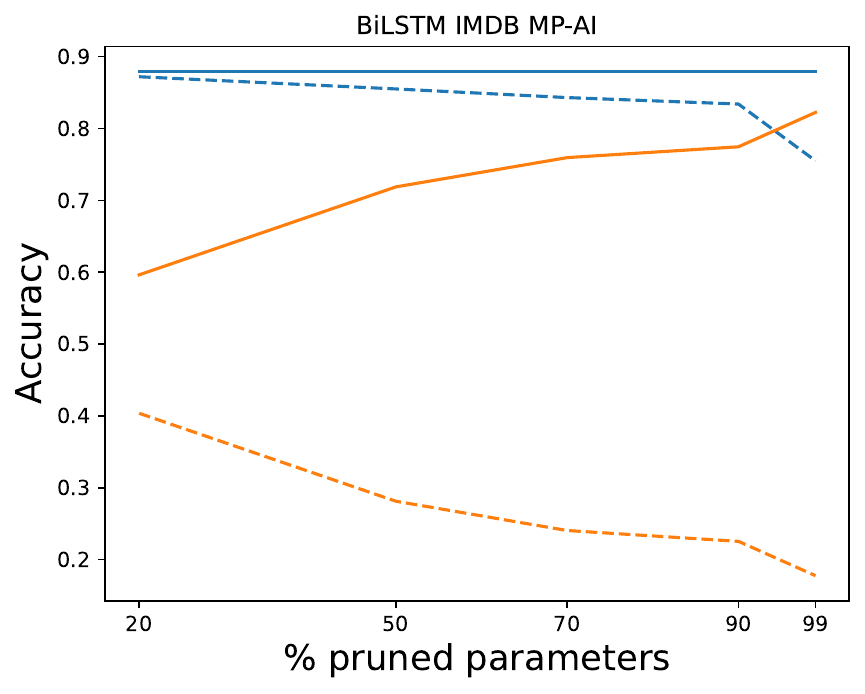} &
 \includegraphics[width=0.223\textwidth]{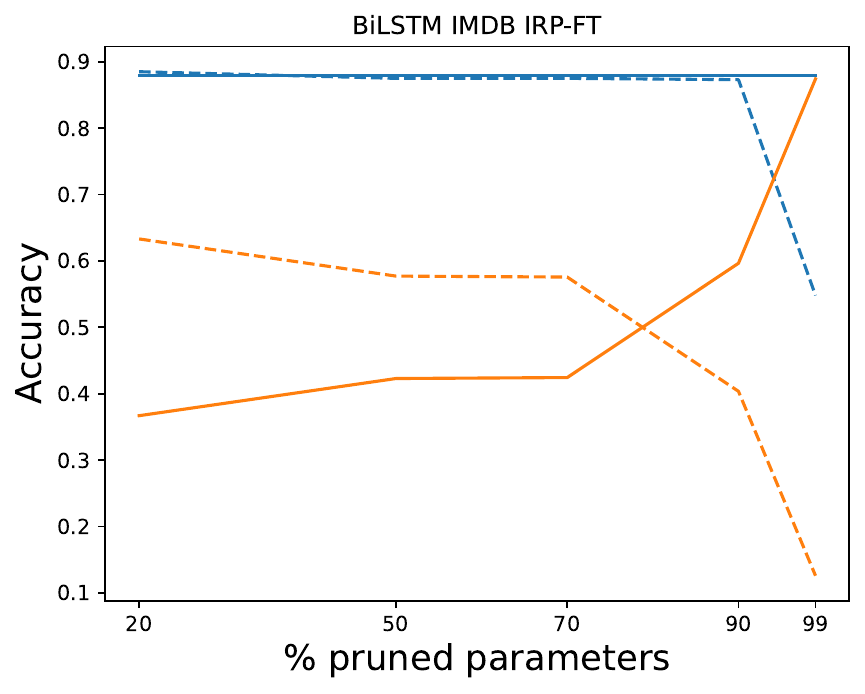} &
 \includegraphics[width=0.223\textwidth]{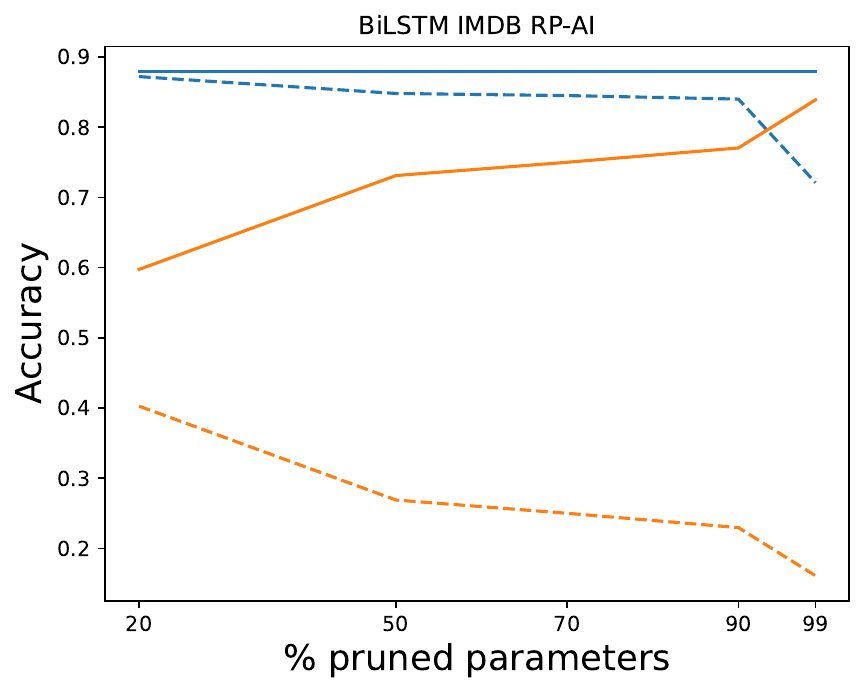}  \\

 \multicolumn{4}{c}{\includegraphics[width=0.7\textwidth]{Images/Effectiveness_all_against_PIEs/legend.pdf}}         
\end{tabular}
\caption{Accuracy of unpruned (black line) and pruned models on PIEs and all samples in the dataset per pruning method, across pruning thresholds (x-axis), over 30 initializations.
}
\label{fig:PIEs_against_all_BiLSTM_IMDB}%
\end{figure*}

\begin{figure*}
\centering
\begin{tabular}{cccc}
 \includegraphics[width=0.223\textwidth]{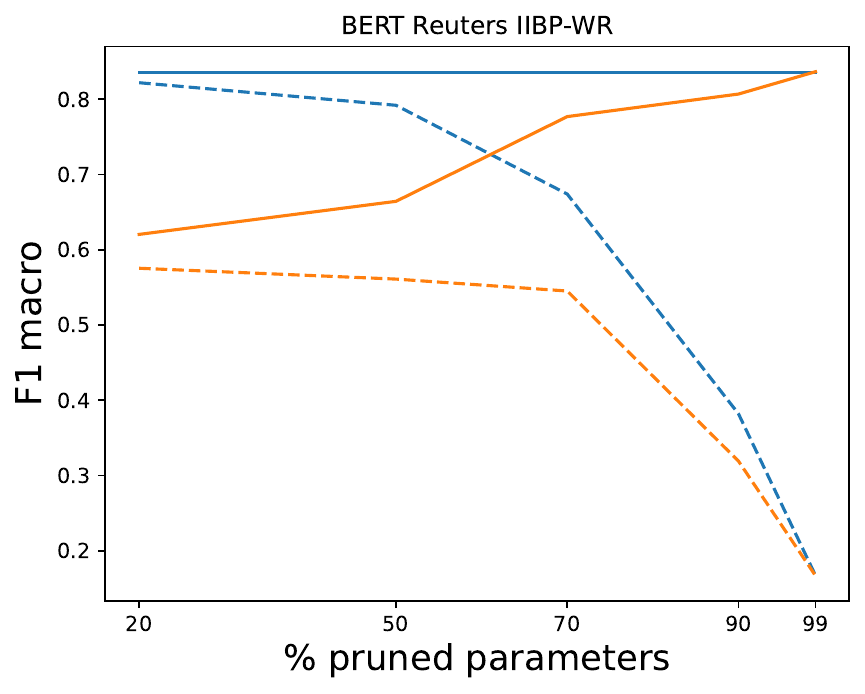} &
 \includegraphics[width=0.223\textwidth]{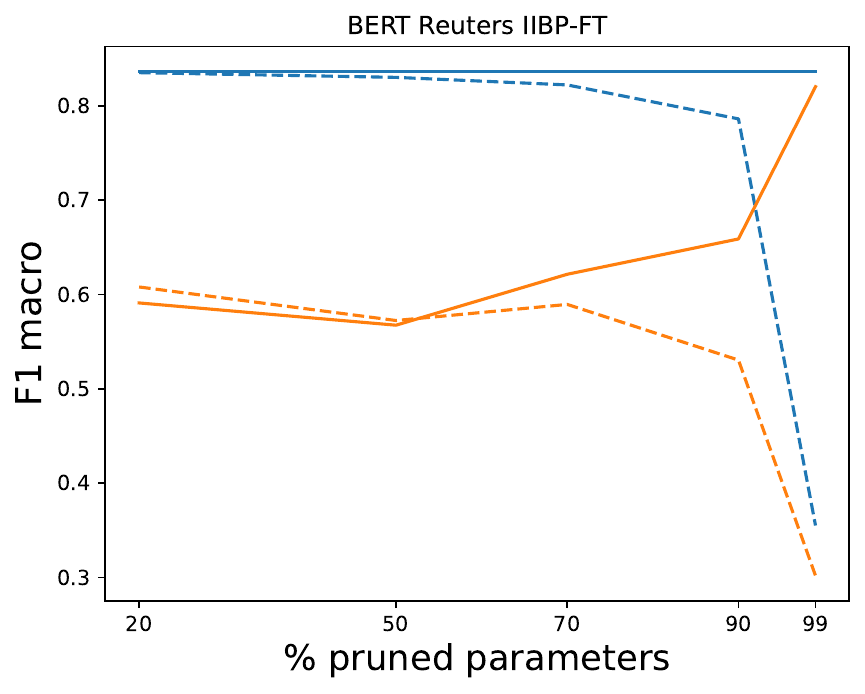} &
 \includegraphics[width=0.223\textwidth]{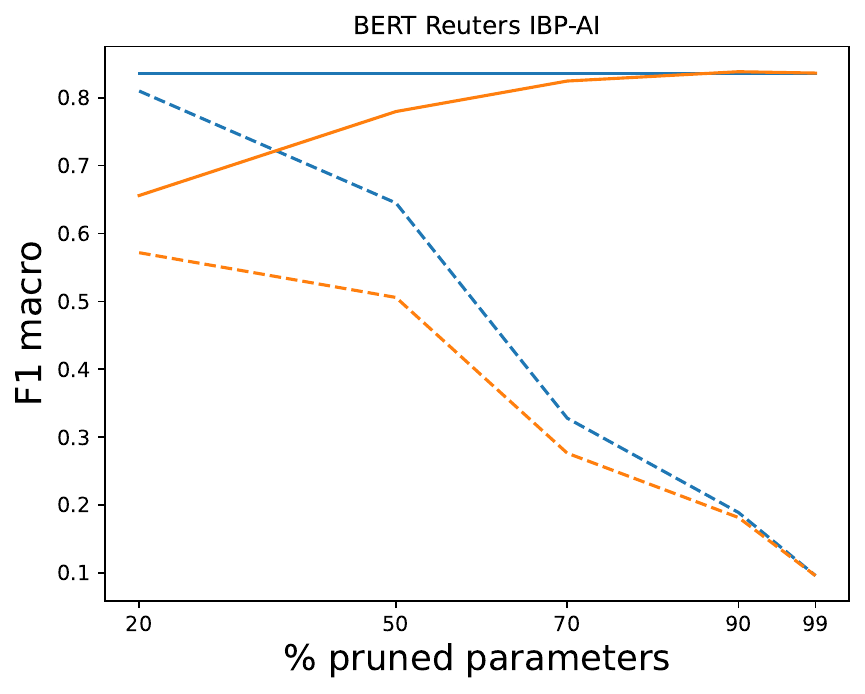} &
 \includegraphics[width=0.223\textwidth]{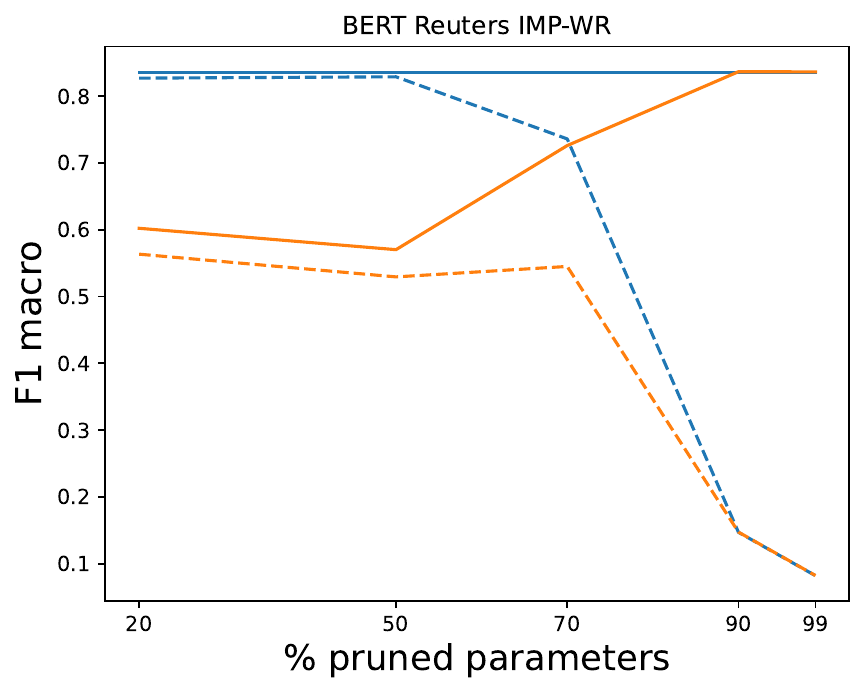} \\

 \includegraphics[width=0.223\textwidth]{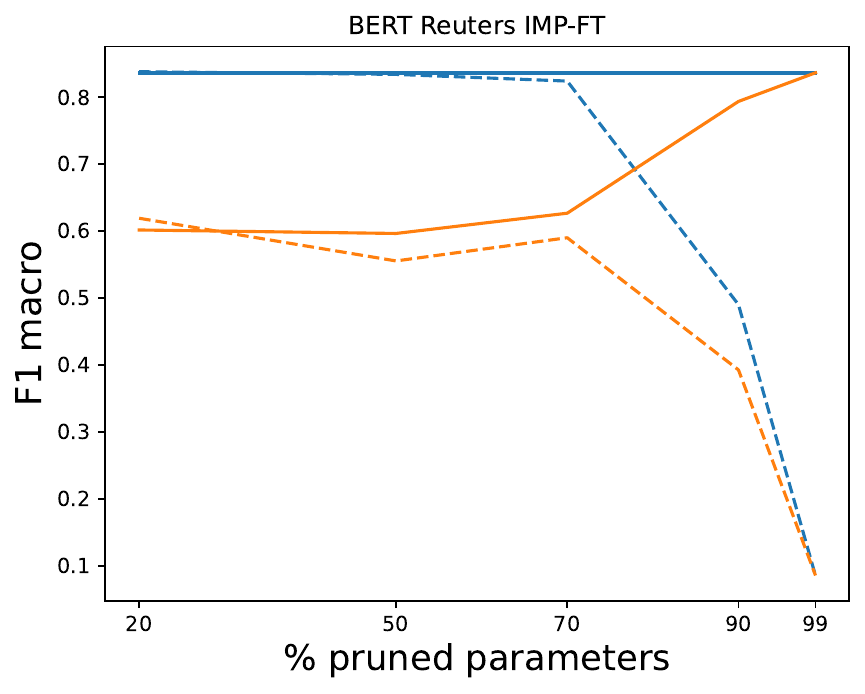} &
 \includegraphics[width=0.223\textwidth]{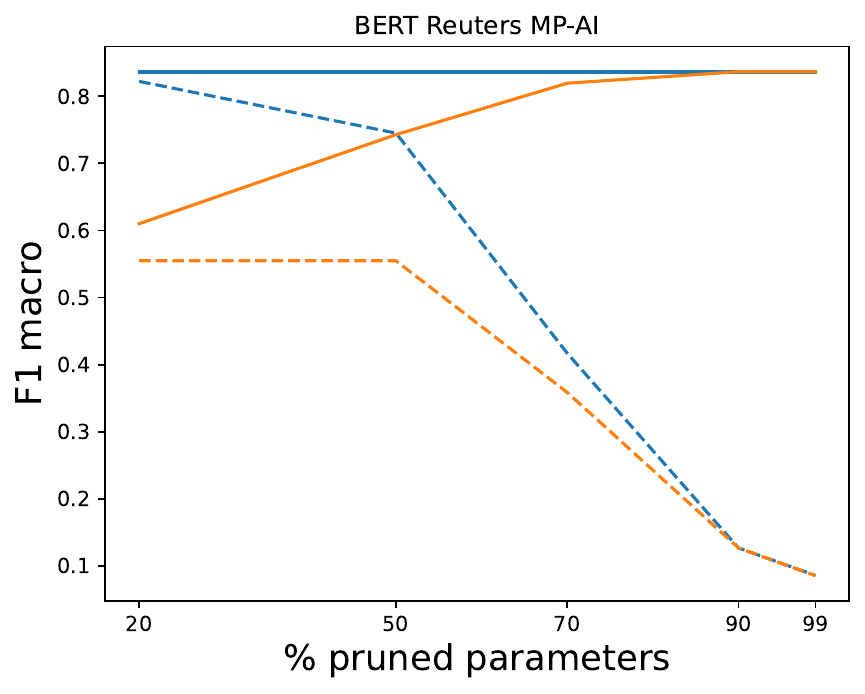} &
 \includegraphics[width=0.223\textwidth]{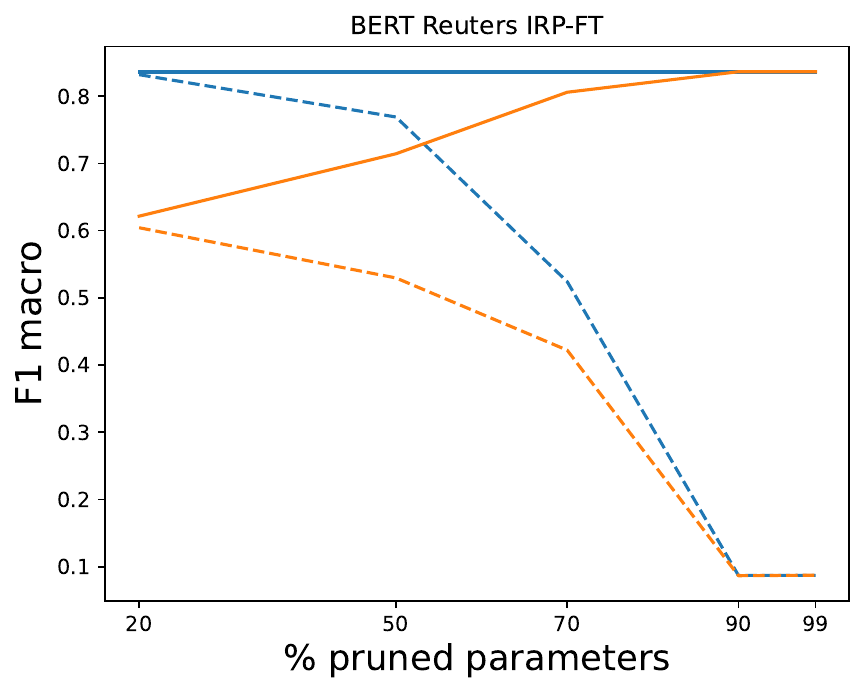} &
 \includegraphics[width=0.223\textwidth]{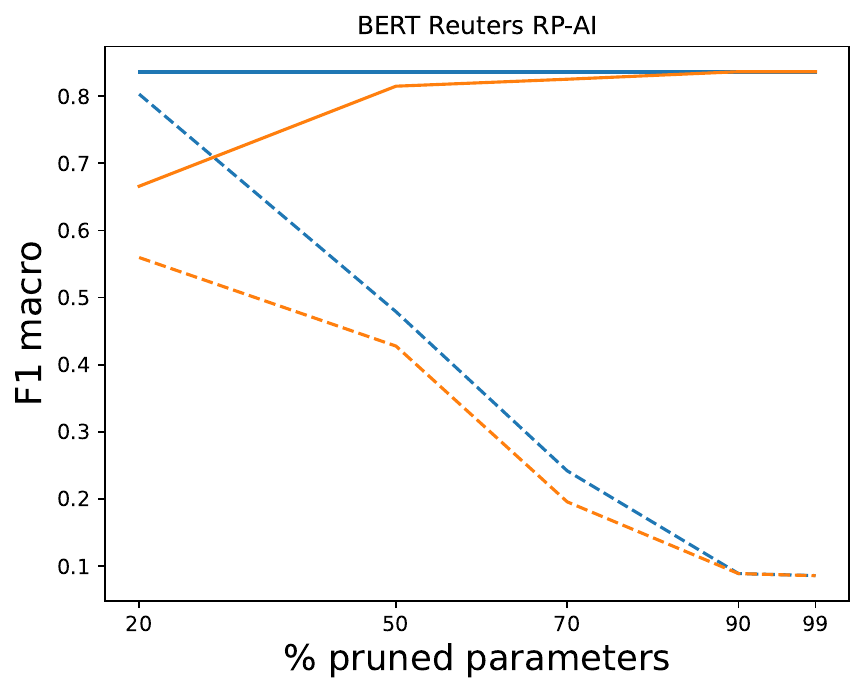}  \\

 \multicolumn{4}{c}{\includegraphics[width=0.7\textwidth]{Images/Effectiveness_all_against_PIEs/legend.pdf}}         
\end{tabular}
\caption{Accuracy of unpruned (black line) and pruned models on PIEs and all samples in the dataset per pruning method, across pruning thresholds (x-axis), over 30 initializations.
}
\label{fig:PIEs_against_all_BERT_Reuters}%
\end{figure*}

\begin{figure*}
\centering
\begin{tabular}{cccc}
 \includegraphics[width=0.223\textwidth]{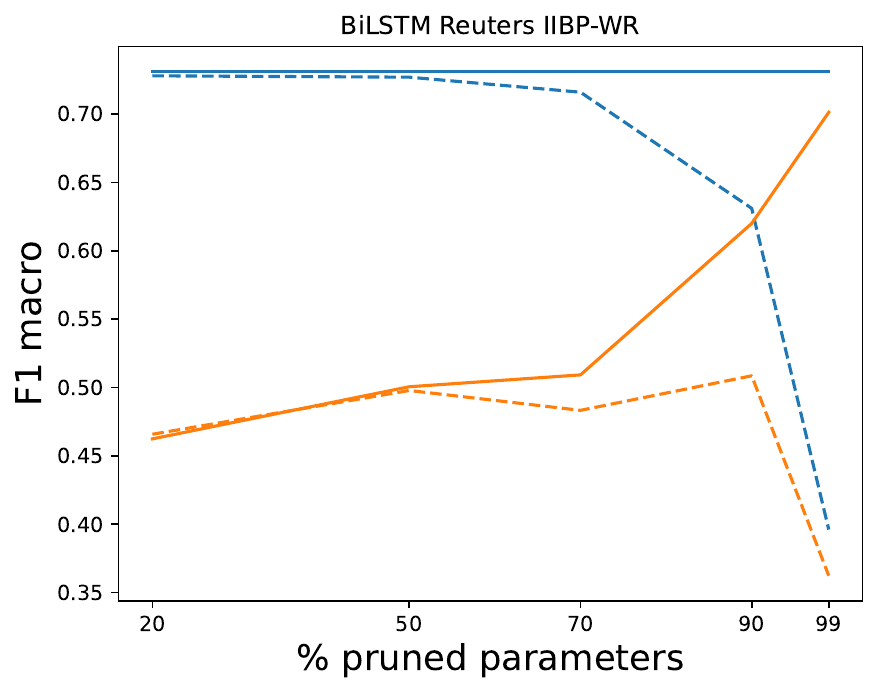} &
 \includegraphics[width=0.223\textwidth]{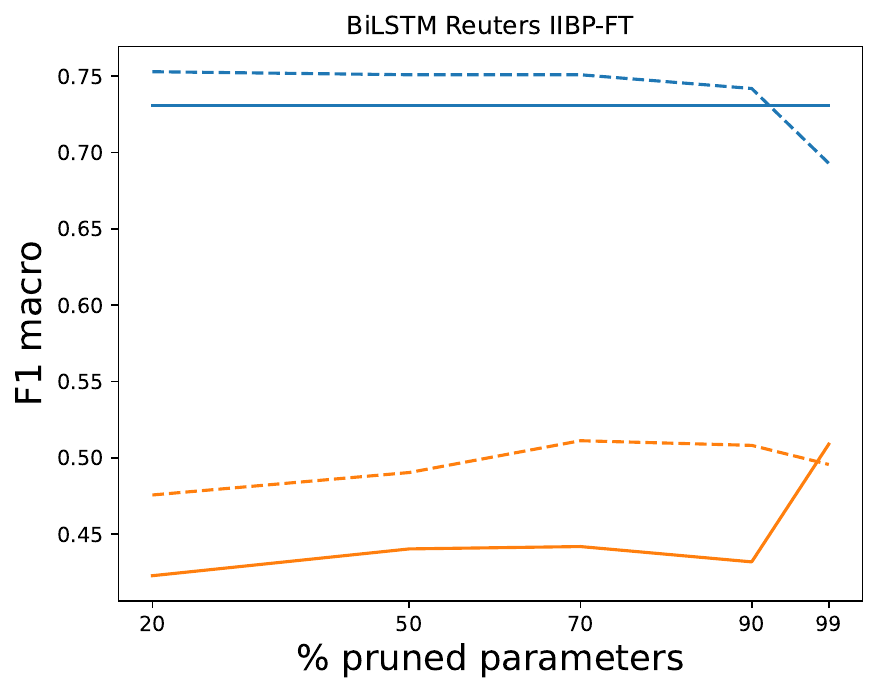} &
 \includegraphics[width=0.223\textwidth]{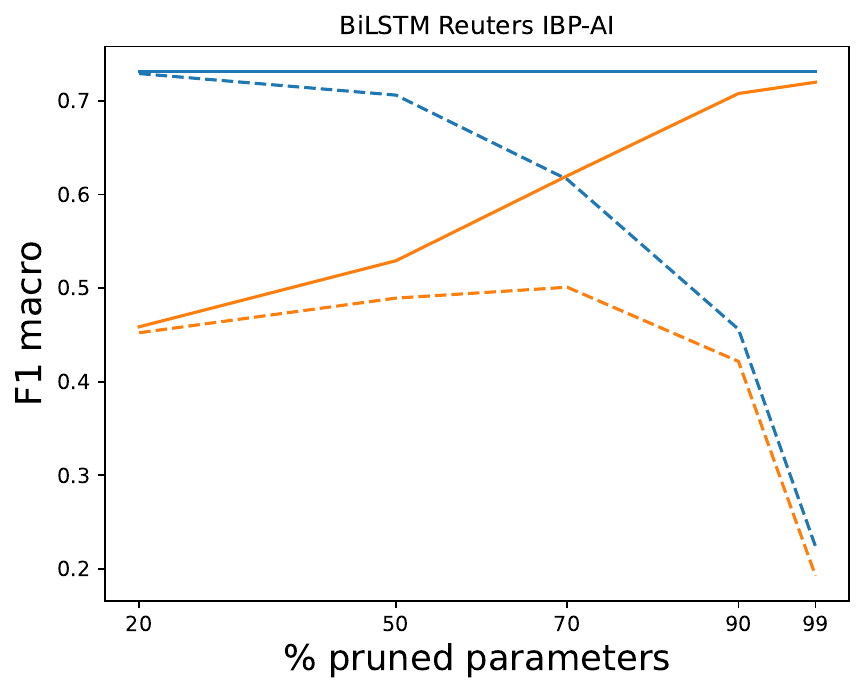} &
 \includegraphics[width=0.223\textwidth]{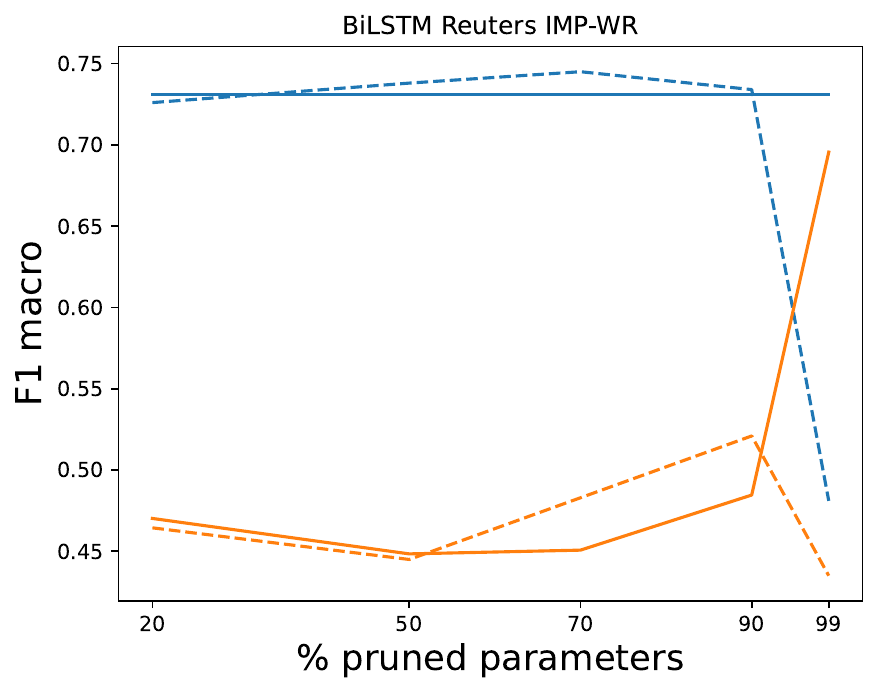} \\

 \includegraphics[width=0.223\textwidth]{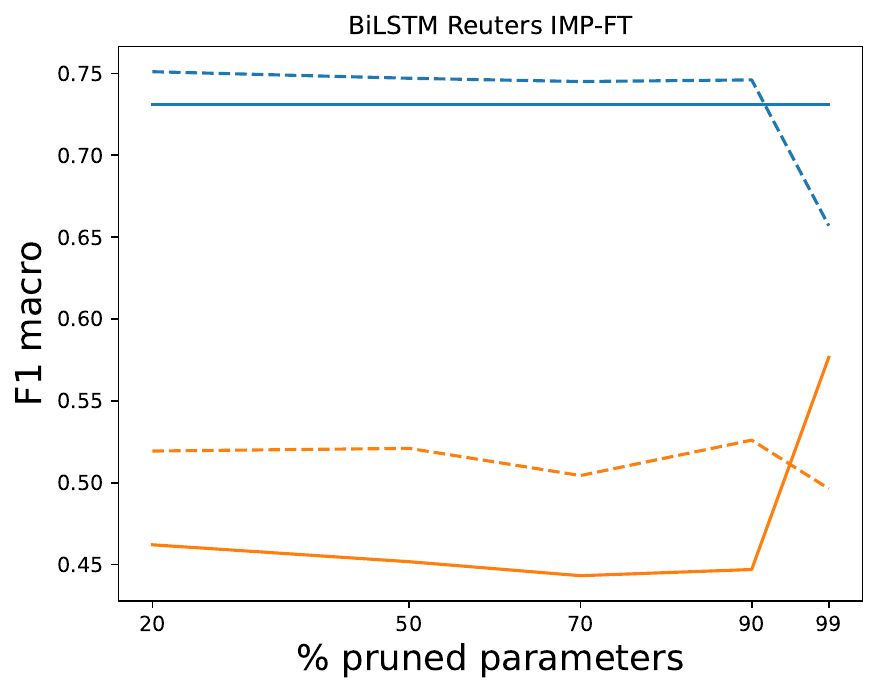} &
 \includegraphics[width=0.223\textwidth]{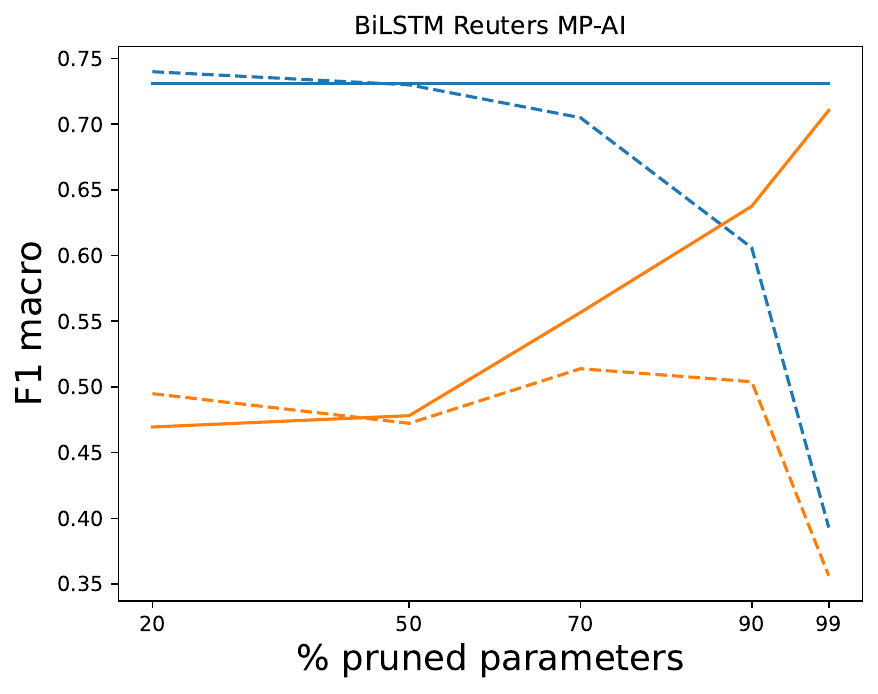} &
 \includegraphics[width=0.223\textwidth]{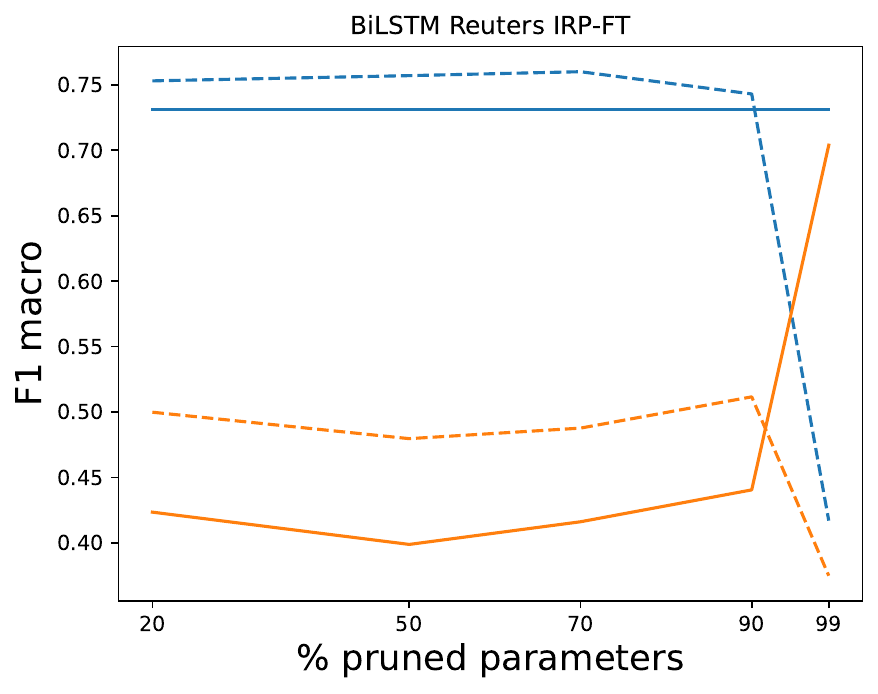} &
 \includegraphics[width=0.223\textwidth]{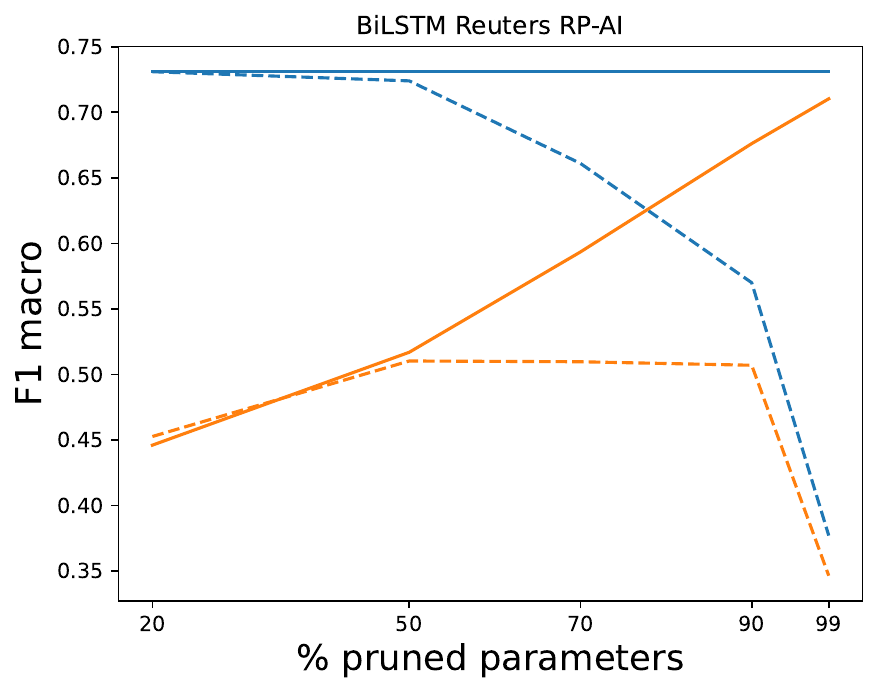}  \\

 \multicolumn{4}{c}{\includegraphics[width=0.7\textwidth]{Images/Effectiveness_all_against_PIEs/legend.pdf}}         
\end{tabular}
\caption{Accuracy of unpruned (black line) and pruned models on PIEs and all samples in the dataset per pruning method, across pruning thresholds (x-axis), over 30 initializations.
}
\label{fig:PIEs_against_all_BiLSTM_Reuters}%
\end{figure*}

\begin{figure*}
\centering
\begin{tabular}{cccc}
 \includegraphics[width=0.223\textwidth]{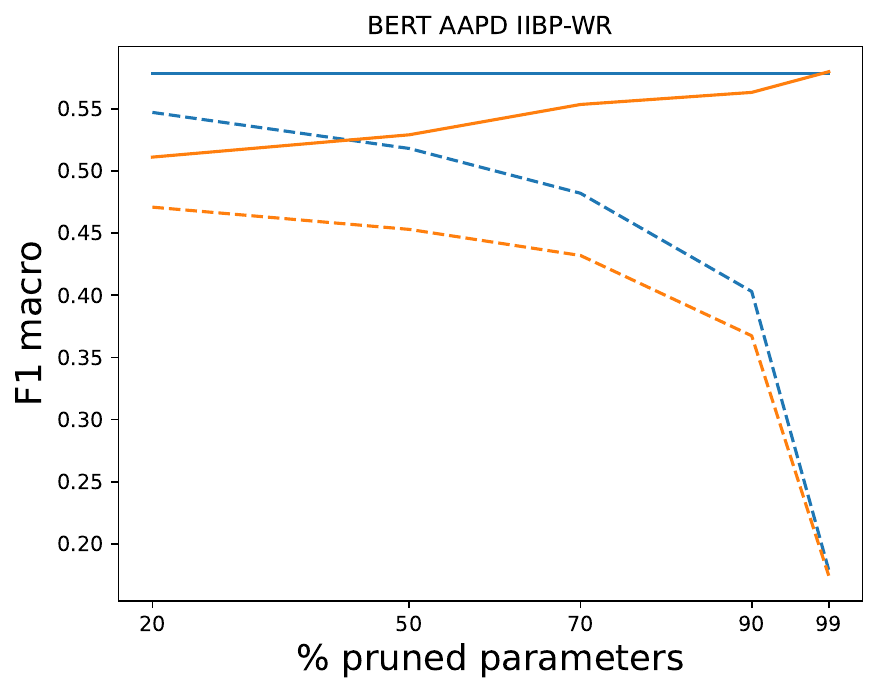} &
 \includegraphics[width=0.223\textwidth]{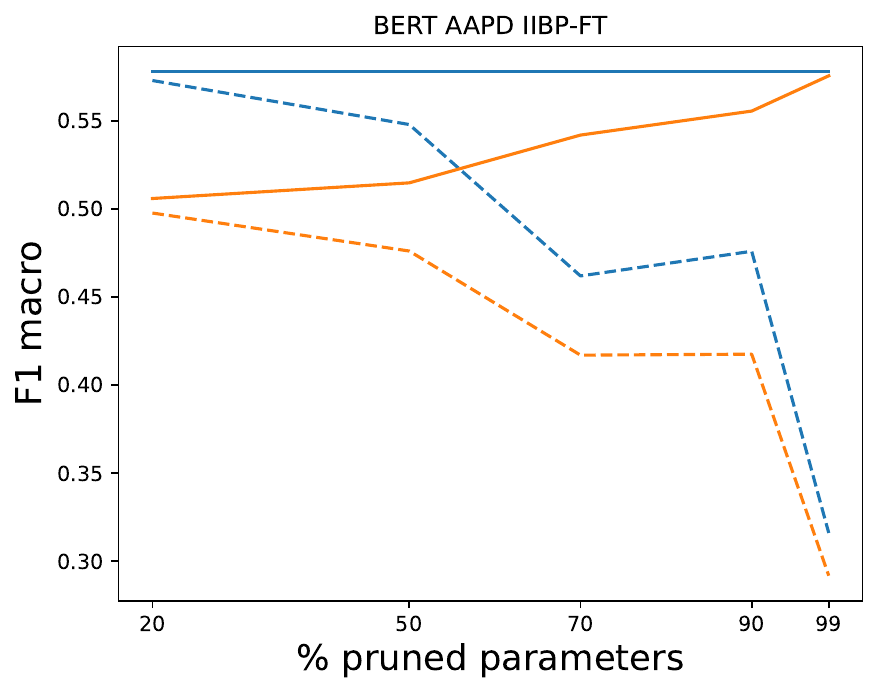} &
 \includegraphics[width=0.223\textwidth]{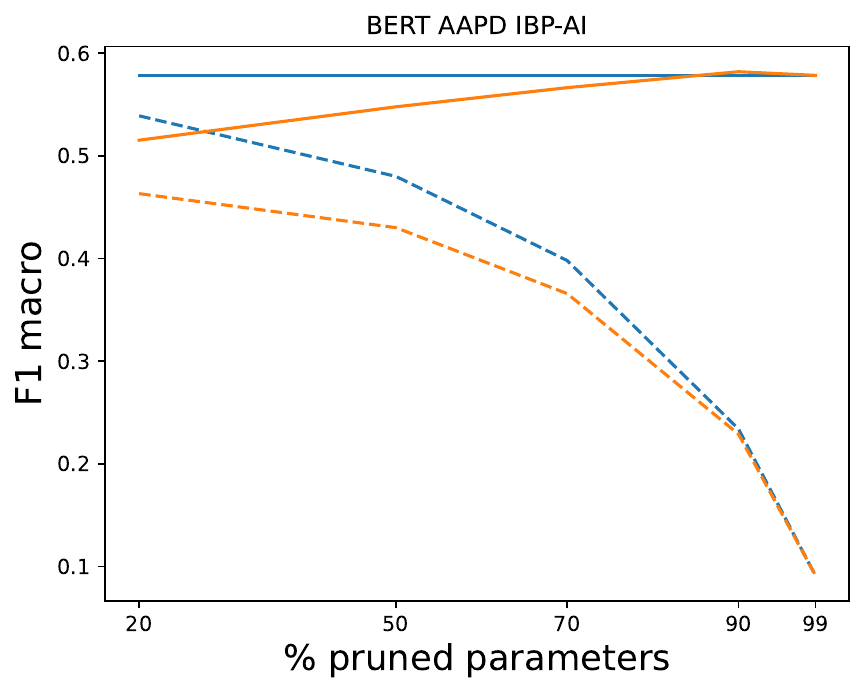} &
 \includegraphics[width=0.223\textwidth]{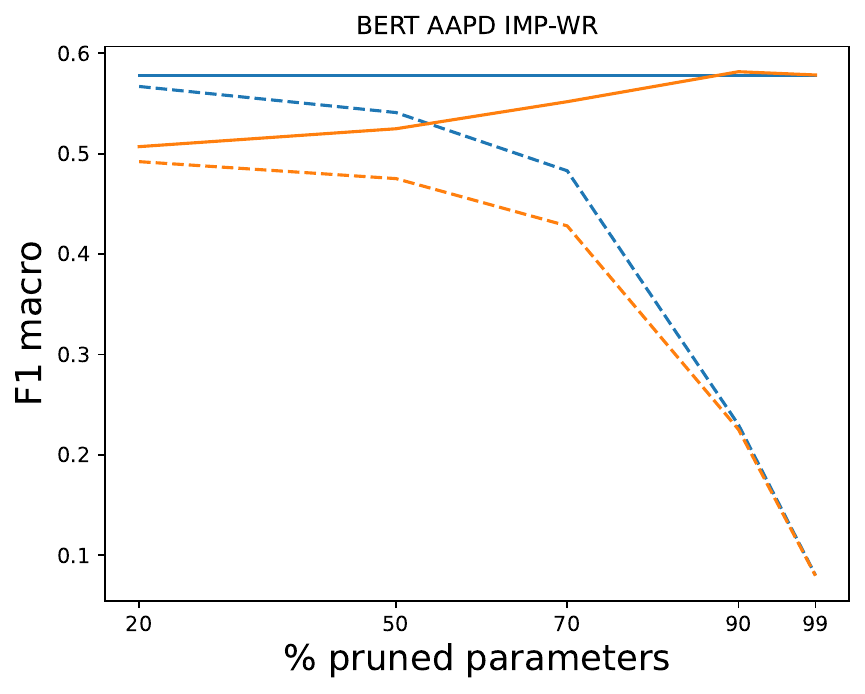} \\

 \includegraphics[width=0.223\textwidth]{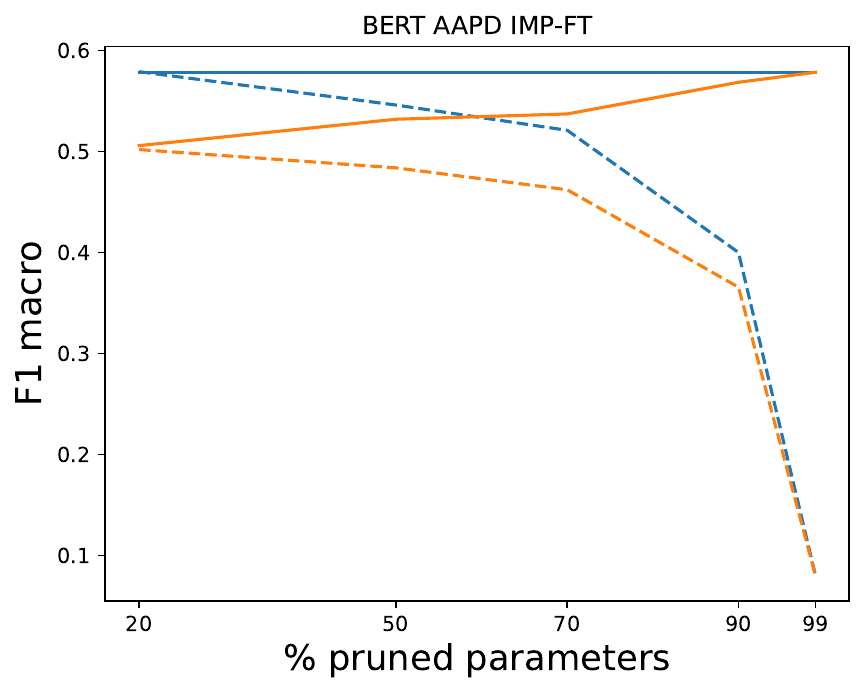} &
 \includegraphics[width=0.223\textwidth]{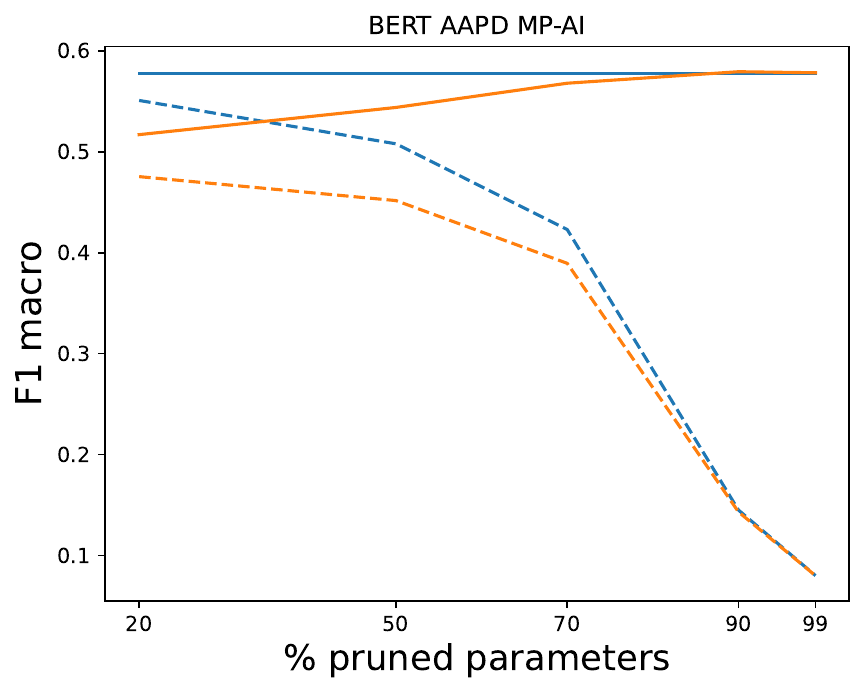} &
 \includegraphics[width=0.223\textwidth]{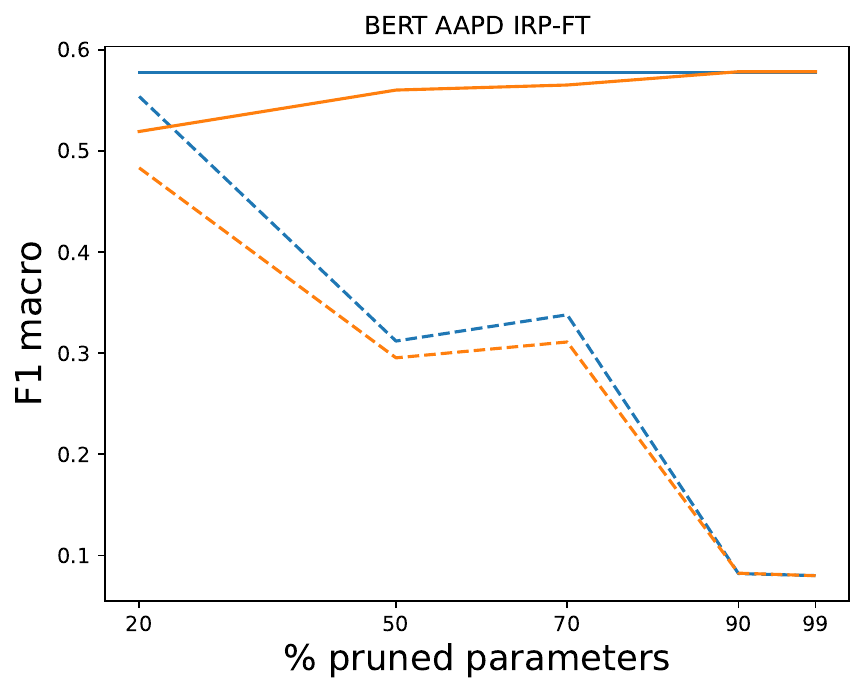} &
 \includegraphics[width=0.223\textwidth]{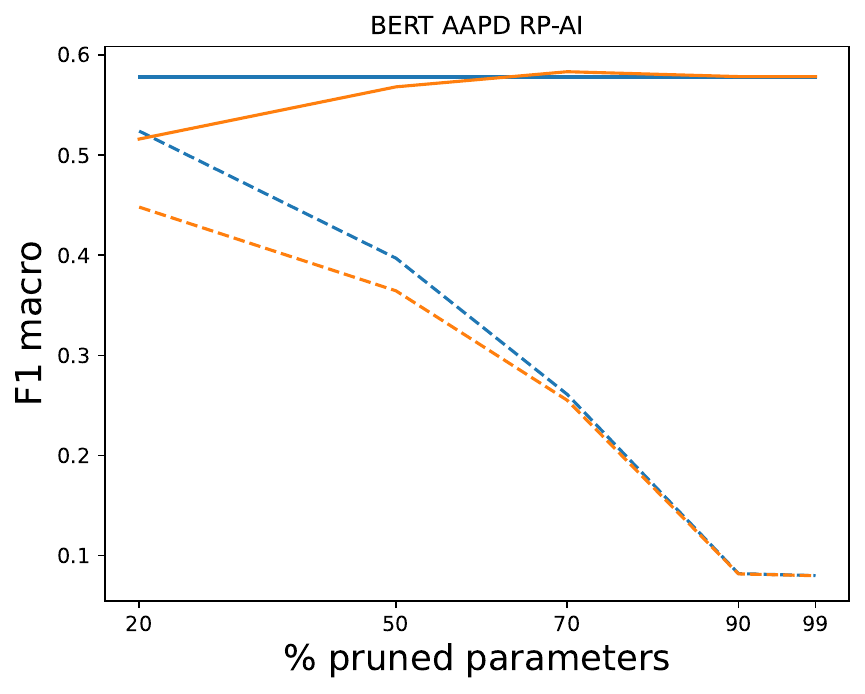}  \\

 \multicolumn{4}{c}{\includegraphics[width=0.7\textwidth]{Images/Effectiveness_all_against_PIEs/legend.pdf}}         
\end{tabular}
\caption{Accuracy of unpruned (black line) and pruned models on PIEs and all samples in the dataset per pruning method, across pruning thresholds (x-axis), over 30 initializations.
}
\label{fig:PIEs_against_all_BERT_AAPD}%
\end{figure*}

\begin{figure*}
\centering
\begin{tabular}{cccc}
 \includegraphics[width=0.223\textwidth]{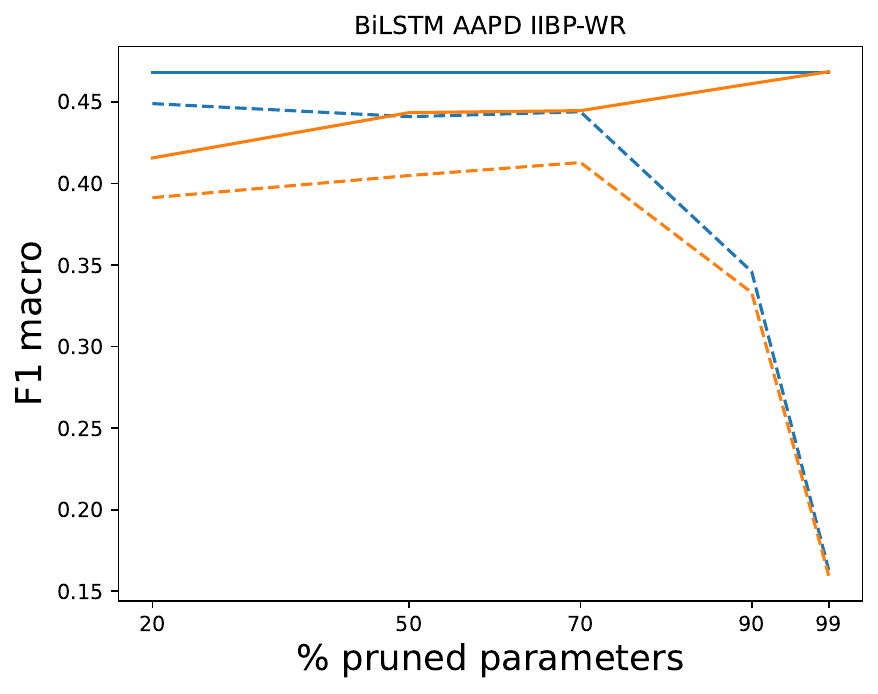} &
 \includegraphics[width=0.223\textwidth]{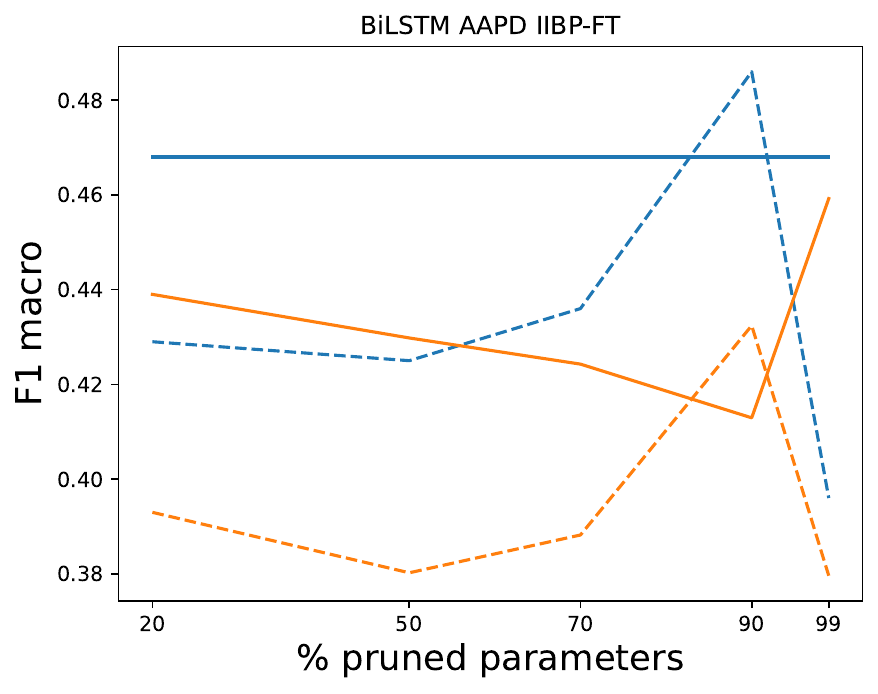} &
 \includegraphics[width=0.223\textwidth]{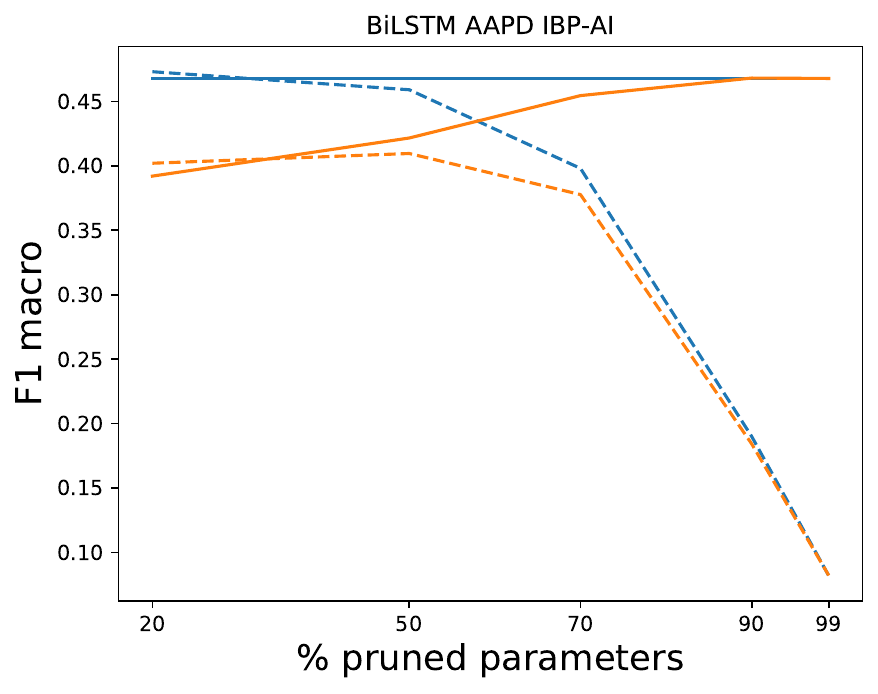} &
 \includegraphics[width=0.223\textwidth]{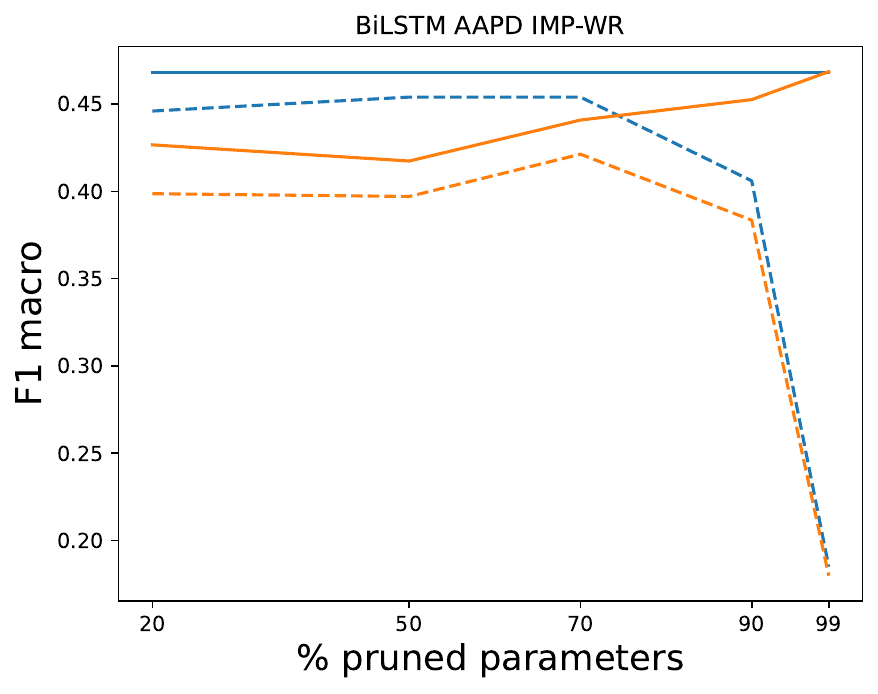} \\

 \includegraphics[width=0.223\textwidth]{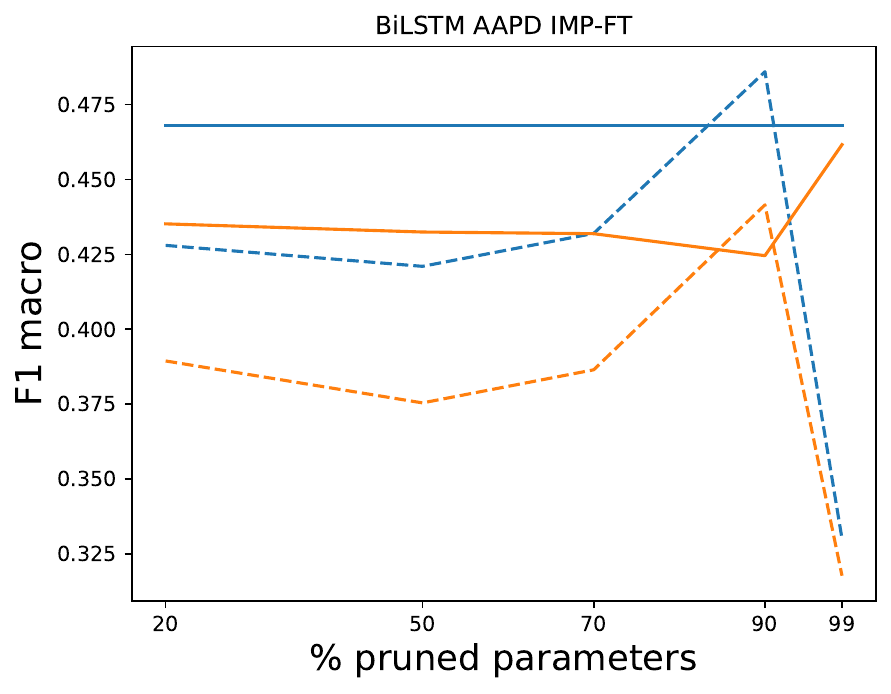} &
 \includegraphics[width=0.223\textwidth]{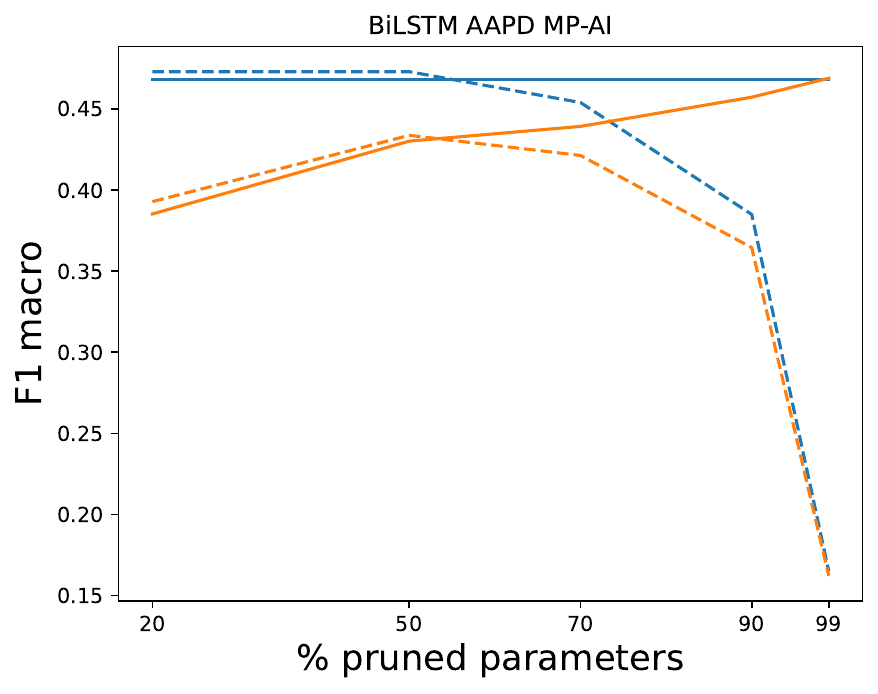} &
 \includegraphics[width=0.223\textwidth]{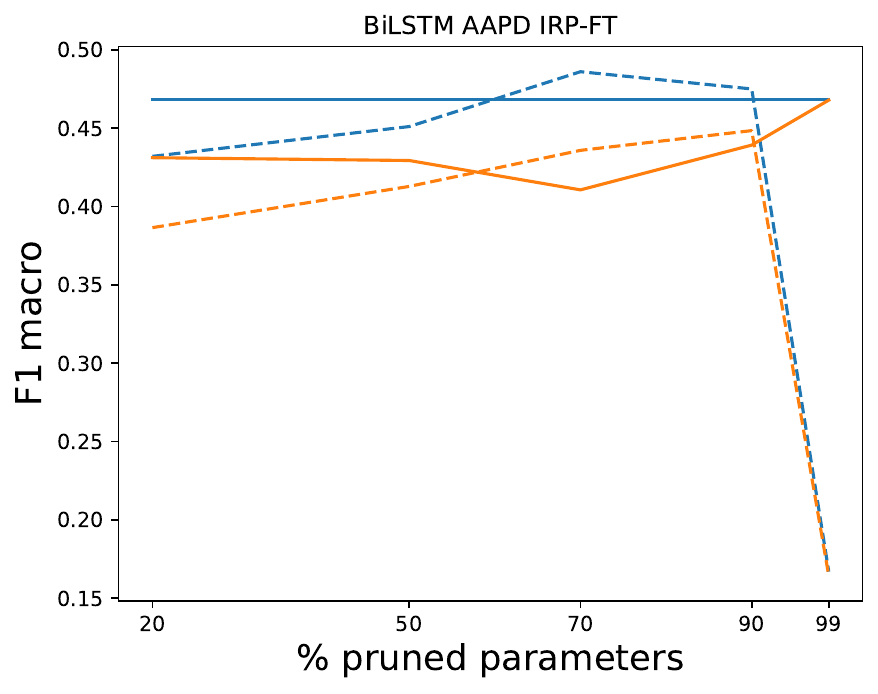} &
 \includegraphics[width=0.223\textwidth]{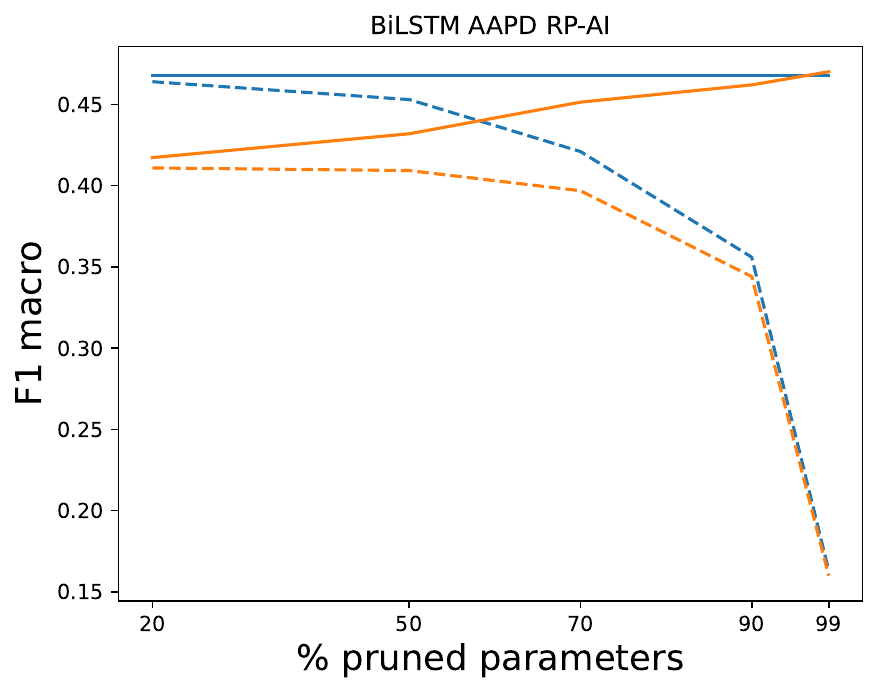}  \\

 \multicolumn{4}{c}{\includegraphics[width=0.7\textwidth]{Images/Effectiveness_all_against_PIEs/legend.pdf}}         
\end{tabular}
\caption{Accuracy of unpruned (black line) and pruned models on PIEs and all samples in the dataset per pruning method, across pruning thresholds (x-axis), over 30 initializations.
}
\label{fig:PIEs_against_all_BiLSTM_AAPD}%
\end{figure*}

In Figures \ref{fig:PIEs_against_all_BiLSTM_SNLI}, \ref{fig:PIEs_against_all_BERT_IMDB}, \ref{fig:PIEs_against_all_BiLSTM_IMDB}, \ref{fig:PIEs_against_all_BERT_Reuters}, \ref{fig:PIEs_against_all_BiLSTM_Reuters}, \ref{fig:PIEs_against_all_BERT_AAPD}, and \ref{fig:PIEs_against_all_BiLSTM_AAPD} we report the accuracy of unpruned and pruned models on PIEs and all samples in the dataset per pruning method, across pruning thresholds. The accuracy on PIEs is lower than the accuracy on all data points for both pruned and unpruned models. The accuracy of the unpruned model on PIEs increases when increasing the amount of pruned parameters, while the accuracy of the pruned model decreases in the same setting. This is because the pruned model misclassifies more samples that are correctly classified by the unpruned model, increasing the amount of disagreement, hence the number of PIEs too.

\subsection{Influential examples in PIEs}\label{app:influential_examples_in_PIEs}

\begin{figure*}
\setlength\tabcolsep{-1pt}
\centering
\begin{tabular}{cccc}

 \includegraphics[width=0.23\textwidth]{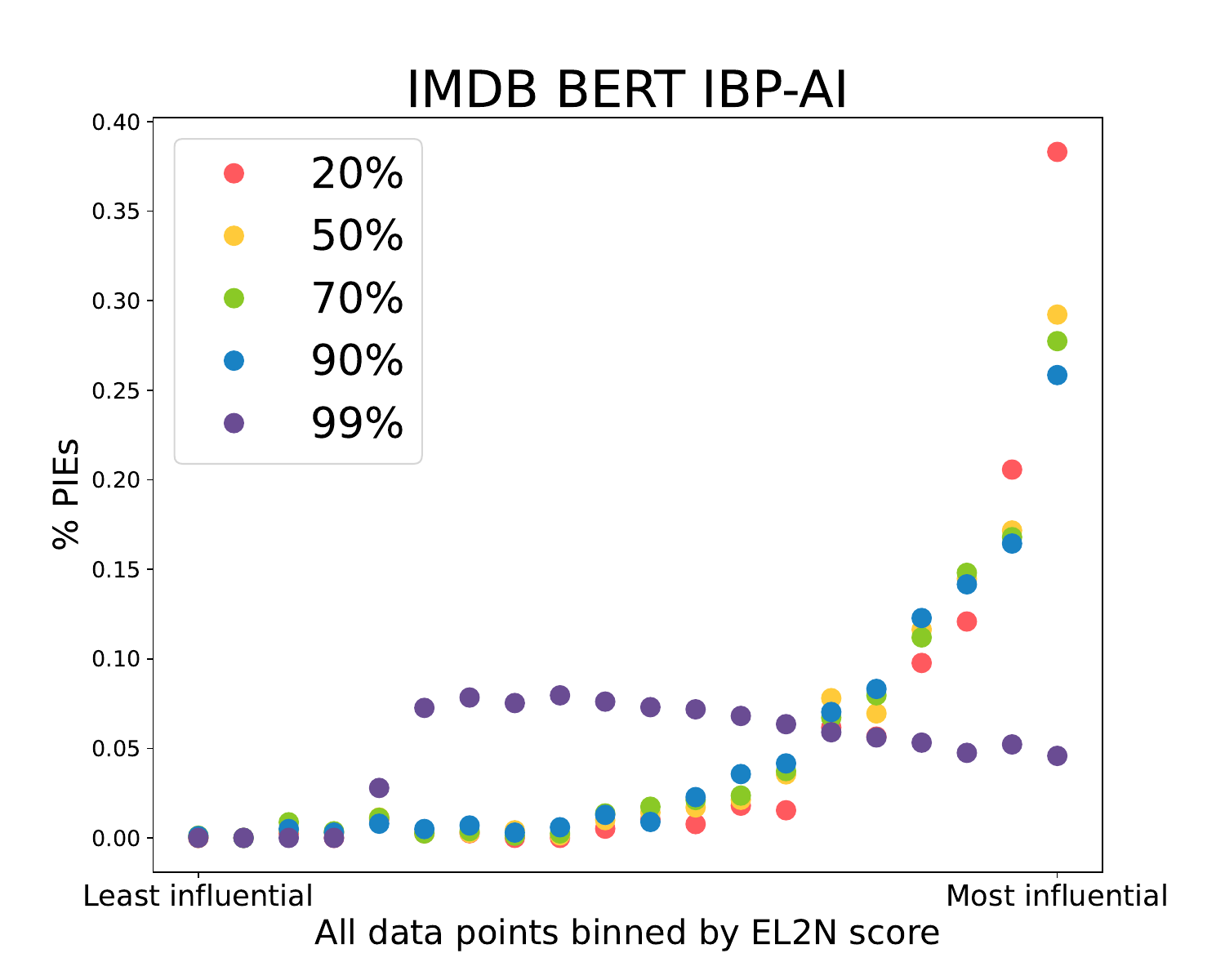} &
 \includegraphics[width=0.23\textwidth]{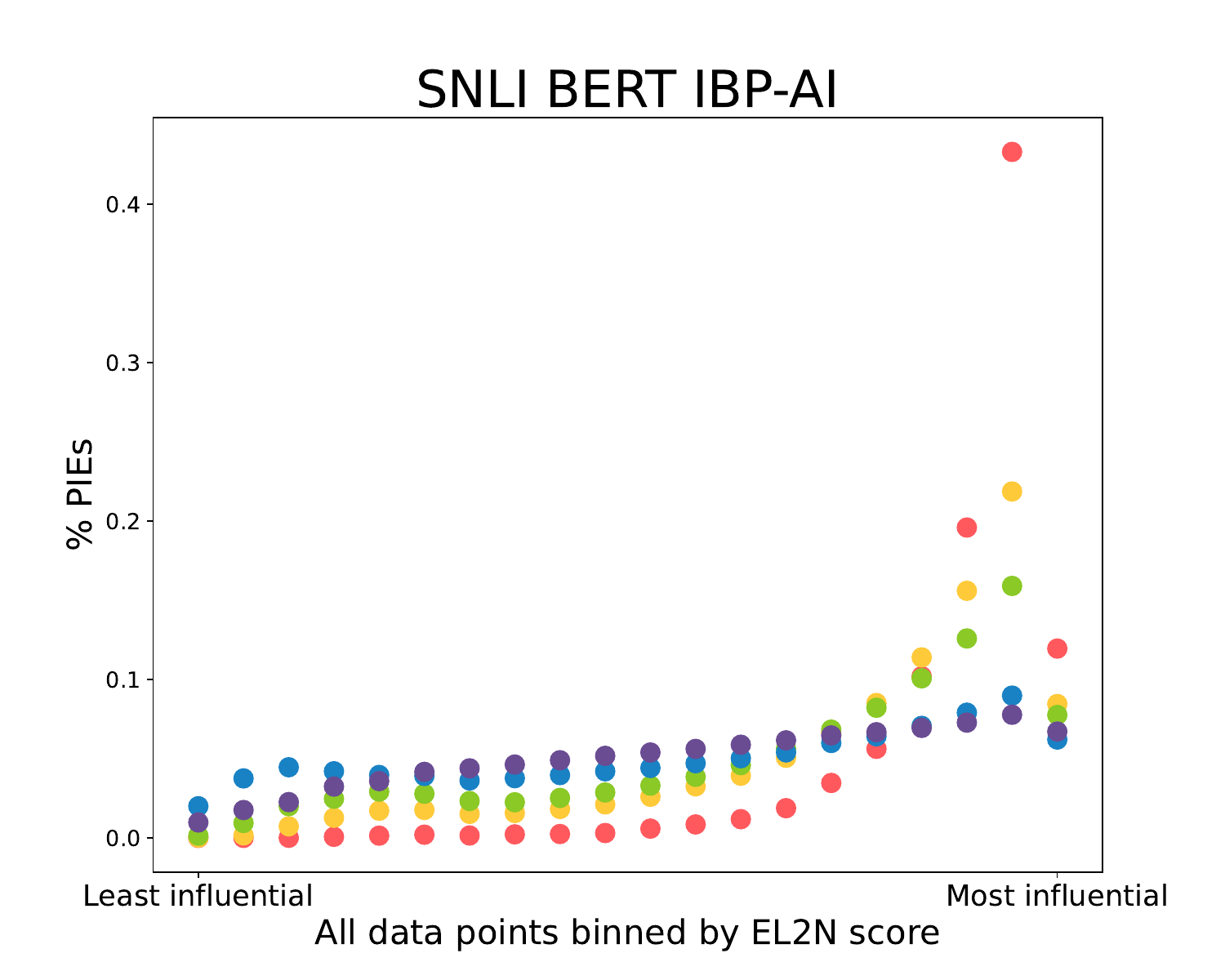} &
 \includegraphics[width=0.23\textwidth]{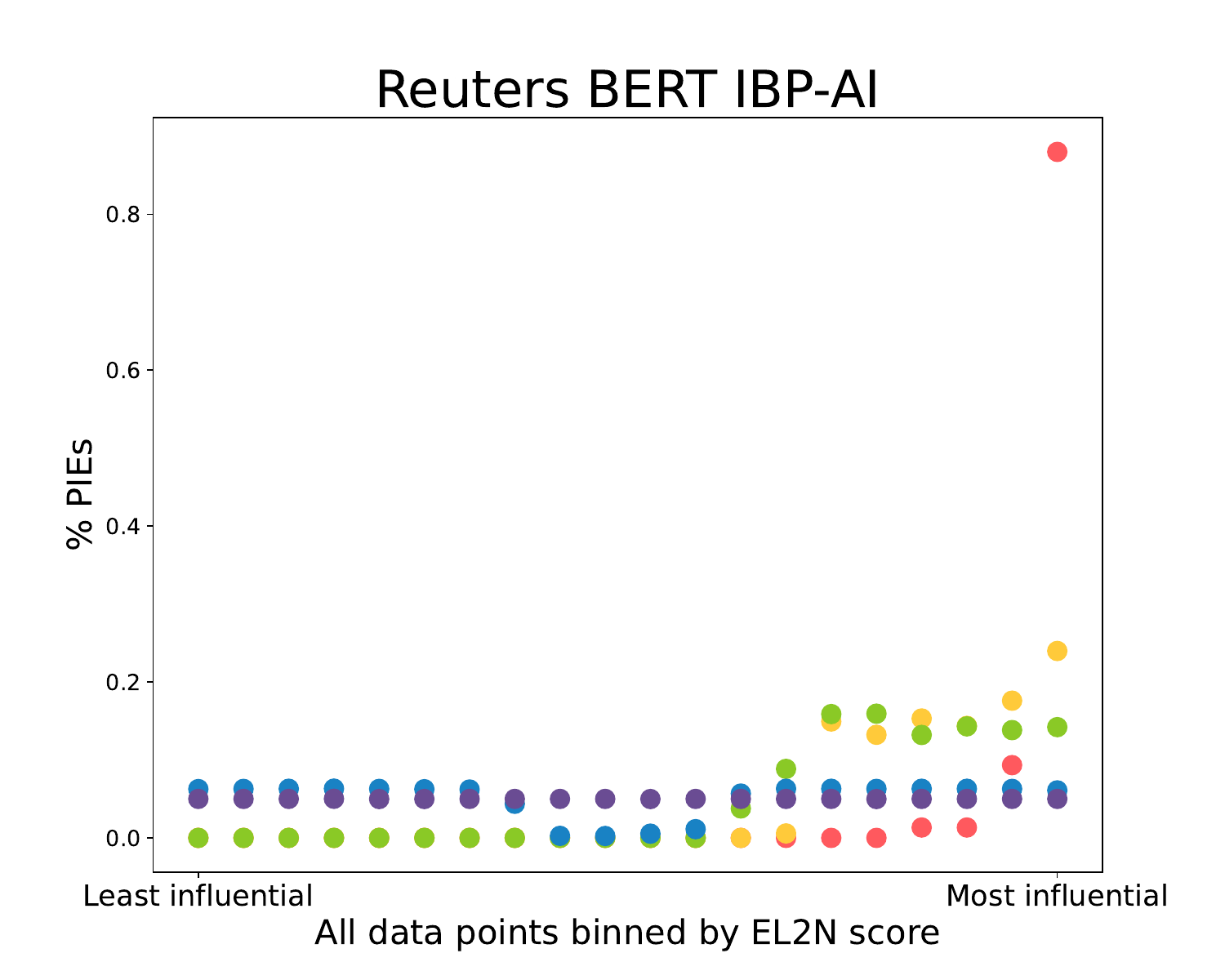} &
 \includegraphics[width=0.23\textwidth]{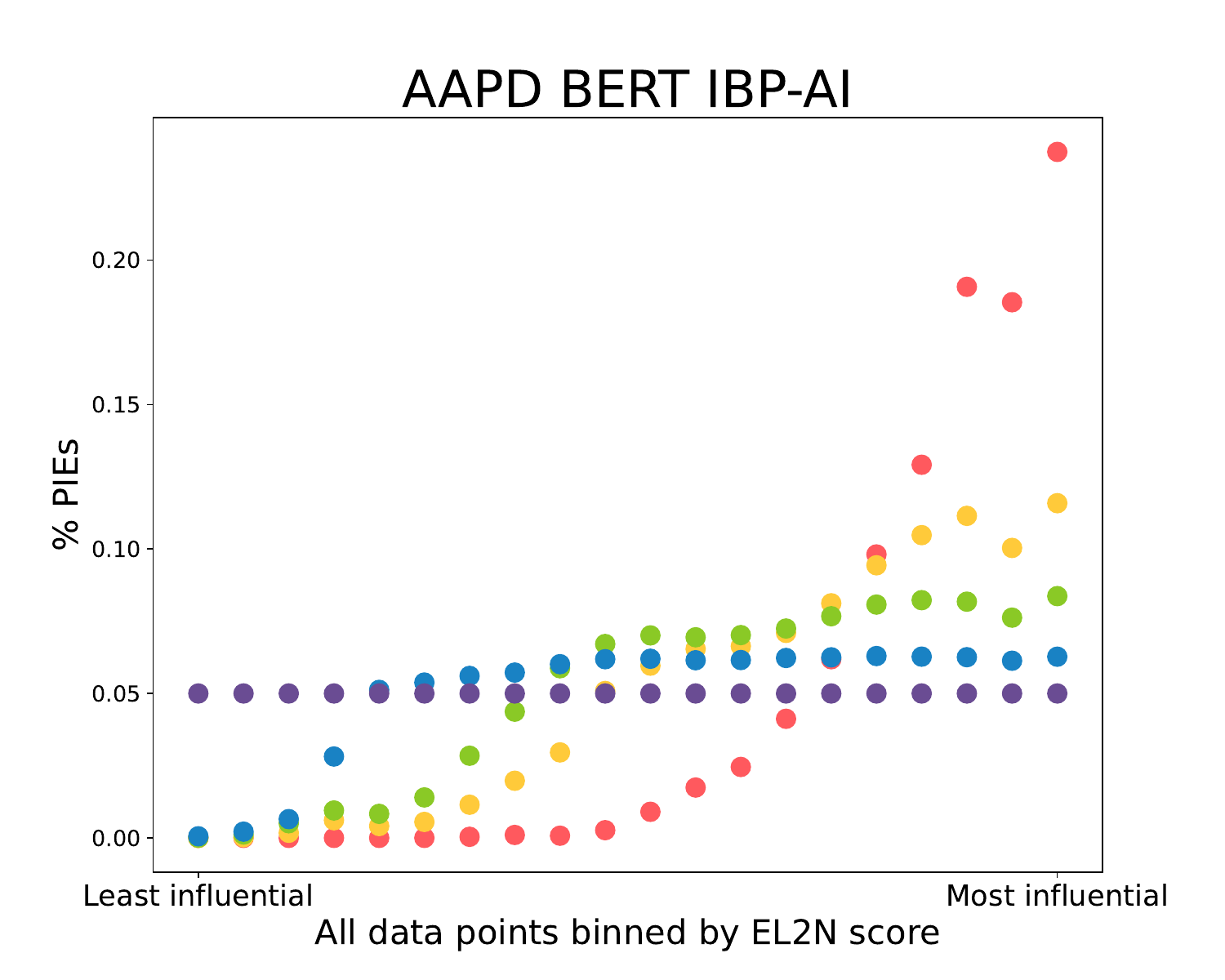} 
  \\

 \includegraphics[width=0.23\textwidth]{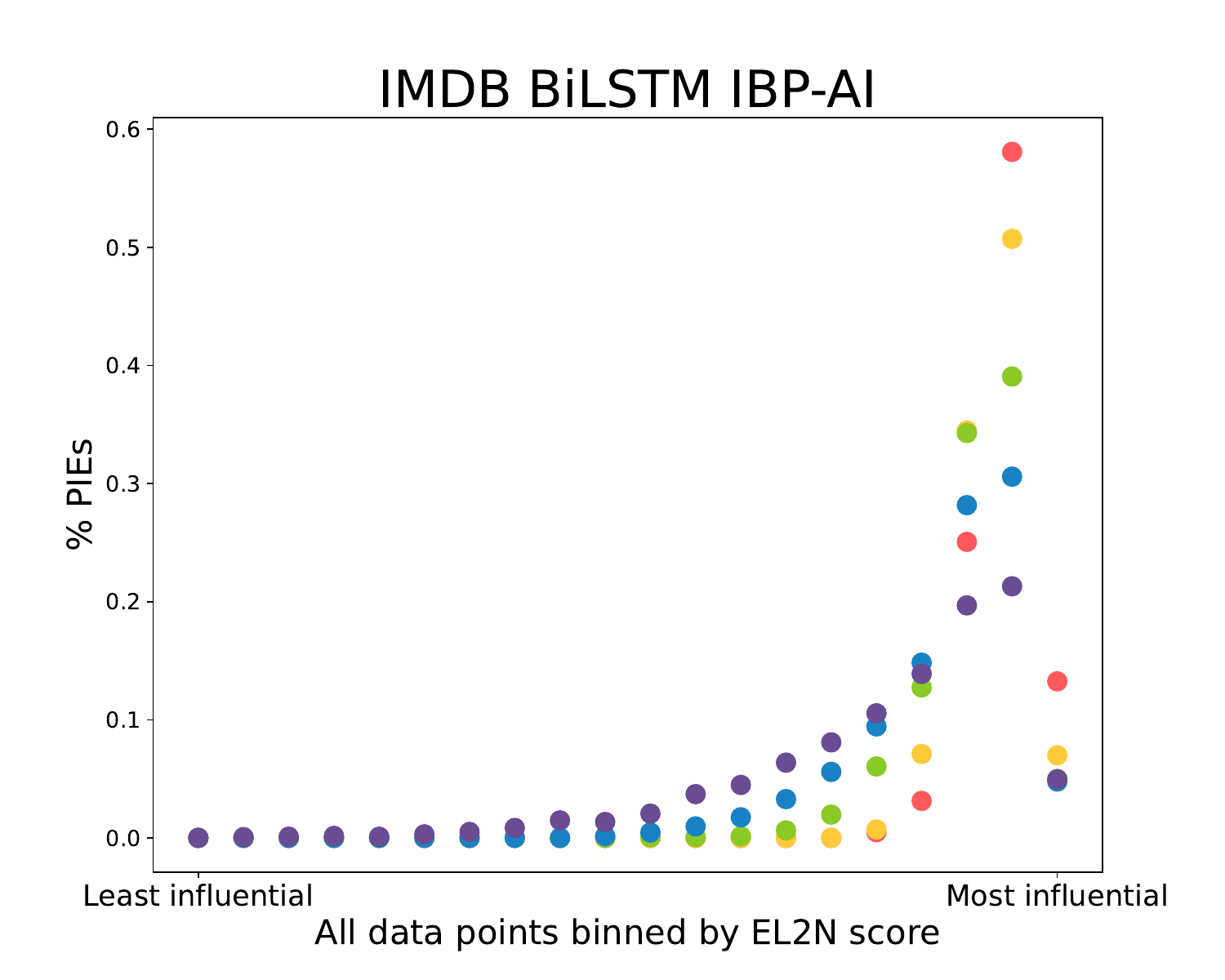} &
 \includegraphics[width=0.23\textwidth]{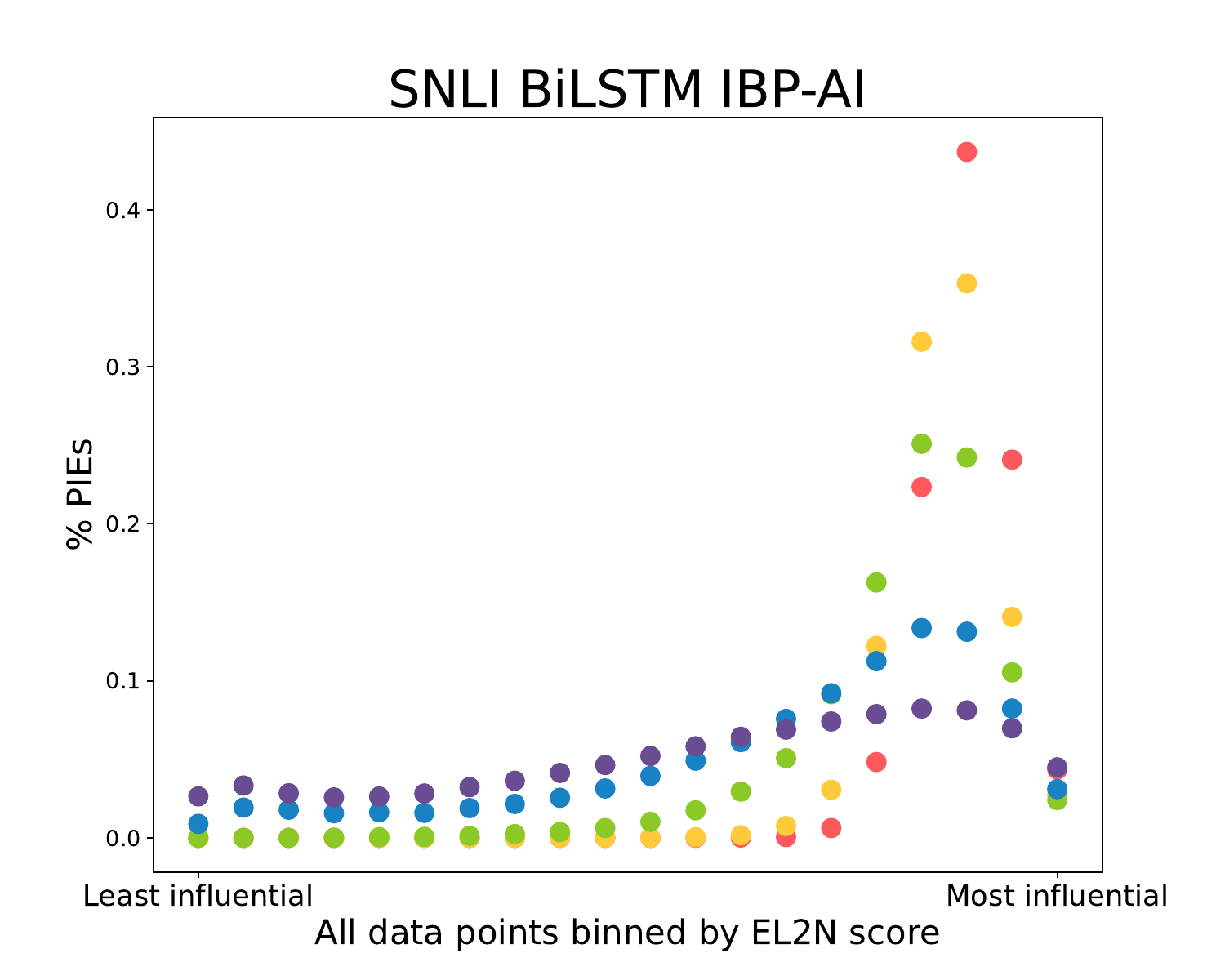} &
 \includegraphics[width=0.23\textwidth]{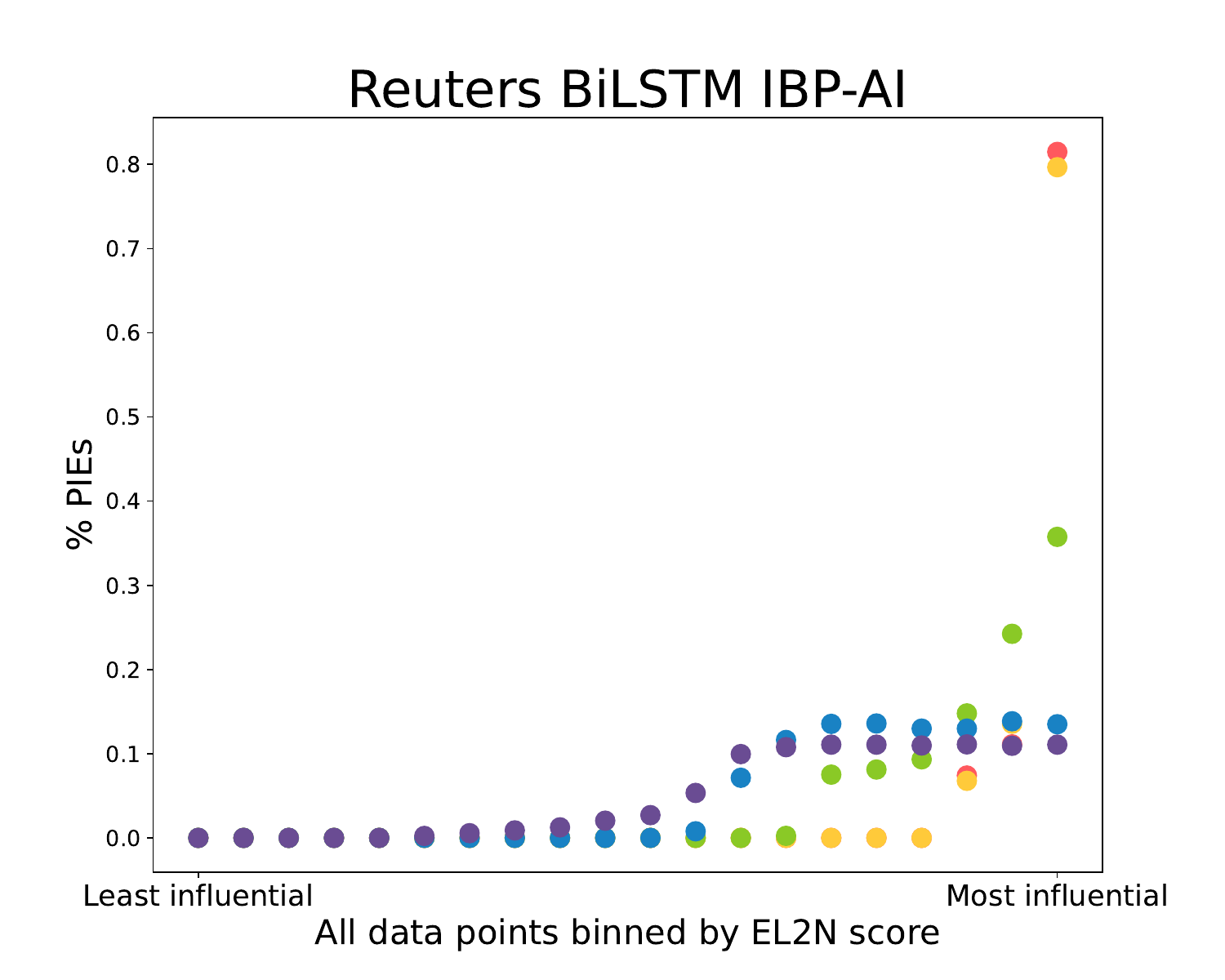} &
 \includegraphics[width=0.23\textwidth]{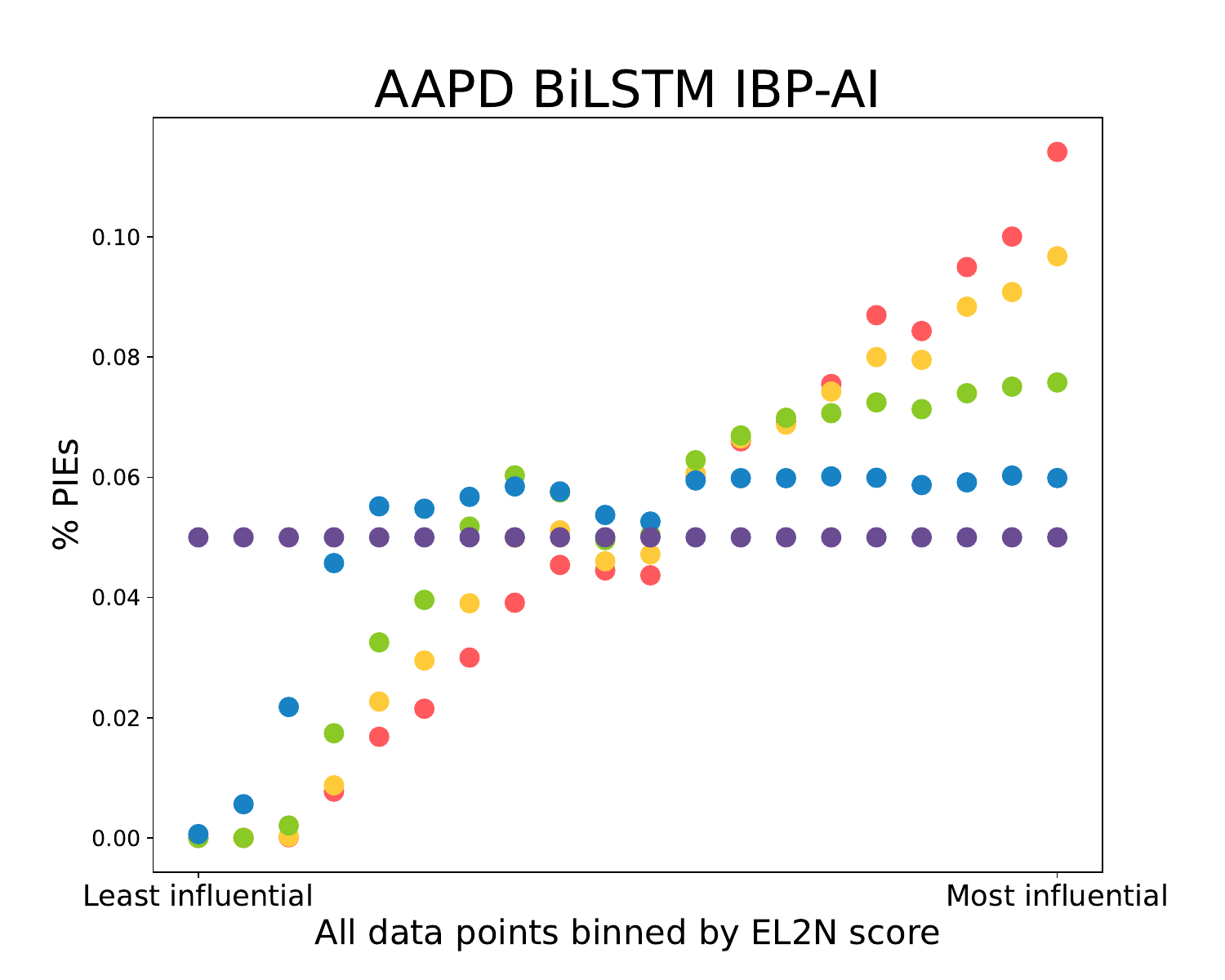} \\

\end{tabular}
\caption{
Percentage of data points that are PIEs (y axis) versus degree of influence (EL2N score) of all data points in the training set (x axis) for IBP-AI.
}
\label{fig:IMDB_BERT_anal4_grid_IBP-AI}%
\end{figure*}

\begin{figure*}
\setlength\tabcolsep{-1pt}
\centering
\begin{tabular}{cccc}

 \includegraphics[width=0.23\textwidth]{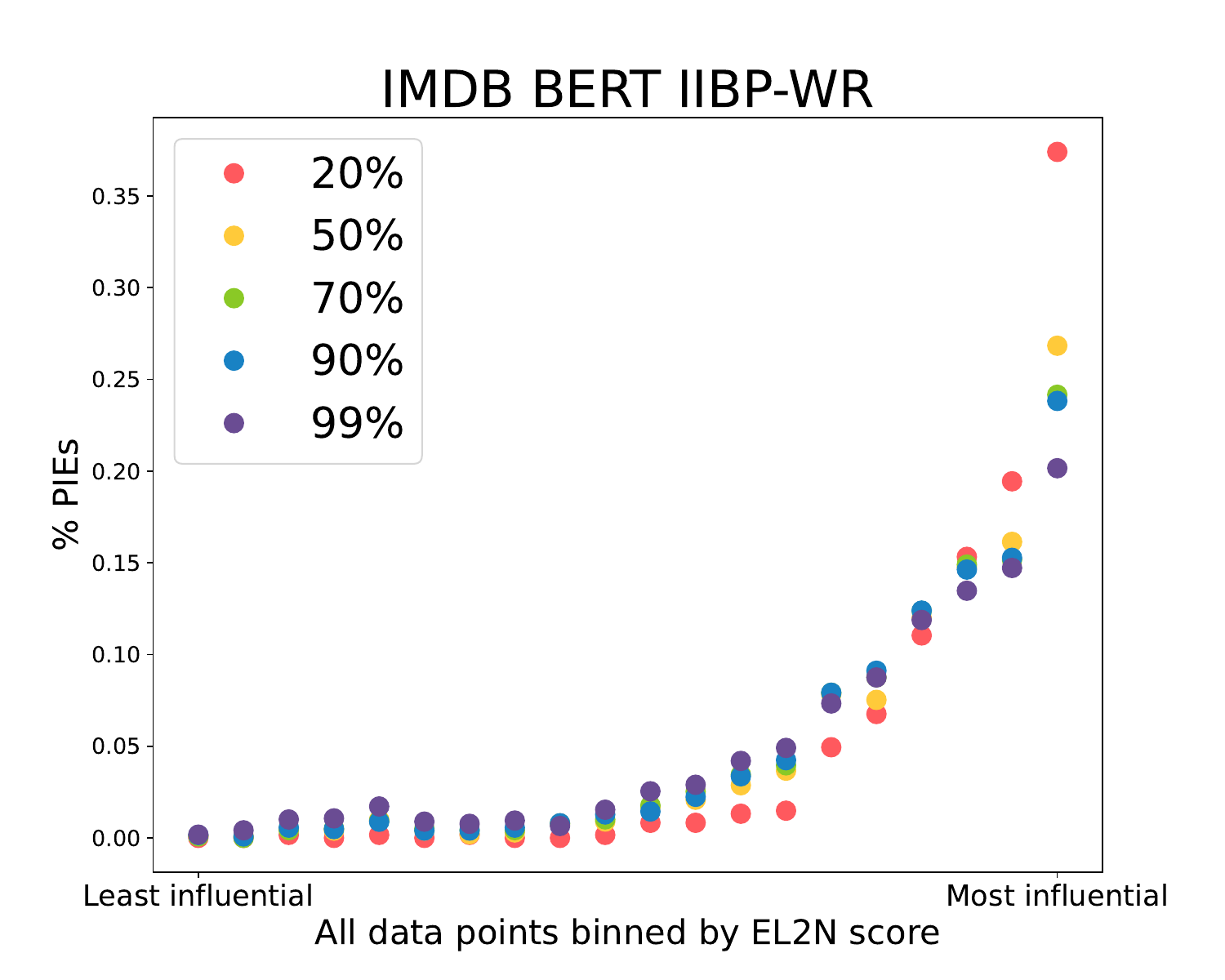} &
 \includegraphics[width=0.23\textwidth]{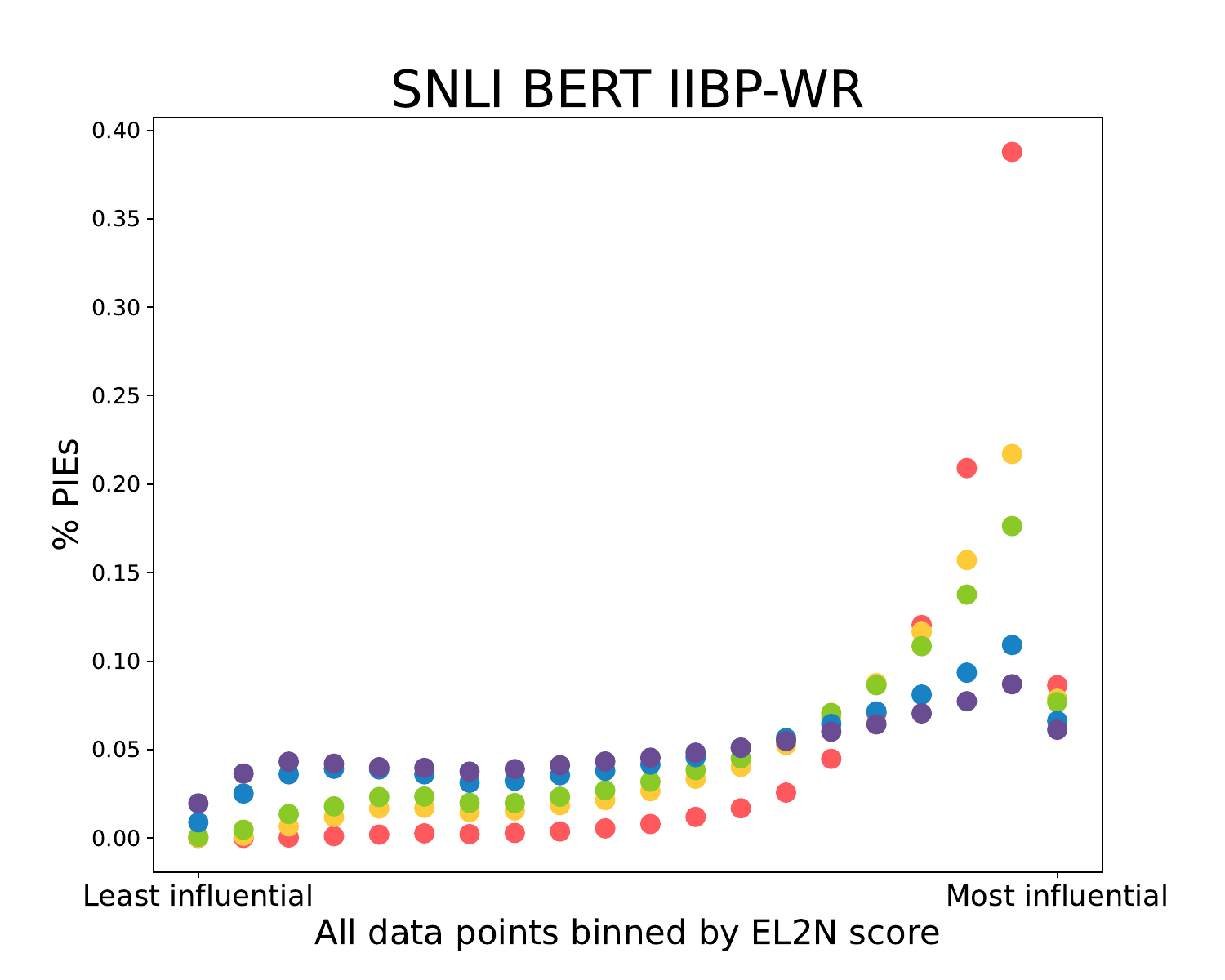} &
 \includegraphics[width=0.23\textwidth]{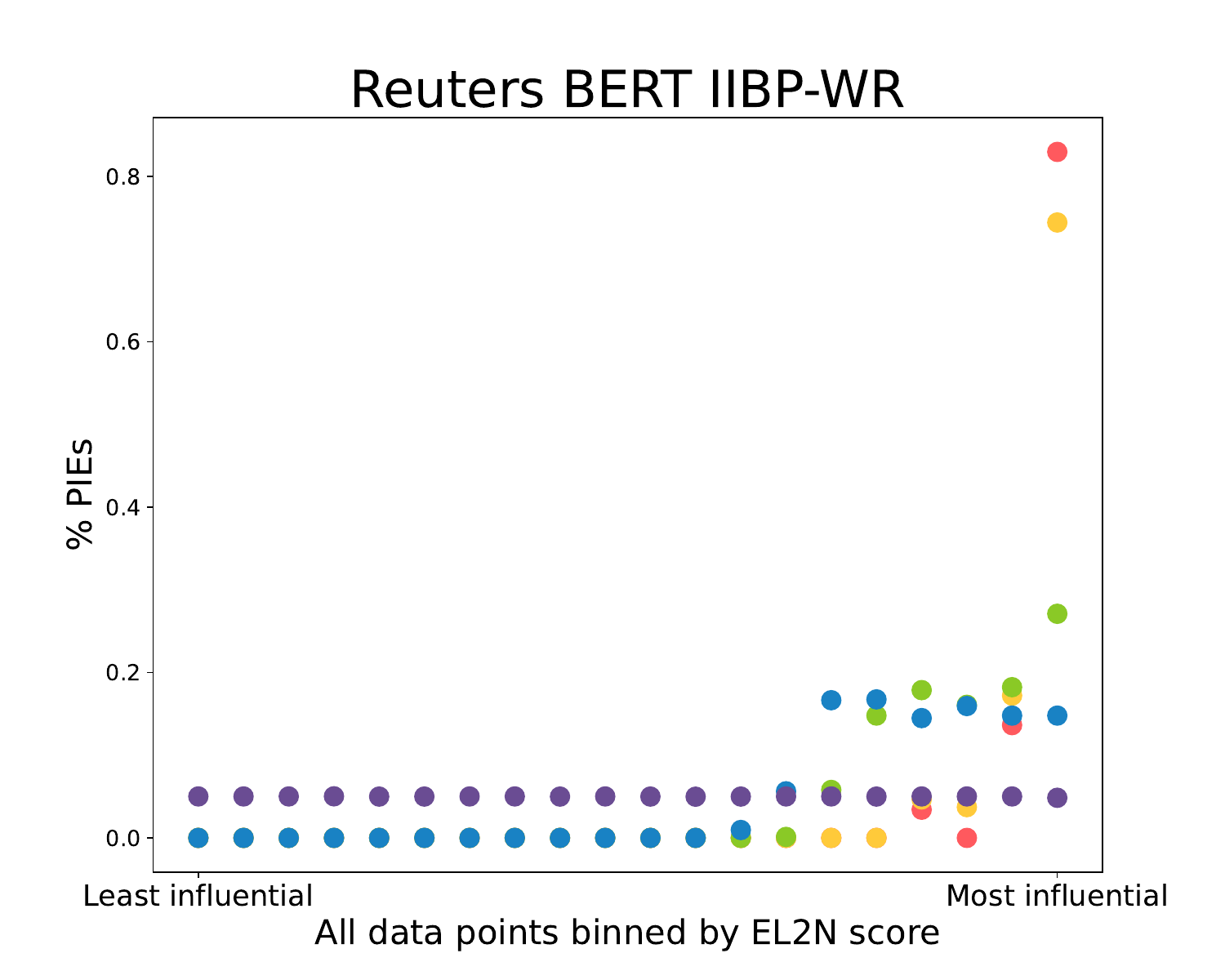} &
 \includegraphics[width=0.23\textwidth]{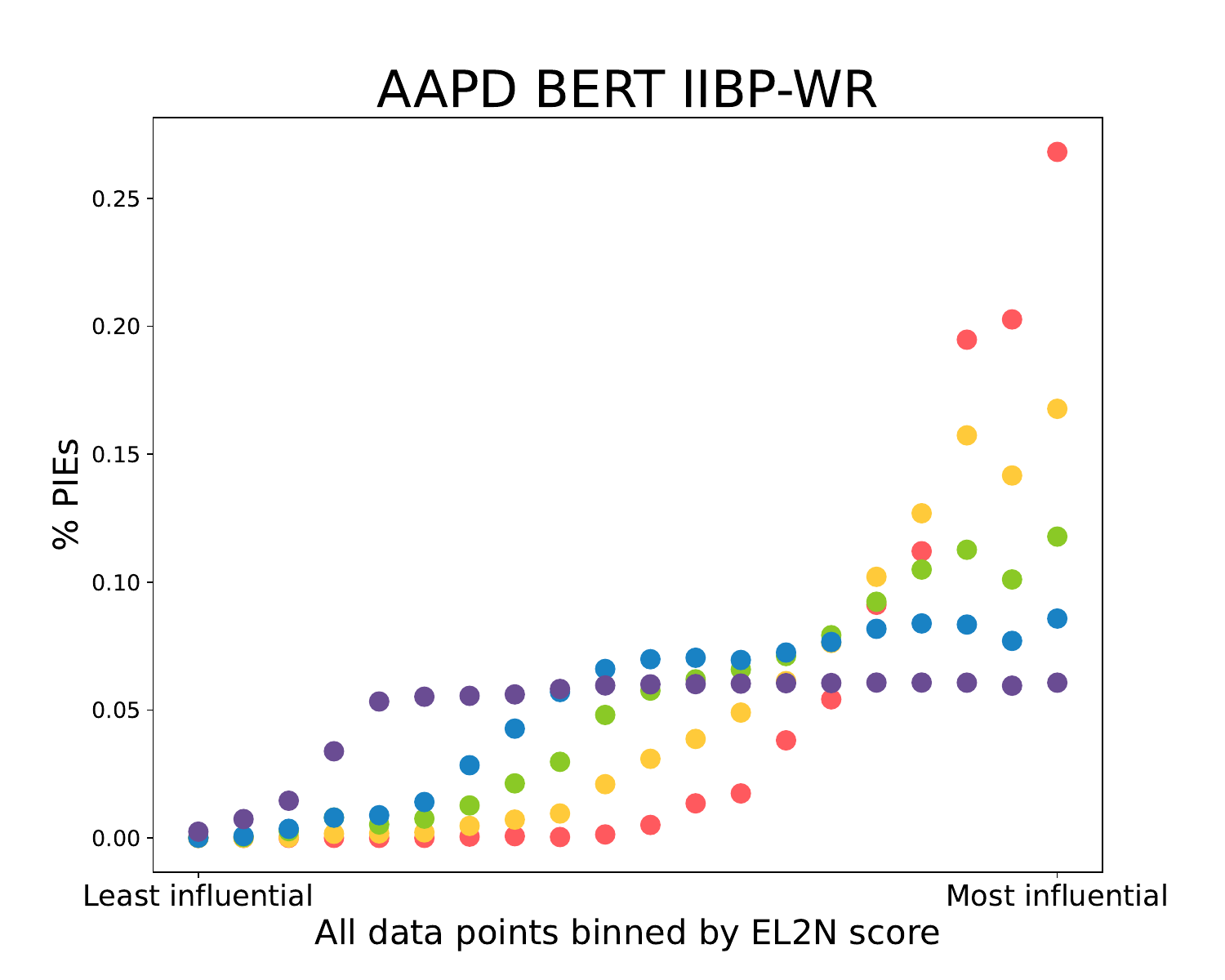} 
  \\

 \includegraphics[width=0.23\textwidth]{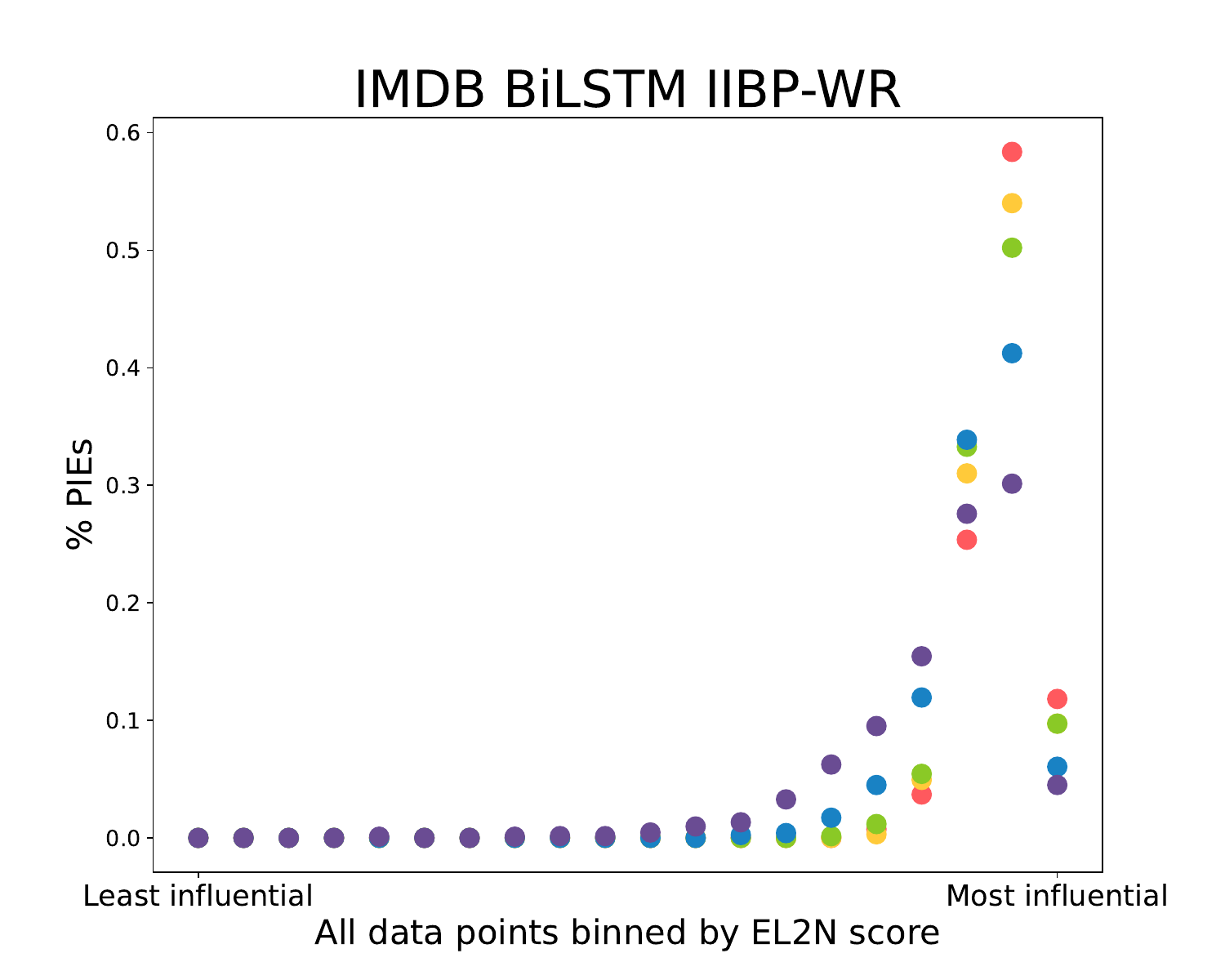} &
 \includegraphics[width=0.23\textwidth]{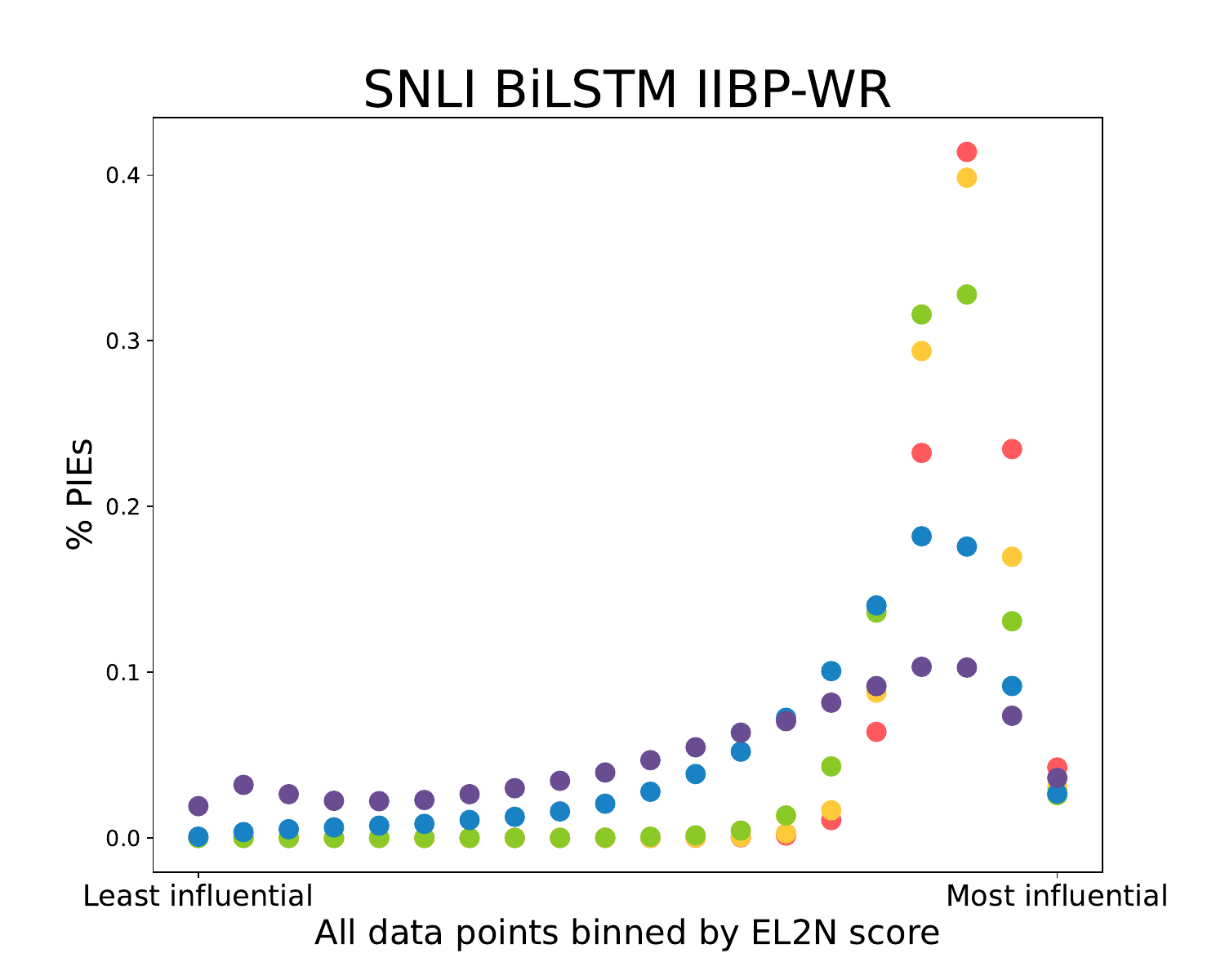} &
 \includegraphics[width=0.23\textwidth]{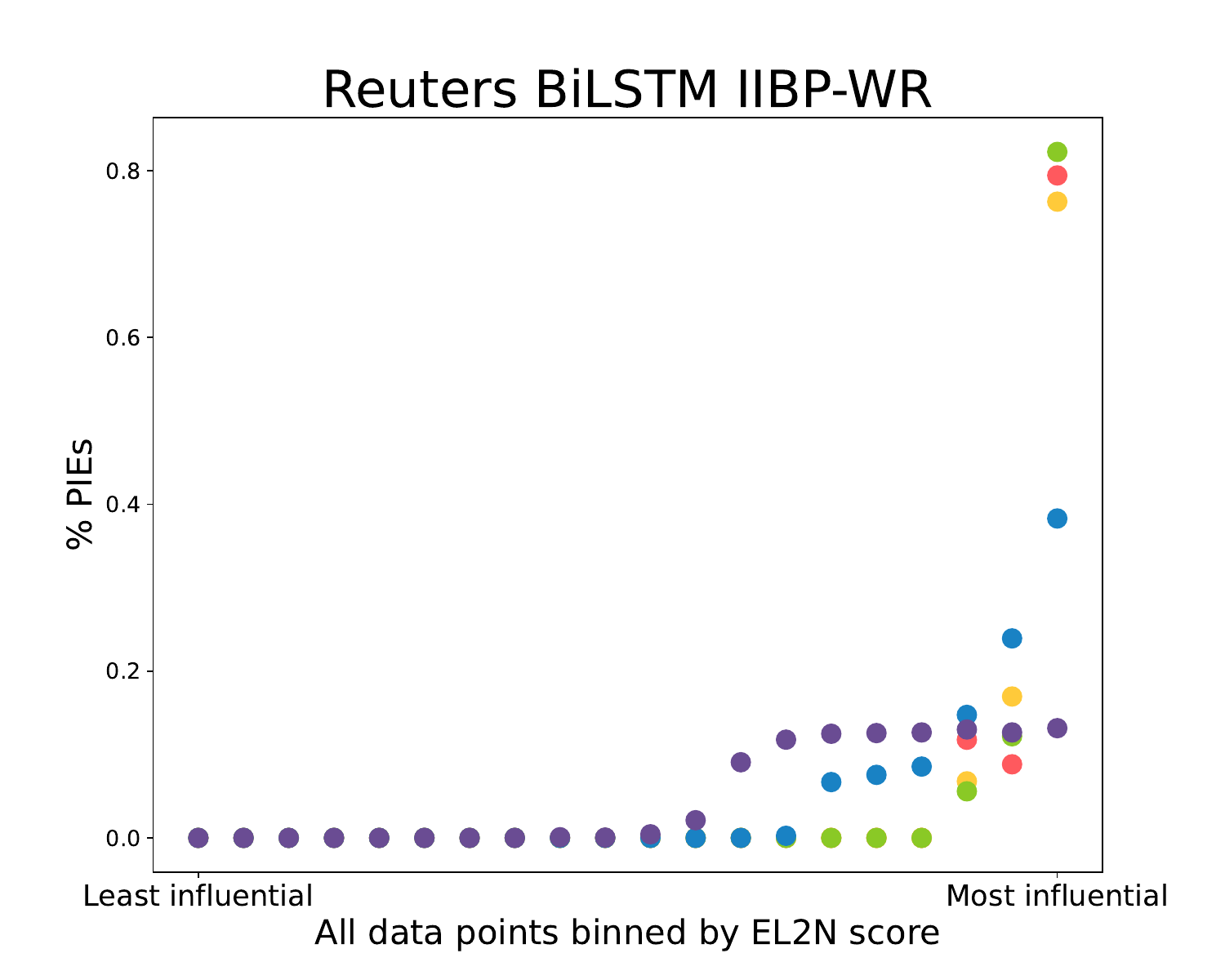} &
 \includegraphics[width=0.23\textwidth]{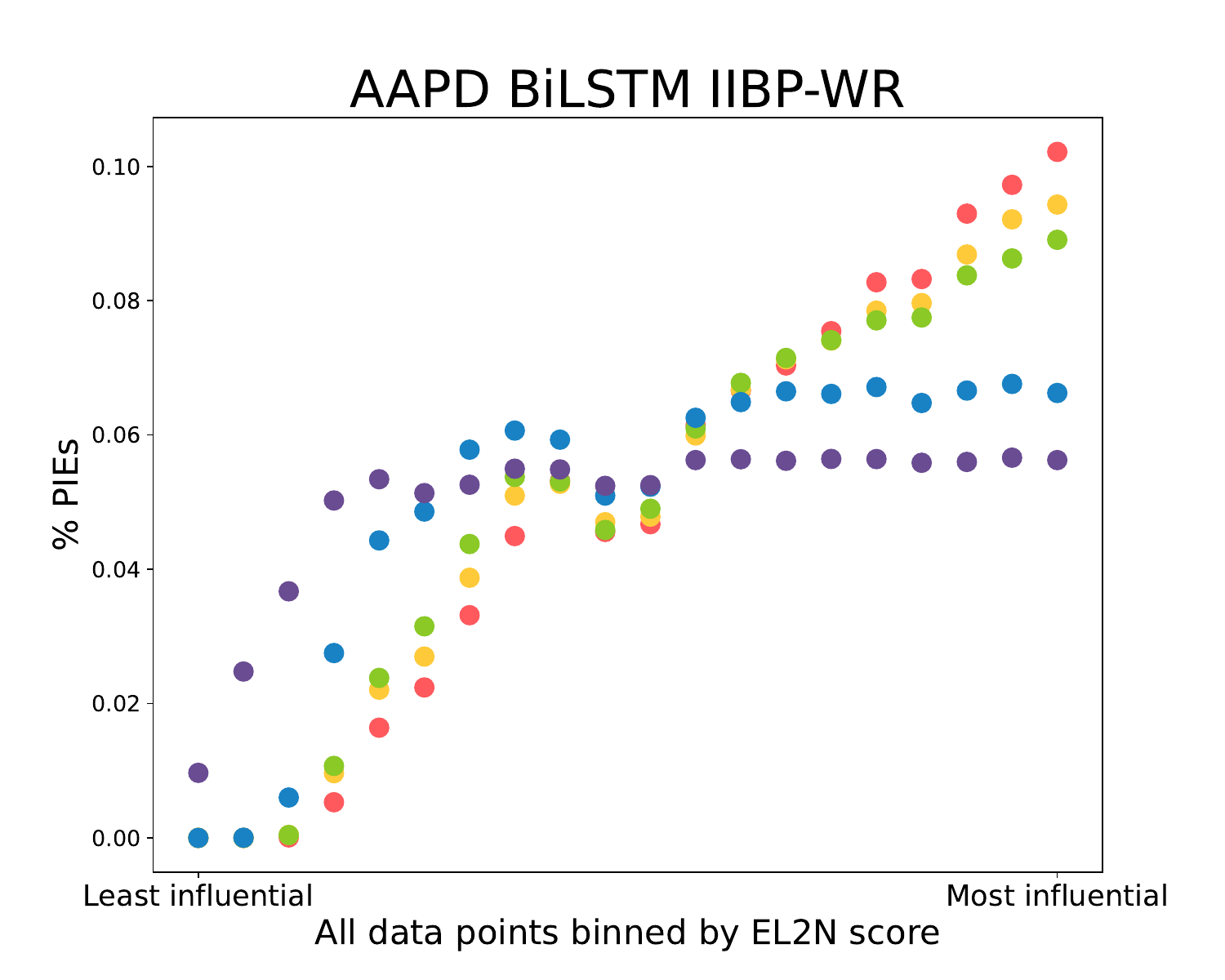} \\

\end{tabular}

\caption{
Percentage of data points that are PIEs (y axis) versus degree of influence (EL2N score) of all data points in the training set (x axis) for IIBP-WR at 20\% and 99\% pruning.
}
\label{fig:IMDB_BERT_anal4_grid_IIBP-WR}%
\end{figure*}

\begin{figure*}
\setlength\tabcolsep{-1pt}%
\centering
\begin{tabular}{cccc}

 \includegraphics[width=0.23\textwidth]{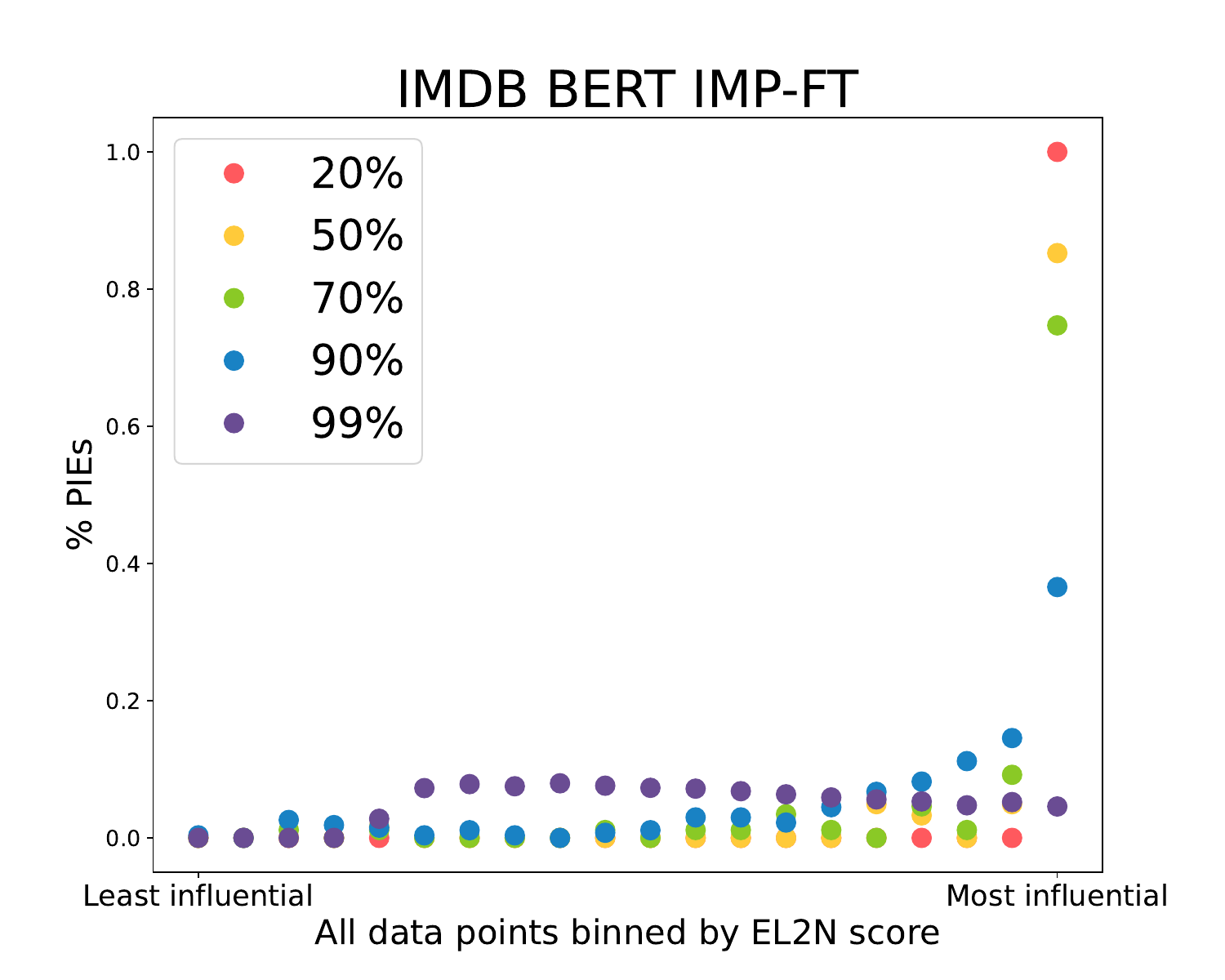} &
 \includegraphics[width=0.23\textwidth]{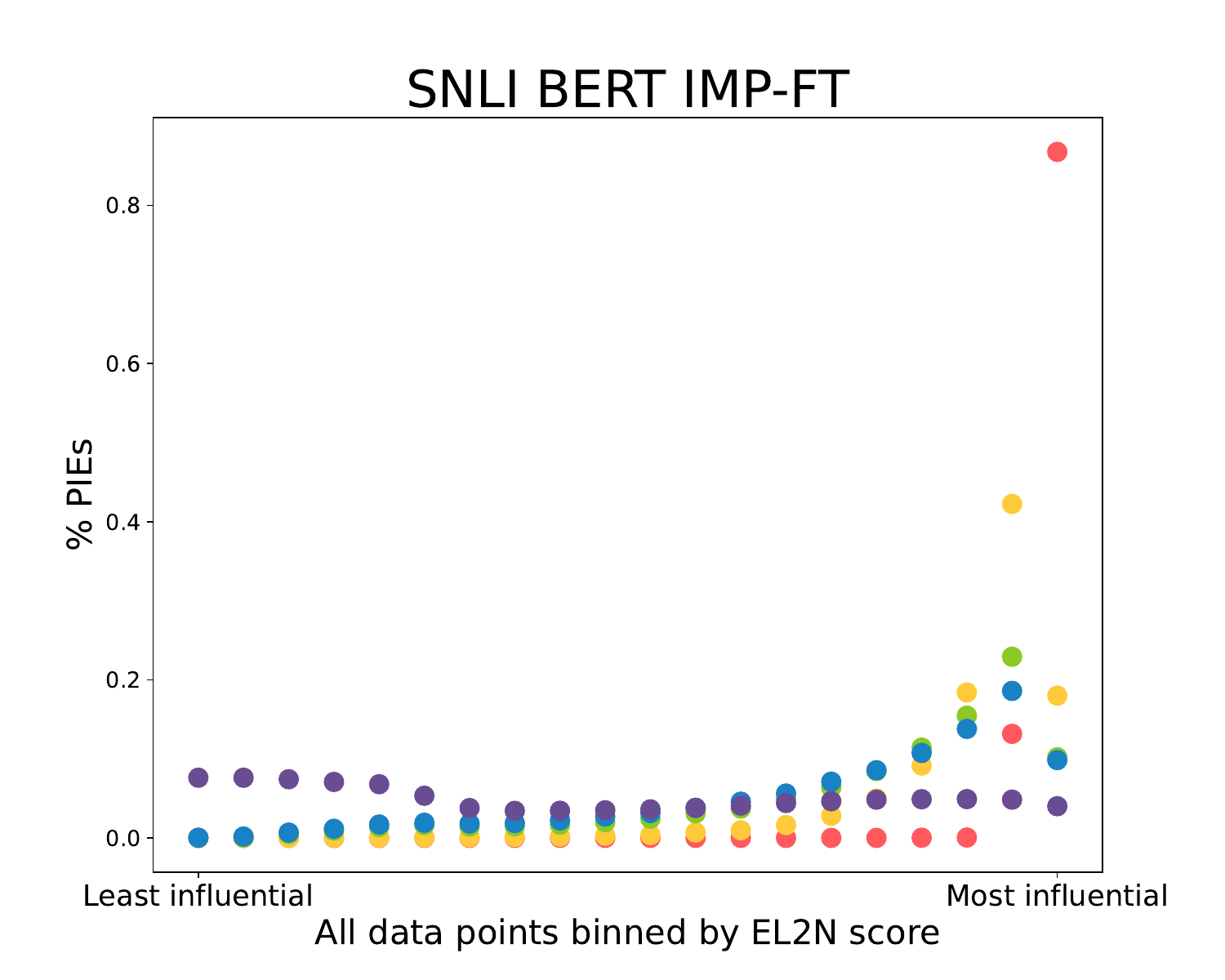} &
 \includegraphics[width=0.23\textwidth]{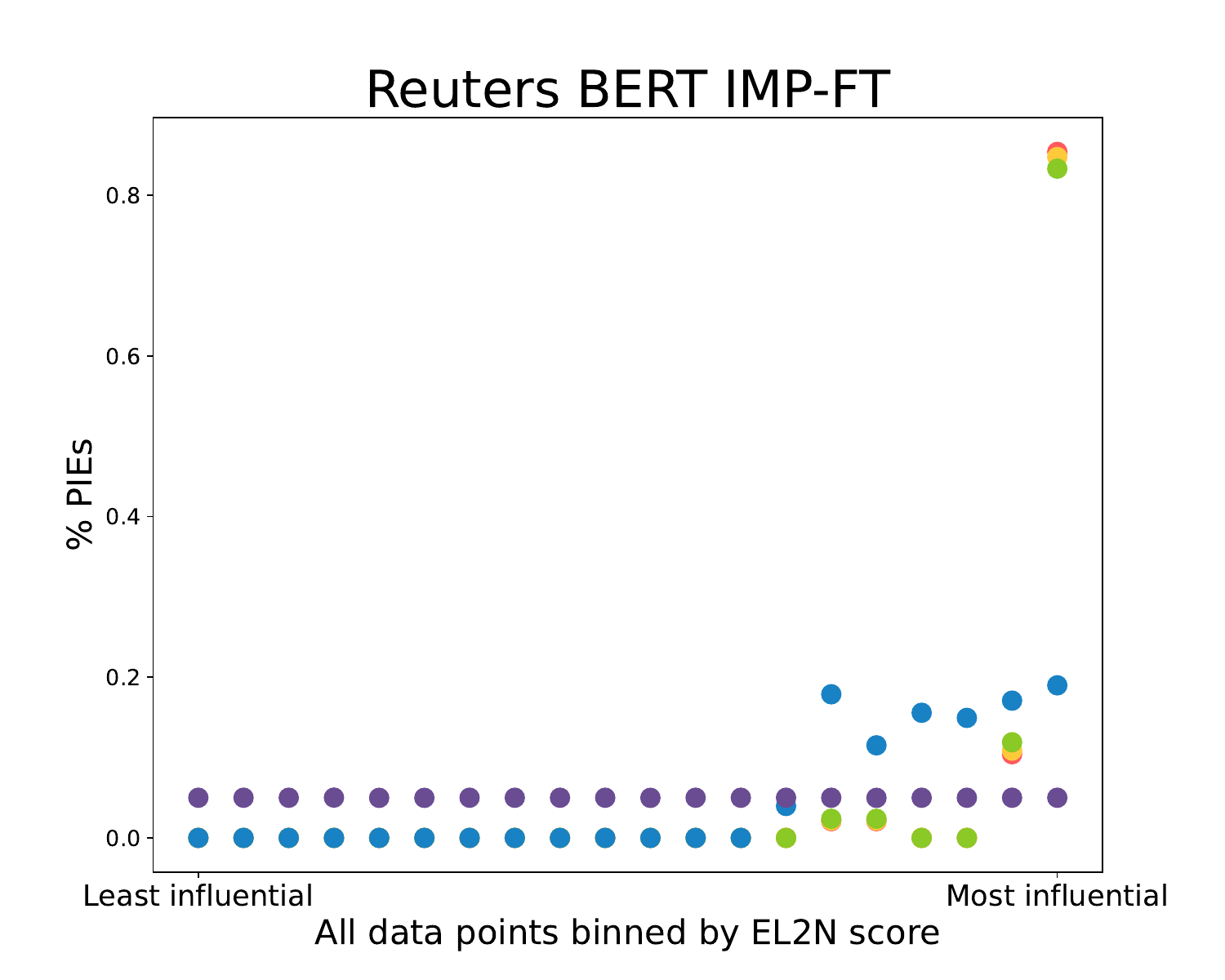} &
 \includegraphics[width=0.23\textwidth]{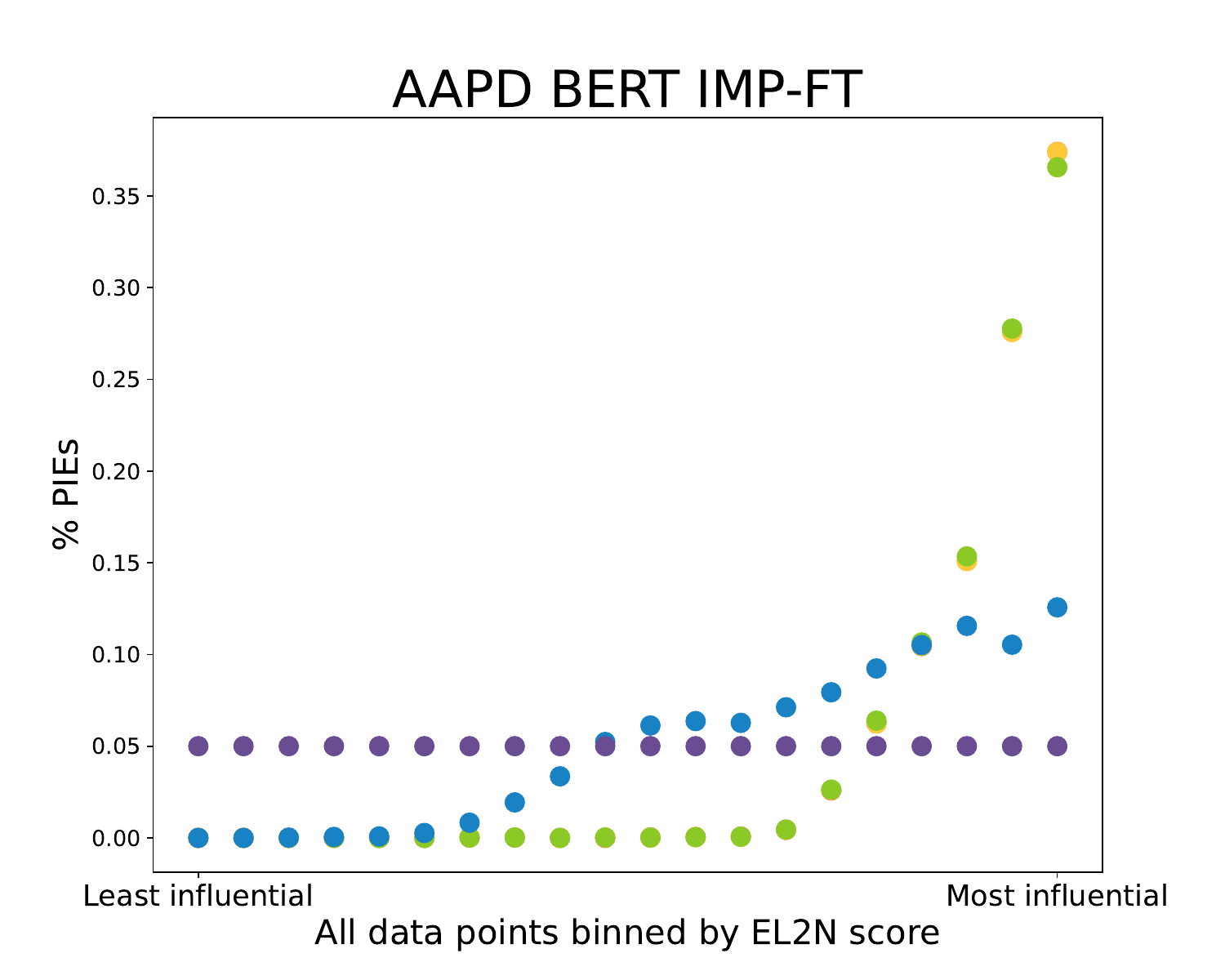} 
  \\

 \includegraphics[width=0.23\textwidth]{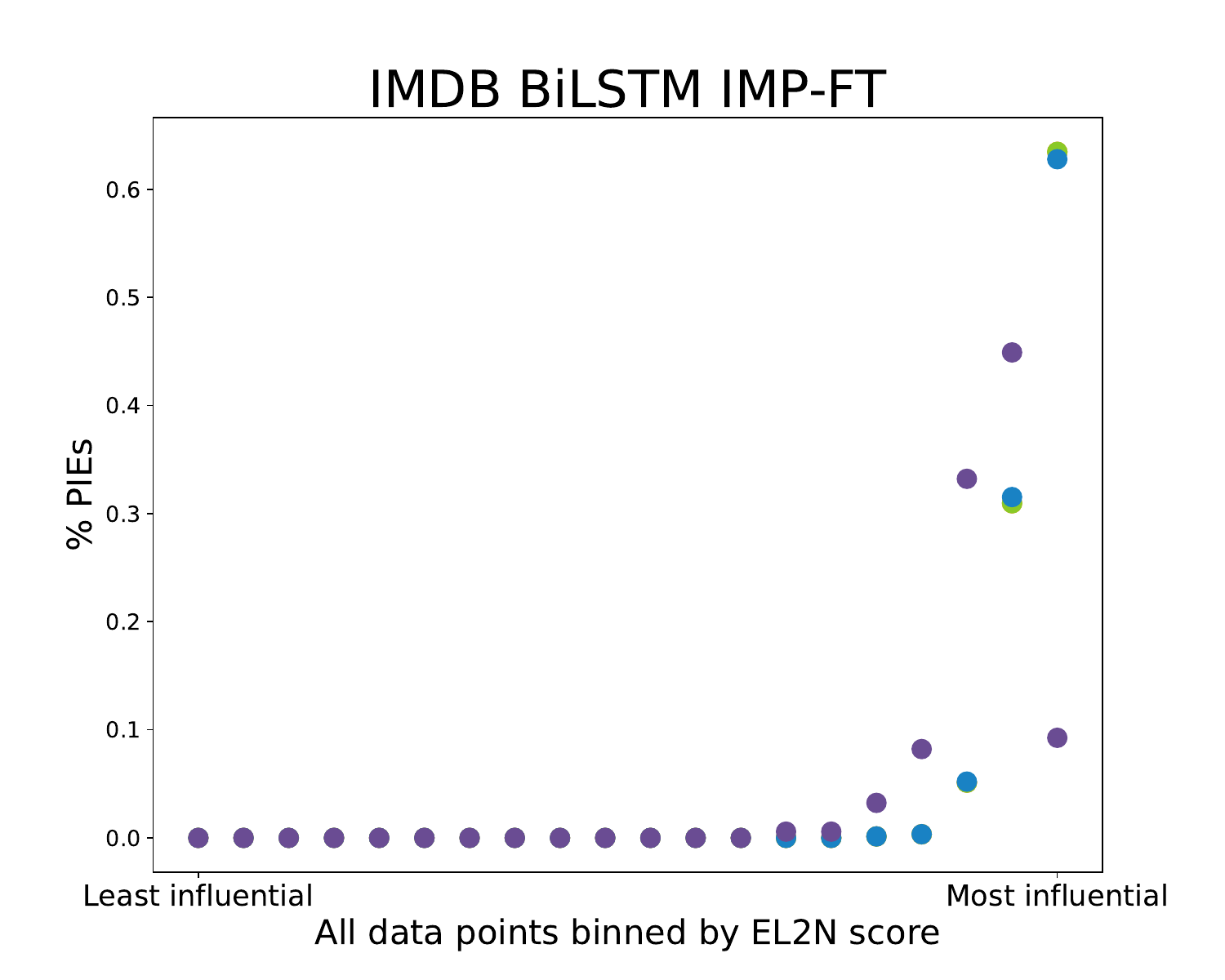} &
 \includegraphics[width=0.23\textwidth]{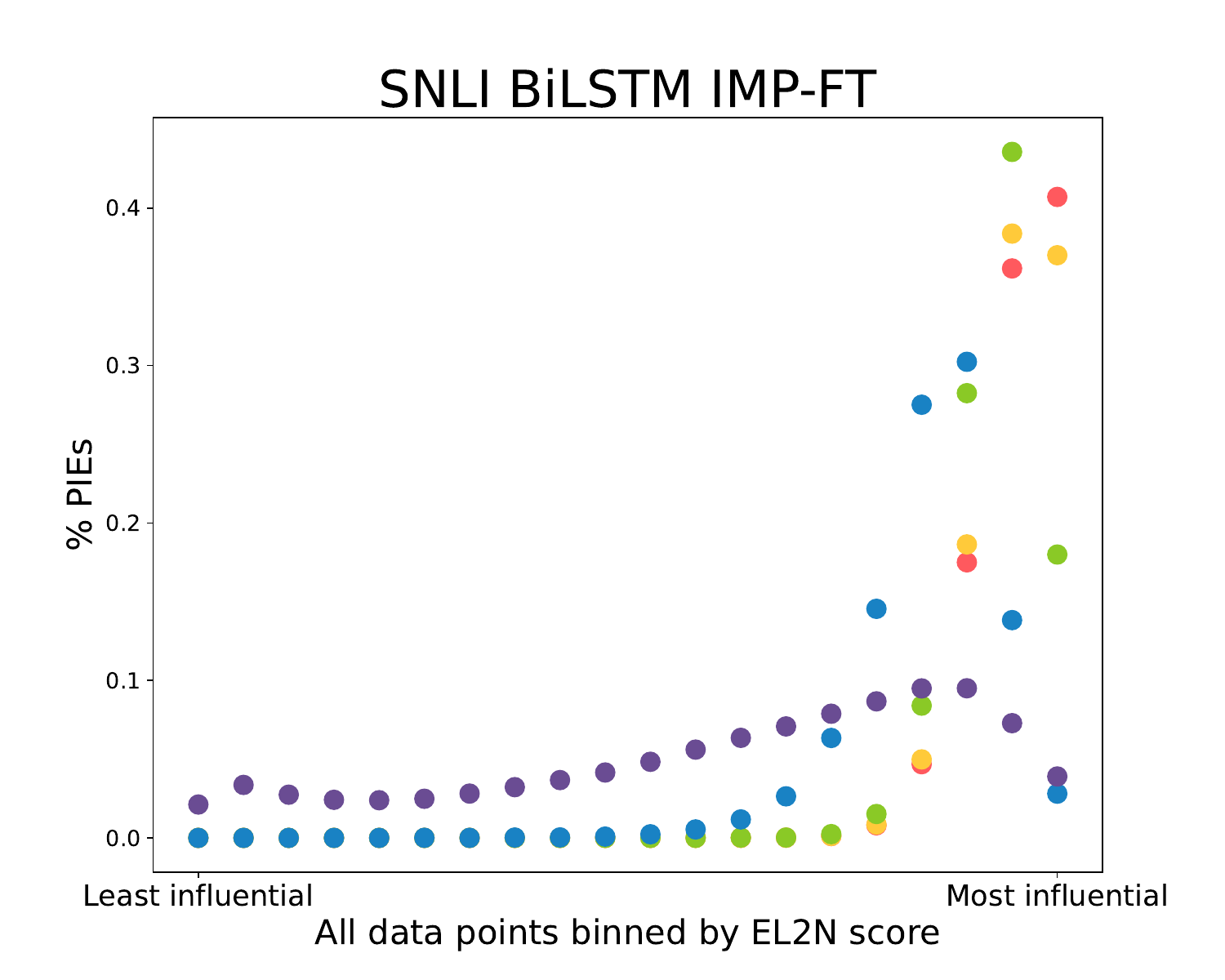} &
 \includegraphics[width=0.23\textwidth]{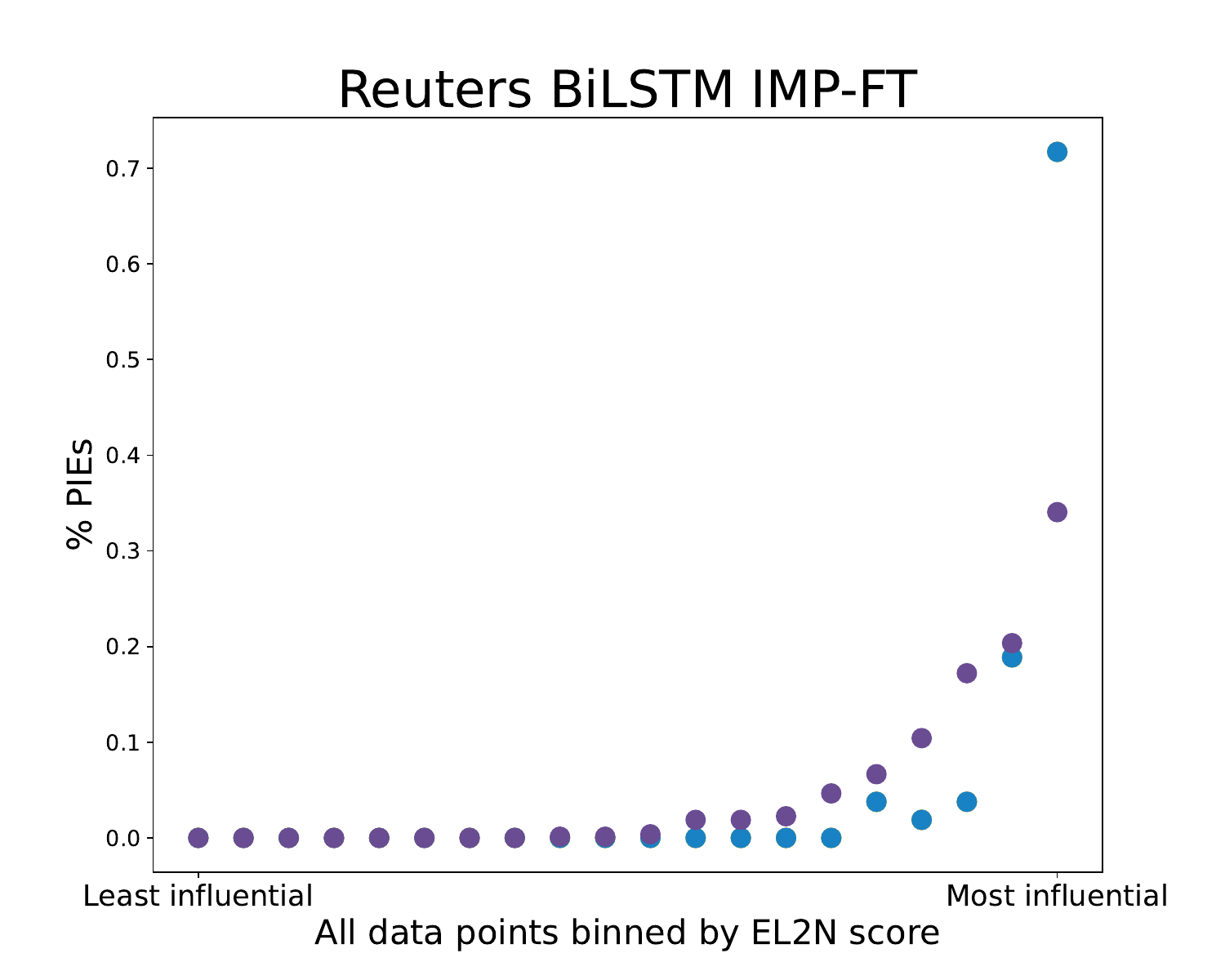} &
 \includegraphics[width=0.23\textwidth]{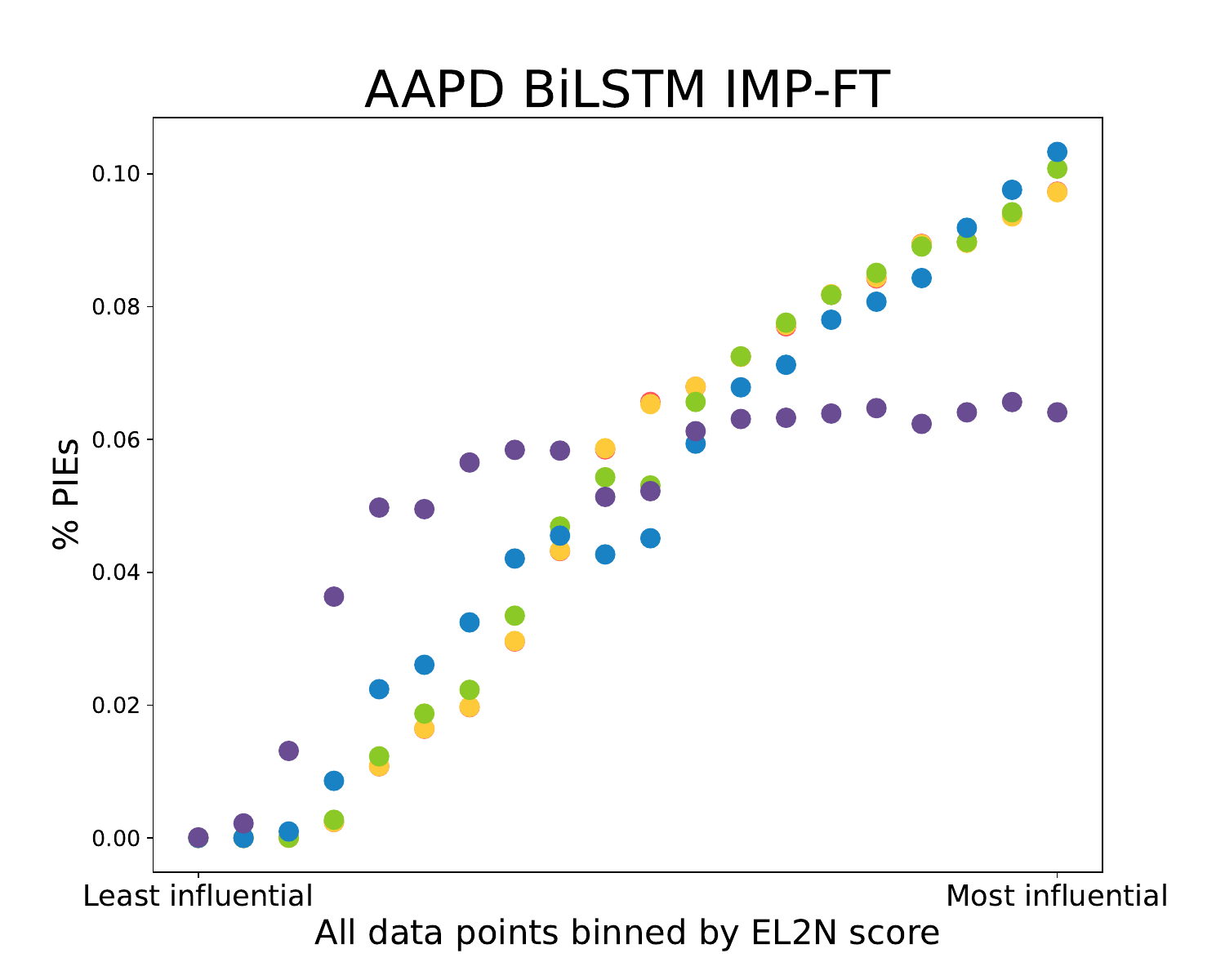} \\

\end{tabular}
\caption{
Percentage of data points that are PIEs (y axis) versus degree of influence (EL2N score) of all data points in the training set (x axis) for IMP-FT. 
}
\label{fig:IMDB_BERT_anal4_grid_IMM}%
\end{figure*}

\begin{figure*}
\setlength\tabcolsep{-1pt}
\centering
\begin{tabular}{cccc}

 \includegraphics[width=0.23\textwidth]{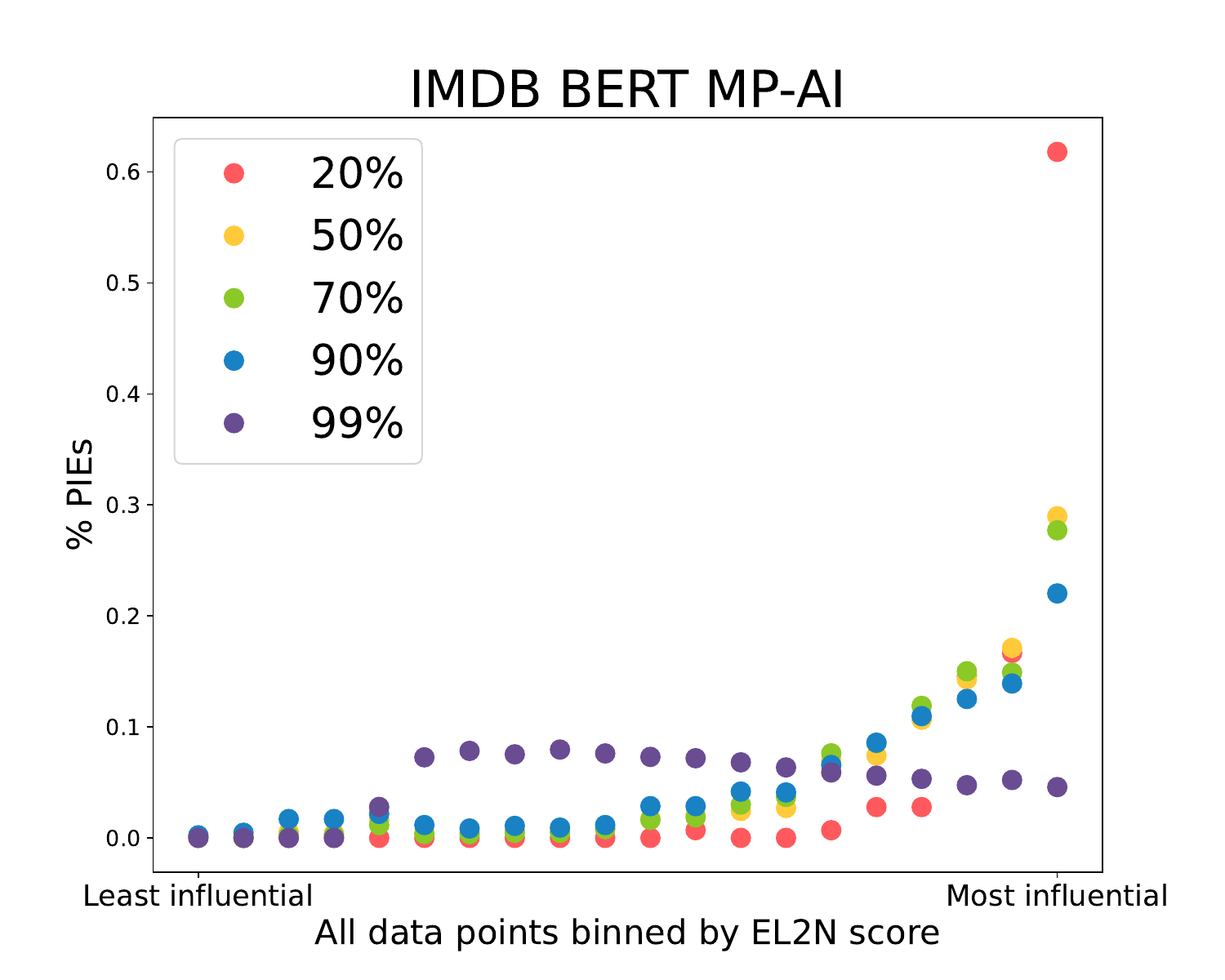} &
 \includegraphics[width=0.23\textwidth]{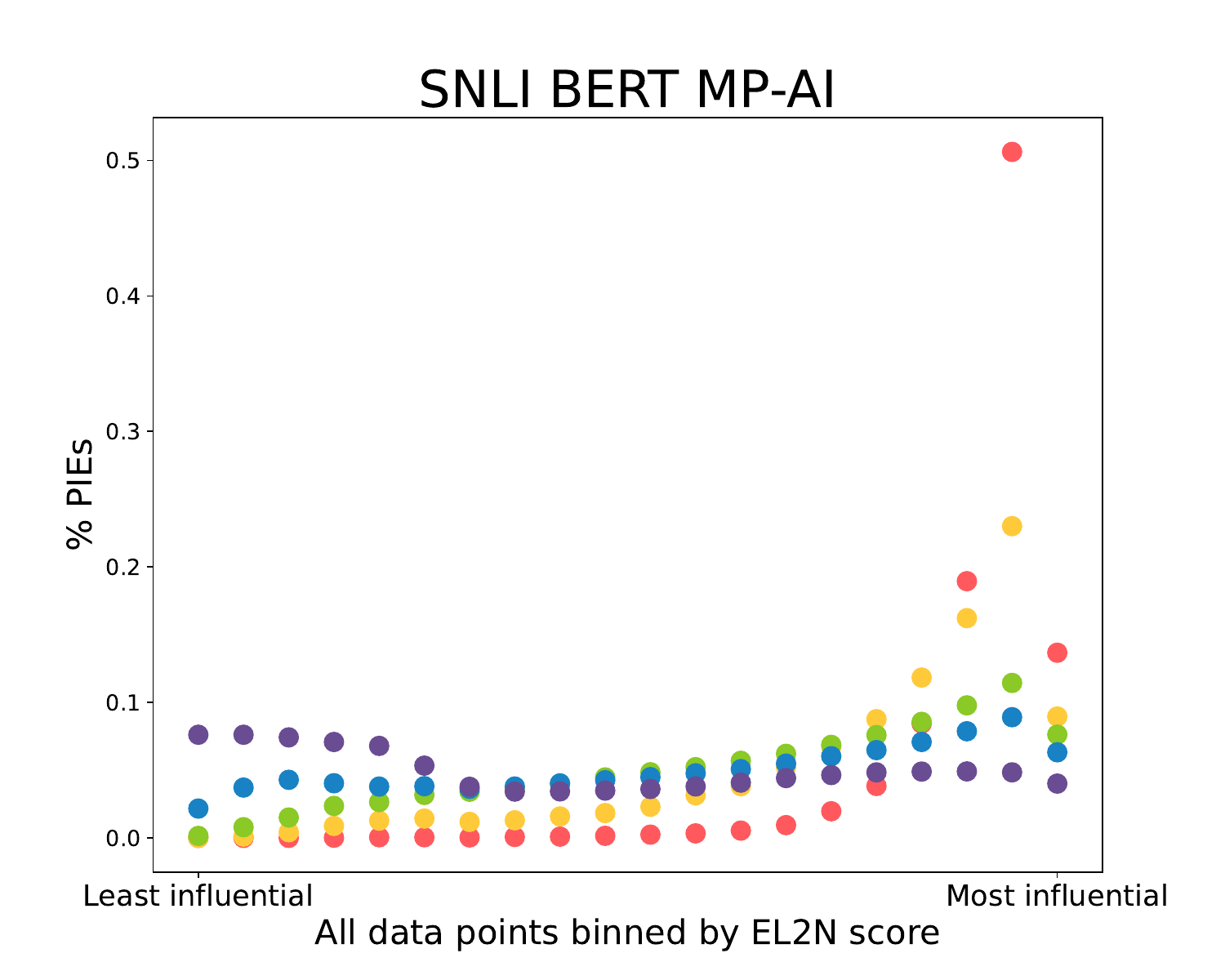} &
 \includegraphics[width=0.23\textwidth]{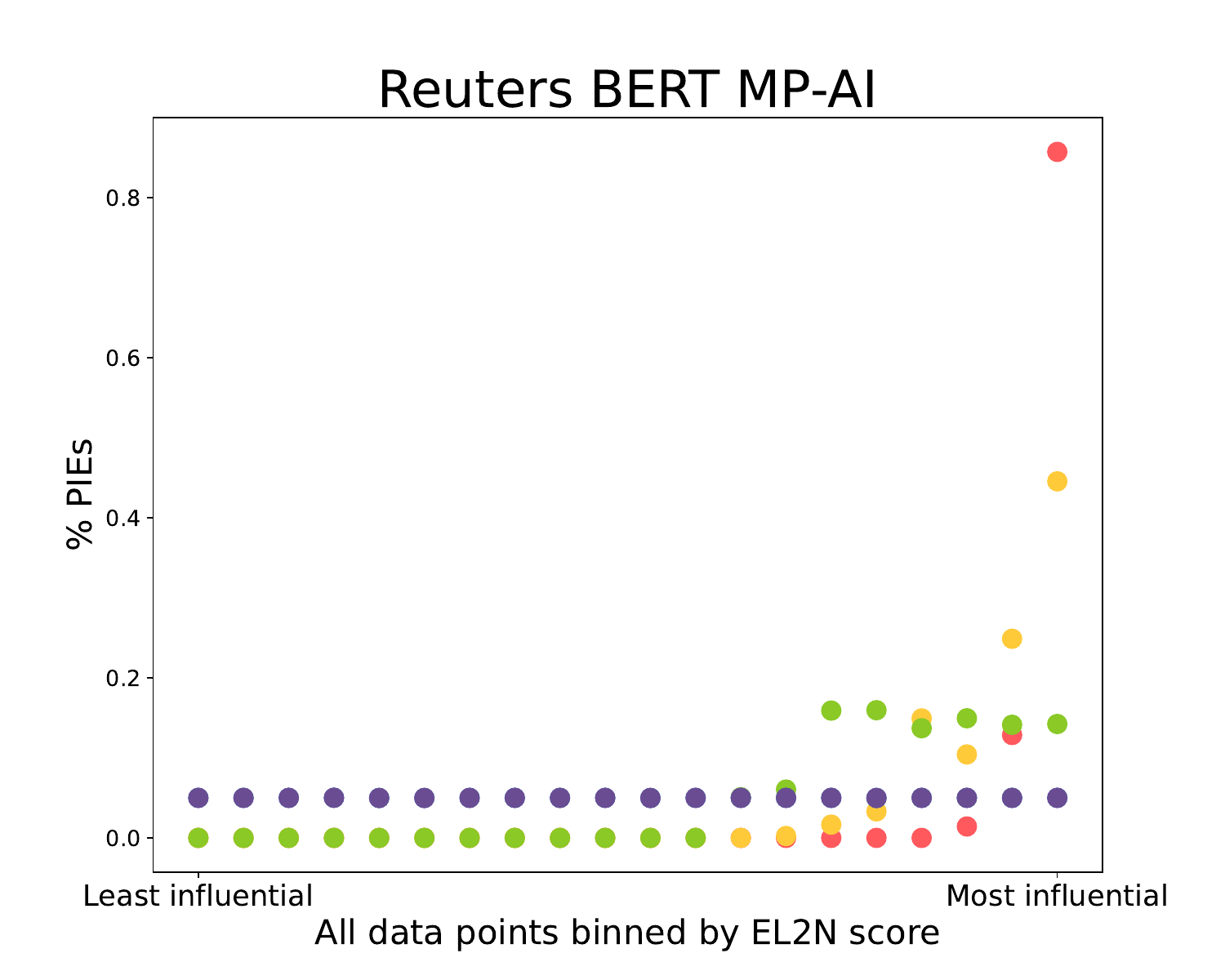} &
 \includegraphics[width=0.23\textwidth]{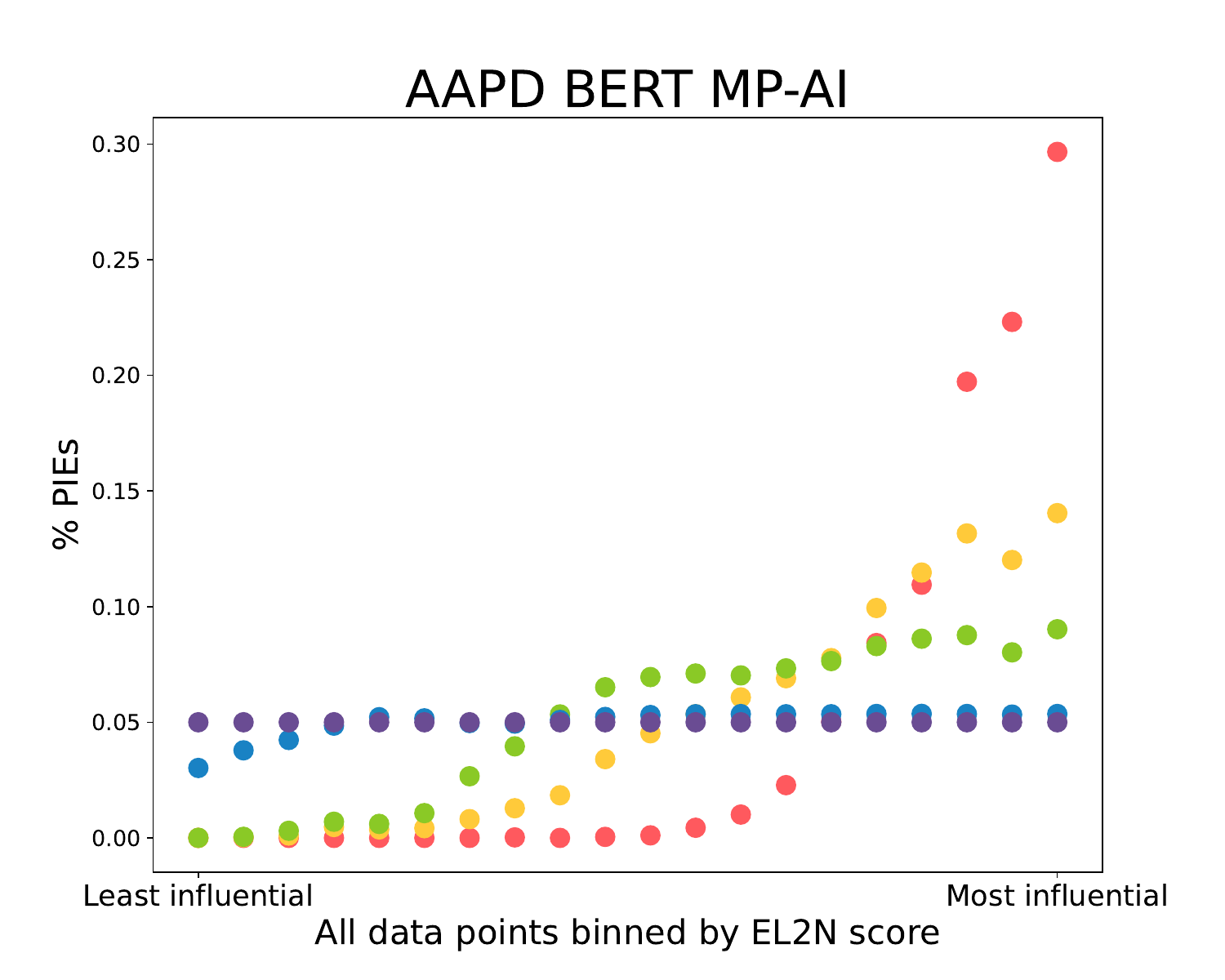} 
  \\

 \includegraphics[width=0.23\textwidth]{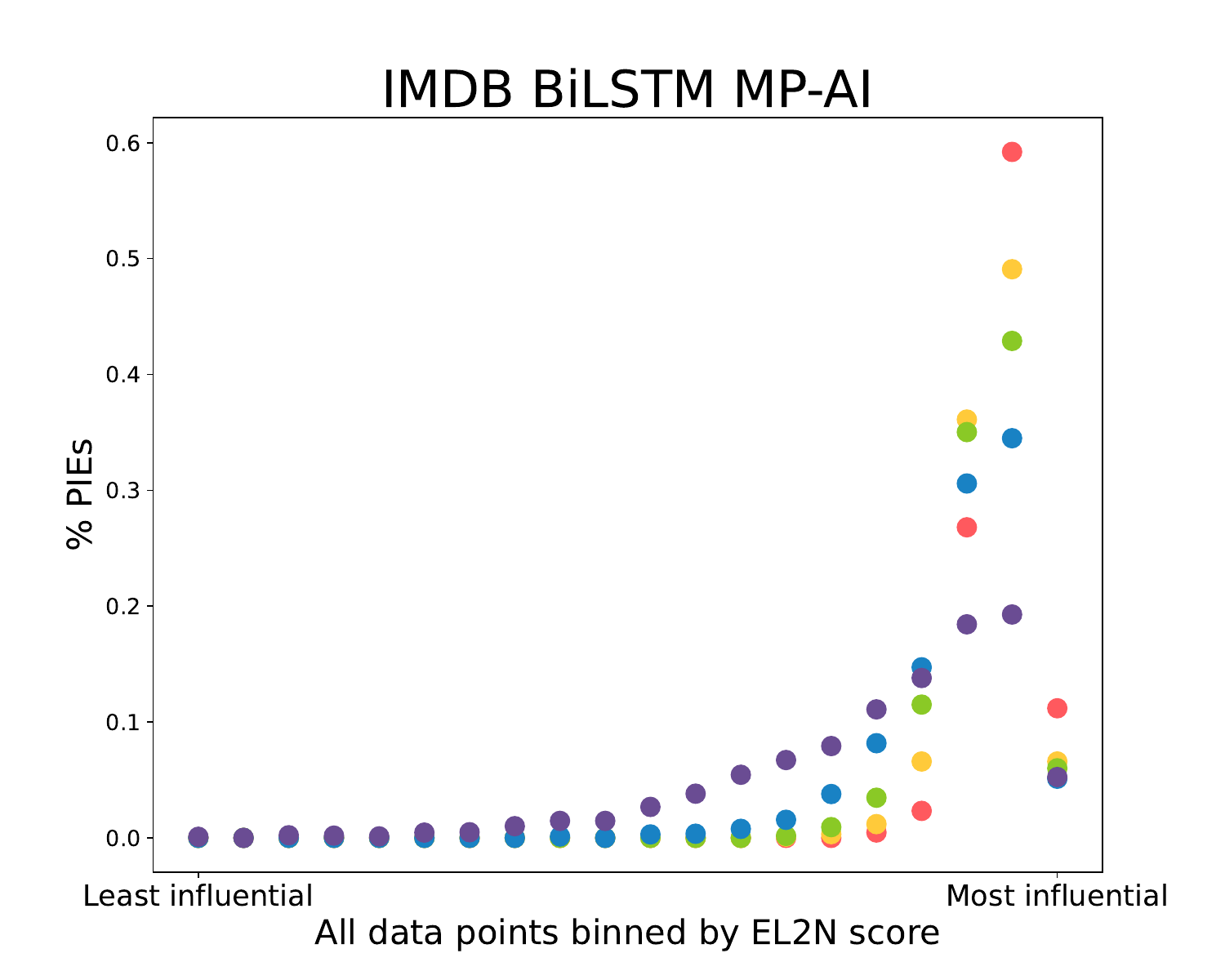} &
 \includegraphics[width=0.23\textwidth]{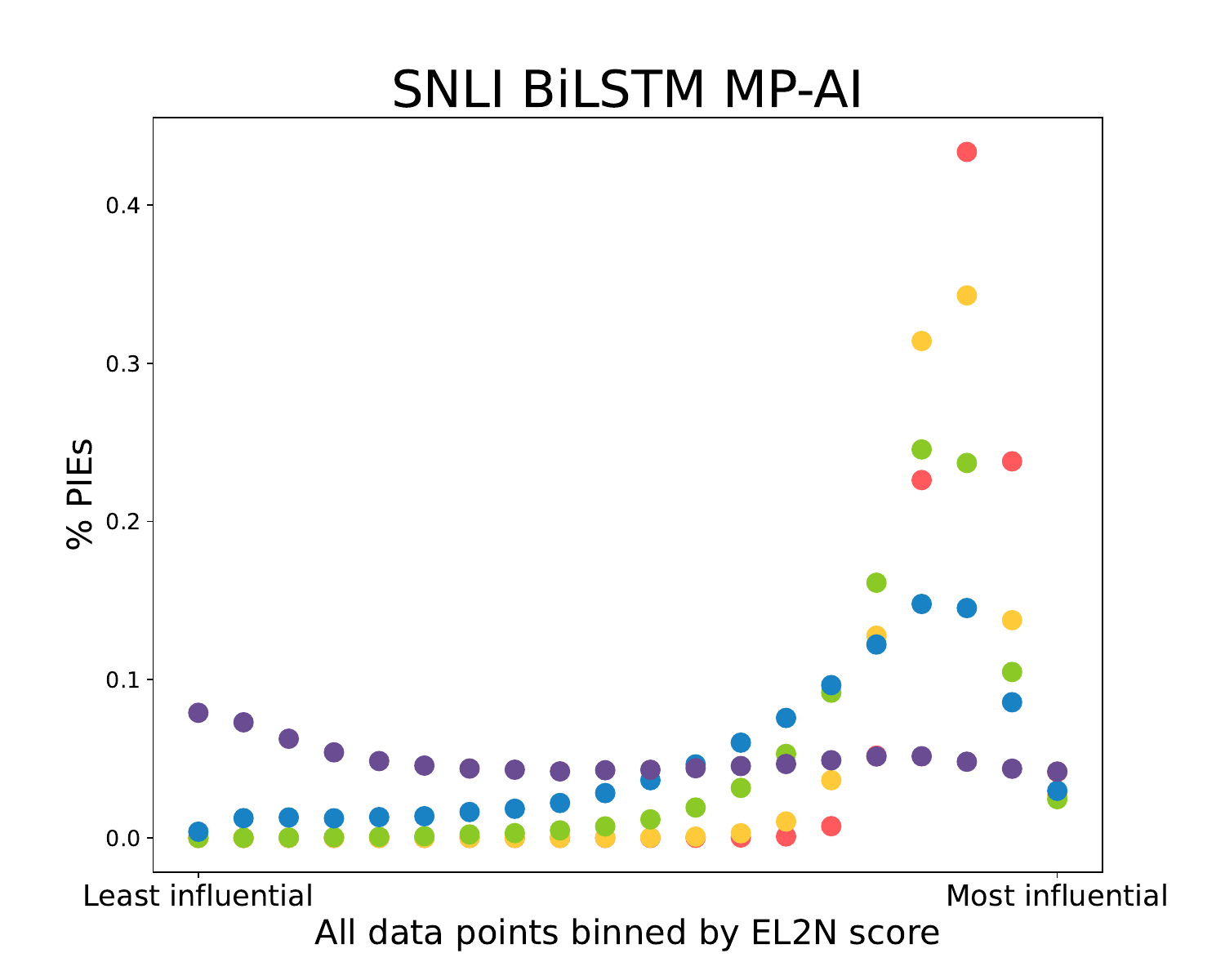} &
 \includegraphics[width=0.23\textwidth]{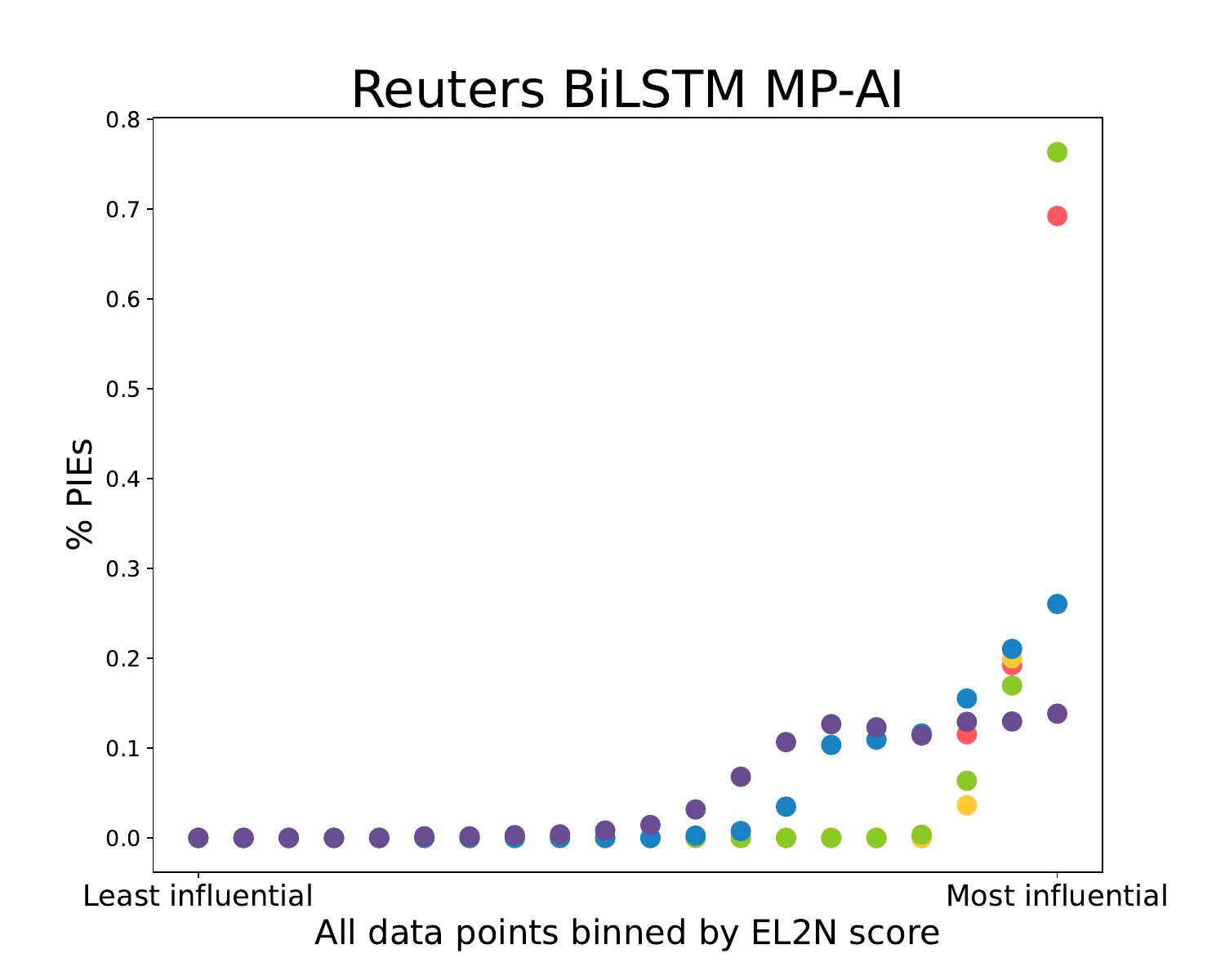} &
 \includegraphics[width=0.23\textwidth]{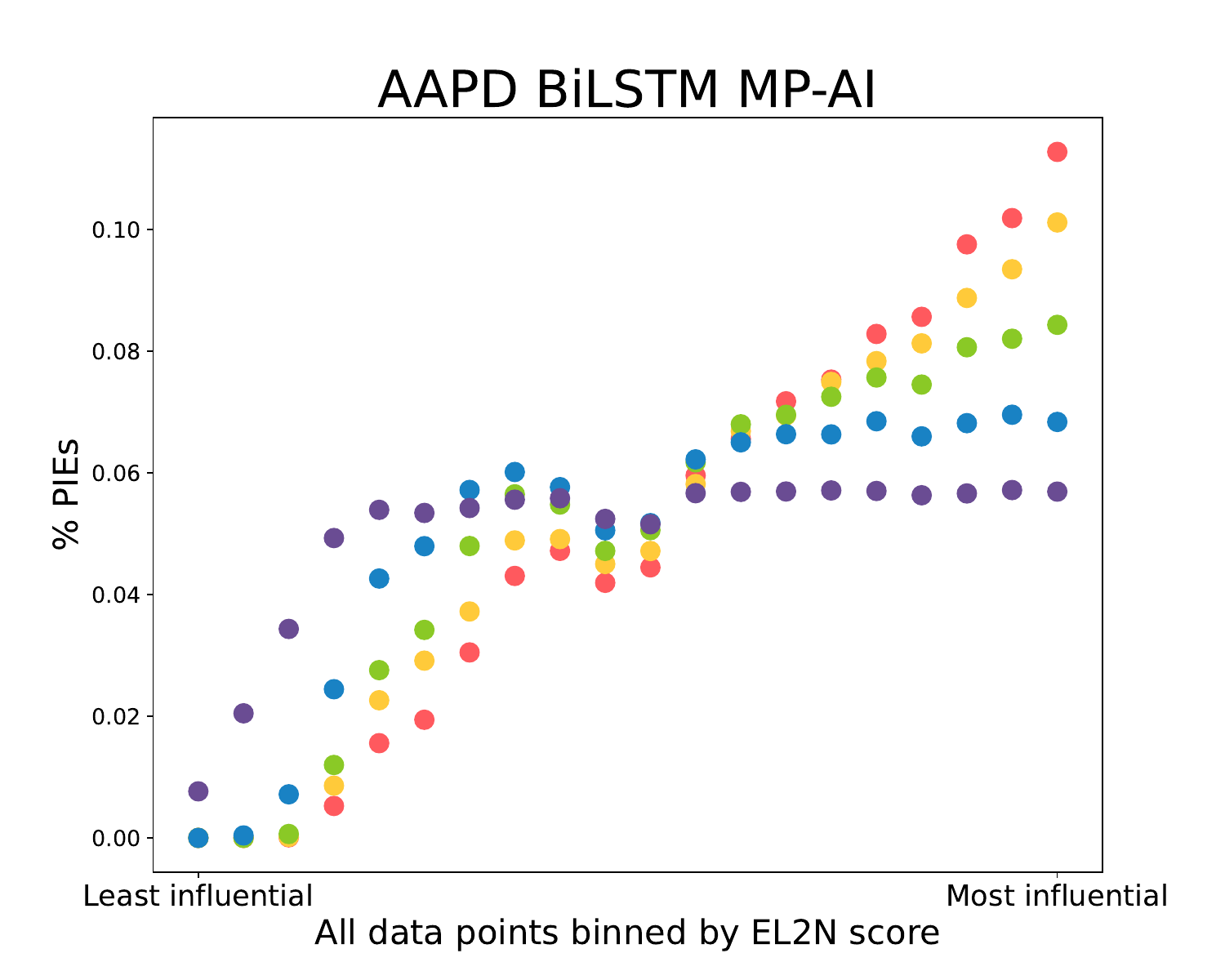} \\

\end{tabular}
\caption{
Percentage of data points that are PIEs (y axis) versus degree of influence (EL2N score) of all data points in the training set (x axis) for IMP-AI.
}
\label{fig:IMDB_BERT_anal4_grid_IMP-AI}%
\end{figure*}

\begin{figure*}
\setlength\tabcolsep{-1pt}
\centering
\begin{tabular}{cccc}

 \includegraphics[width=0.23\textwidth]{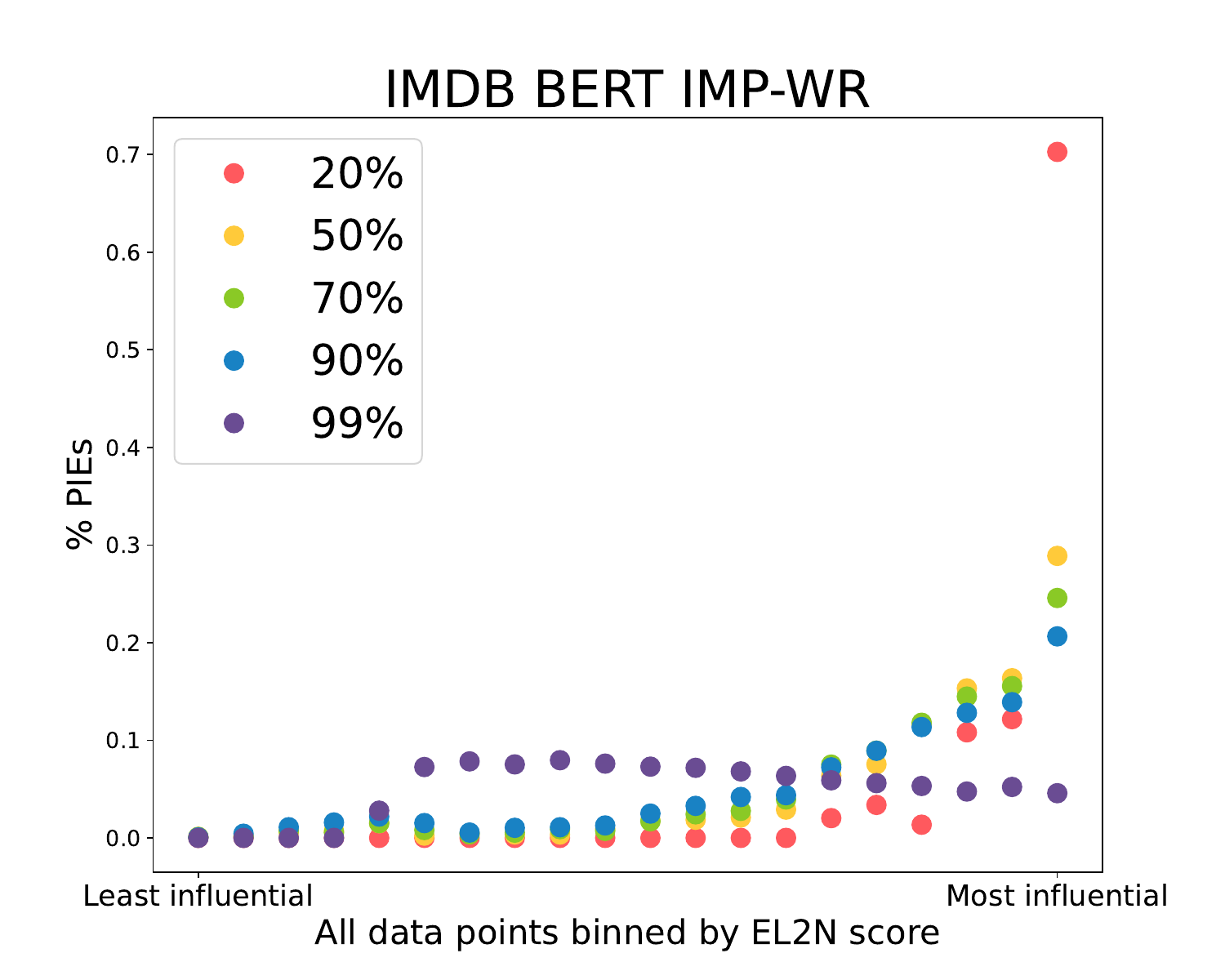} &
 \includegraphics[width=0.23\textwidth]{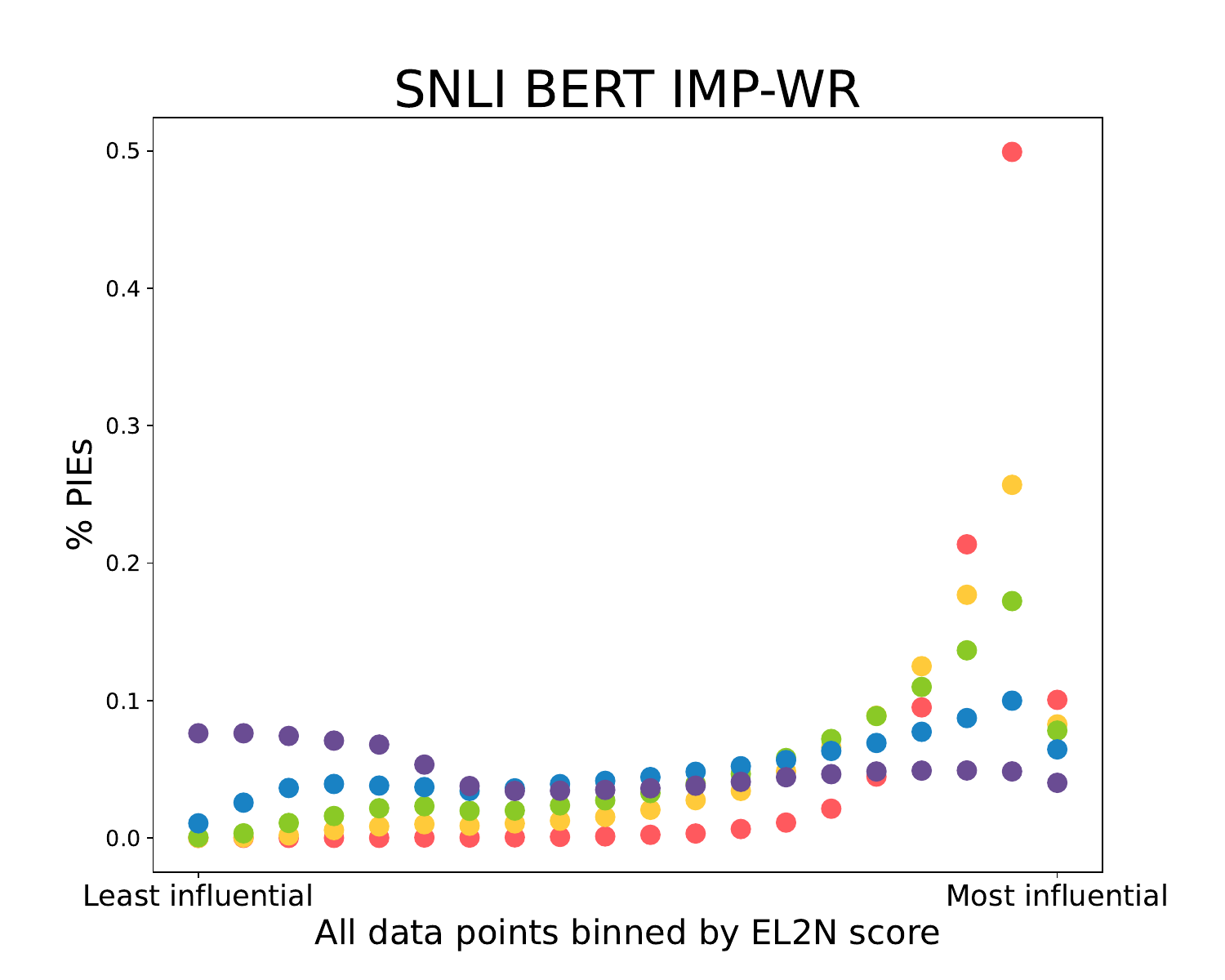} &
 \includegraphics[width=0.23\textwidth]{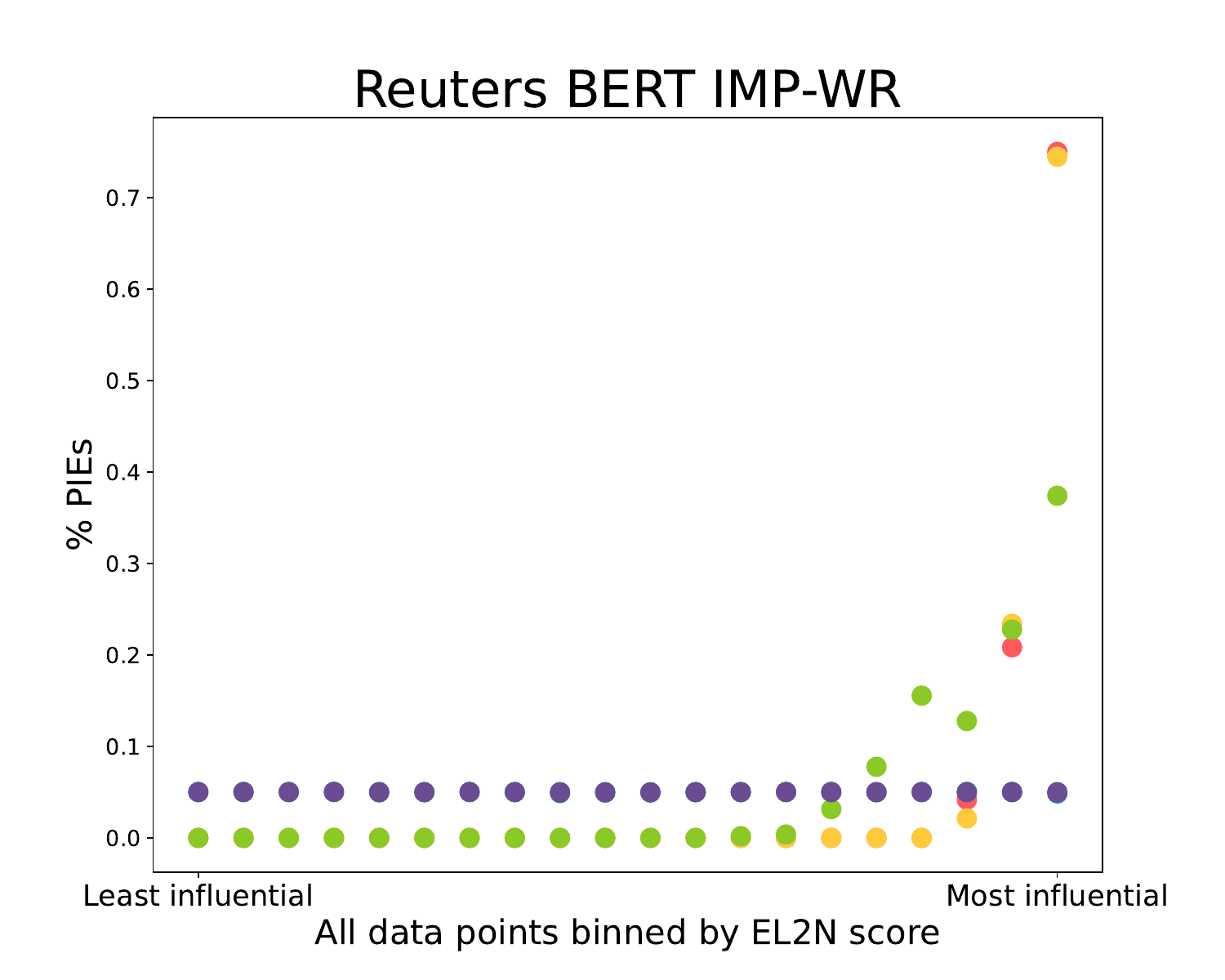} &
 \includegraphics[width=0.23\textwidth]{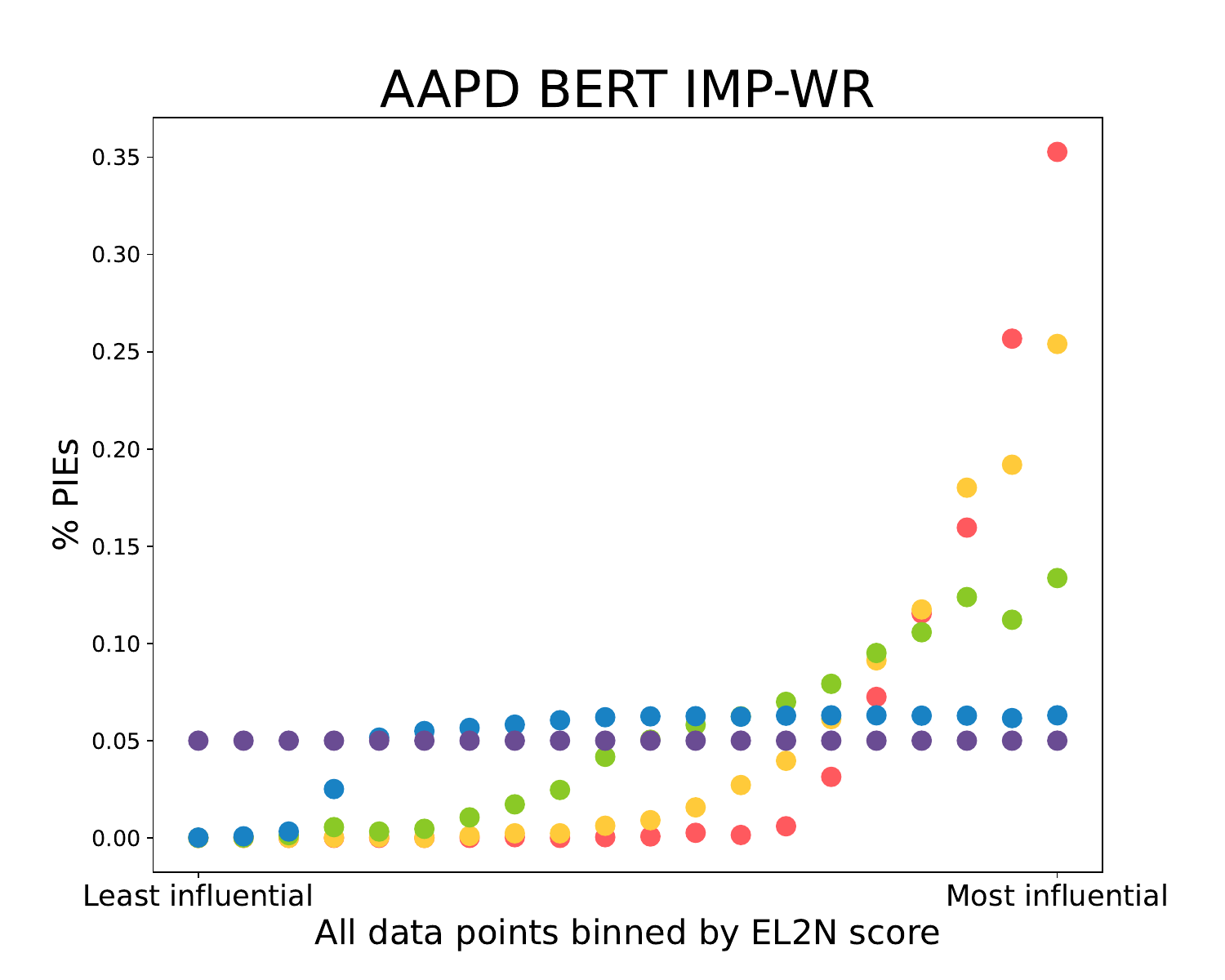} 
  \\

 \includegraphics[width=0.23\textwidth]{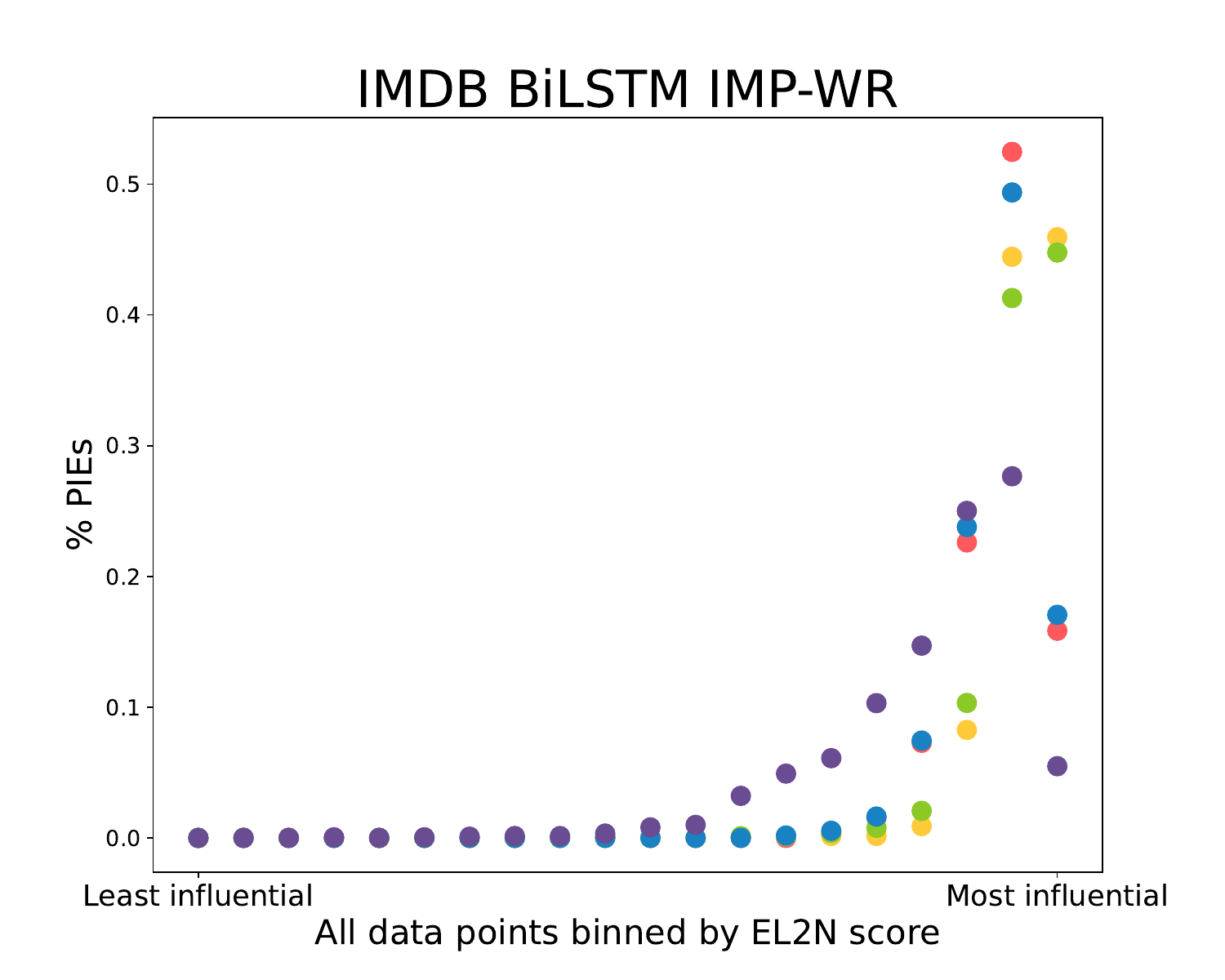} &
 \includegraphics[width=0.23\textwidth]{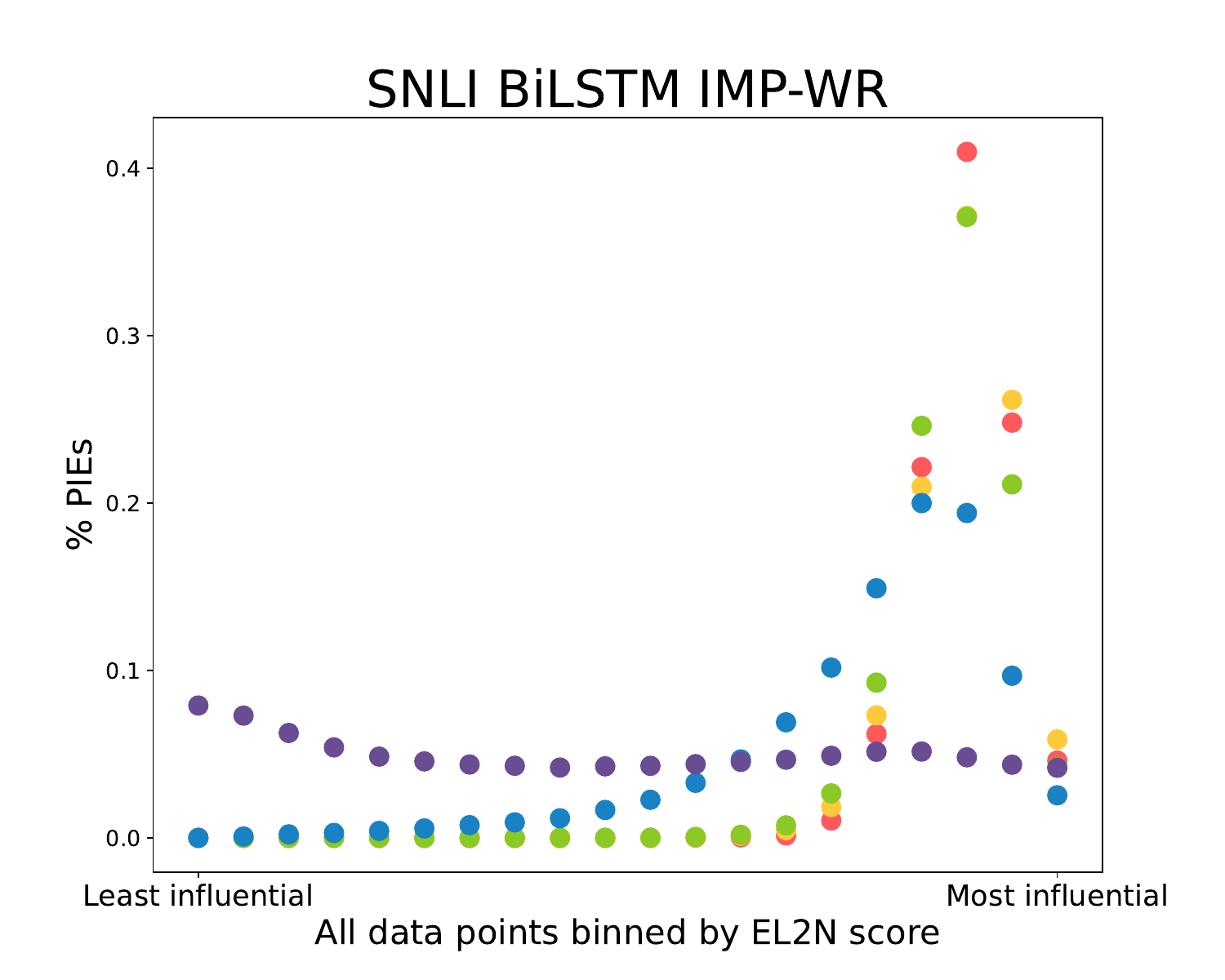} &
 \includegraphics[width=0.23\textwidth]{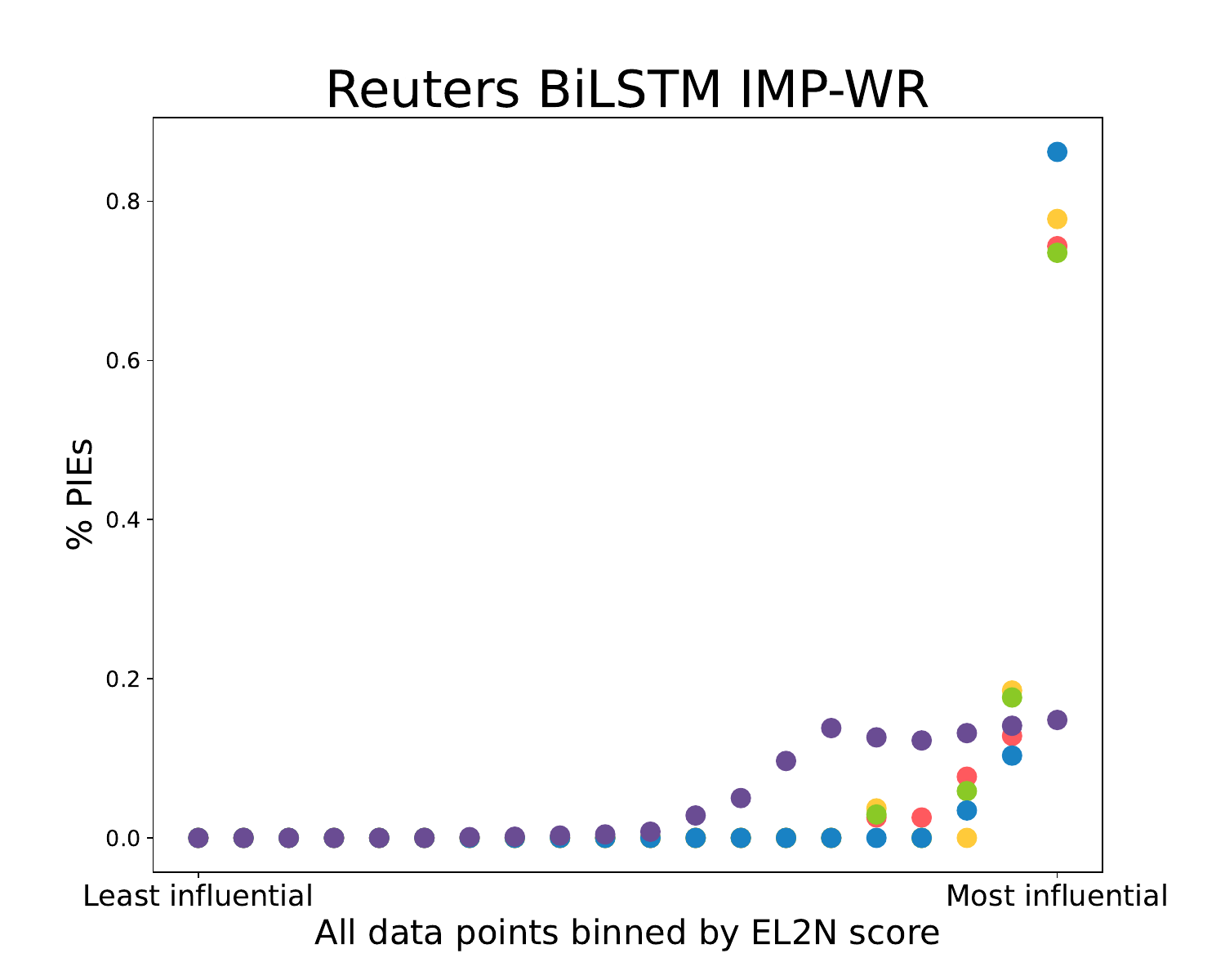} &
 \includegraphics[width=0.23\textwidth]{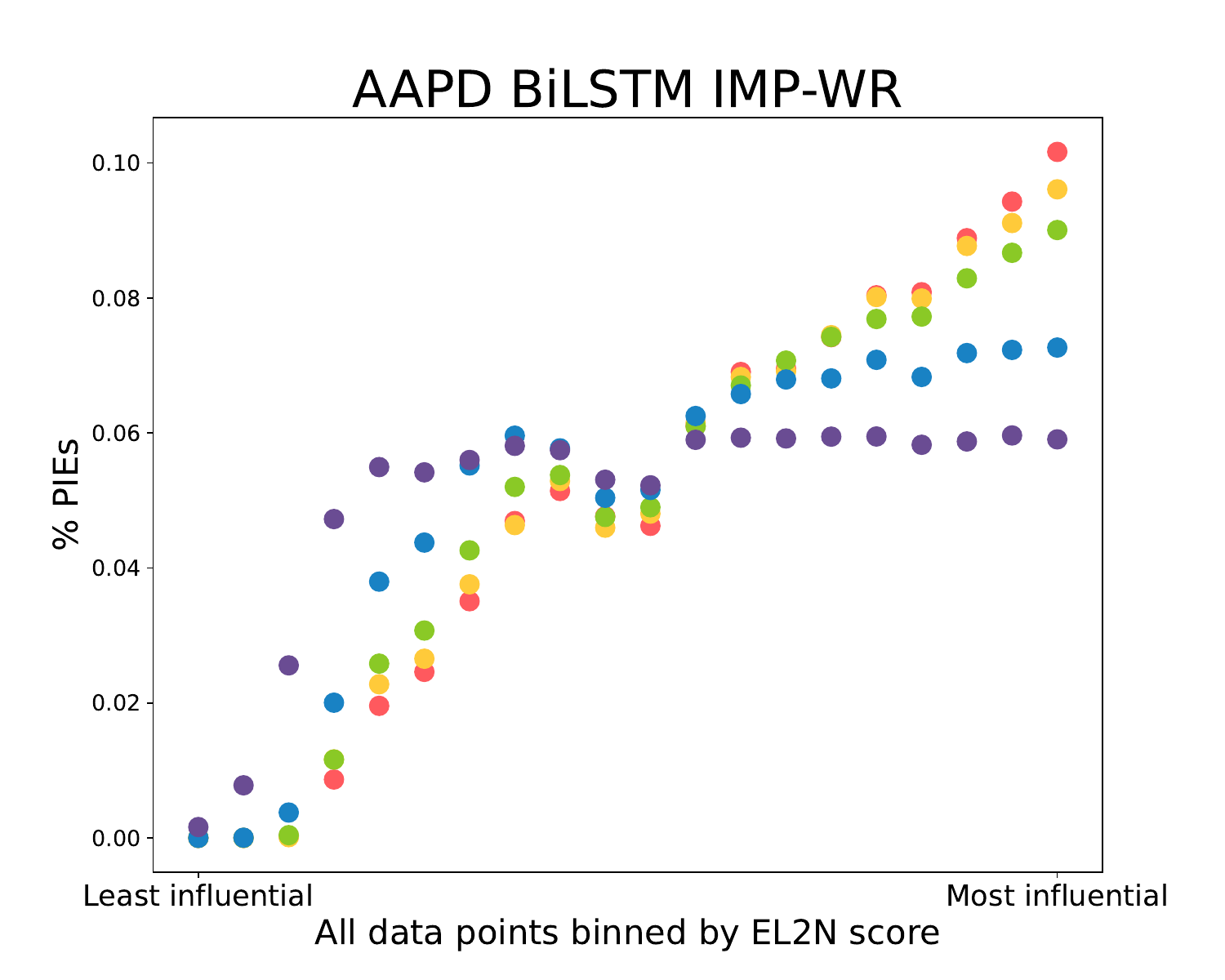} \\

\end{tabular}
\caption{
Percentage of data points that are PIEs (y axis) versus degree of influence (EL2N score) of all data points in the training set (x axis) for IMP-WR.
}
\label{fig:IMDB_BERT_anal4_grid_IMP-WR}%
\end{figure*}

\begin{figure*}
\setlength\tabcolsep{-1pt}
\centering
\begin{tabular}{cccc}

 \includegraphics[width=0.23\textwidth]{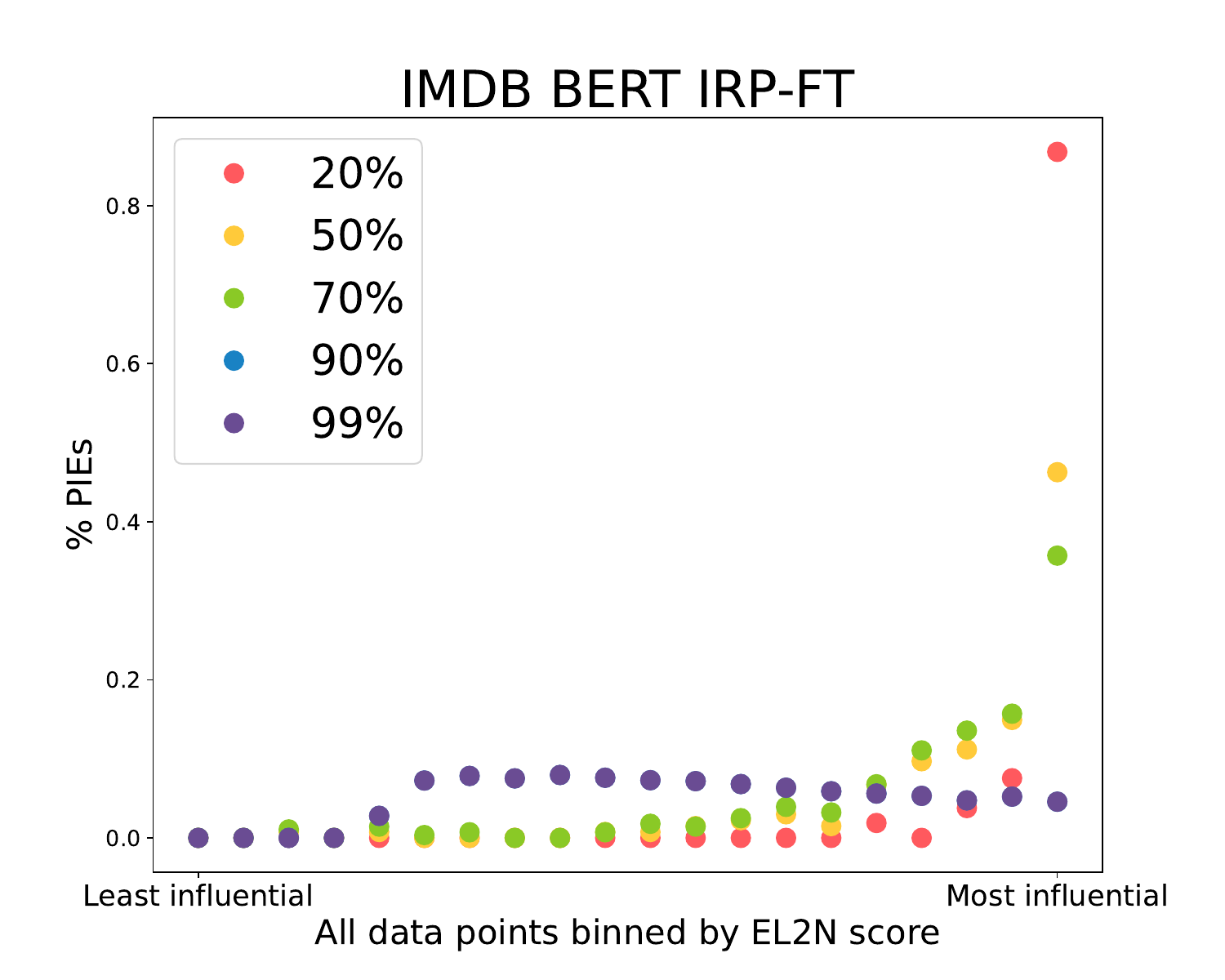} &
 \includegraphics[width=0.23\textwidth]{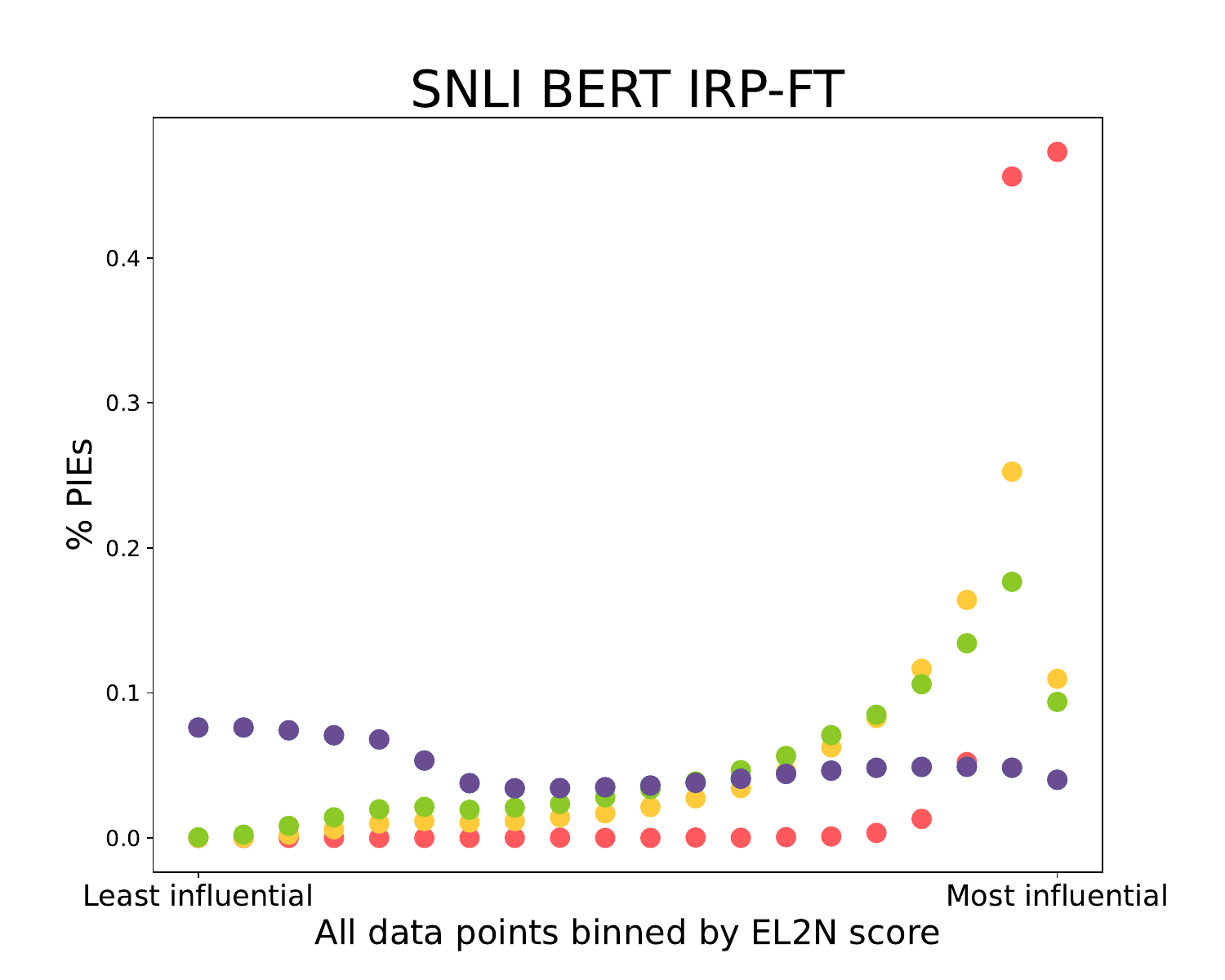} &
 \includegraphics[width=0.23\textwidth]{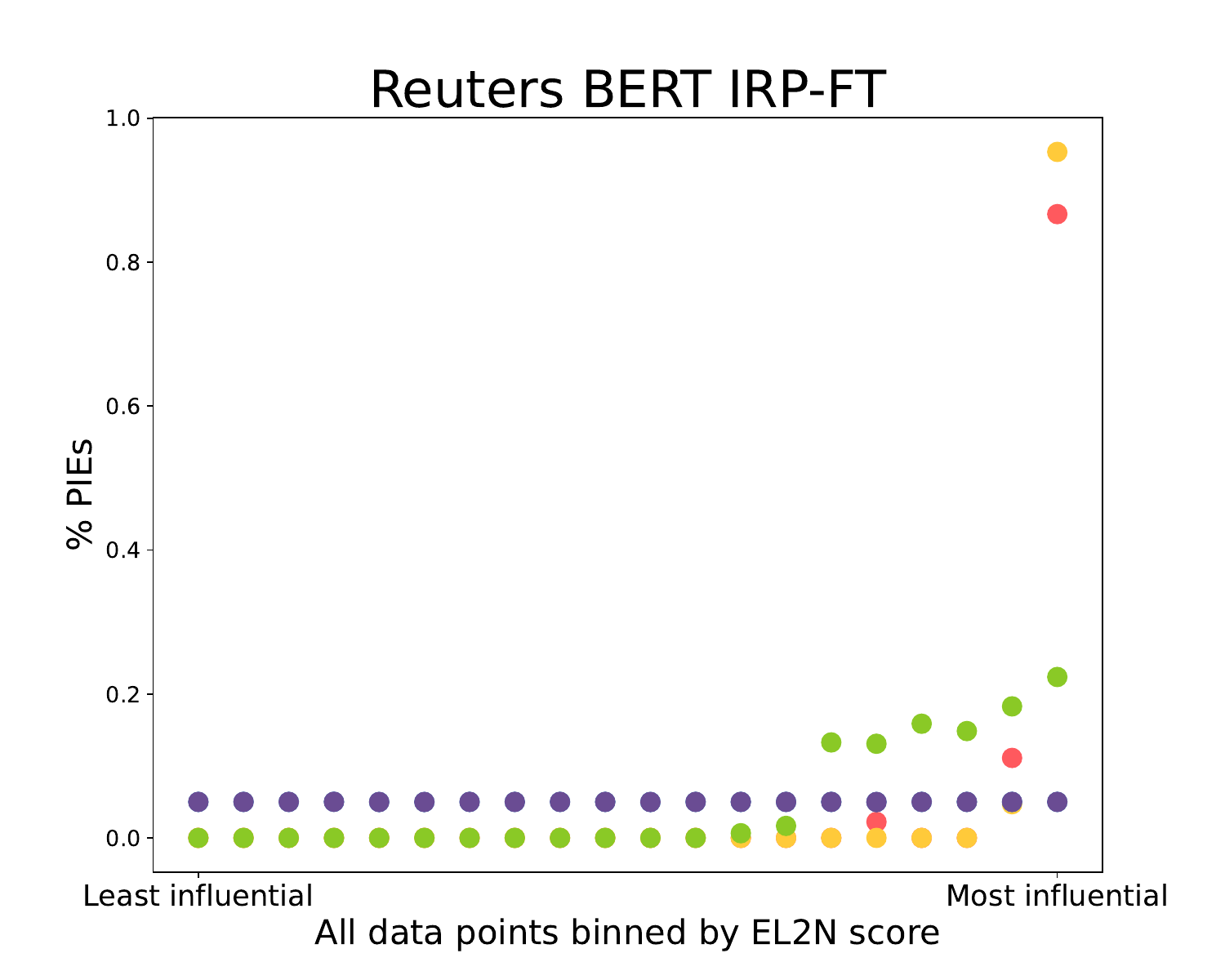} &
 \includegraphics[width=0.23\textwidth]{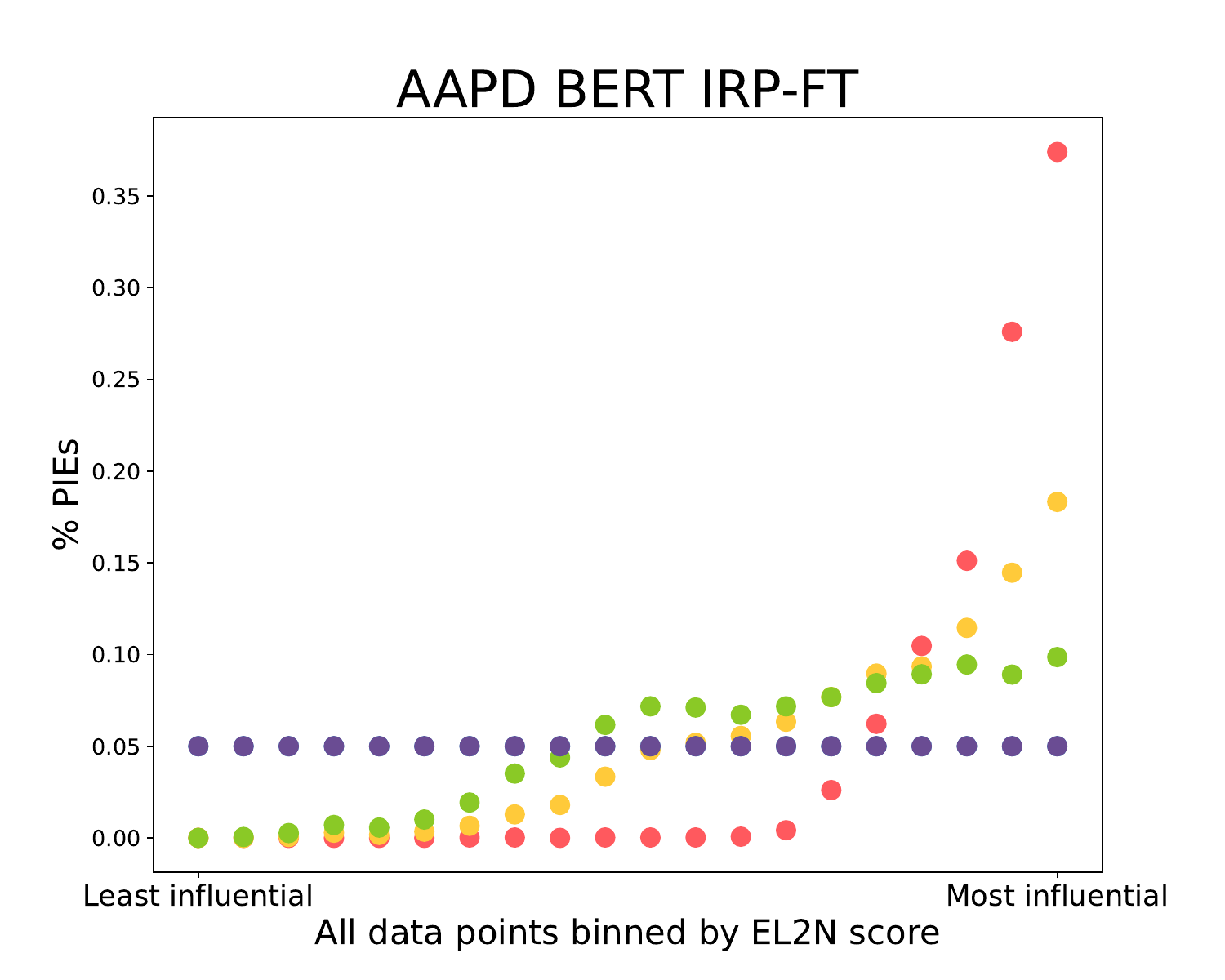} 
  \\

 \includegraphics[width=0.23\textwidth]{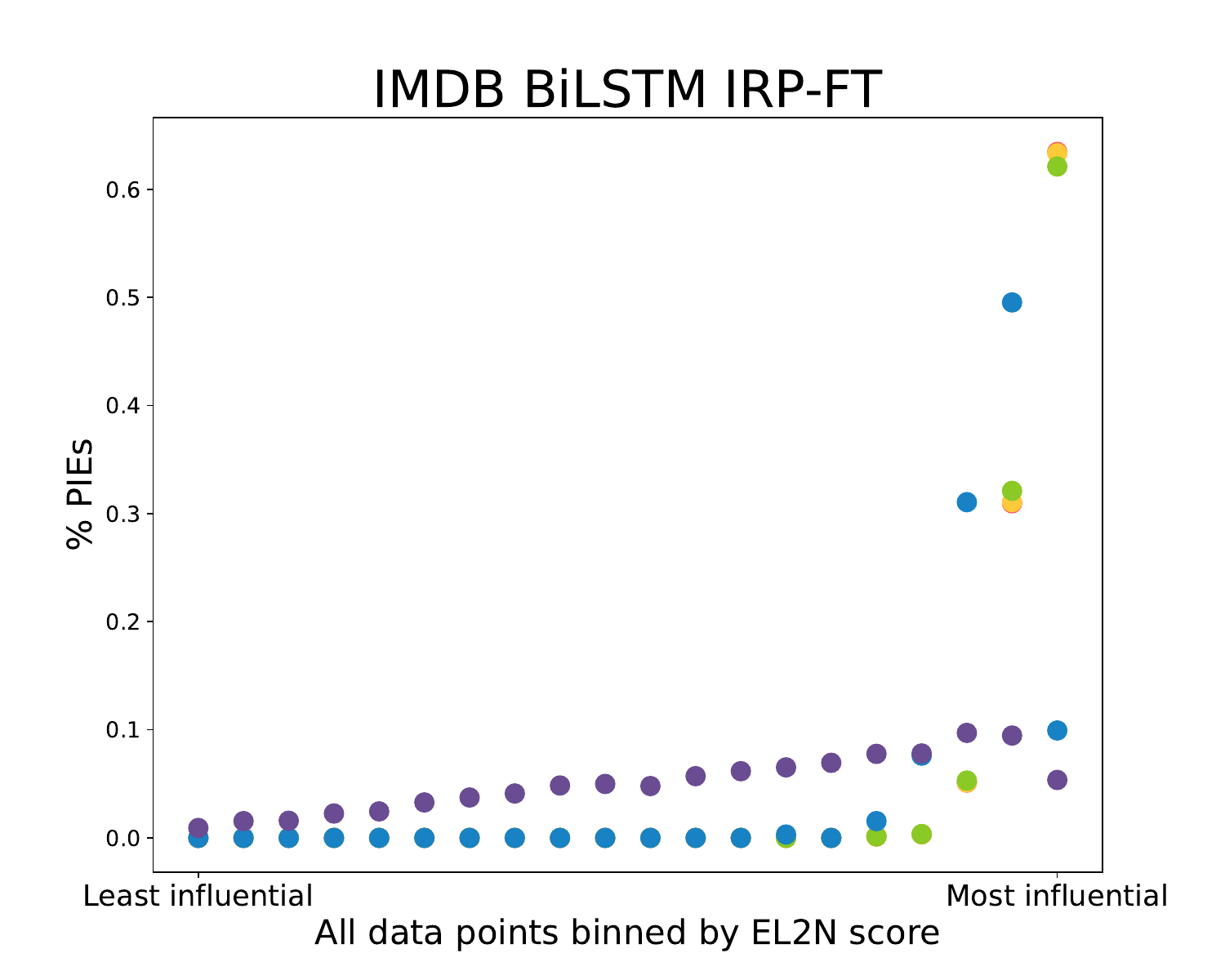} &
 \includegraphics[width=0.23\textwidth]{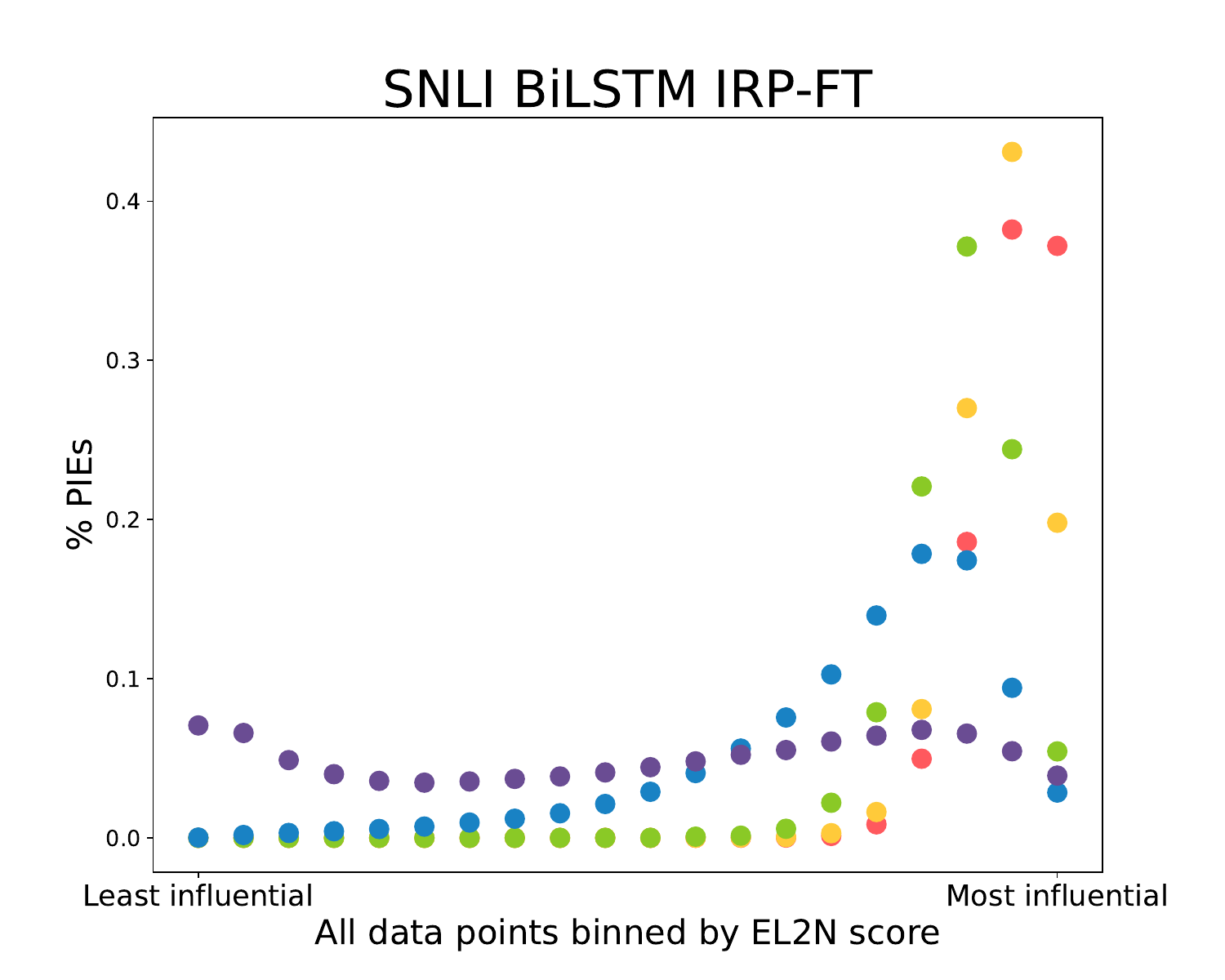} &
 \includegraphics[width=0.23\textwidth]{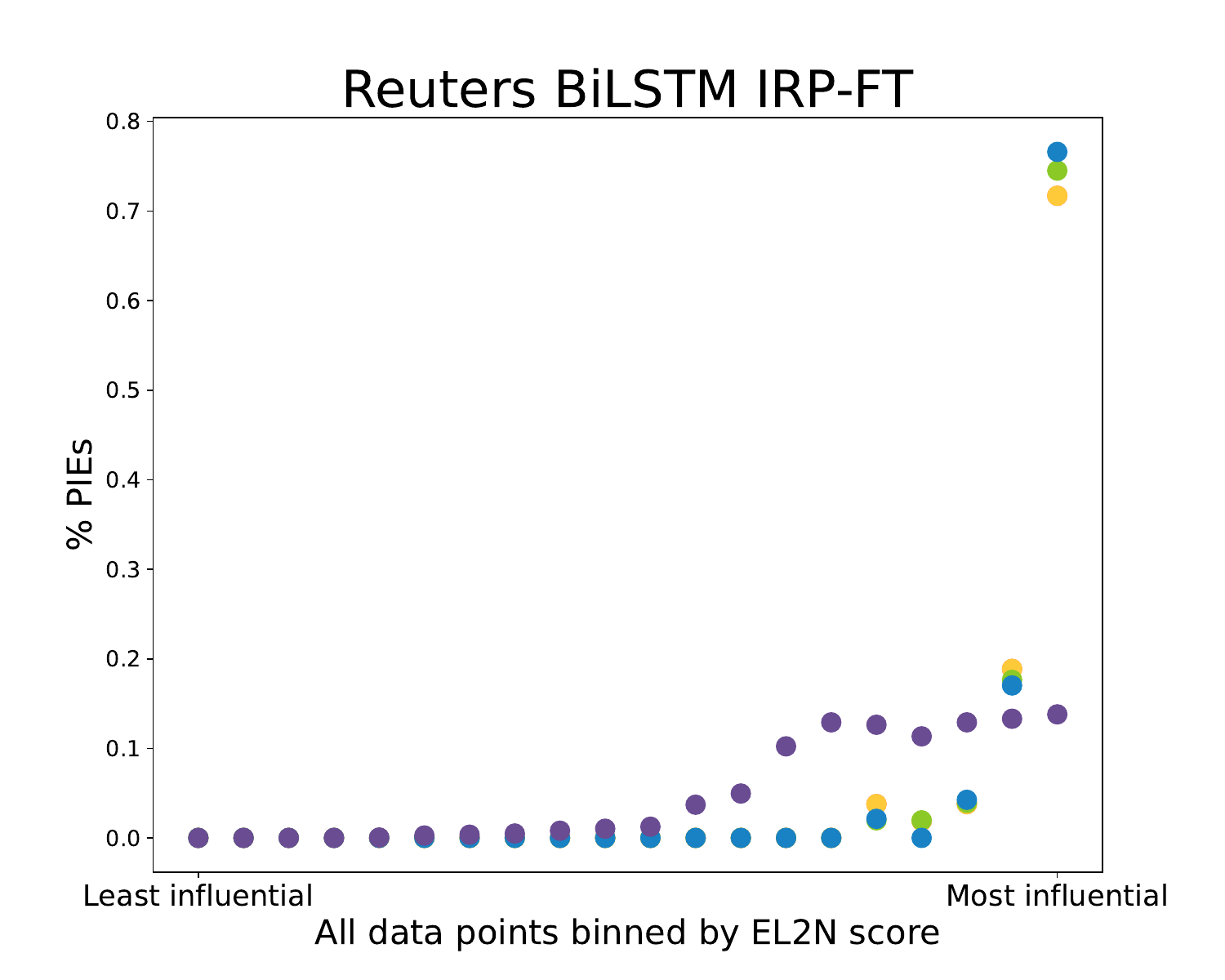} &
 \includegraphics[width=0.23\textwidth]{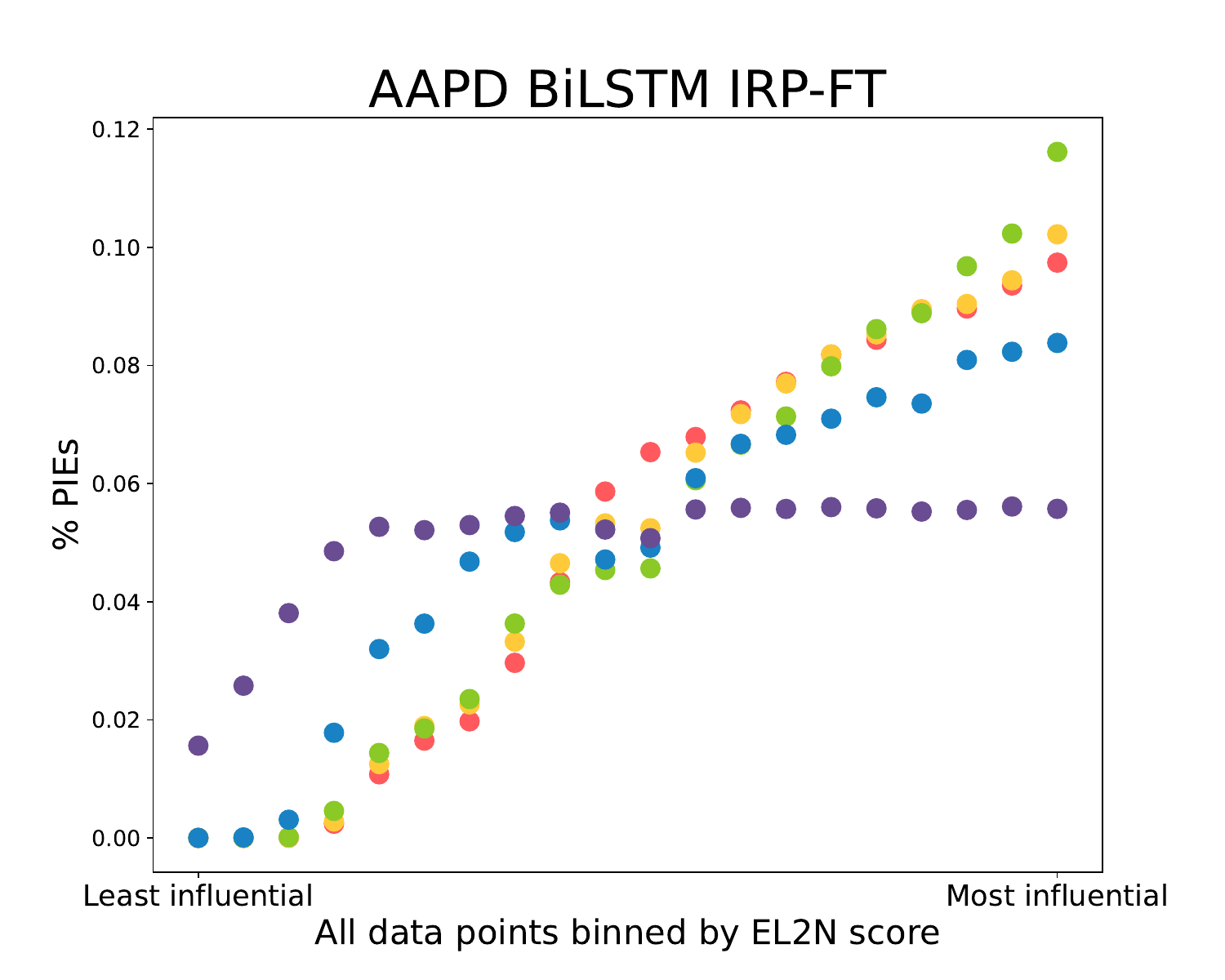} \\

\end{tabular}
\caption{
Percentage of data points that are PIEs (y axis) versus degree of influence (EL2N score) of all data points in the training set (x axis) for IRP-FT.
}
\label{fig:IMDB_BERT_anal4_grid_IRP}%
\end{figure*}

\begin{figure*}
\setlength\tabcolsep{-1pt}
\centering
\begin{tabular}{cccc}

 \includegraphics[width=0.23\textwidth]{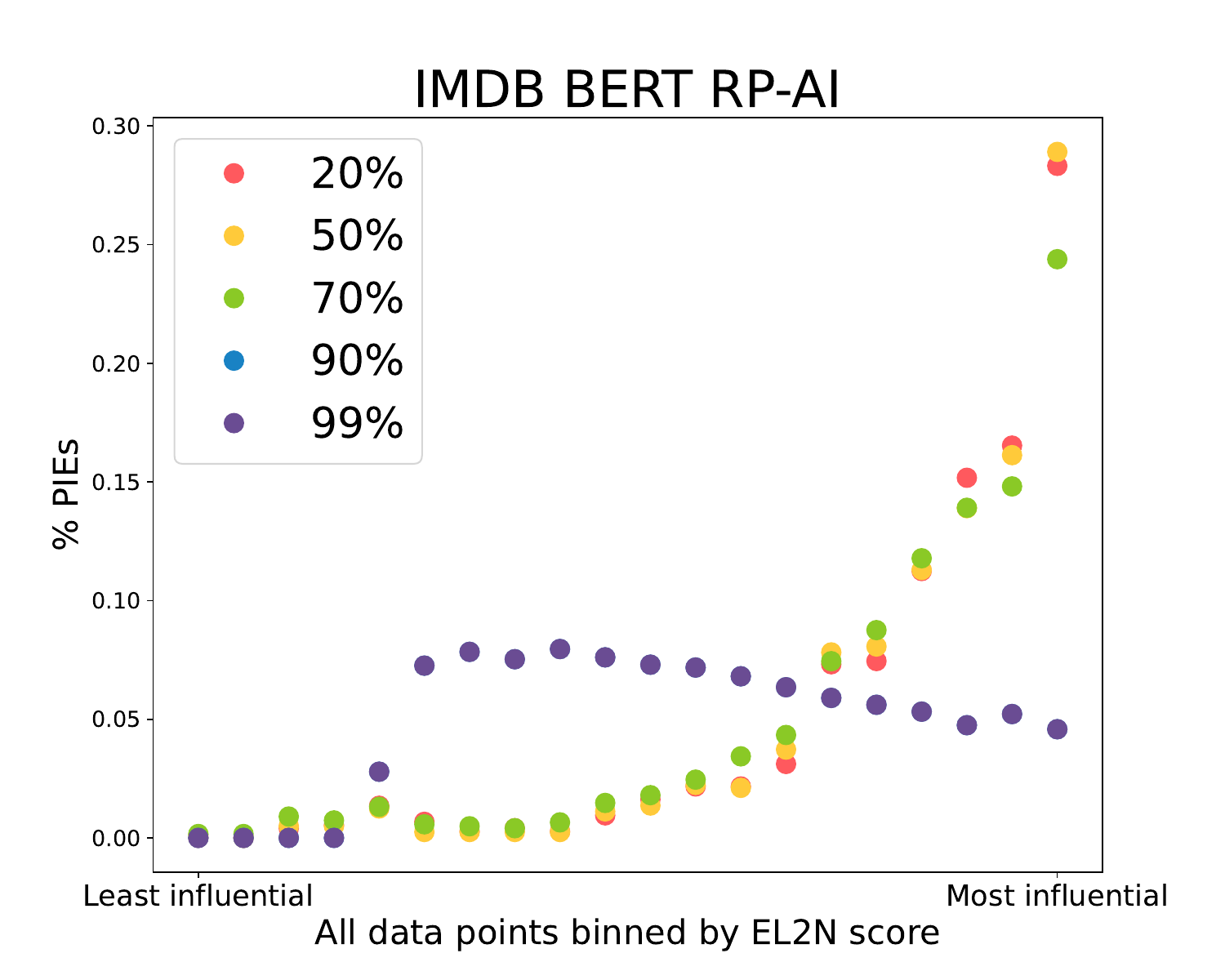} &
 \includegraphics[width=0.23\textwidth]{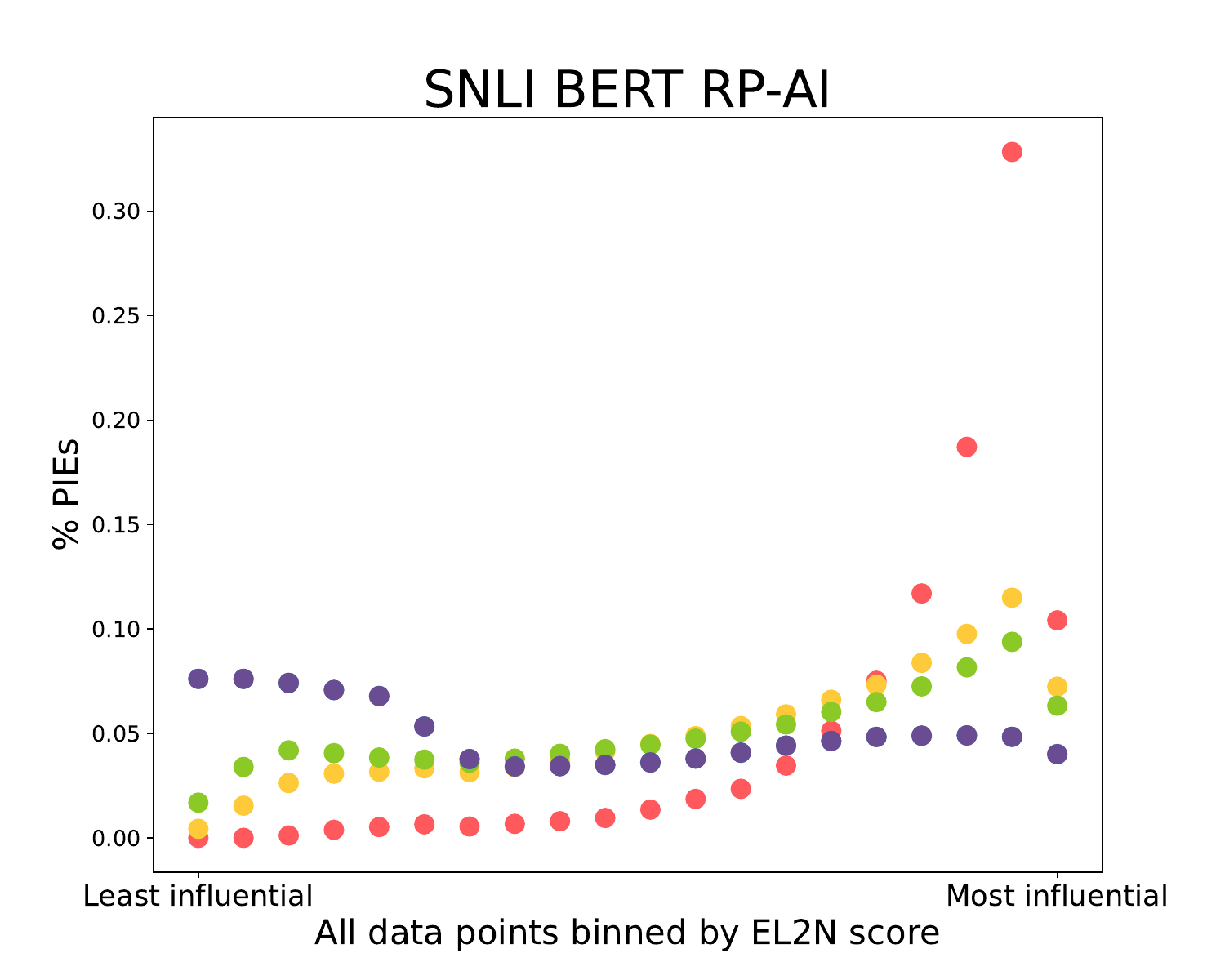} &
 \includegraphics[width=0.23\textwidth]{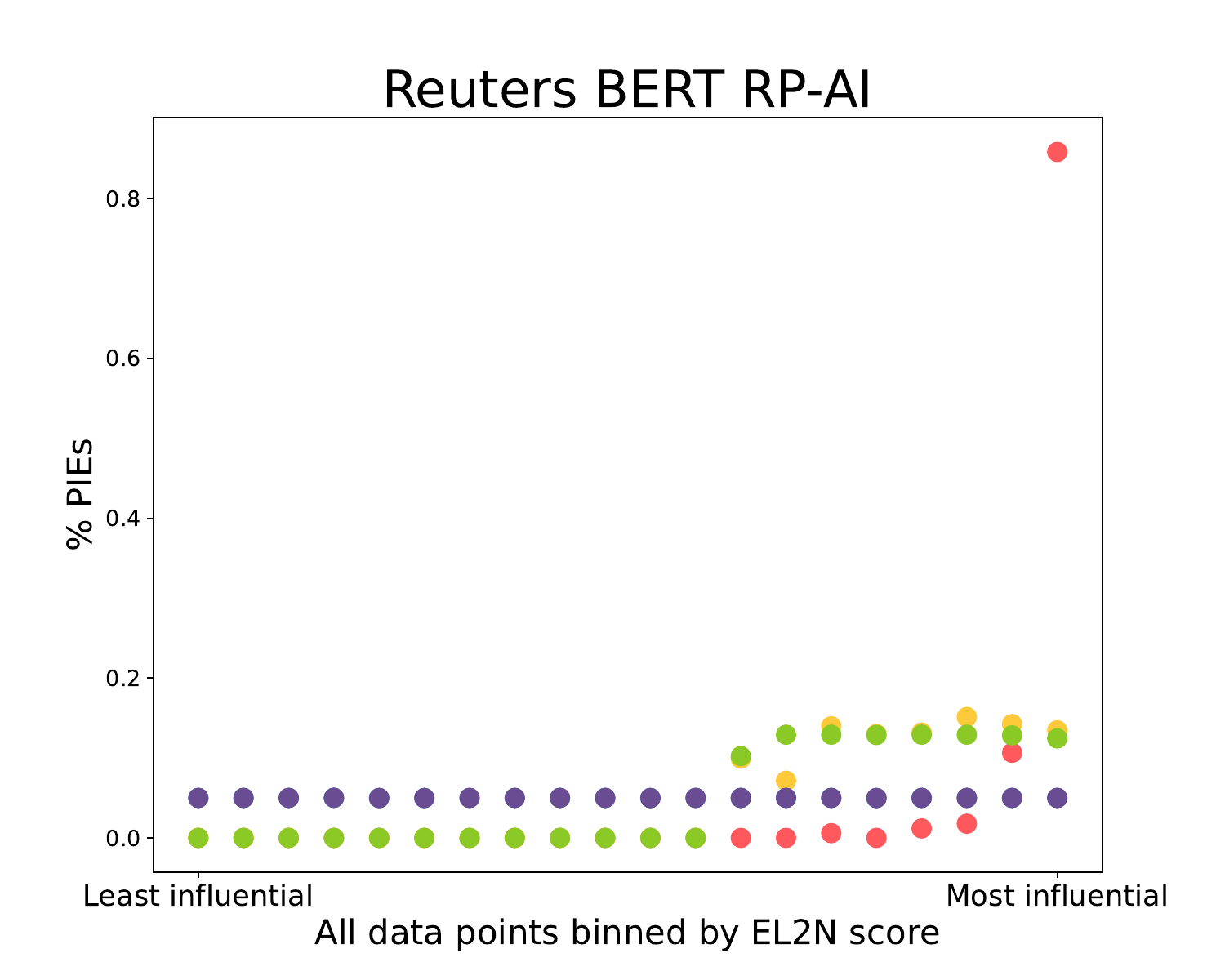} &
 \includegraphics[width=0.23\textwidth]{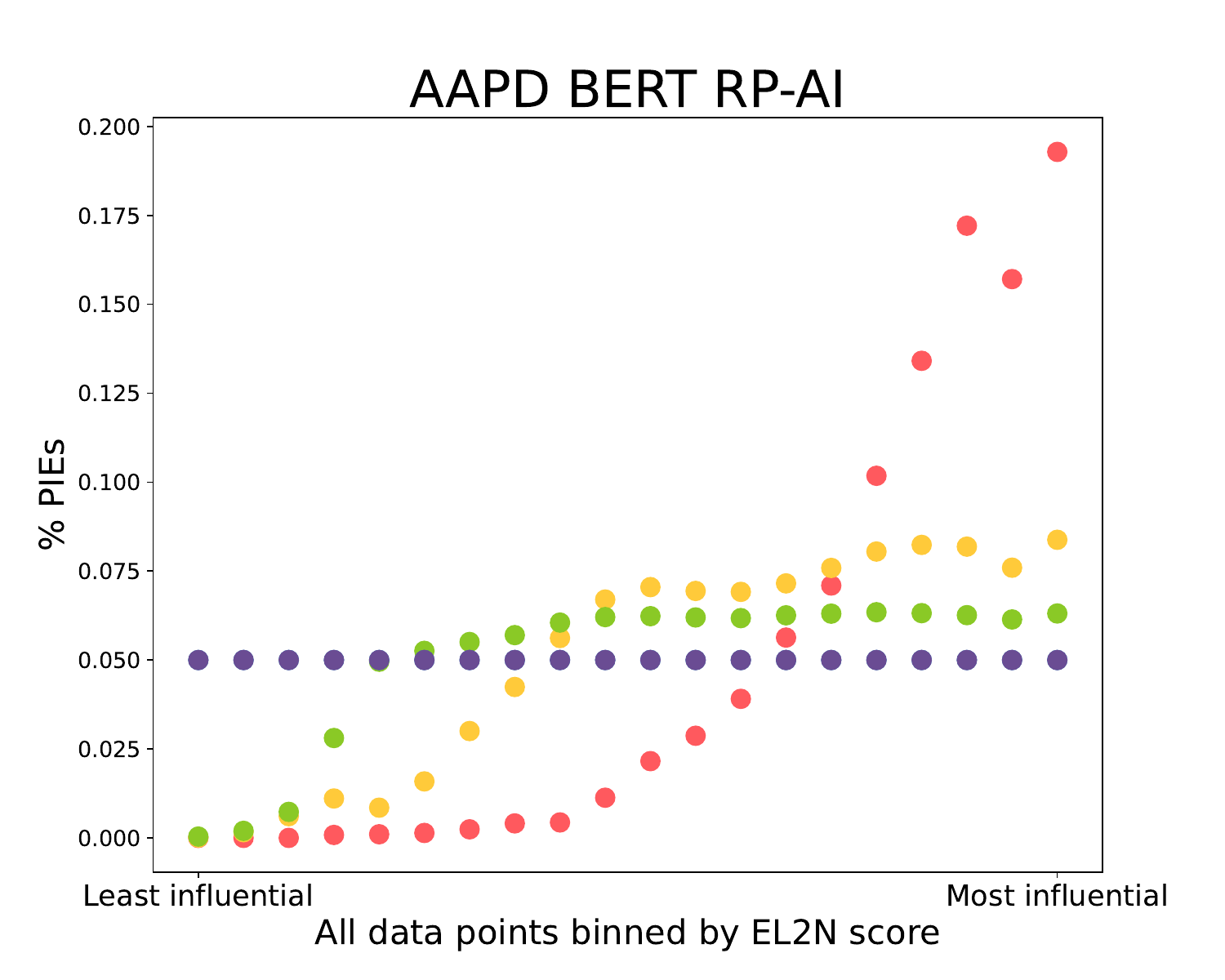} 
  \\

 \includegraphics[width=0.23\textwidth]{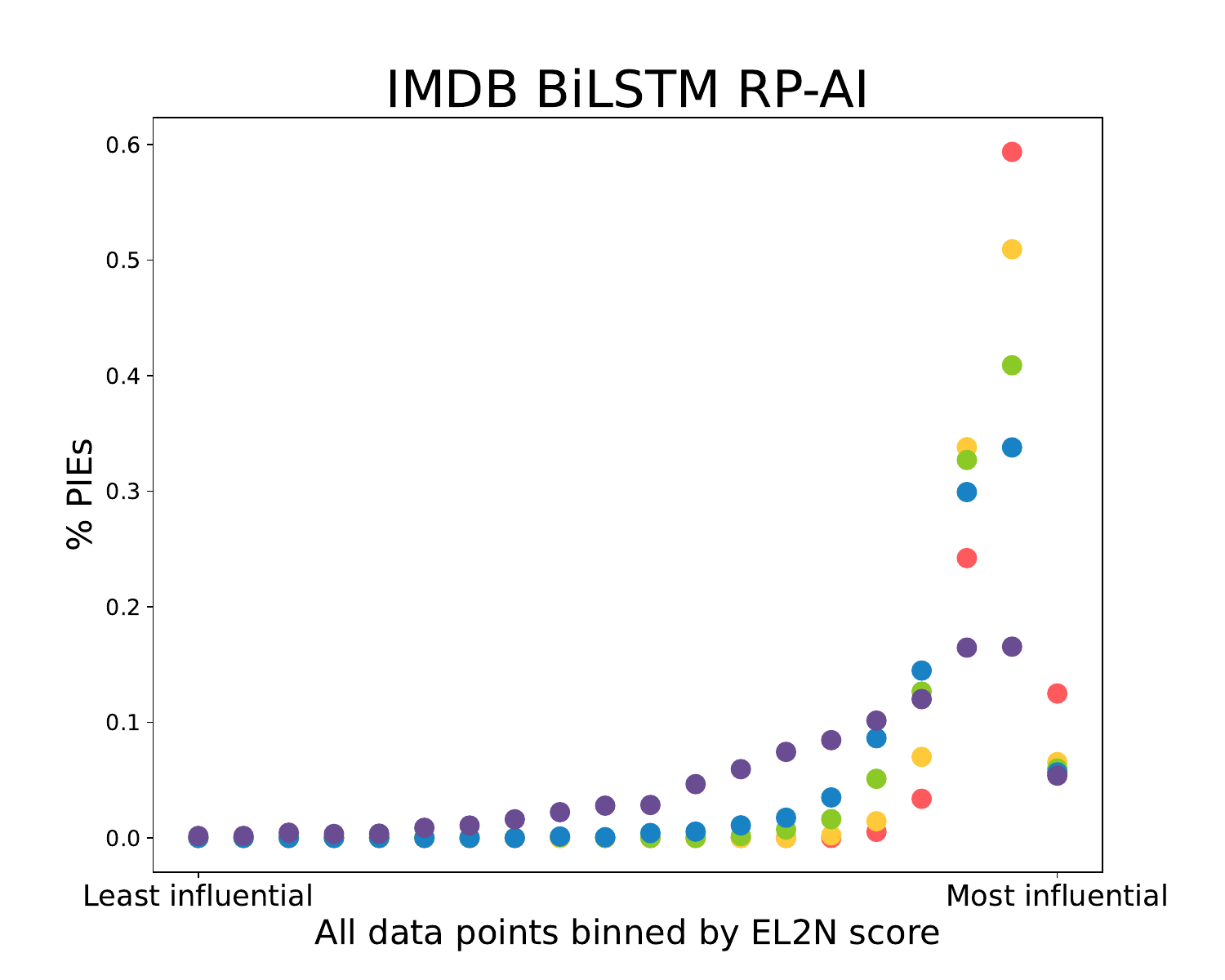} &
 \includegraphics[width=0.23\textwidth]{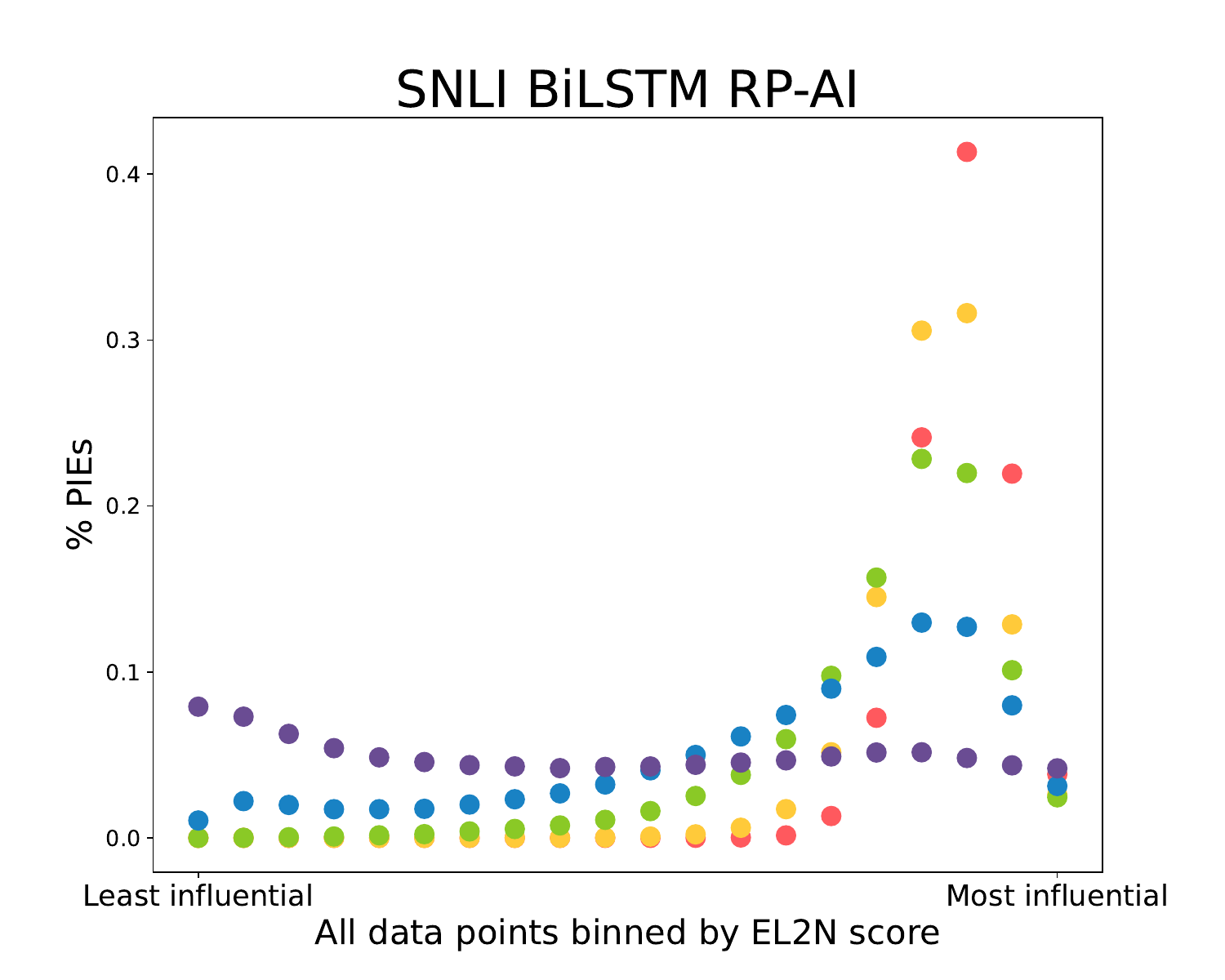} &
 \includegraphics[width=0.23\textwidth]{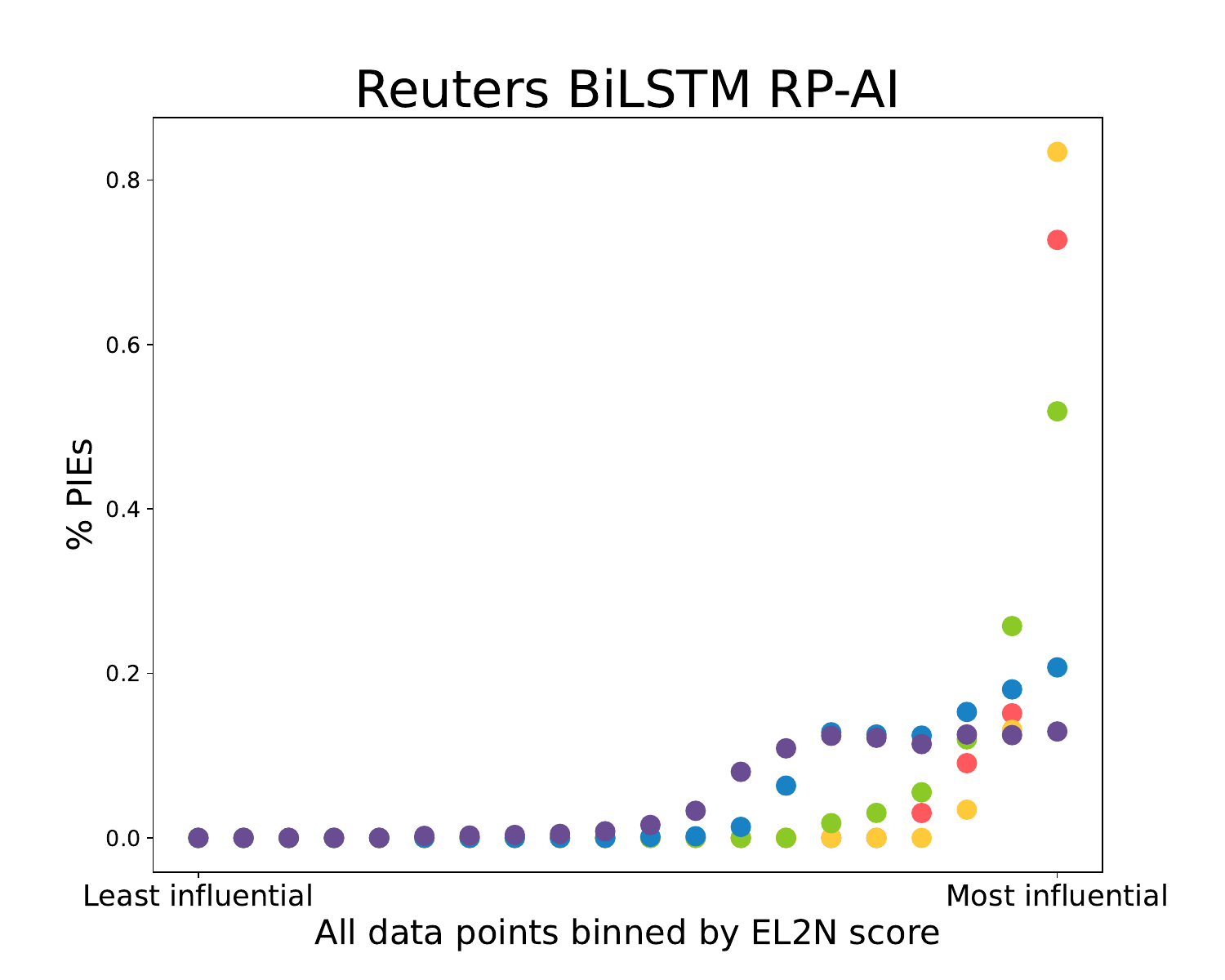} &
 \includegraphics[width=0.23\textwidth]{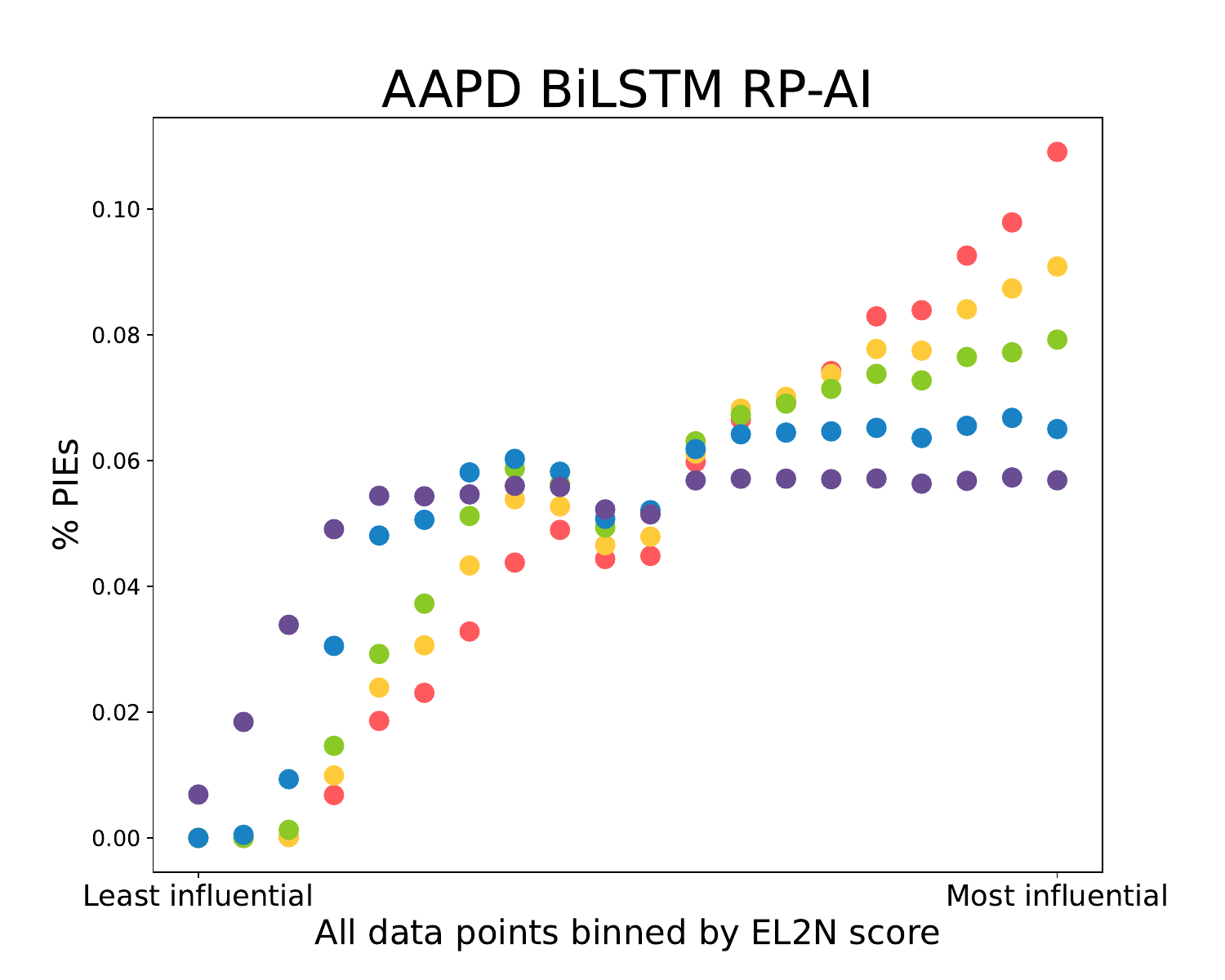} \\

\end{tabular}
\caption{
Percentage of data points that are PIEs (y axis) versus degree of influence (EL2N score) of all data points in the training set (x axis) for RP-AI.
}
\label{fig:IMDB_BERT_anal4_grid_RP-AI}%
\end{figure*}

We report here the additional results of Section \ref{sec:exp4}. 
Figures \ref{fig:IMDB_BERT_anal4_grid_IBP-AI}, \ref{fig:IMDB_BERT_anal4_grid_IIBP-WR},  \ref{fig:IMDB_BERT_anal4_grid_IMM}, \ref{fig:IMDB_BERT_anal4_grid_IMP-AI}, \ref{fig:IMDB_BERT_anal4_grid_IMP-WR},
\ref{fig:IMDB_BERT_anal4_grid_IRP}, and
\ref{fig:IMDB_BERT_anal4_grid_RP-AI} report the percentage of data points that are PIEs versus the degree of influence of all data points in the training set, for each pruning algorithm. PIEs are concentrated on the most influential examples. The higher the amount of pruned parameters, the more PIEs are distributed across examples with different influence on model generalization.

\subsection{Textual characteristics of PIEs}\label{app:textual_characteristics_of_PIEs}

We report here the additional results of Section \ref{sec:Textual_characteristics_of_PIEs}. 
In most cases, 
PIEs require a higher formal education level to be understood than all data points, with the exception of AAPD.
AAPD leads to significant disagreement between pruned and unpruned models, even with 20\% parameter pruning (See Table \ref{tab:percentage_pies_app})). This is due to our extension of PIEs for multi-label settings, which considers a sample as a PIE if there is prediction disagreement on any class.
The more classes in the dataset, the higher the chance of samples being labelled as PIEs. AAPD has 53 classes, the highest class count of all our datasets.
As shown in the remaining settings, the more the disagreement between pruned and unpruned model predictions, the harder it is to observe a difference between the formal education level needed to understand PIEs and the dataset. Hence, on AAPD, we do not observe the same behaviour obtained in the three remaining datasets.

PIEs are overall longer than the average text lenght of all data points. PIEs can have up to $1.13$ and $1.9$ more tokens than the average number of tokens for a sample in the dataset for IMDB, and Reuters respectively.
The behaviour can be observed with both BERT and BiLSTM models.
About the ratio between the average number of tokens for the PIEs and in all the samples of the dataset on SNLI and AAPD datasets: we do not see the same behaviour as in IMDB and Reuters. SNLI is mostly made of short samples, hence it is harder to observe the behaviour on such a dataset, even if the trend is the same.
On AAPD, the same observation on the needed education level holds when discussing text length.

\begin{figure}
\centering

\resizebox{\columnwidth}{!}{%
\begin{tabular}{cc}

 \includegraphics[width=0.5\textwidth]{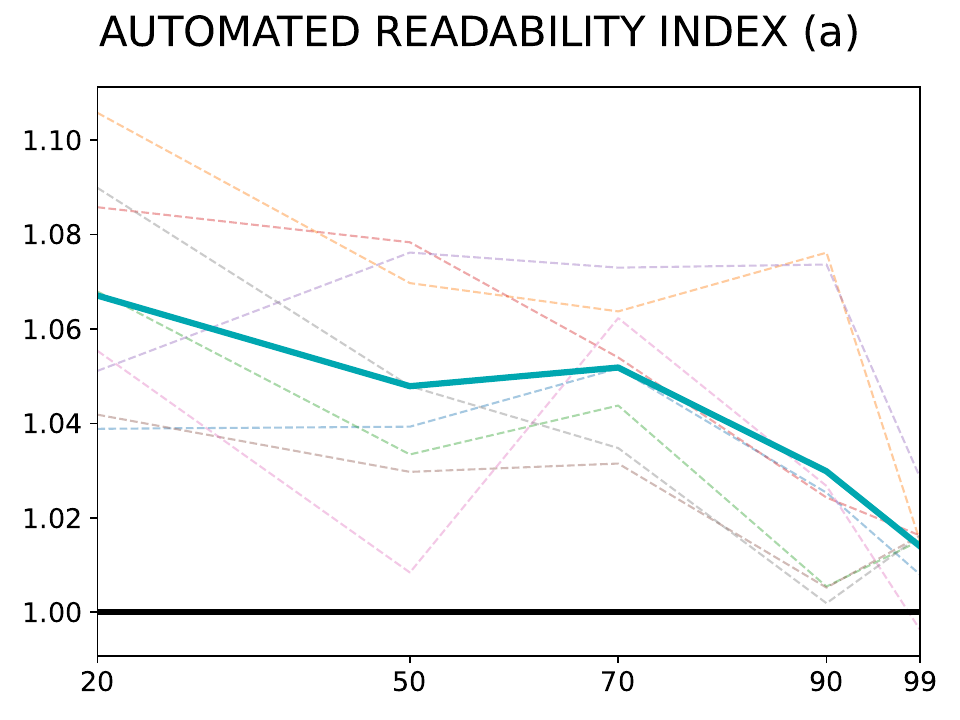} &
 \includegraphics[width=0.5\textwidth]{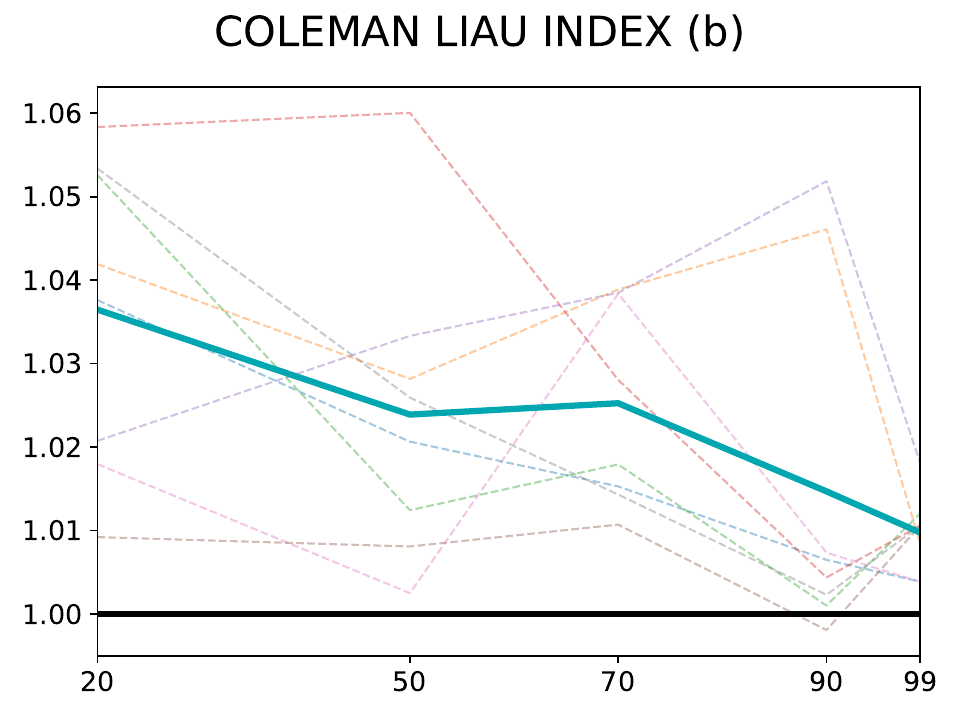} \\
 \includegraphics[width=0.5\textwidth]{Images/Readability_scores/flesch_kincaid_grade/SNLI_BERT_ALL_DIFF.pdf} &
 \includegraphics[width=0.5\textwidth]{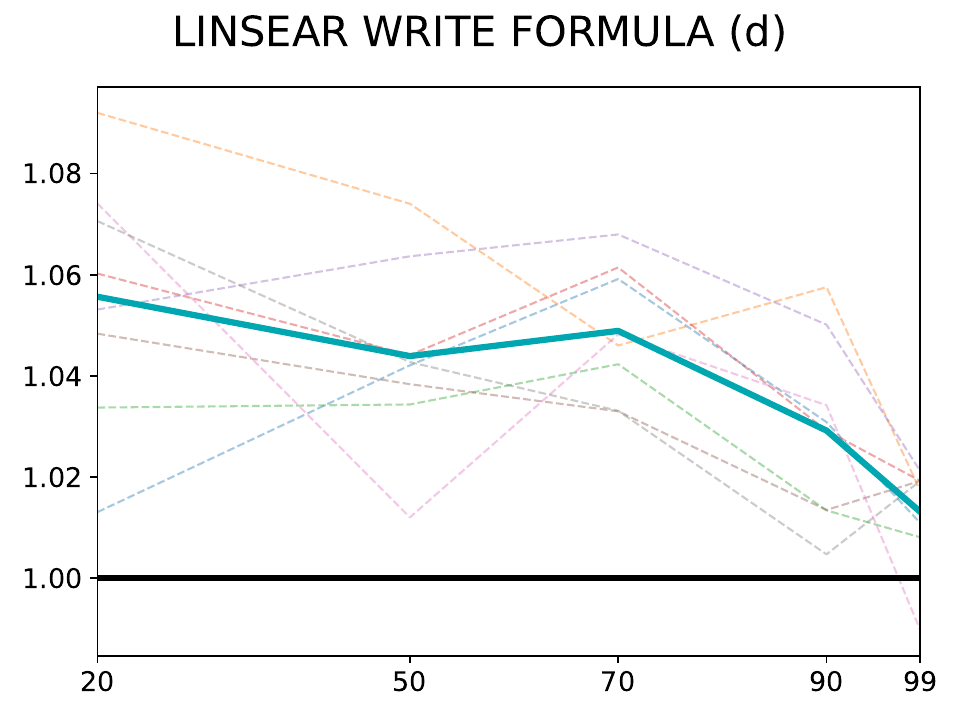} \\
 \includegraphics[width=0.5\textwidth]{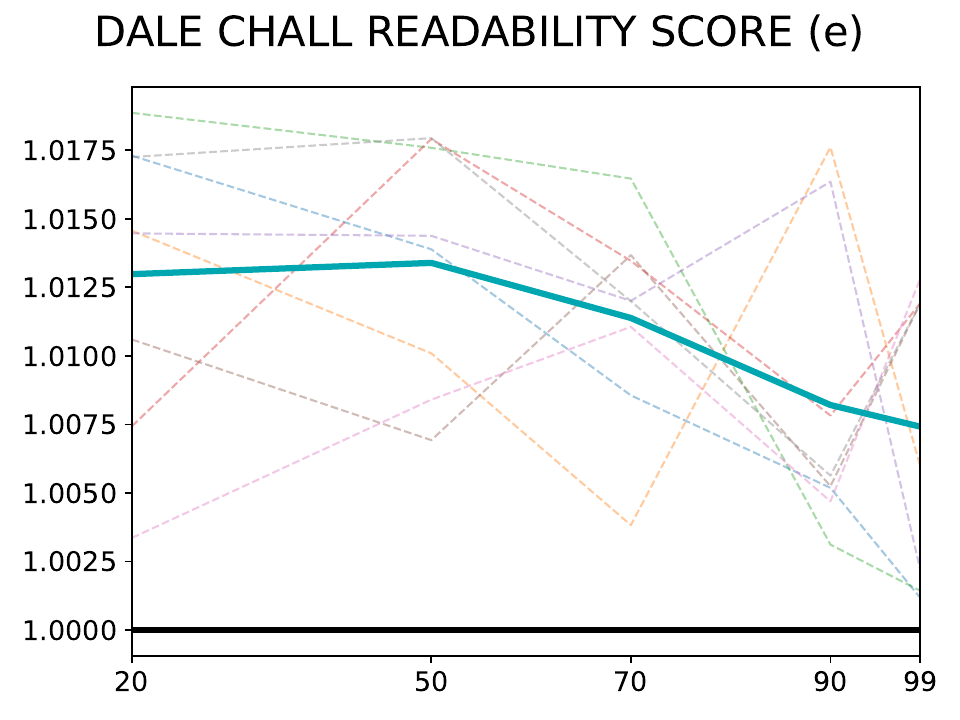} &
 \includegraphics[width=0.5\textwidth]{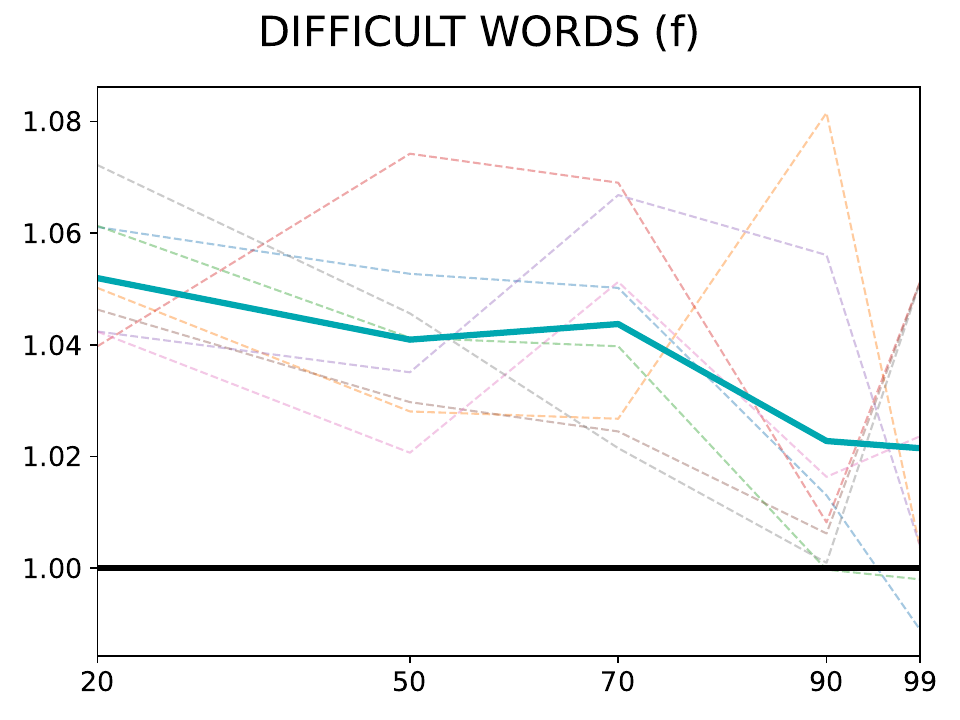} \\
 \includegraphics[width=0.5\textwidth]{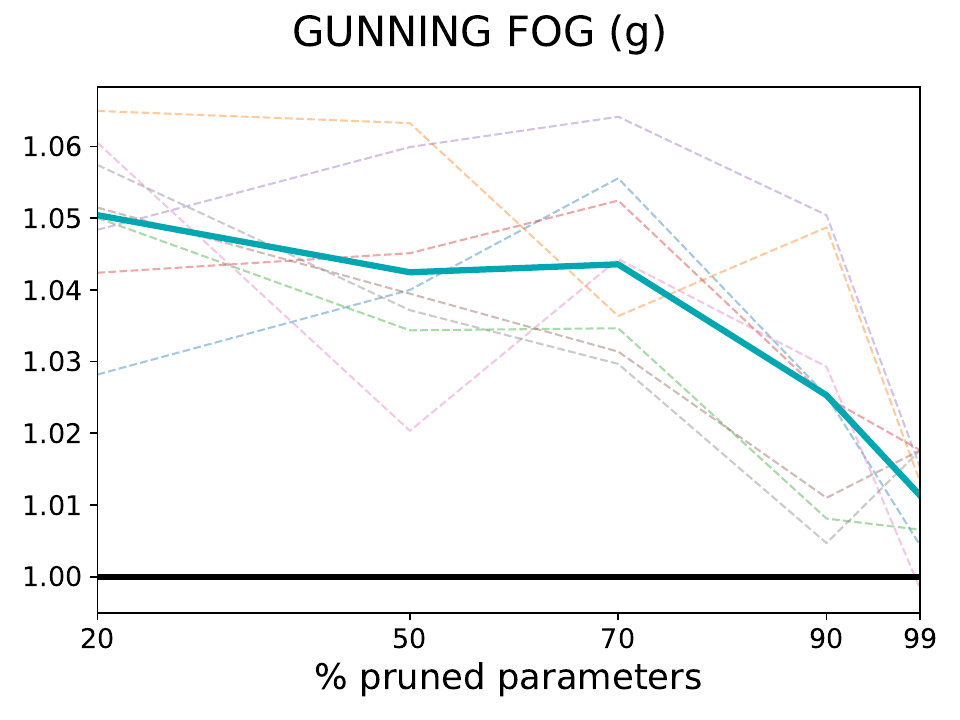} &
 \includegraphics[width=0.5\textwidth]{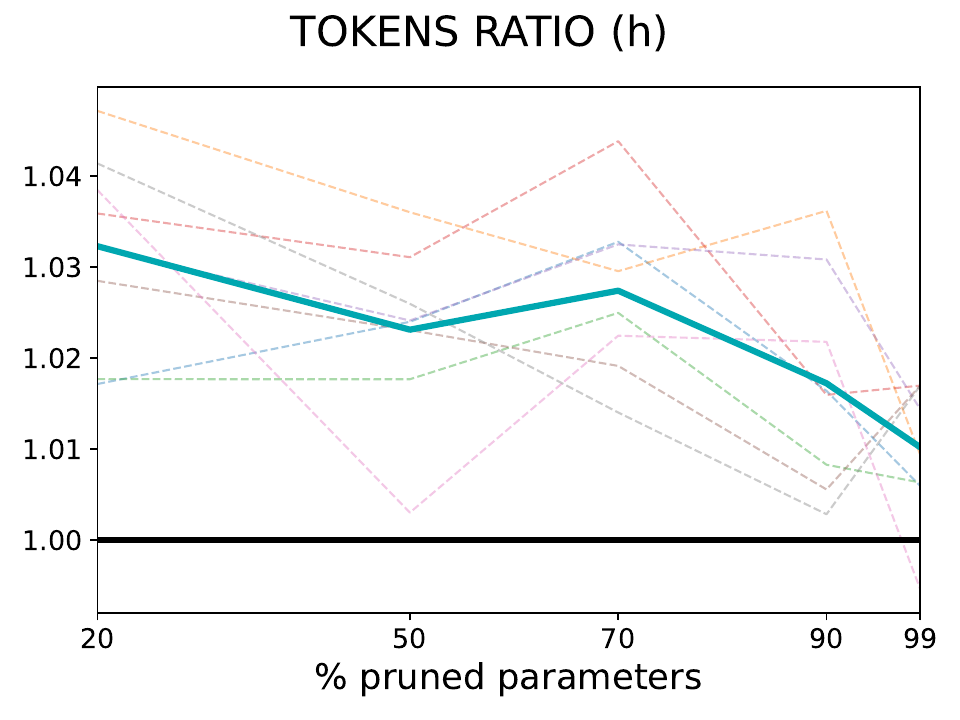} \\
 \multicolumn{2}{c}{\includegraphics[width=0.9\textwidth]{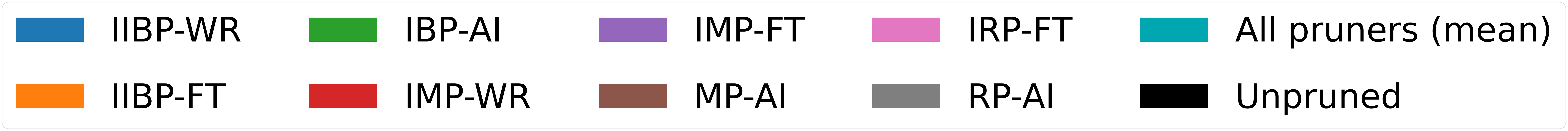}}    
\end{tabular}
}
\caption{How the text of PIEs differs from the text of all data points, according to 7 readability scores (plots (a)-(g)) and text length (plot (h)). 
Ratio between the scores of PIEs and the scores of all data points (y axis), across pruning thresholds (x axis), for BiLSTM and SNLI.
The solid black horizontal line represents equal scores in PIEs and all data points. The solid turquoise line is the mean score of all pruners. Any line above the solid black line means that PIEs are harder to understand (plots (a)-(g)) or have longer text (plot (h)), on average, than all data points. 
}
\label{fig:readability_SNLI_BiLSTM}%
\end{figure}

\begin{figure}
\centering

\resizebox{\columnwidth}{!}{%
\begin{tabular}{cc}

 \includegraphics[width=0.5\textwidth]{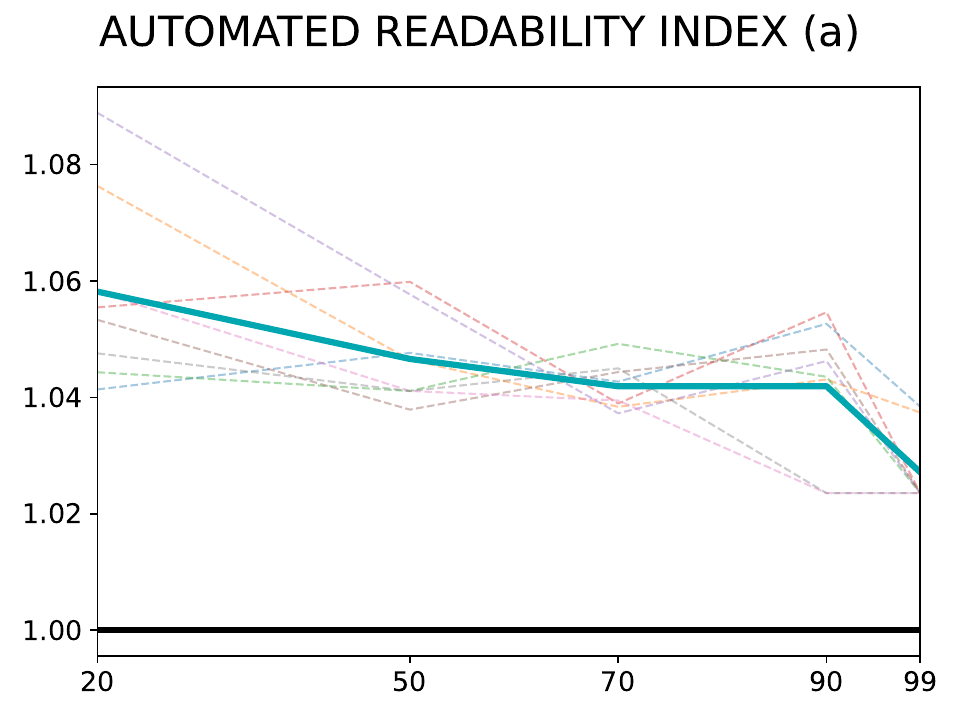} &
 \includegraphics[width=0.5\textwidth]{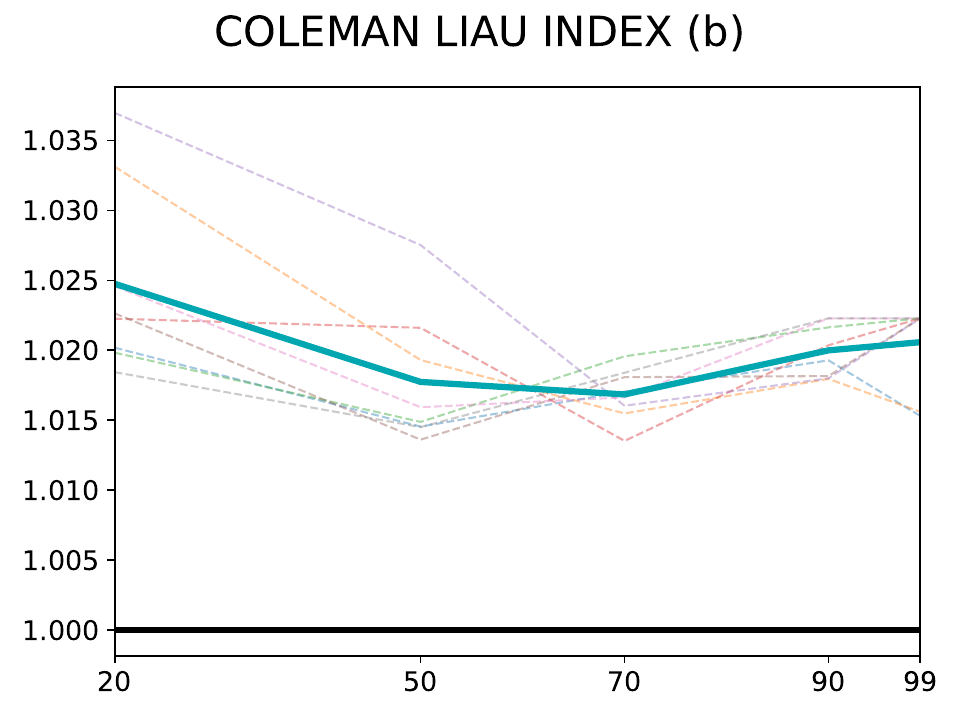} \\
 \includegraphics[width=0.5\textwidth]{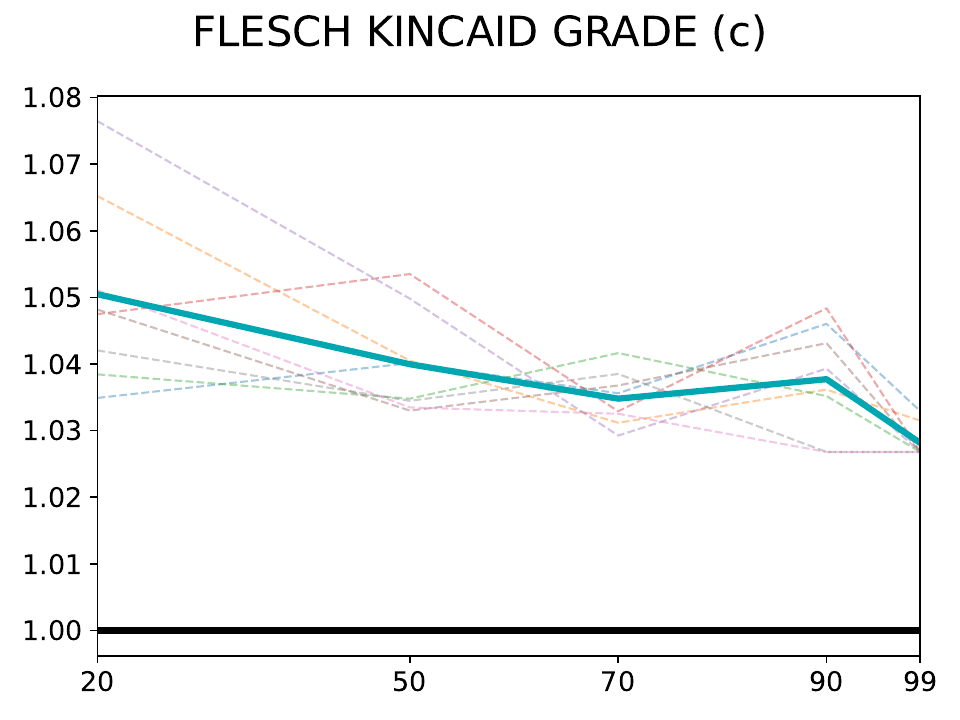} &
 \includegraphics[width=0.5\textwidth]{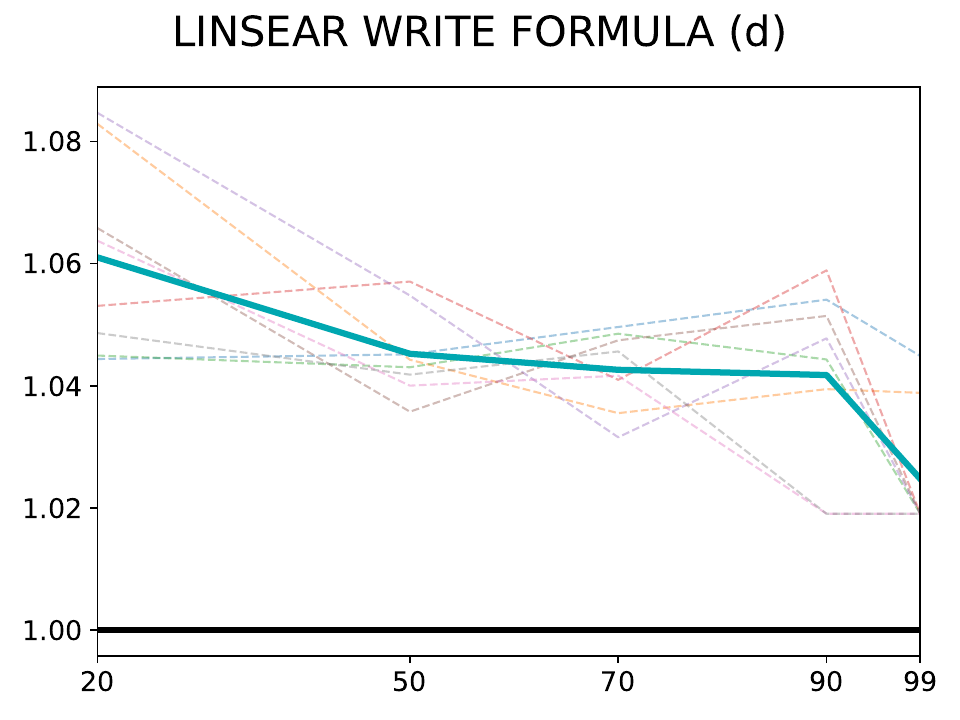} \\
 \includegraphics[width=0.5\textwidth]{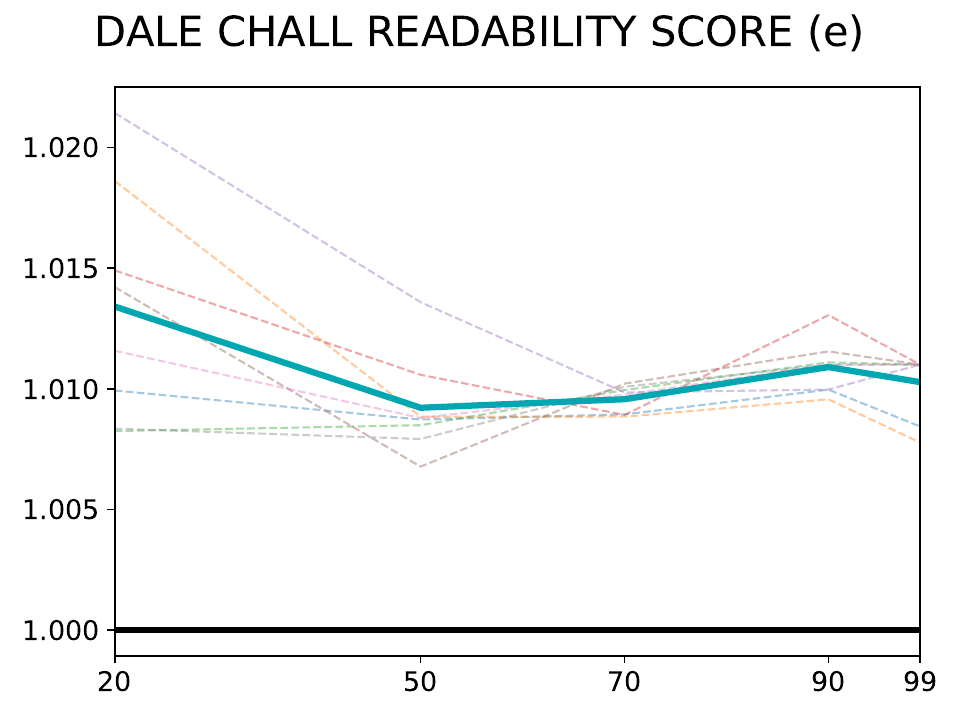} &
 \includegraphics[width=0.5\textwidth]{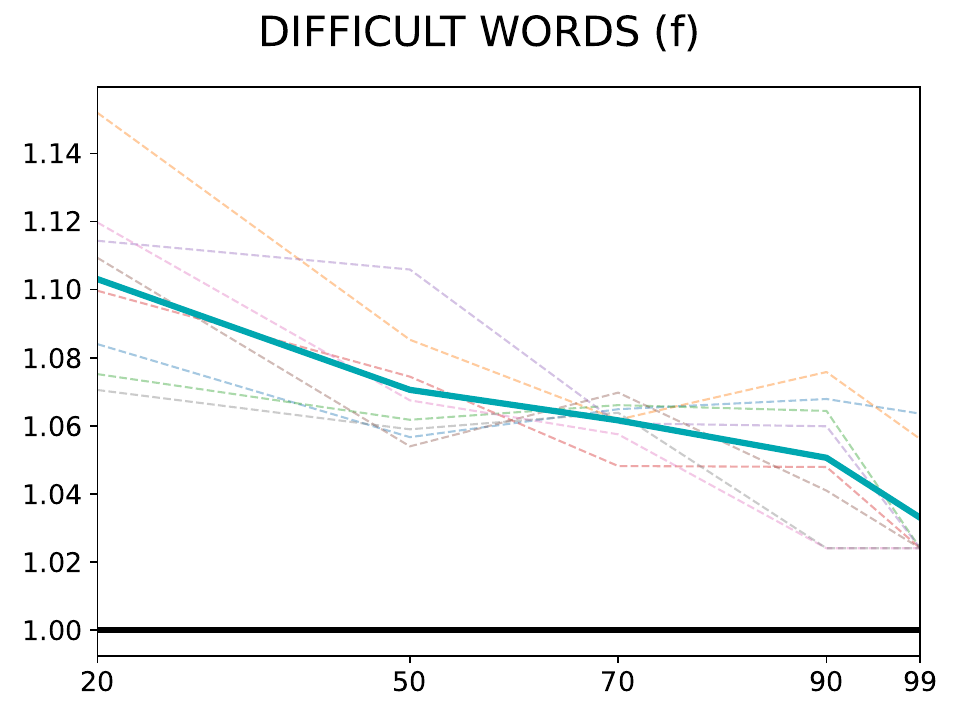} \\
 \includegraphics[width=0.5\textwidth]{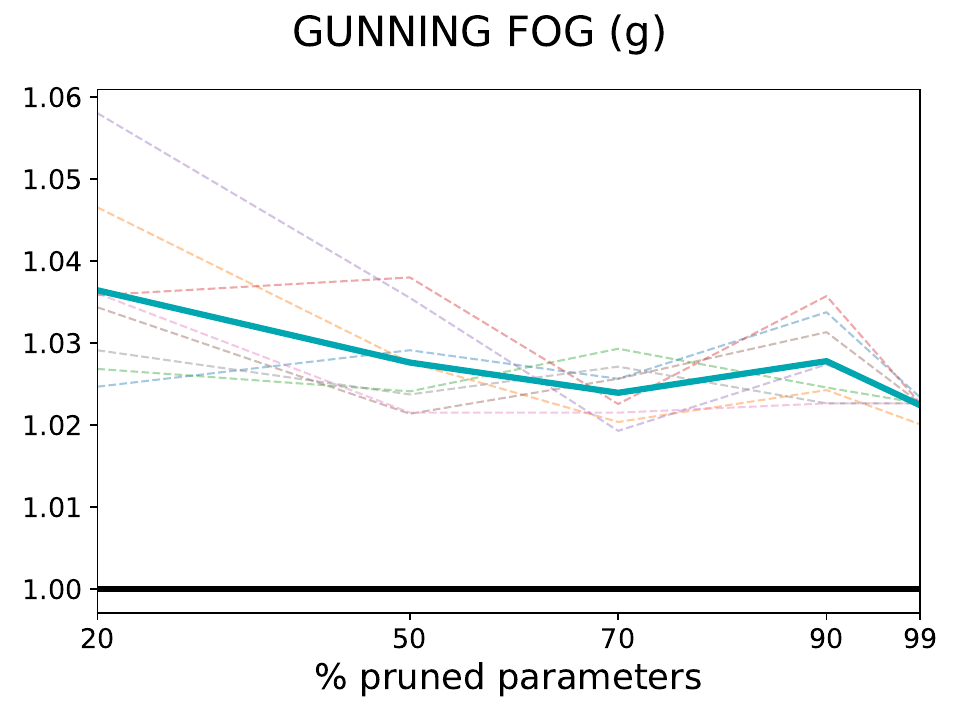} &
 \includegraphics[width=0.5\textwidth]{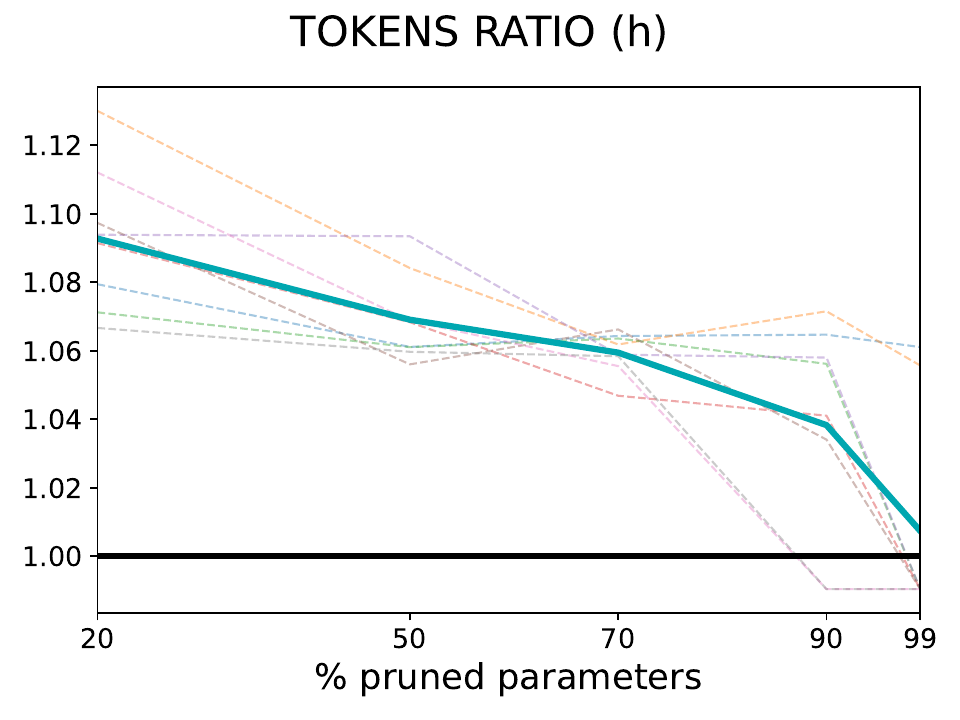} \\
 \multicolumn{2}{c}{\includegraphics[width=0.9\textwidth]{Images/PIEs/Text_length/legend.pdf}}    
\end{tabular}
}
\caption{How the text of PIEs differs from the text of all data points, according to 7 readability scores (plots (a)-(g)) and text length (plot (h)). 
Ratio between the scores of PIEs and the scores of all data points (y axis), across pruning thresholds (x axis), for BERT and IMDB.
The solid black horizontal line represents equal scores in PIEs and all data points. The solid turquoise line is the mean score of all pruners. Any line above the solid black line means that PIEs are harder to understand (plots (a)-(g)) or have longer text (plot (h)), on average, than all data points.
}
\label{fig:readability_IMDB_BERT}%
\end{figure}

\begin{figure}
\centering

\resizebox{\columnwidth}{!}{%
\begin{tabular}{cc}

 \includegraphics[width=0.5\textwidth]{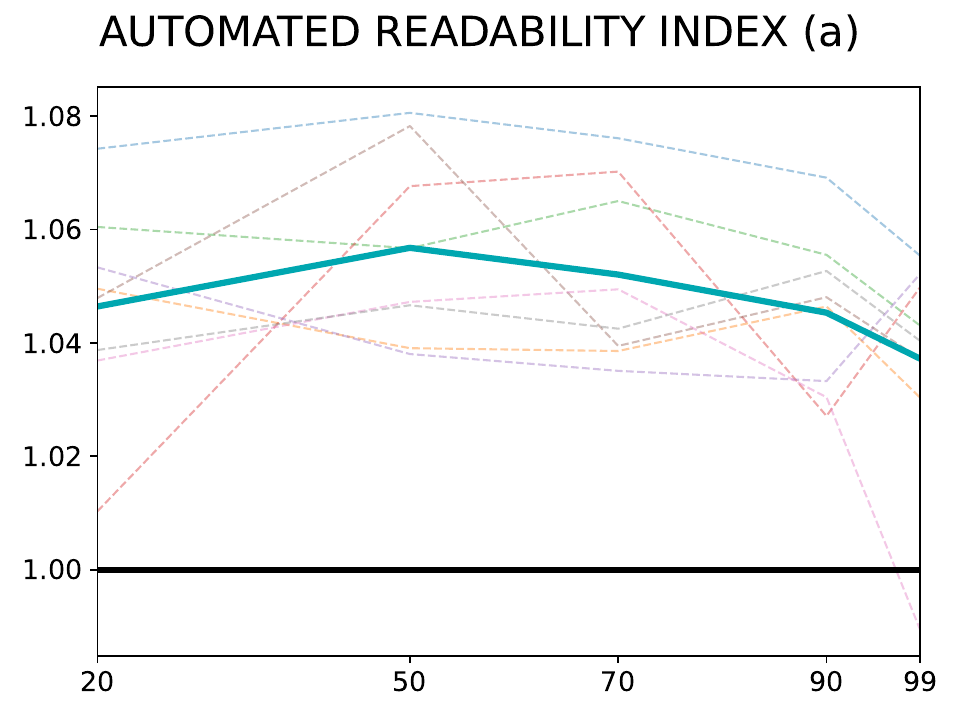} &
 \includegraphics[width=0.5\textwidth]{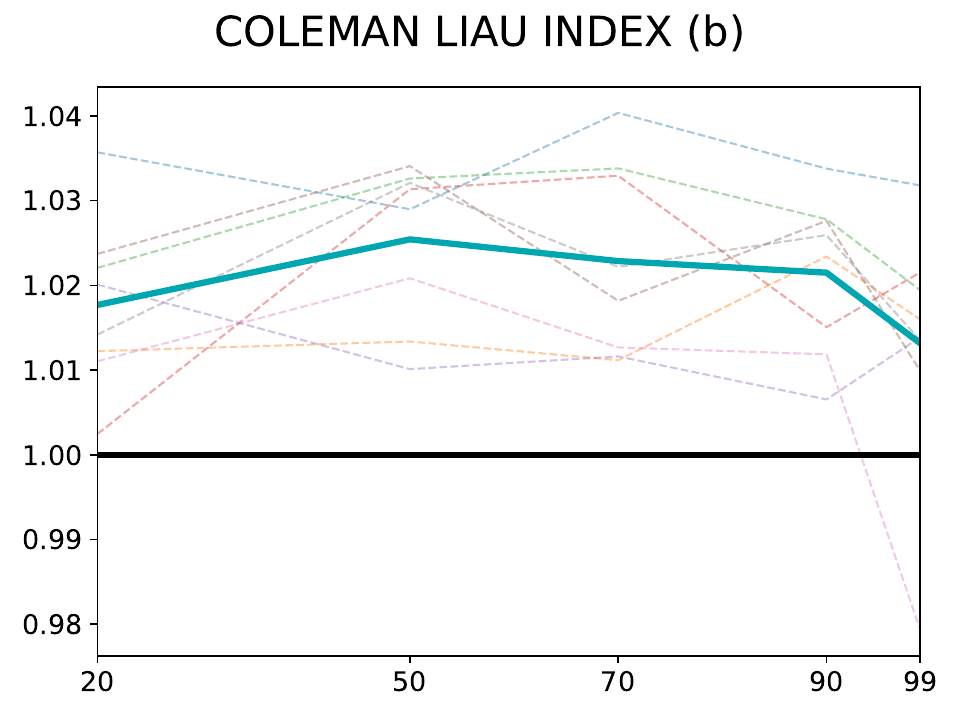} \\
 \includegraphics[width=0.5\textwidth]{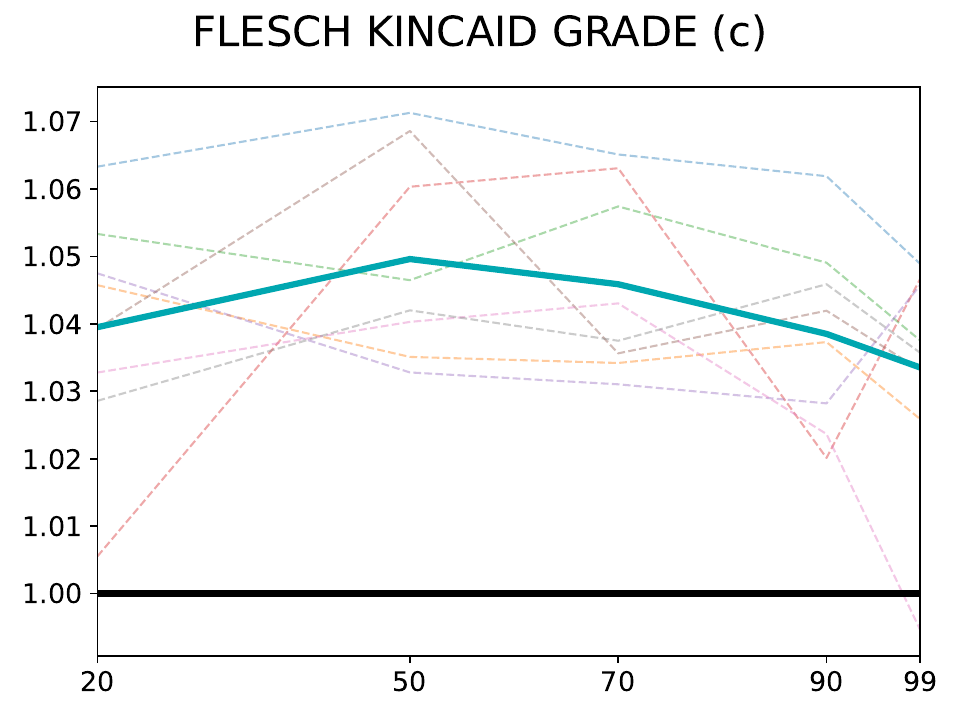} &
 \includegraphics[width=0.5\textwidth]{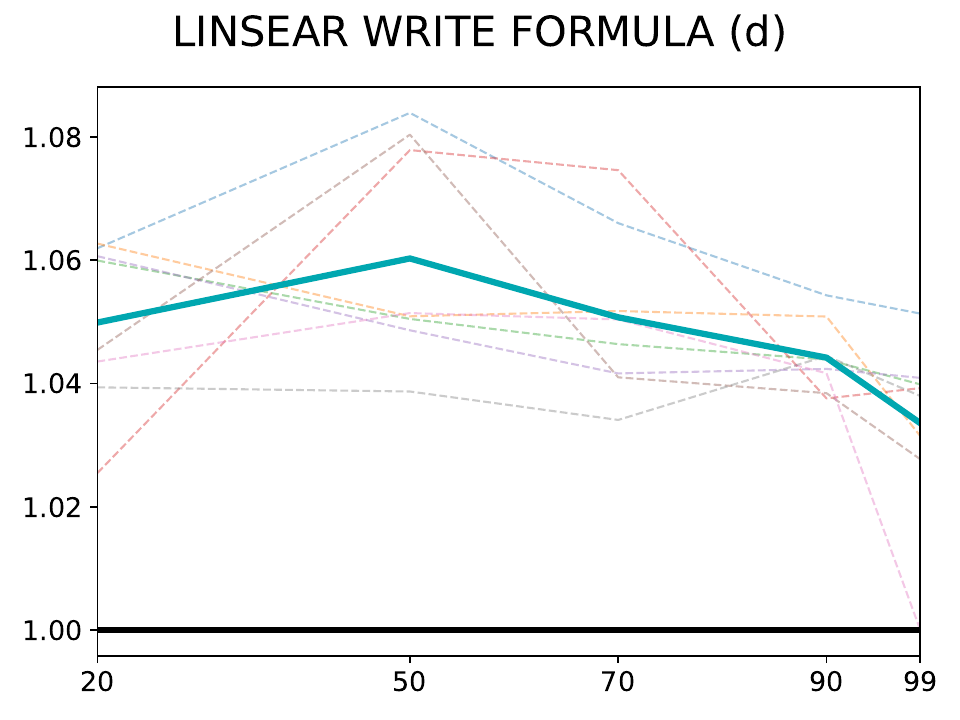} \\
 \includegraphics[width=0.5\textwidth]{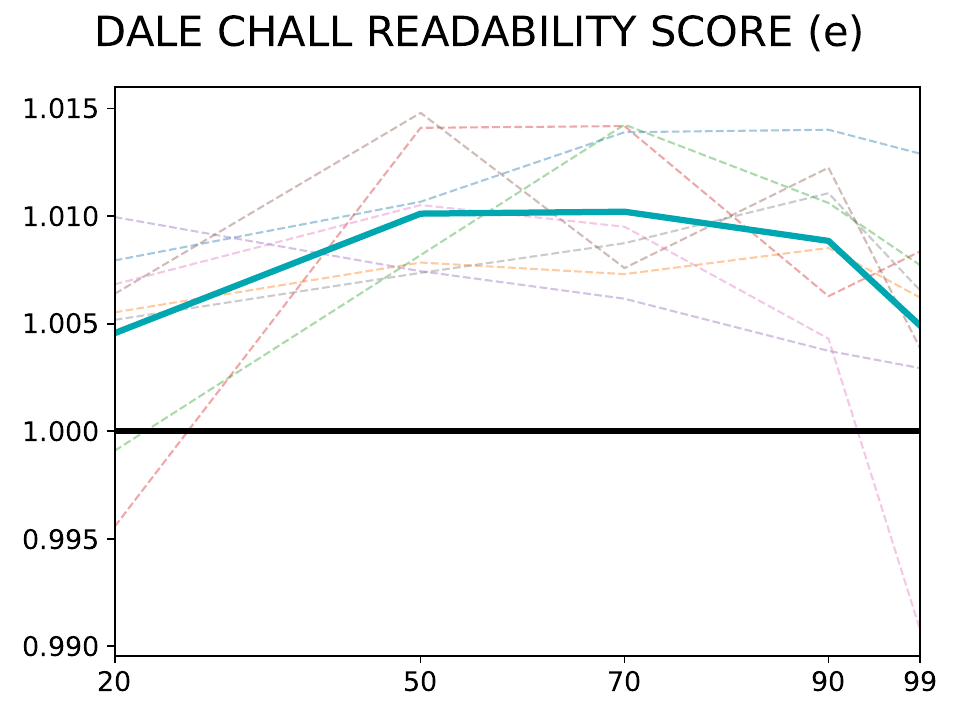} &
 \includegraphics[width=0.5\textwidth]{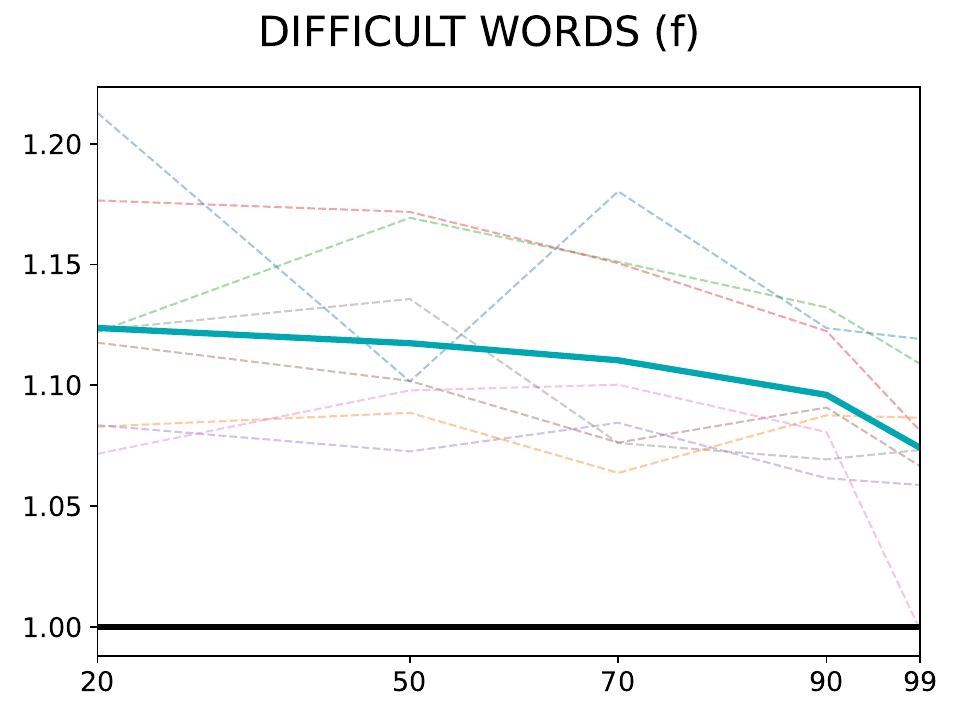} \\
 \includegraphics[width=0.5\textwidth]{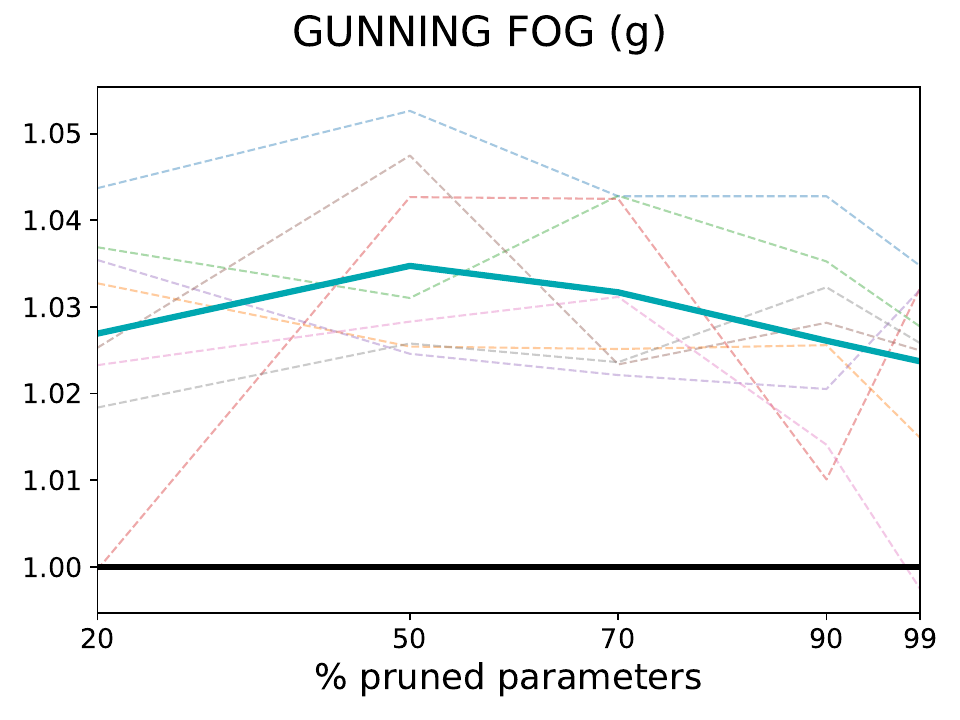} &
 \includegraphics[width=0.5\textwidth]{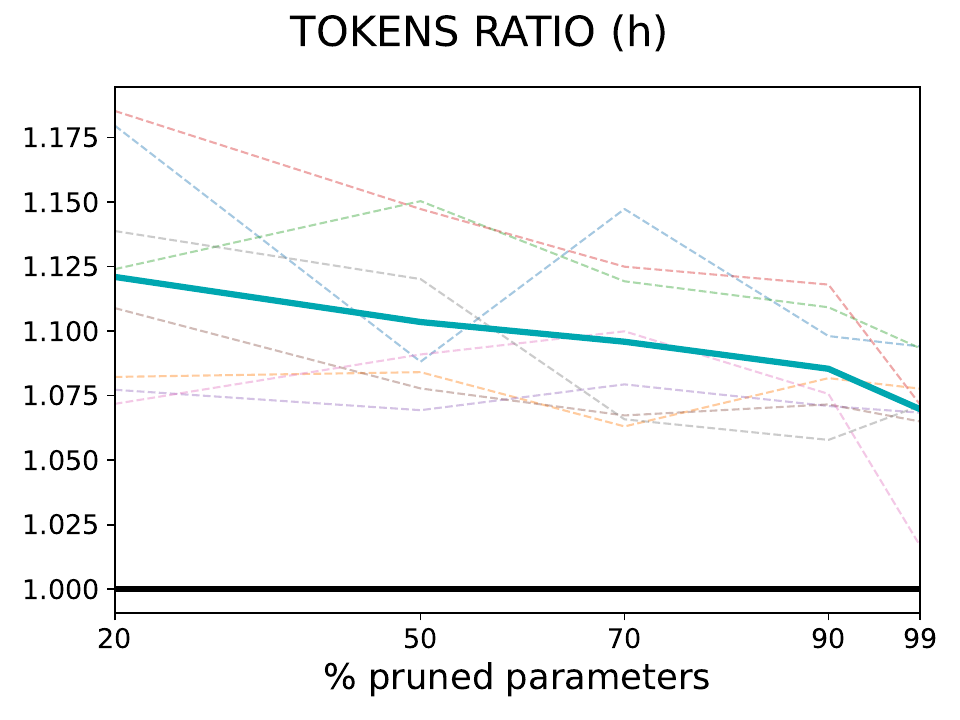} \\
 \multicolumn{2}{c}{\includegraphics[width=0.9\textwidth]{Images/PIEs/Text_length/legend.pdf}}    
\end{tabular}
}
\caption{How the text of PIEs differs from the text of all data points, according to 7 readability scores (plots (a)-(g)) and text length (plot (h)). 
Ratio between the scores of PIEs and the scores of all data points (y axis), across pruning thresholds (x axis), for BiLSTM and IMDB.
The solid black horizontal line represents equal scores in PIEs and all data points. The solid turquoise line is the mean score of all pruners. Any line above the solid black line means that PIEs are harder to understand (plots (a)-(g)) or have longer text (plot (h)), on average, than all data points. 
}
\label{fig:readability_IMDB_BiLSTM}%
\end{figure}

\begin{figure}
\centering

\resizebox{\columnwidth}{!}{%
\begin{tabular}{cc}

 \includegraphics[width=0.5\textwidth]{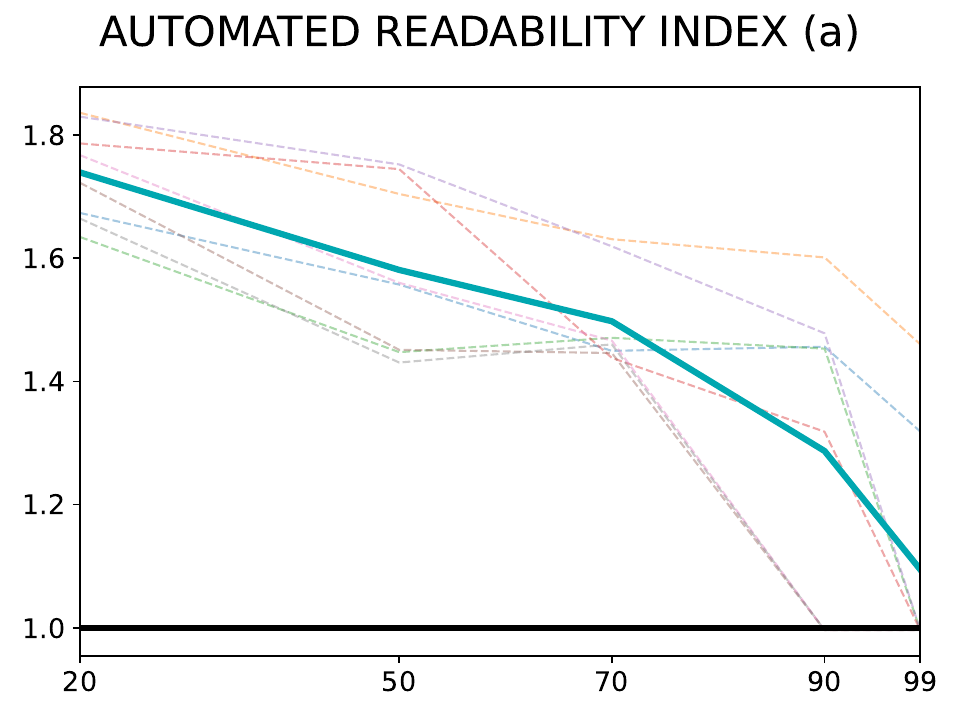} &
 \includegraphics[width=0.5\textwidth]{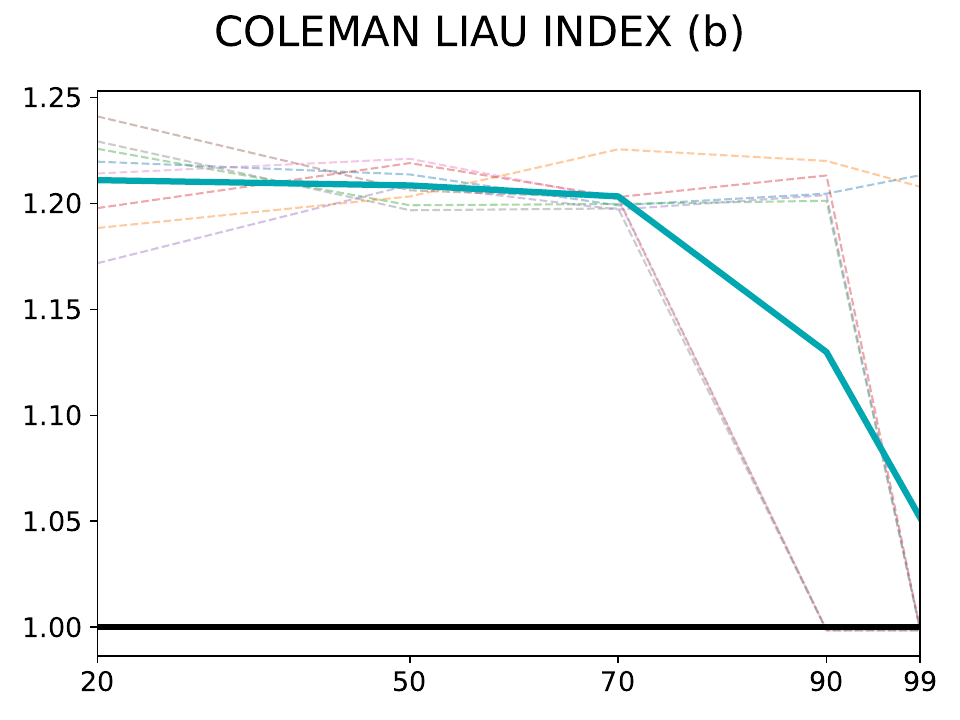} \\
 \includegraphics[width=0.5\textwidth]{Images/Readability_scores/flesch_kincaid_grade/SNLI_BERT_ALL_DIFF.pdf} &
 \includegraphics[width=0.5\textwidth]{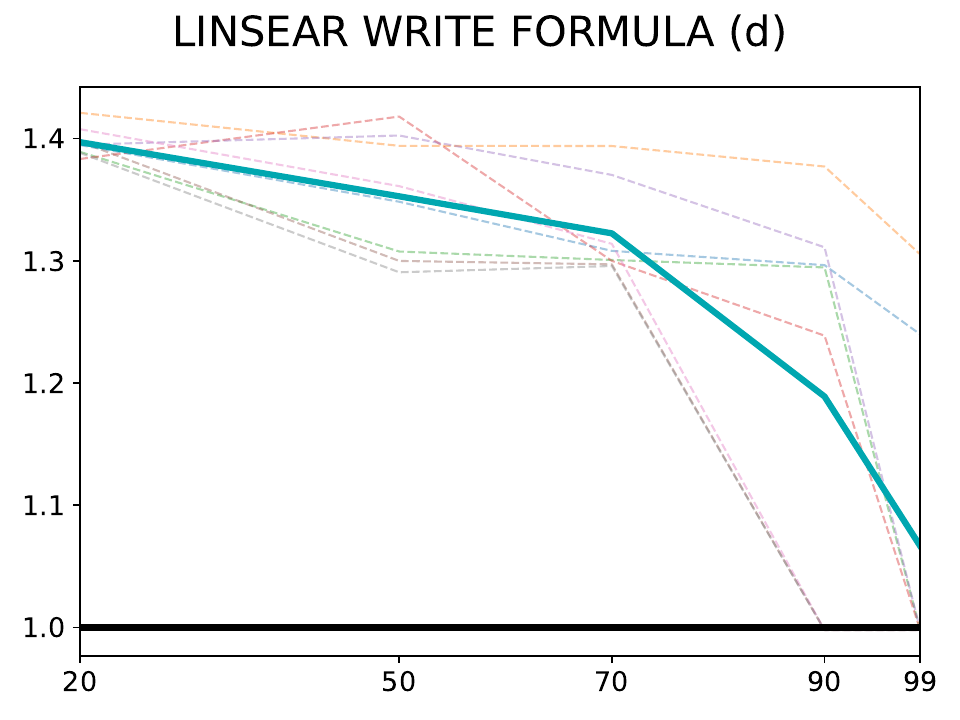} \\
 \includegraphics[width=0.5\textwidth]{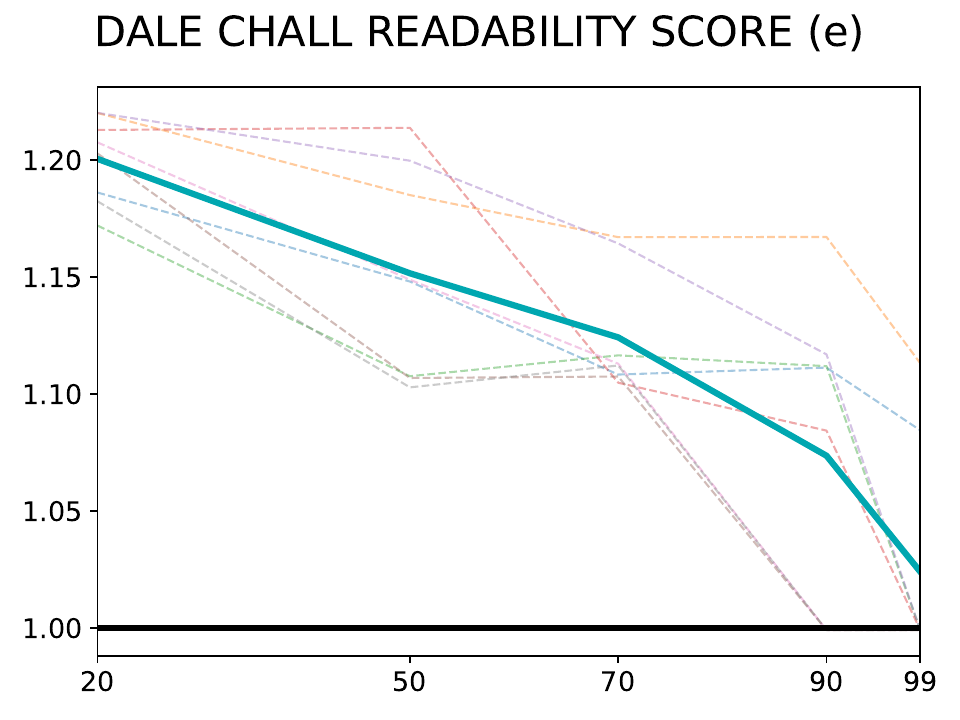} &
 \includegraphics[width=0.5\textwidth]{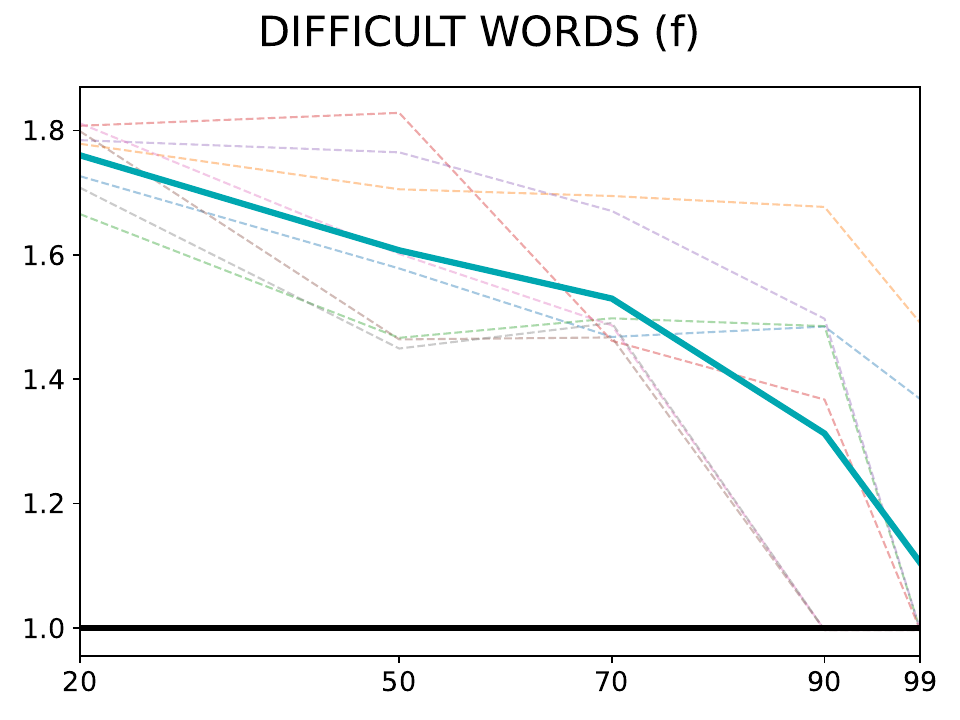} \\
 \includegraphics[width=0.5\textwidth]{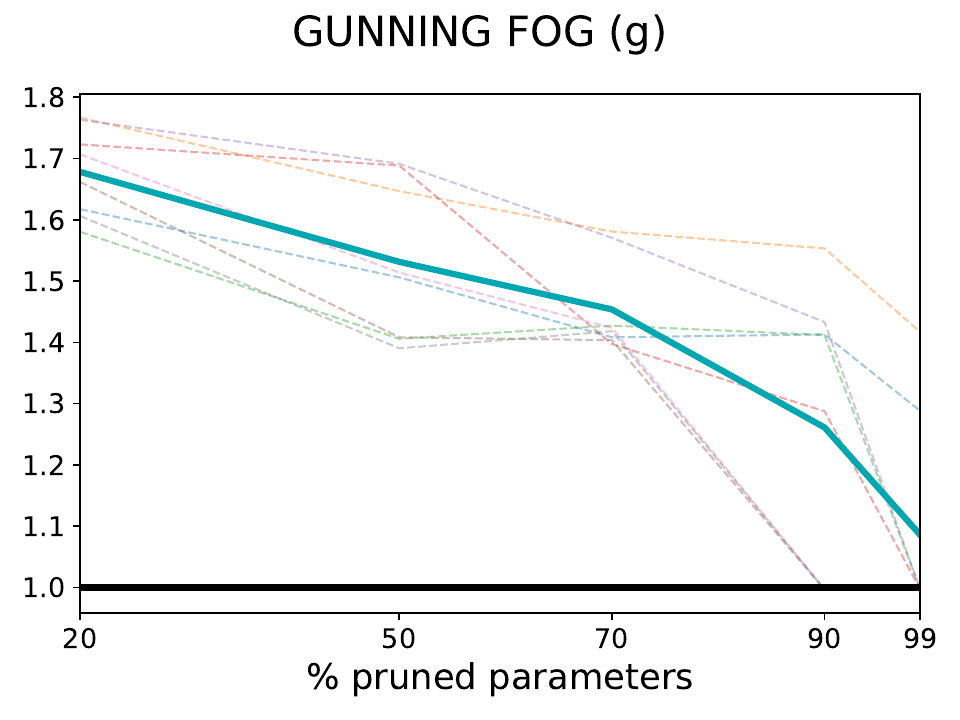} &
 \includegraphics[width=0.5\textwidth]{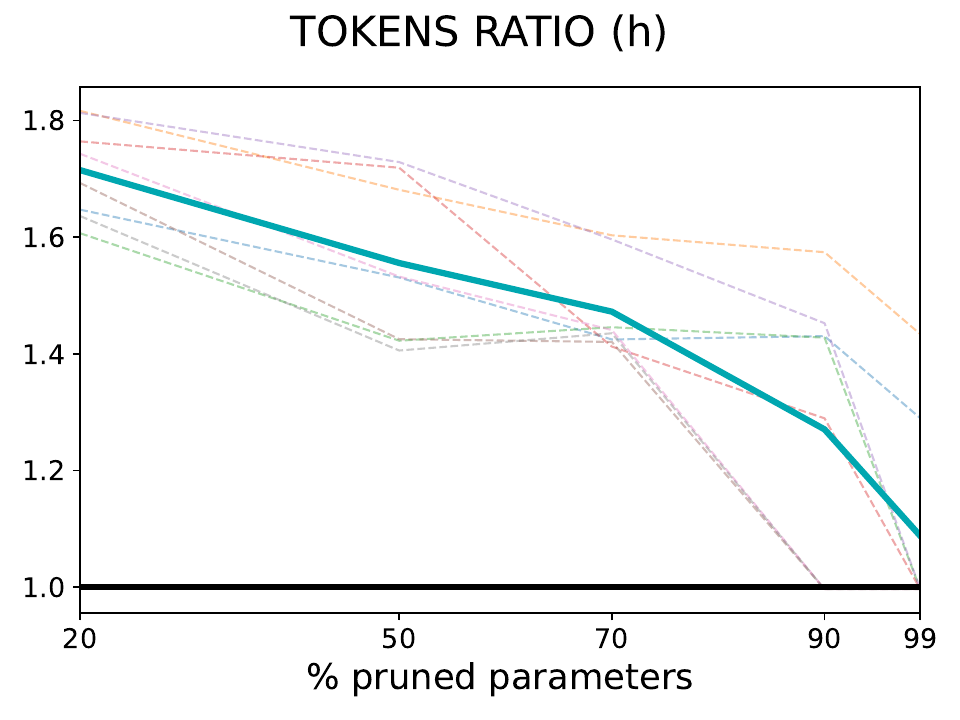} \\
 
 \multicolumn{2}{c}{\includegraphics[width=0.9\textwidth]{Images/PIEs/Text_length/legend.pdf}}    
\end{tabular}
}
\caption{How the text of PIEs differs from the text of all data points, according to 7 readability scores (plots (a)-(g)) and text length (plot (h)). 
Ratio between the scores of PIEs and the scores of all data points (y axis), across pruning thresholds (x axis), for BERT and Reuters.
The solid black horizontal line represents equal scores in PIEs and all data points. The solid turquoise line is the mean score of all pruners. Any line above the solid black line means that PIEs are harder to understand (plots (a)-(g)) or have longer text (plot (h)), on average, than all data points.
}
\label{fig:readability_Reuters_BERT}%
\end{figure}

\begin{figure}
\centering

\resizebox{\columnwidth}{!}{%
\begin{tabular}{ccccc}

 \includegraphics[width=0.5\textwidth]{Images/Readability_scores/automated_readability_index/Reuters_BERT_ALL_DIFF.pdf} &
 \includegraphics[width=0.5\textwidth]{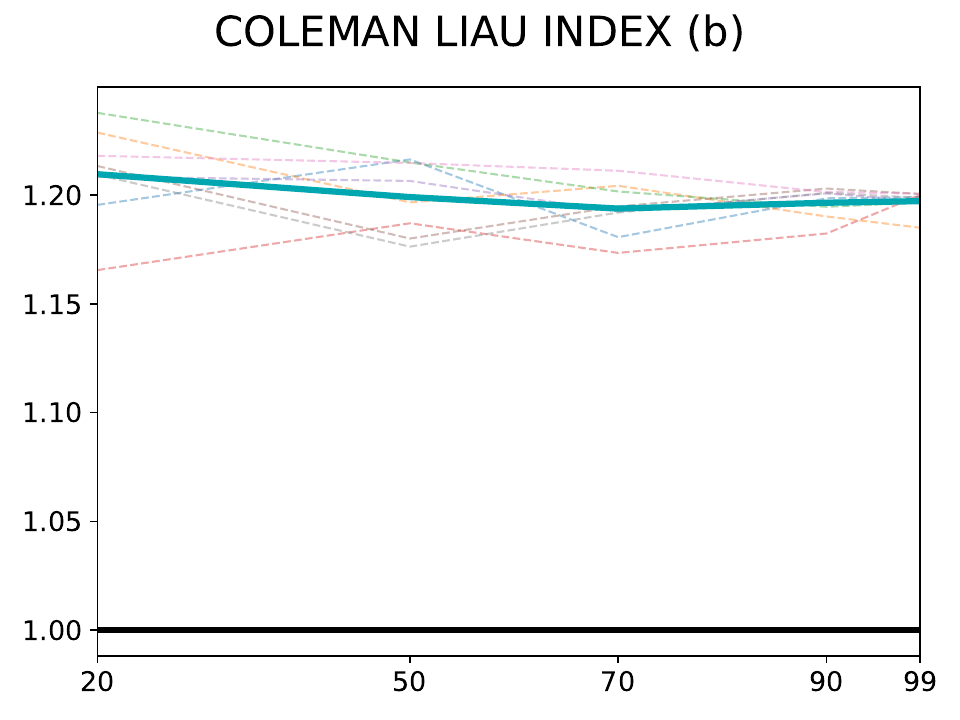} \\
 \includegraphics[width=0.5\textwidth]{Images/Readability_scores/flesch_kincaid_grade/SNLI_BERT_ALL_DIFF.pdf} &
 \includegraphics[width=0.5\textwidth]{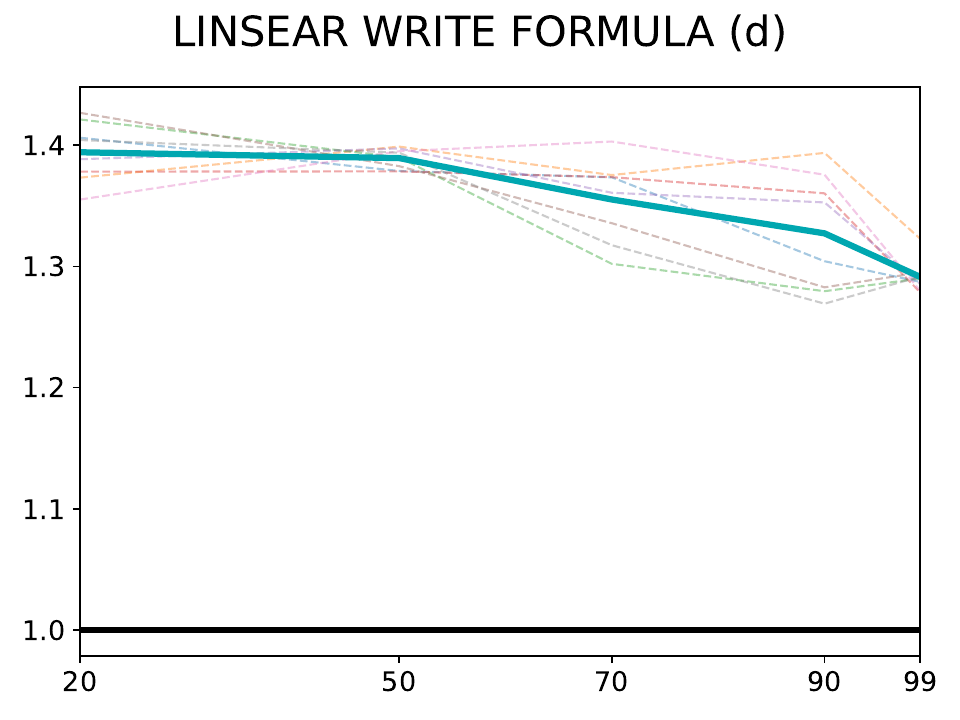} \\
 \includegraphics[width=0.5\textwidth]{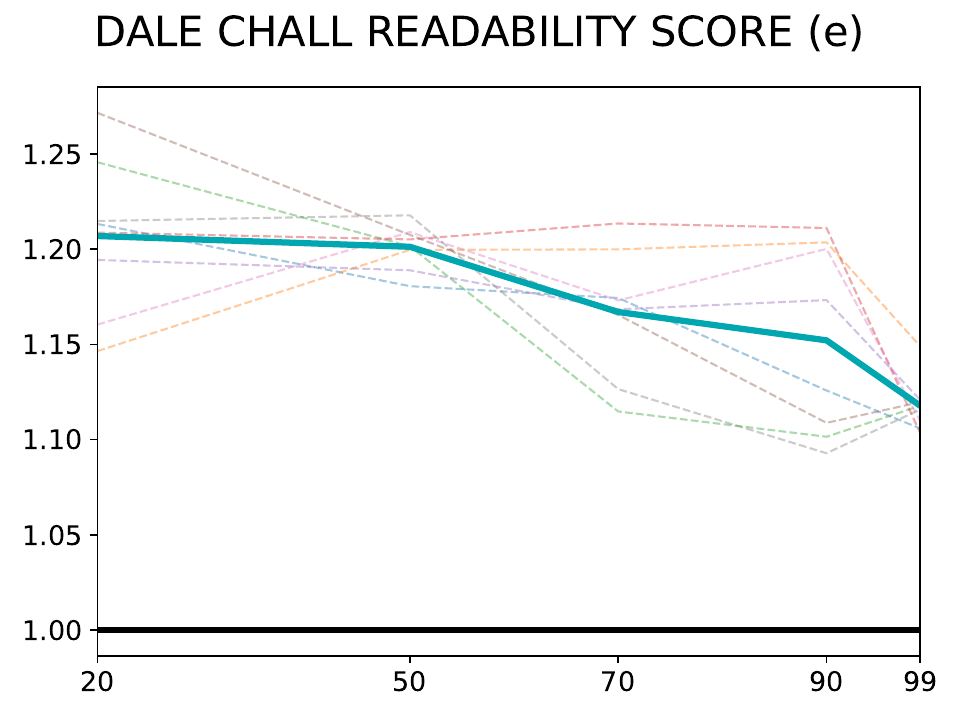} &
 \includegraphics[width=0.5\textwidth]{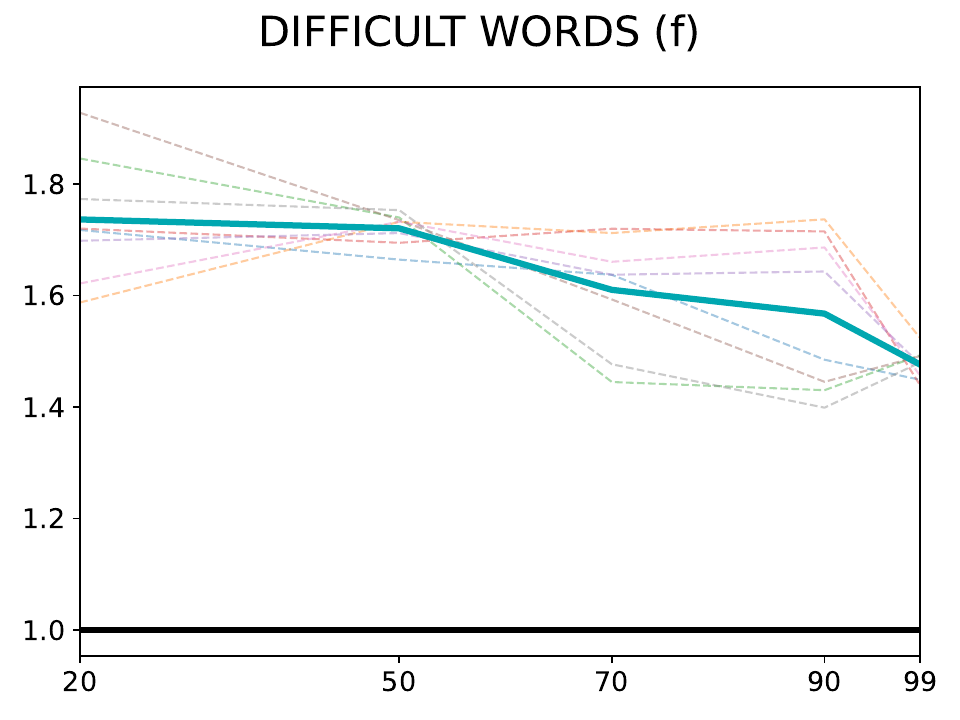} \\
 \includegraphics[width=0.5\textwidth]{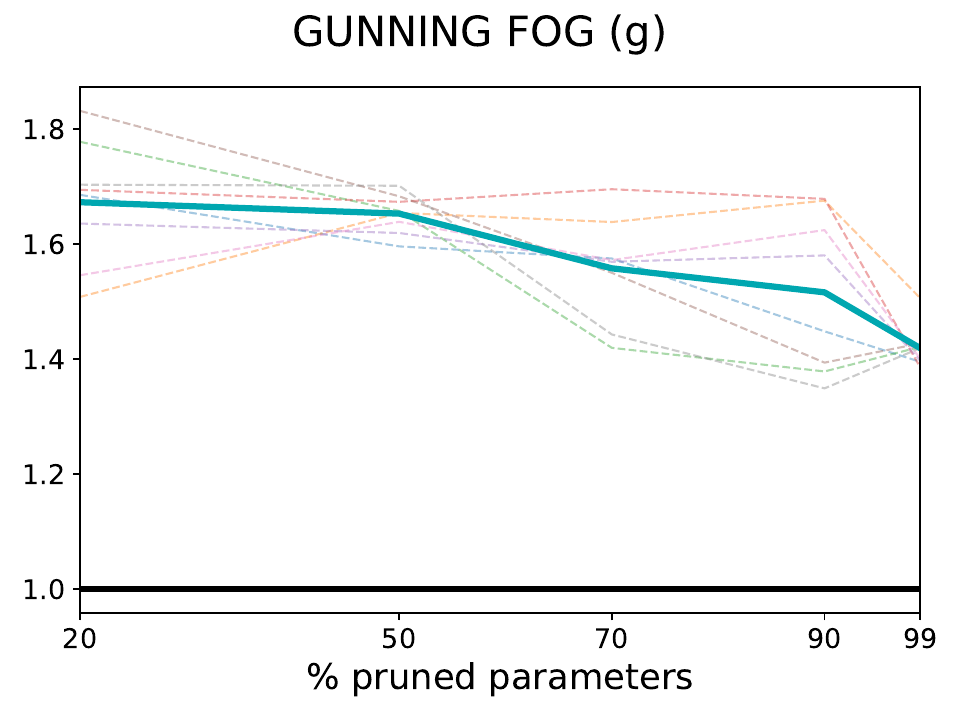} &
 \includegraphics[width=0.5\textwidth]{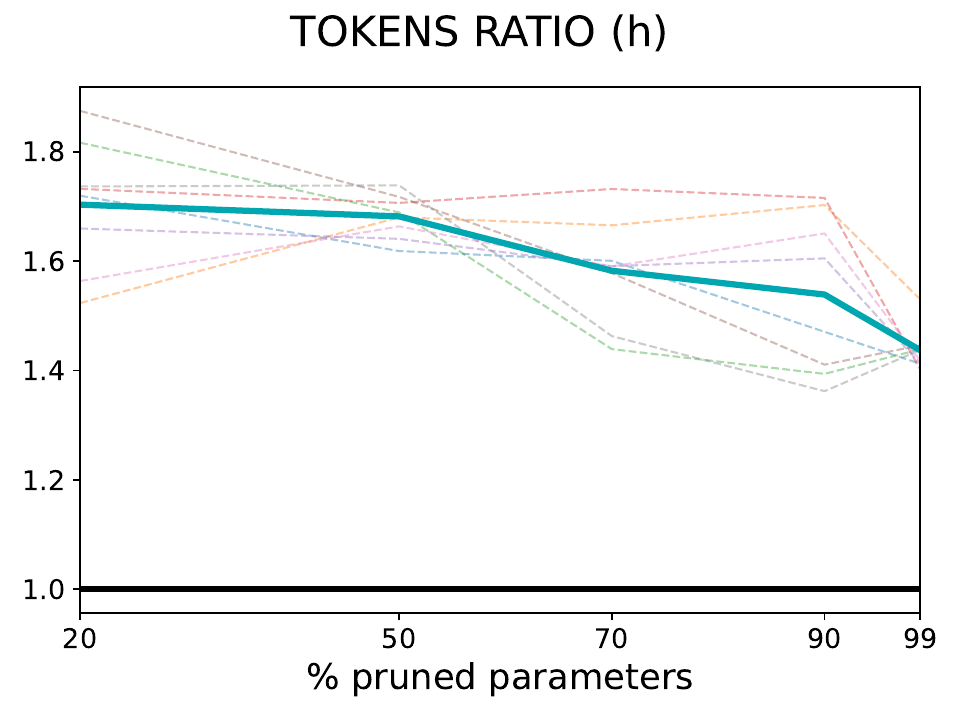} \\
 \multicolumn{2}{c}{\includegraphics[width=0.9\textwidth]{Images/PIEs/Text_length/legend.pdf}}    
\end{tabular}
}
\caption{How the text of PIEs differs from the text of all data points, according to 7 readability scores (plots (a)-(g)) and text length (plot (h)). 
Ratio between the scores of PIEs and the scores of all data points (y axis), across pruning thresholds (x axis), for BiLSTM and Reuters.
The solid black horizontal line represents equal scores in PIEs and all data points. The solid turquoise line is the mean score of all pruners. Any line above the solid black line means that PIEs are harder to understand (plots (a)-(g)) or have longer text (plot (h)), on average, than all data points. 
}
\label{fig:readability_Reuters_BiLSTM}%
\end{figure}

\begin{figure}
\centering

\resizebox{\columnwidth}{!}{%
\begin{tabular}{cc}

 \includegraphics[width=0.5\textwidth]{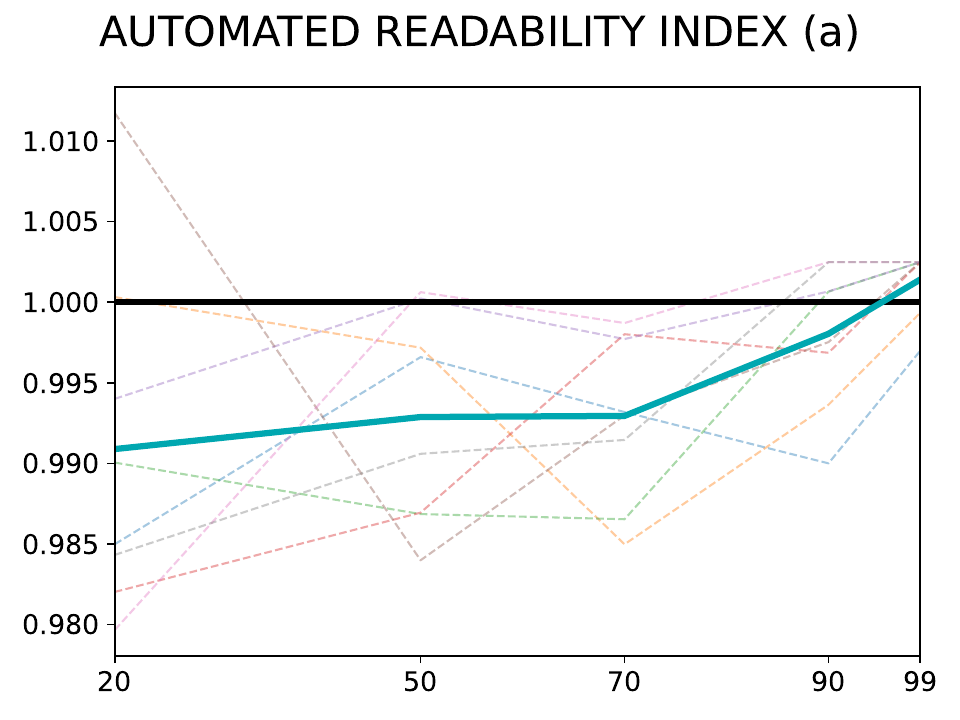} &
 \includegraphics[width=0.5\textwidth]{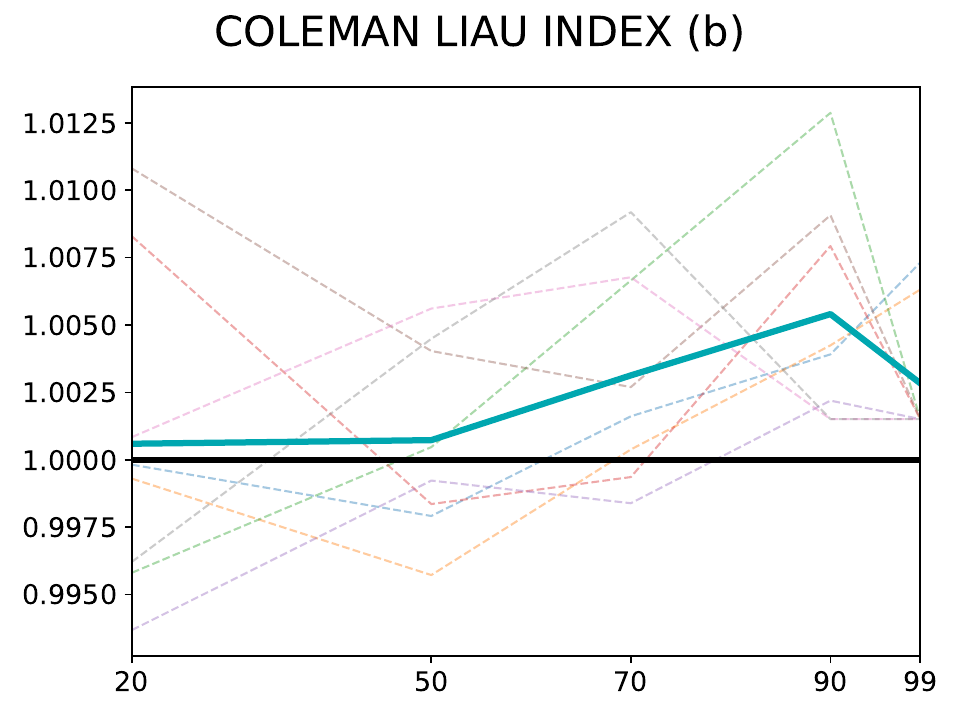} \\
 \includegraphics[width=0.5\textwidth]{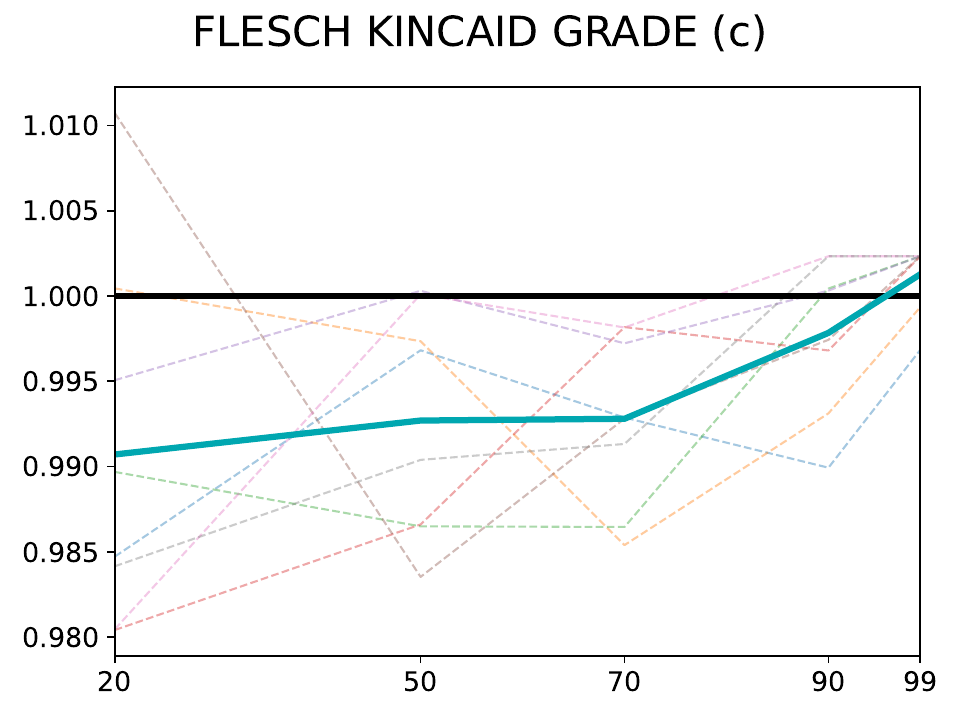} &
 \includegraphics[width=0.5\textwidth]{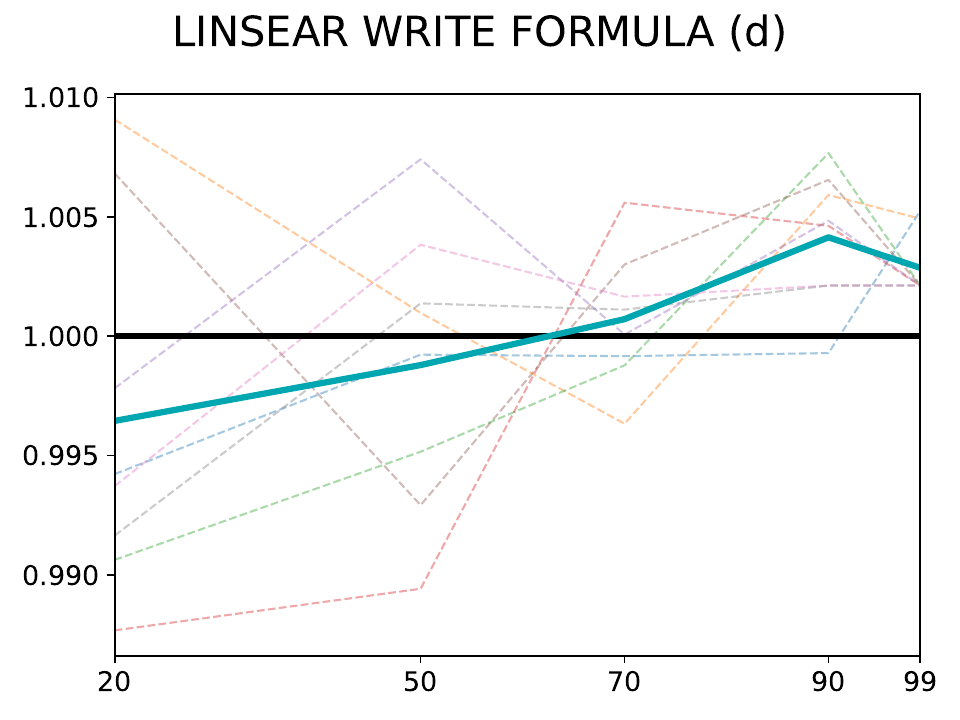} \\
 \includegraphics[width=0.5\textwidth]{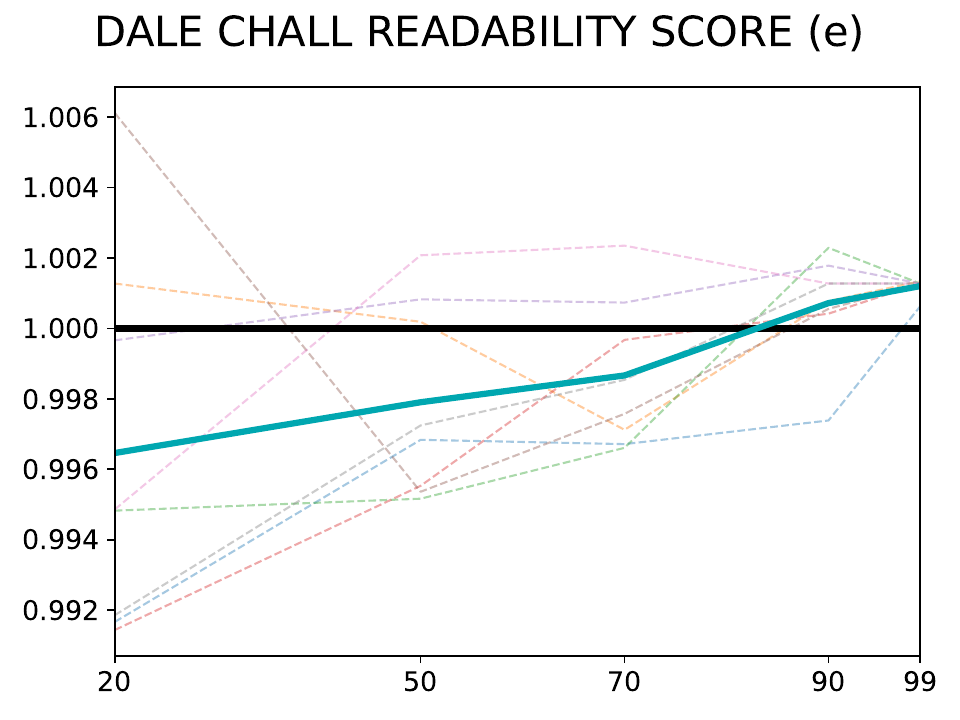} &
 \includegraphics[width=0.5\textwidth]{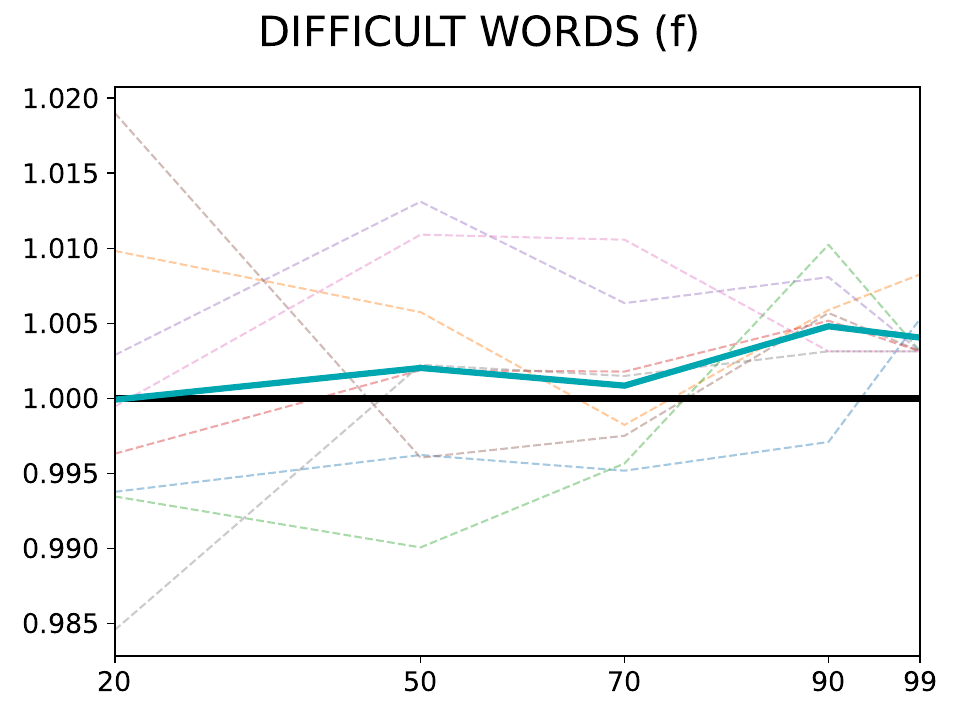} \\
 \includegraphics[width=0.5\textwidth]{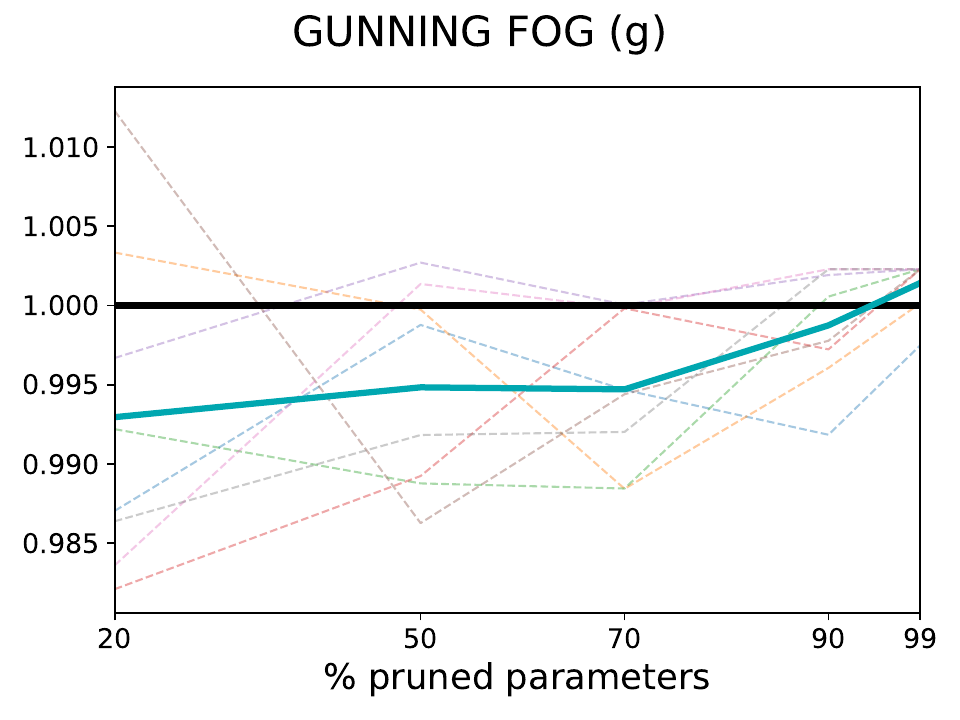} &
 \includegraphics[width=0.5\textwidth]{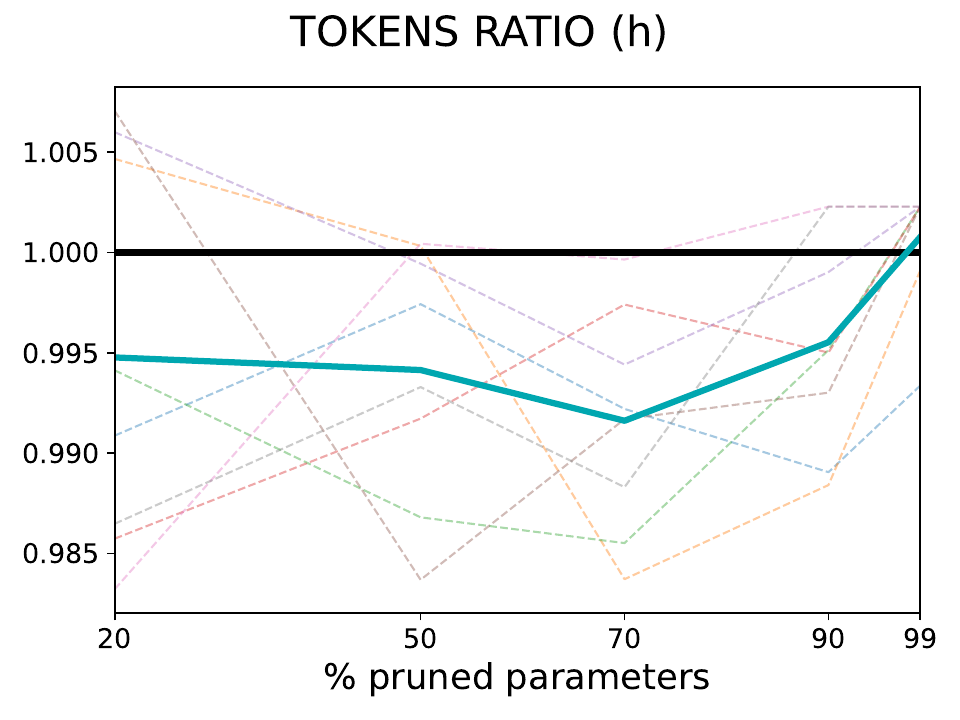} \\
 \multicolumn{2}{c}{\includegraphics[width=0.9\textwidth]{Images/PIEs/Text_length/legend.pdf}}    
\end{tabular}
}
\caption{How the text of PIEs differs from the text of all data points, according to 7 readability scores (plots (a)-(g)) and text length (plot (h)). 
Ratio between the scores of PIEs and the scores of all data points (y axis), across pruning thresholds (x axis), for BERT and AAPD.
The solid black horizontal line represents equal scores in PIEs and all data points. The solid turquoise line is the mean score of all pruners. Any line above the solid black line means that PIEs are harder to understand (plots (a)-(g)) or have longer text (plot (h)), on average, than all data points. 
}
\label{fig:readability_AAPD_BERT}%
\end{figure}

\begin{figure}
\centering

\resizebox{\columnwidth}{!}{%
\begin{tabular}{cc}

 \includegraphics[width=0.5\textwidth]{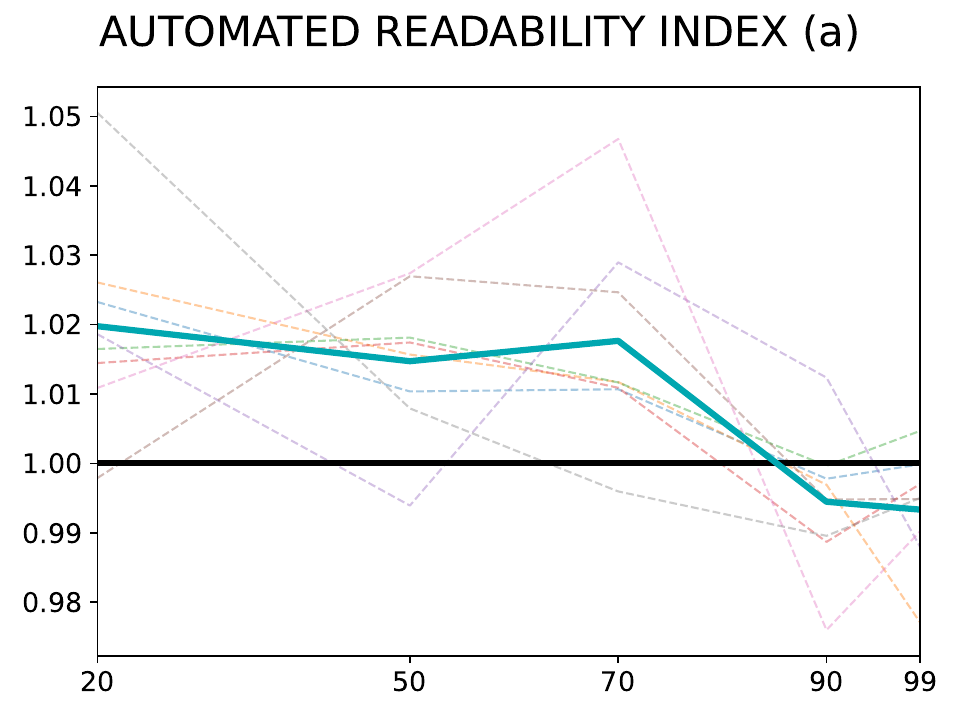} &
 \includegraphics[width=0.5\textwidth]{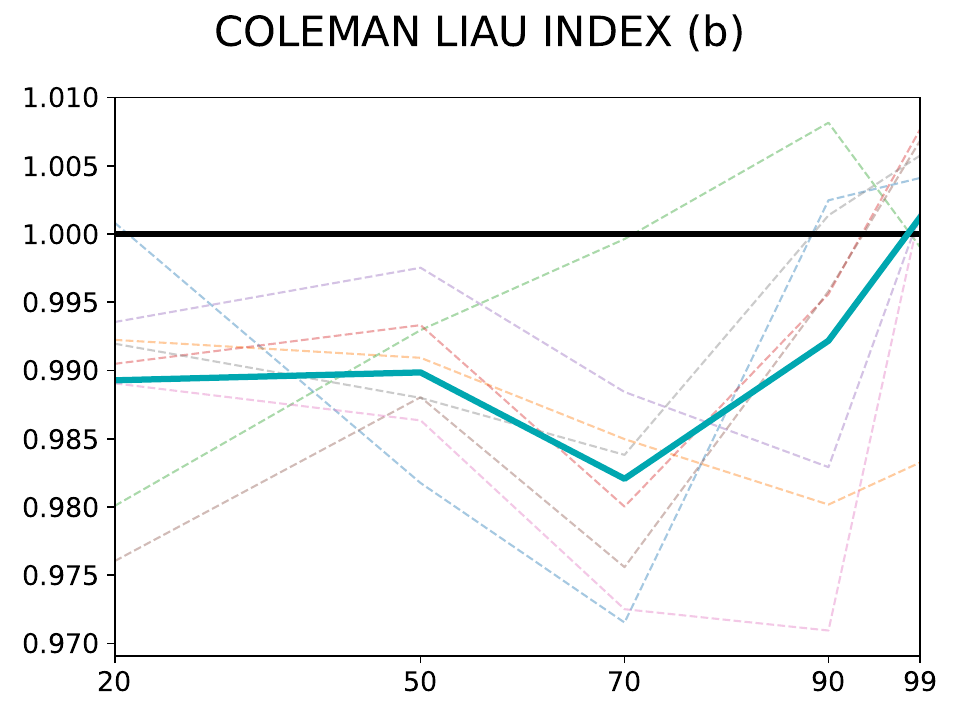} \\
 \includegraphics[width=0.5\textwidth]{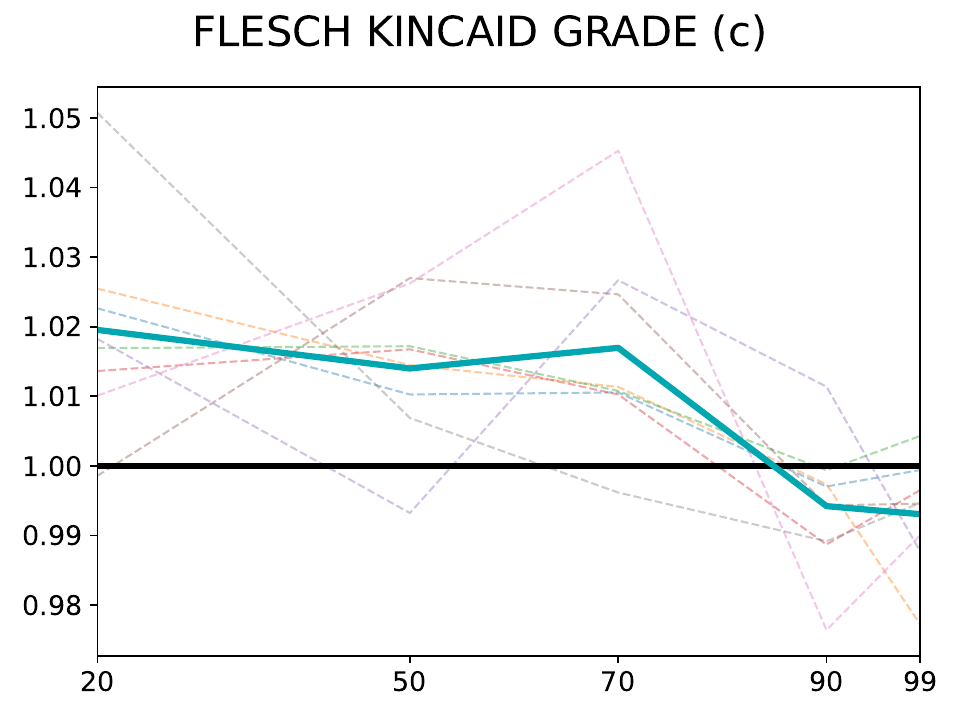} &
 \includegraphics[width=0.5\textwidth]{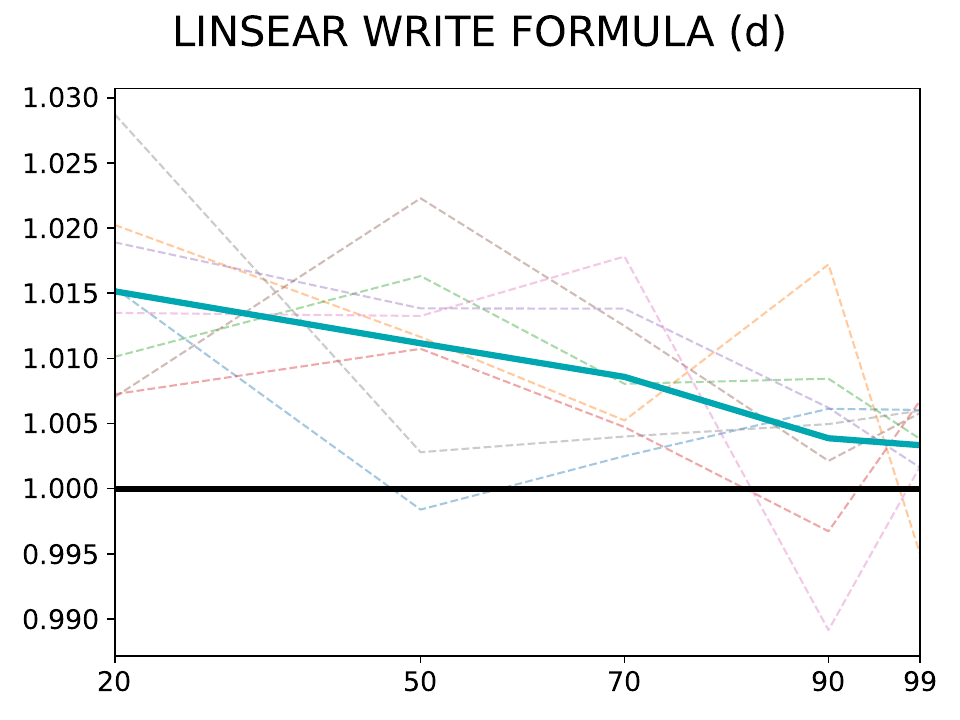} \\
 \includegraphics[width=0.5\textwidth]{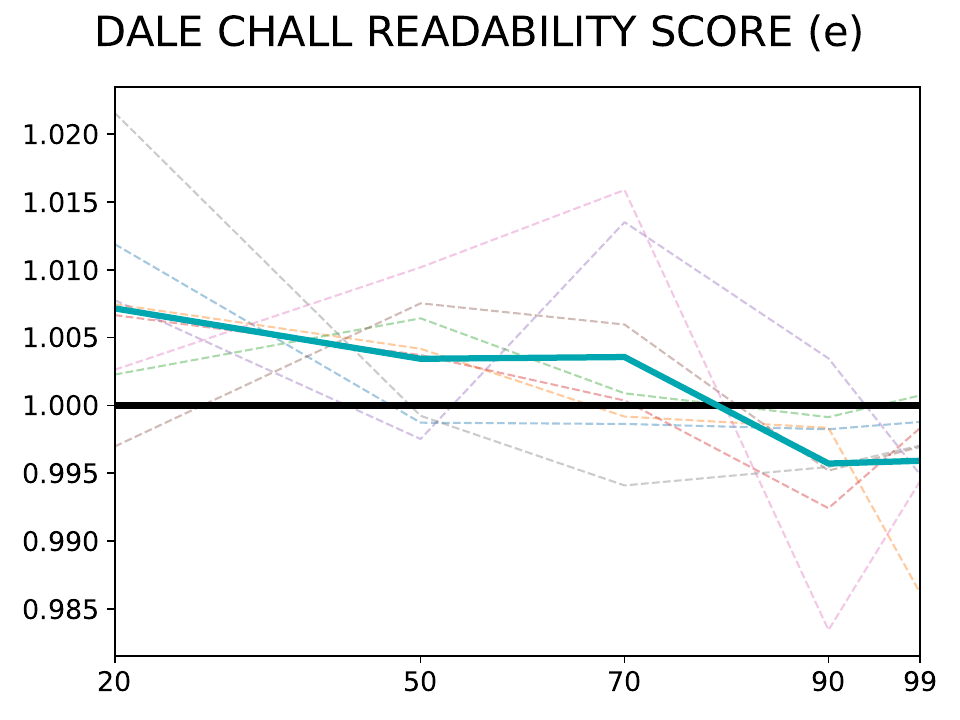} &
 \includegraphics[width=0.5\textwidth]{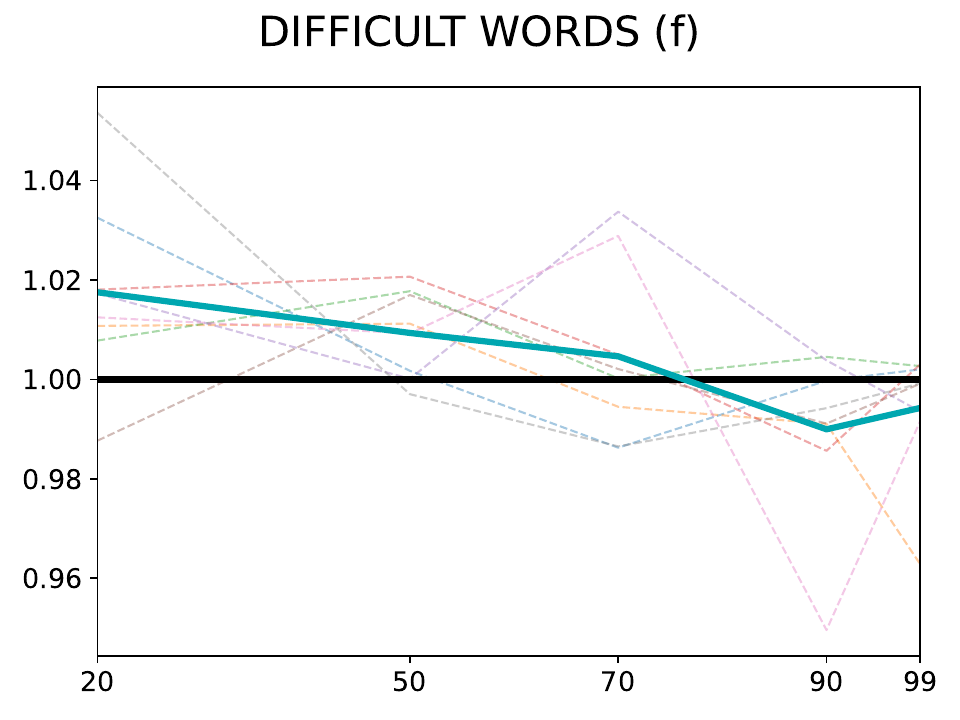} \\
 \includegraphics[width=0.5\textwidth]{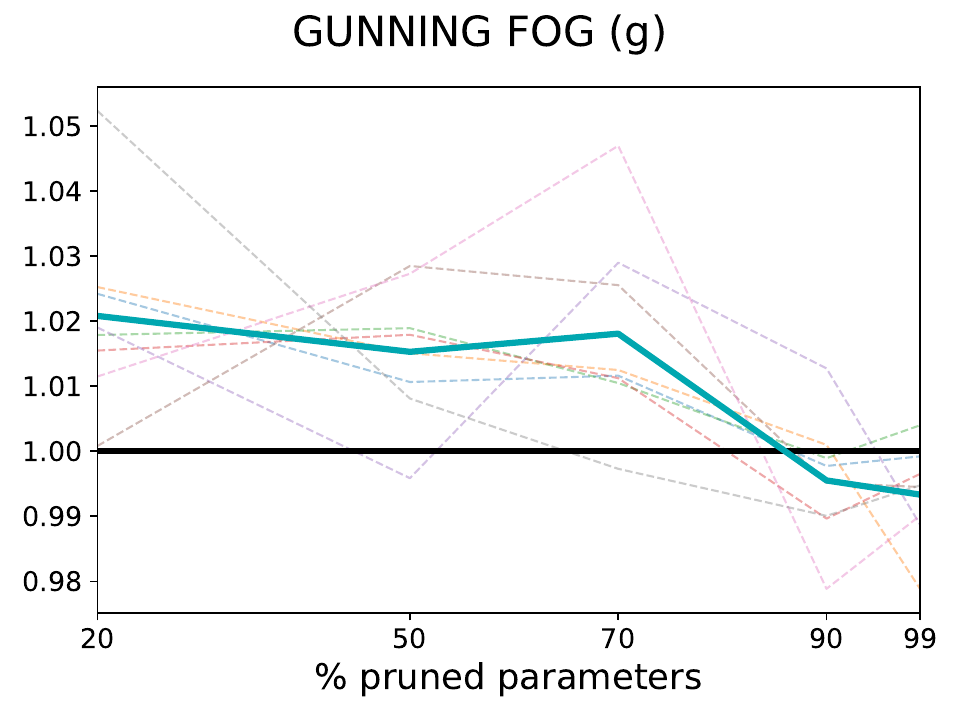} &
 \includegraphics[width=0.5\textwidth]{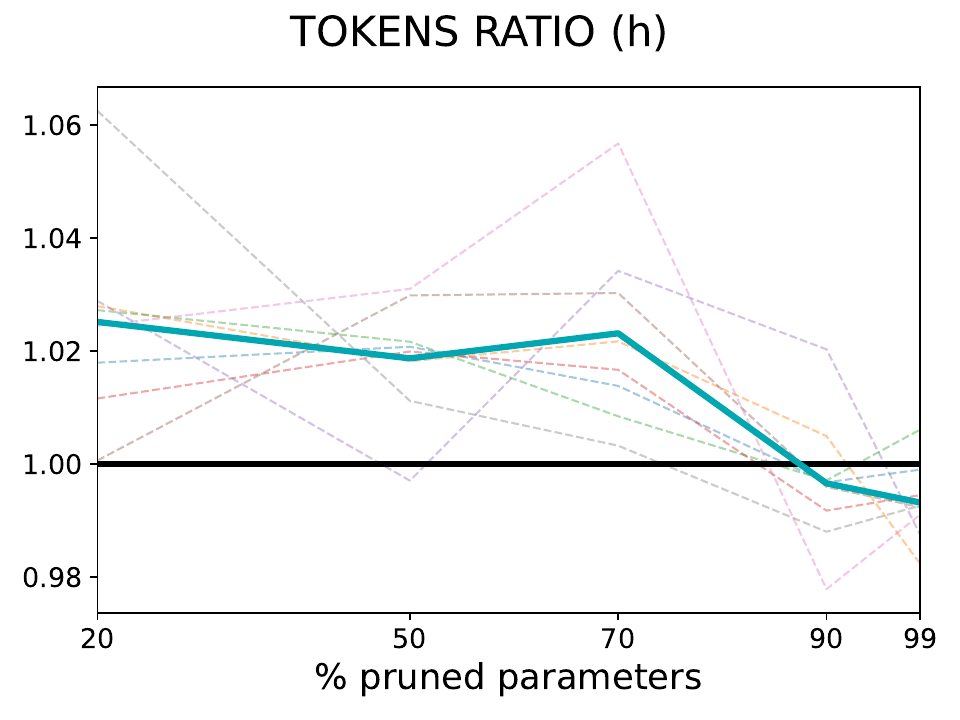} \\
 \multicolumn{2}{c}{\includegraphics[width=0.9\textwidth]{Images/PIEs/Text_length/legend.pdf}}    
\end{tabular}
}
\caption{How the text of PIEs differs from the text of all data points, according to 7 readability scores (plots (a)-(g)) and text length (plot (h)). 
Ratio between the scores of PIEs and the scores of all data points (y axis), across pruning thresholds (x axis), for BiLSTM and AAPD.
The solid black horizontal line represents equal scores in PIEs and all data points. The solid turquoise line is the mean score of all pruners. Any line above the solid black line means that PIEs are harder to understand (plots (a)-(g)) or have longer text (plot (h)), on average, than all data points.
}
\label{fig:readability_AAPD_BiLSTM}%
\end{figure}

\end{document}

%% file: Tables/PIEs/Examples/PIEs_examples.tex
\resizebox{0.955\columnwidth}{!}{%
\begin{tabular}{|l|}
\hline
\textbf{Class:} \colorbox{lime}{neutral} \\ \hline
\begin{tabular}[c]{@{}l@{}}\textbf{Premise:}\\ A group of seven individuals wearing rafting gear,\\white water raft down a river.\\ \textbf{Hypothesis:}\\ Seven men and women are in a yellow boat.\end{tabular} \\ \hline
\textbf{Unpruned model prediction:} entailment \\ \hline
\textbf{Pruned model prediction:} \colorbox{lime}{neutral} \\ \specialrule{.15em}{.075em}{.075em} 

\textbf{Class:} \colorbox{lime}{entailment} \\ \hline
\begin{tabular}[c]{@{}l@{}}\textbf{Premise:}\\ A woman is painting a mural of a woman's face.\\ \textbf{Hypothesis:}\\ There is a woman painting.\end{tabular} \\ \hline
\textbf{Unpruned model prediction:} \colorbox{lime}{entailment} \\ \hline
\textbf{Pruned model prediction:} contradiction \\ \hline
\end{tabular}
}

%% file: Tables/Setup/Dataset/stats.tex
\resizebox{\columnwidth}{!}{%
\begin{tabular}{lrrrrc}
\textbf{Dataset} & \textbf{\# train} & \textbf{\# test} & \textbf{\# val} & \textbf{\# classes} & \textbf{Classification} \\ \hline
IMDB    & 20000  & 25000 & 5000 &  2  & single-label \\
SNLI    & 549367 & 9824  & 9842 &   3 & single-label \\
Reuters & 6737   & 1429  & 1440 &  23 & multi-label \\
AAPD    & 53840  & 1000  & 1000 &  54 & multi-label \\ \hline
\end{tabular}
}

%% file: Tables/Setup/Pruning_algorithms/algorithms.tex
\centering
\resizebox{0.99\columnwidth}{!}{%
\begin{tabular}{@{}l|ccc@{}}
\textbf{\begin{tabular}[c]{@{}l@{}}Scoring $\rightarrow$\\ Scheduling $\downarrow$\end{tabular}} & \textbf{Impact} & \textbf{Magnitude} & \textbf{Random} \\ \midrule

\begin{tabular}[l]{@{}l@{}}\textbf{Iterative + Weight} \\ \textbf{Rewinding} \end{tabular} & IIBP-WR & IMP-WR & - \\

\begin{tabular}[l]{@{}l@{}}\textbf{Iterative + Fine} \\ \textbf{tuning} \end{tabular}  & IIBP-FT & IMP-FT & IRP-FT \\

\textbf{At Initialization} & IBP-AI & MP-AI & RP-AI \\

\end{tabular}
}

%% file: Tables/PIEs/merged_PIEs_percentage_orginal.tex
\resizebox{0.78\columnwidth}{!}{%
\begin{tabular}{lclrrrrr}
\multicolumn{8}{c}{\textbf{Single-label}} \\
 & \multicolumn{1}{l}{} & \multicolumn{1}{c}{\textbf{Pruner}} & \textbf{20\%} & \textbf{50\%} & \textbf{70\%} & \textbf{90\%} & \textbf{99\%} \\ \hline
\multirow{16}{*}{\begin{rotate}{90}IMDB\end{rotate}} & \multirow{8}{*}{\begin{rotate}{90}BERT\end{rotate}} & IIBP-WR & 7 & 10 & 10 & \cellcolor{green!25}10 & 11 \\
\multicolumn{1}{c}{} &  & IIBP-FT & 4 & 10 & 13 & 13 & \cellcolor{green!25}11 \\
\multicolumn{1}{c}{} &  & IBP-AI & 8 & 10 & 10 & 10 & 50 \\
\multicolumn{1}{c}{} &  & IMP-WR & 4 & \cellcolor{green!25}9 & \cellcolor{green!25}11 & 12 & \cellcolor{gray!25}50 \\
\multicolumn{1}{c}{} &  & IMP-FT & \cellcolor{green!25}3 & 10 & 14 & 13 & \cellcolor{gray!25}50 \\
\multicolumn{1}{c}{} &  & MP-AI & 6 & 10 & 10 & 12 & \cellcolor{gray!25}50 \\
\multicolumn{1}{c}{} &  & IRP-FT & 8 & \cellcolor{gray!25}13 & \cellcolor{gray!25}12 &\cellcolor{gray!25} 50 &\cellcolor{gray!25} 50 \\
\multicolumn{1}{c}{} &  & RP-AI & \cellcolor{gray!25}10 & 10 & 10 & \cellcolor{gray!25}50 & \cellcolor{gray!25}50 \\ \cline{2-8} 
\multicolumn{1}{c}{} & \multirow{8}{*}{\begin{rotate}{90}BiLSTM\end{rotate}} & IIBP-WR & 2 & 3 & 4 & 7 & 10 \\
\multicolumn{1}{c}{} &  & IIBP-FT & 5 & 5 & 5 & 5 & \cellcolor{green!25}3 \\
\multicolumn{1}{c}{} &  & IBP-AI & 2 & \cellcolor{gray!25}5 & \cellcolor{gray!25}7 & \cellcolor{gray!25}10 & 16 \\
\multicolumn{1}{c}{} &  & IMP-WR & 2 & 3 & 4 & 3 & 11 \\
\multicolumn{1}{c}{} &  & IMP-FT & \cellcolor{green!25}5 & \cellcolor{green!25}5 & \cellcolor{green!25}5 & \cellcolor{green!25}5 & 5 \\
\multicolumn{1}{c}{} &  & MP-AI & \cellcolor{gray!25}2 & 5 & 6 & 9 & 16 \\
\multicolumn{1}{c}{} &  & IRP-FT & 5 & 5 & 5 & 3 & \cellcolor{gray!25}29 \\
\multicolumn{1}{c}{} &  & RP-AI & \cellcolor{gray!25}2 & \cellcolor{gray!25}4 & 6 & 9 & 18 \\ \hline
\multicolumn{1}{c}{\multirow{16}{*}{\begin{rotate}{90}SNLI\end{rotate}}} & \multirow{8}{*}{\begin{rotate}{90}BERT\end{rotate}} & IIBP-WR & \cellcolor{gray!25}7 & 13 & 16 & 27 & 35 \\
\multicolumn{1}{c}{} &  & IIBP-FT & 3 &\cellcolor{green!25} 5 &\cellcolor{green!25} 8 &\cellcolor{green!25} 13 & \cellcolor{green!25}28 \\
\multicolumn{1}{c}{} &  & IBP-AI & 6 & 12 & 18 & 34 & 47 \\
\multicolumn{1}{c}{} &  & IMP-WR & 5 & 11 & 16 & 30 & 66 \\
\multicolumn{1}{c}{} &  & IMP-FT & \cellcolor{green!25}3 & 6 & 12 & 16 & 66 \\
\multicolumn{1}{c}{} &  & MP-AI & 5 & 12 & 27 & 35 & \cellcolor{gray!25} 66 \\
\multicolumn{1}{c}{} &  & IRP-FT & 4 & 11 & 17 & \cellcolor{gray!25} 66 & 66 \\
\multicolumn{1}{c}{} &  & RP-AI & 8 & \cellcolor{gray!25} 26 & \cellcolor{gray!25} 32 & 66 & \cellcolor{gray!25}66 \\ \cline{2-8} 
\multicolumn{1}{c}{} & \multirow{8}{*}{\begin{rotate}{90}BiLSTM\end{rotate}} & IIBP-WR & \cellcolor{green!25} 4 & 5 & 7 & 16 & 29 \\
\multicolumn{1}{c}{} &  & IIBP-FT & 6 & 5 & 5 &\cellcolor{green!25} 6 & \cellcolor{green!25}23 \\
\multicolumn{1}{c}{} &  & IBP-AI & 4 & 7 & 11 & 23 & 39 \\
\multicolumn{1}{c}{} &  & IMP-WR & 5 & \cellcolor{green!25}5 & \cellcolor{green!25}5 & 14 & \cellcolor{gray!25}62 \\
\multicolumn{1}{c}{} &  & IMP-FT & \cellcolor{gray!25} 6 & \cellcolor{gray!25} 6 & 5 & 8 & 32 \\
\multicolumn{1}{c}{} &  & MP-AI & 4 & 7 & 12 & 20 & \cellcolor{gray!25} 62 \\
\multicolumn{1}{c}{} &  & IRP-FT & 6 & 5 & 5 & 16 & 46 \\
\multicolumn{1}{c}{} &  & RP-AI & 4 & 8 &\cellcolor{gray!25} 13 & \cellcolor{gray!25} 23 & \cellcolor{gray!25} 62 \\ \hline
\\
\multicolumn{8}{c}{\textbf{Multi-label}} \\
 & \multicolumn{1}{l}{} & \multicolumn{1}{c}{\textbf{Pruner}} & \textbf{20\%} & \textbf{50\%} & \textbf{70\%} & \textbf{90\%} & \textbf{99\%} \\ \hline
\multirow{16}{*}{\begin{rotate}{90}Reuters\end{rotate}} & \multirow{8}{*}{\begin{rotate}{90}BERT\end{rotate}} & IIBP-WR & 5 & 8 & 16 & 31 & 100 \\
 &  & IIBP-FT & 5 & 6 & 7 & \cellcolor{green!25}10 & \cellcolor{green!25}32 \\
 &  & IBP-AI & 6 & 18 & 35 & 84 & 100 \\
 &  & IMP-WR & 4 & 6 & 14 & 100 & \cellcolor{gray!25} 100 \\
 &  & IMP-FT & \cellcolor{green!25}5 & \cellcolor{green!25}6 &\cellcolor{green!25} 8 & 23 & 100 \\
 &  & MP-AI & 5 & 12 & 32 & 100 & 100 \\
 &  & IRP-FT & 5 & 10 & 24 &\cellcolor{gray!25} 100 & 100 \\
 &  & RP-AI &\cellcolor{gray!25} 7 & \cellcolor{gray!25}34 & \cellcolor{gray!25} 41 & 100 & 100 \\ \cline{2-8} 
 & \multirow{8}{*}{\begin{rotate}{90}BiLSTM\end{rotate}} & IIBP-WR & 4 & 5 & 7 & 14 & 37 \\
 &  & IIBP-FT & \cellcolor{green!25}4 & 5 & 5 & 5 & \cellcolor{green!25}9 \\
 &  & IBP-AI & 4 & \cellcolor{gray!25} 7 & \cellcolor{gray!25}14 & \cellcolor{gray!25}34 &\cellcolor{gray!25} 44 \\
 &  & IMP-WR &\cellcolor{gray!25} 5 & 5 & 6 & 7 & 29 \\
 &  & IMP-FT & 5 & 4 & 4 &\cellcolor{green!25} 6 & 13 \\
 &  & MP-AI & 5 & 6 & 9 & 19 & 33 \\
 &  & IRP-FT &\cellcolor{green!25} 5 & \cellcolor{green!25}5 & \cellcolor{green!25}6 & 7 & 31 \\
 &  & RP-AI & 4 & 7 & 10 & 22 & 35 \\ \hline
\multirow{16}{*}{\begin{rotate}{90}AAPD\end{rotate}} & \multirow{8}{*}{\begin{rotate}{90}BERT\end{rotate}} & IIBP-WR & 31 & 40 & 48 & 59 & 81 \\
 &  & IIBP-FT & 29 & \cellcolor{green!25}37 & 45 & \cellcolor{green!25}51 & \cellcolor{green!25}63 \\
 &  & IBP-AI & 34 & 48 & 59 & 78 & 100 \\
 &  & IMP-WR & 31 & 38 & 49 & 79 & \cellcolor{gray!25}100 \\
 &  & IMP-FT & \cellcolor{green!25}28 & 40 & \cellcolor{green!25}45 & 56 & \cellcolor{gray!25}100 \\
 &  & MP-AI & 34 & 47 & 57 & 94 & \cellcolor{gray!25}100 \\
 &  & IRP-FT & 33 &\cellcolor{gray!25} 63 & 62 &\cellcolor{gray!25} 100 &\cellcolor{gray!25} 100 \\
 &  & RP-AI &\cellcolor{gray!25} 38 & 59 & \cellcolor{gray!25}76 & \cellcolor{gray!25}100 & \cellcolor{gray!25}100 \\ \cline{2-8} 
 & \multirow{8}{*}{\begin{rotate}{90}BiLSTM\end{rotate}} & IIBP-WR & 26 & 34 & 39 & 64 & 88 \\
 &  & IIBP-FT & 41 & 40 & 37 & \cellcolor{green!25}28 &\cellcolor{green!25} 59 \\
 &  & IBP-AI & \cellcolor{green!25}22 & 33 & 53 &\cellcolor{gray!25} 82 &\cellcolor{gray!25} 100 \\
 &  & IMP-WR & 26 & 32 & 38 & 56 & 83 \\
 &  & IMP-FT & \cellcolor{gray!25}39 & \cellcolor{gray!25}41 & 37 & \cellcolor{green!25}34 & 69 \\
 &  & MP-AI & \cellcolor{green!25}21 & \cellcolor{green!25}30 & 41 & 62 & 86 \\
 &  & IRP-FT & 41 & 36 &\cellcolor{green!25} 30 & 44 & 88 \\
 &  & RP-AI & 24 & 35 & \cellcolor{gray!25}49 & 67 & 87 \\ \hline
\end{tabular}
}

%% file: Tables/Setup/Architectures/stats.tex
\resizebox{\columnwidth}{!}{%
\begin{tabular}{ccrlllll}
\hline
\textbf{LM} &
  \textbf{Dataset} &
  \multicolumn{1}{c}{\textbf{\begin{tabular}[c]{@{}c@{}}\# parameters\end{tabular}}} &
  \multicolumn{1}{c}{\textbf{20\%}} &
  \multicolumn{1}{c}{\textbf{50\%}} &
  \multicolumn{1}{c}{\textbf{70\%}} &
  \multicolumn{1}{c}{\textbf{90\%}} &
  \multicolumn{1}{c}{\textbf{99\%}} \\ \hline
\multirow{4}{*}{\textbf{BERT}}   & \textbf{IMDB}    & 109,483,778 & \multirow{4}{*}{15\%} & \multirow{4}{*}{39\%} & \multirow{4}{*}{55\%} & \multirow{4}{*}{70\%} & \multirow{4}{*}{77\%}\\
                                 & \textbf{SNLI}    & 109,484,547\\
                                 & \textbf{Reuters} & 109,499,927\\
                                 & \textbf{AAPD}    & 109,523,766\\ \hline
                                 
\multirow{4}{*}{\textbf{BiLSTM}} & \textbf{IMDB}    & 647,810    & \multirow{4}{*}{20\%} & \multirow{4}{*}{50\%} & \multirow{4}{*}{69\%} & \multirow{4}{*}{89\%} & \multirow{4}{*}{98\%}\\
                                 & \textbf{SNLI}    & 647,939\\
                                 & \textbf{Reuters} & 650,519\\
                                 & \textbf{AAPD}    & 654,518\\ \hline
\end{tabular}
}

%% file: Tables/Setup/Dataset/stats_appendix.tex
\resizebox{1.9\columnwidth}{!}{%
\begin{tabular}{lrrrccrrrrcc}
\textbf{Dataset} & \textbf{\# train} & \textbf{\# test} & \textbf{\# val} & \textbf{Mean/median} & \textbf{Min/max len} & \begin{tabular}[c]{@{}c@{}} \textbf{Std}\\ \textbf{len}\end{tabular} & \begin{tabular}[c]{@{}c@{}} \textbf{Tokens}\\ \textbf{85\%}\end{tabular} & \begin{tabular}[c]{@{}c@{}} \textbf{Max}\\ \textbf{tokens}\end{tabular} & \begin{tabular}[c]{@{}c@{}} \textbf{\#}\\ \textbf{classes}\end{tabular} & \textbf{Task} & \textbf{Classification} \\ \hline
\textbf{IMDB}    & 20000  & 25000 & 5000 & 268/201 & 8/2753 & 197 & 430 & 512 & 2  & \begin{tabular}[c]{@{}c@{}}Sentiment \\ analysis\end{tabular}         & single-label \\
\textbf{SNLI}    & 549367 & 9824  & 9842 & 23/22   & 5/124 & 7  & 30  & 128 & 3  & \begin{tabular}[c]{@{}c@{}}Natural language \\ inference\end{tabular} & single-label \\
\textbf{Reuters} & 6737   & 1429  & 1440 & 126/79  & 5/1305 & 137 & 232 & 256 & 23 & \begin{tabular}[c]{@{}c@{}}Document \\ categorization\end{tabular}    & multi-label \\
\textbf{AAPD}    & 53840  & 1000  & 1000 & 167/161 & 1/599 & 70  & 242 & 256 & 54 & \begin{tabular}[c]{@{}c@{}}Document \\ categorization\end{tabular}    & multi-label \\ \hline
\end{tabular}
}

%% file: Tables/Setup/Hyperparameters_tuning/stats.tex
\begin{table}
\centering
\resizebox{\columnwidth}{!}{%
\begin{tabular}{clrrrrrrr}
\multirow{2}{*}{\textbf{Dataset}} &
  \multicolumn{1}{c}{\multirow{2}{*}{\textbf{Architecture}}} &
  \multicolumn{1}{l}{\multirow{2}{*}{\textbf{Batch size}}} &
  \multicolumn{2}{c}{\textbf{Epochs}} &
  \multicolumn{1}{c}{\multirow{2}{*}{\textbf{Best epoch}}} &
  \multicolumn{2}{c}{\textbf{lr}} &
  \multicolumn{1}{c}{\multirow{2}{*}{\textbf{Best lr}}} \\
 &
  \multicolumn{1}{c}{} &
  \multicolumn{1}{l}{} &
  \multicolumn{1}{c}{\textbf{Min}} &
  \multicolumn{1}{c}{\textbf{Max}} &
  \multicolumn{1}{c}{} &
  \multicolumn{1}{c}{\textbf{Min}} &
  \multicolumn{1}{c}{\textbf{Max}} &
  \multicolumn{1}{c}{} \\ \hline
\textbf{IMDB}                     & \textbf{BERT}   & 32   & 2  & 6   & 3  & 2e-5 & 2e-4 & 0.00003 \\
\textbf{}                         & \textbf{BiLSTM} & 1024 & 10 & 30  & 26 & 2e-4 & 2e-3 & 0.00148 \\ \hline
\multirow{2}{*}{\textbf{SNLI}}    & \textbf{BERT}   & 256  & 2  & 6   & 2  & 2e-5 & 2e-4 & 0.00004 \\
                                  & \textbf{BiLSTM} & 4096 & 30 & 50  & 39 & 2e-4 & 2e-3 & 0.00070 \\ \hline
\multirow{2}{*}{\textbf{Reuters}} & \textbf{BERT}   & 128  & 5  & 15  & 14 & 2e-5 & 2e-4 & 0.00005 \\
                                  & \textbf{BiLSTM} & 512  & 30 & 100 & 72 & 2e-4 & 2e-3 & 0.00186 \\ \hline
\multirow{2}{*}{\textbf{AAPD}}    & \textbf{BERT}   & 256  & 5  & 15  & 13 & 2e-5 & 2e-4 & 0.00008 \\
                                  & \textbf{BiLSTM} & 2048 & 30 & 60  & 50 & 2e-4 & 2e-3 & 0.00200 \\ \hline
\end{tabular}
}
\caption{Search space and best configuration for the hyperparameter tuning of the models. Min and Max epochs represent the range of epochs used to perform hyperparameter tuning. Best epoch is the best epoch found with hyperparameter tuning. Min and Max lr are the range learning rate is tuned on. Best lr is the best learning rate found during hyperparameter optimization. The batch size is set to maximize the GPU usage.}
\label{tab:hyperparam_tuning}
\end{table}

%% file: Tables/Effectiveness/effectiveness_merged.tex
\begin{table*}
\centering

\resizebox{1.65\columnwidth}{!}{%
\begin{tabular}{cclcccccc}

\multicolumn{9}{c}{\textbf{Single-label}: Accuracy} \\
\multicolumn{3}{c}{\% pruned parameters}  & 0\% & 20\% & 50\% & 70\% & 90\% & 99\% \\
dataset & model & pruning algo &  &  &  &  &  &  \\
\cline{1-9} \multirow[c]{16}{*}{\begin{rotate}{90}IMDB\end{rotate}} & \multirow[c]{8}{*}{\begin{rotate}{90}BERT\end{rotate}} & IIBP-WR & \bfseries .932 $\pm$ .005 & .892 $\pm$ .009 & .870 $\pm$ .016 & .864 $\pm$ .026 & \bfseries .863 $\pm$ .011 & .742 $\pm$ .136 \\
 &  & IIBP-FT & \bfseries .932 $\pm$ .005 & .919 $\pm$ .004 & .869 $\pm$ .008 & .848 $\pm$ .007 & .843 $\pm$ .010 & \bfseries .828 $\pm$ .064 \\
 &  & IBP-AI & \bfseries .932 $\pm$ .005 & .882 $\pm$ .010 & .864 $\pm$ .021 & .865 $\pm$ .016 & .841 $\pm$ .069 & .526 $\pm$ .079 \\
 &  & IMP-WR & \bfseries .932 $\pm$ .005 & .911 $\pm$ .009 & \bfseries .880 $\pm$ .006 & \bfseries .870 $\pm$ .009 & .857 $\pm$ .007 & .500 $\pm$ .000 \\
 &  & IMP-FT & \bfseries .932 $\pm$ .005 & \bfseries .924 $\pm$ .004 & .873 $\pm$ .007 & .845 $\pm$ .004 & .850 $\pm$ .007 & .500 $\pm$ .000 \\
 &  & MP-AI & \bfseries .932 $\pm$ .005 & .904 $\pm$ .008 & .871 $\pm$ .009 & .867 $\pm$ .010 & .852 $\pm$ .011 & .500 $\pm$ .000 \\
 &  & IRP-FT & \bfseries .932 $\pm$ .005 & .877 $\pm$ .011 & .846 $\pm$ .009 & .778 $\pm$ .141 & .500 $\pm$ .000 & .500 $\pm$ .000 \\
 &  & RP-AI & \bfseries .932 $\pm$ .005 & .874 $\pm$ .004 & .866 $\pm$ .012 & .828 $\pm$ .114 & .500 $\pm$ .000 & .500 $\pm$ .000 \\
 \cline{2-9} & \multirow[c]{8}{*}{\begin{rotate}{90}BiLSTM\end{rotate}}  &  IIBP-WR & \bfseries .879 $\pm$ .016 & .868 $\pm$ .021 & .861 $\pm$ .026 & .856 $\pm$ .027 & .837 $\pm$ .025 & .806 $\pm$ .026 \\
 &  & IIBP-FT & \bfseries .879 $\pm$ .016 & .883 $\pm$ .011 & .880 $\pm$ .013 & .878 $\pm$ .010 & .872 $\pm$ .011 & \bfseries .872 $\pm$ .013 \\
 &  & IBP-AI & \bfseries .879 $\pm$ .016 & .874 $\pm$ .017 & .848 $\pm$ .019 & .820 $\pm$ .032 & .805 $\pm$ .029 & .755 $\pm$ .022 \\
 &  & IMP-WR & \bfseries .879 $\pm$ .016 & .875 $\pm$ .020 & .881 $\pm$ .012 & .876 $\pm$ .011 & .873 $\pm$ .025 & .804 $\pm$ .018 \\
 &  & IMP-FT & \bfseries .879 $\pm$ .016 & \bfseries .886 $\pm$ .010 & \bfseries .887 $\pm$ .010 & \bfseries .882 $\pm$ .009 & \bfseries .878 $\pm$ .007 & .862 $\pm$ .013 \\
 &  & MP-AI & \bfseries .879 $\pm$ .016 & .872 $\pm$ .019 & .855 $\pm$ .018 & .843 $\pm$ .023 & .834 $\pm$ .015 & .755 $\pm$ .021 \\
 &  & IRP-FT & \bfseries .879 $\pm$ .016 & .885 $\pm$ .010 & .875 $\pm$ .011 & .875 $\pm$ .012 & .873 $\pm$ .017 & .548 $\pm$ .073 \\
 &  & RP-AI & \bfseries .879 $\pm$ .016 & .872 $\pm$ .037 & .848 $\pm$ .027 & .845 $\pm$ .026 & .840 $\pm$ .016 & .721 $\pm$ .024 \\
\cline{1-9} \multirow[c]{16}{*}{\begin{rotate}{90}SNLI\end{rotate}} & \multirow[c]{9}{*}{\begin{rotate}{90}BERT\end{rotate}}  &  IIBP-WR & \bfseries .901 $\pm$ .002 & .849 $\pm$ .098 & .822 $\pm$ .004 & .794 $\pm$ .007 & .683 $\pm$ .044 & .578 $\pm$ .053 \\
 &  & IIBP-FT & \bfseries .901 $\pm$ .002 & .892 $\pm$ .002 & \bfseries .876 $\pm$ .003 & \bfseries .857 $\pm$ .003 & \bfseries .806 $\pm$ .090 & \bfseries .654 $\pm$ .071 \\
 &  & IBP-AI & \bfseries .901 $\pm$ .002 & .872 $\pm$ .002 & .824 $\pm$ .005 & .768 $\pm$ .028 & .625 $\pm$ .016 & .395 $\pm$ .086 \\
 &  & IMP-WR & \bfseries .901 $\pm$ .002 & .883 $\pm$ .003 & .847 $\pm$ .004 & .799 $\pm$ .004 & .646 $\pm$ .033 & .336 $\pm$ .008 \\
 &  & IMP-FT & \bfseries .901 $\pm$ .002 & \bfseries .895 $\pm$ .002 & .875 $\pm$ .002 & .835 $\pm$ .004 & .799 $\pm$ .005 & .336 $\pm$ .008 \\
 &  & MP-AI & \bfseries .901 $\pm$ .002 & .882 $\pm$ .002 & .833 $\pm$ .003 & .691 $\pm$ .016 & .616 $\pm$ .011 & .335 $\pm$ .007 \\
 &  & IRP-FT & \bfseries .901 $\pm$ .002 & .885 $\pm$ .003 & .836 $\pm$ .004 & .785 $\pm$ .008 & .342 $\pm$ .034 & .336 $\pm$ .008 \\
 &  & RP-AI & \bfseries .901 $\pm$ .002 & .854 $\pm$ .004 & .695 $\pm$ .007 & .647 $\pm$ .005 & .366 $\pm$ .069 & .335 $\pm$ .007 \\
 \cline{2-9} & \multirow[c]{9}{*}{\begin{rotate}{90}BiLSTM\end{rotate}}  &  IIBP-WR & \bfseries .778 $\pm$ .004 & \bfseries .780 $\pm$ .004 & .774 $\pm$ .005 & .763 $\pm$ .005 & .715 $\pm$ .007 & .614 $\pm$ .007 \\
 &  & IIBP-FT & \bfseries .778 $\pm$ .004 & .742 $\pm$ .004 & .750 $\pm$ .004 & .762 $\pm$ .003 & \bfseries .771 $\pm$ .004 & \bfseries .657 $\pm$ .011 \\
 &  & IBP-AI & \bfseries .778 $\pm$ .004 & .776 $\pm$ .004 & .766 $\pm$ .004 & .743 $\pm$ .007 & .669 $\pm$ .009 & .431 $\pm$ .104 \\
 &  & IMP-WR & \bfseries .778 $\pm$ .004 & .779 $\pm$ .004 & \bfseries .782 $\pm$ .004 & \bfseries .782 $\pm$ .004 & .726 $\pm$ .009 & .336 $\pm$ .007 \\
 &  & IMP-FT & \bfseries .778 $\pm$ .004 & .741 $\pm$ .004 & .746 $\pm$ .004 & .766 $\pm$ .003 & .765 $\pm$ .004 & .574 $\pm$ .019 \\
 &  & MP-AI & \bfseries .778 $\pm$ .004 & .776 $\pm$ .004 & .764 $\pm$ .006 & .743 $\pm$ .005 & .687 $\pm$ .007 & .336 $\pm$ .007 \\
 &  & IRP-FT & \bfseries .778 $\pm$ .004 & .746 $\pm$ .004 & .769 $\pm$ .004 & .779 $\pm$ .004 & .712 $\pm$ .007 & .389 $\pm$ .070 \\
 &  & RP-AI & \bfseries .778 $\pm$ .004 & .776 $\pm$ .003 & .762 $\pm$ .005 & .739 $\pm$ .006 & .667 $\pm$ .017 & .336 $\pm$ .007 \\
\hline

& & & & & & & & \\
\multicolumn{9}{c}{\textbf{Multi-label}: F1 Macro} \\
\multicolumn{3}{c}{\% pruned parameters}  & 0\% & 20\% & 50\% & 70\% & 90\% & 99\% \\
dataset & model & pruning algo &  &  &  &  &  &  \\
\cline{1-9} \multirow[c]{16}{*}{\begin{rotate}{90}Reuters\end{rotate}} & \multirow[c]{10}{*}{\begin{rotate}{90}BERT\end{rotate}}  & IIBP-WR & \bfseries .836 $\pm$ .004 & .822 $\pm$ .011 & .792 $\pm$ .018 & .674 $\pm$ .064 & .382 $\pm$ .046 & .167 $\pm$ .041 \\
 &  & IIBP-FT & \bfseries .836 $\pm$ .004 & .835 $\pm$ .005 & .830 $\pm$ .005 & .822 $\pm$ .008 & \bfseries .786 $\pm$ .029 & \bfseries .355 $\pm$ .061 \\
 &  & IBP-AI & \bfseries .836 $\pm$ .004 & .810 $\pm$ .008 & .645 $\pm$ .048 & .328 $\pm$ .048 & .189 $\pm$ .027 & .096 $\pm$ .018 \\
 &  & IMP-WR & \bfseries .836 $\pm$ .004 & .827 $\pm$ .006 & .829 $\pm$ .005 & .736 $\pm$ .015 & .147 $\pm$ .025 & .082 $\pm$ .008 \\
 &  & IMP-FT & \bfseries .836 $\pm$ .004 & \bfseries .838 $\pm$ .005 & \bfseries .834 $\pm$ .005 & \bfseries .824 $\pm$ .006 & .490 $\pm$ .086 & .085 $\pm$ .005 \\
 &  & MP-AI & \bfseries .836 $\pm$ .004 & .822 $\pm$ .006 & .745 $\pm$ .021 & .417 $\pm$ .057 & .127 $\pm$ .031 & .086 $\pm$ .004 \\
 &  & IRP-FT & \bfseries .836 $\pm$ .004 & .832 $\pm$ .005 & .769 $\pm$ .013 & .524 $\pm$ .075 & .087 $\pm$ .001 & .087 $\pm$ .002 \\
 &  & RP-AI & \bfseries .836 $\pm$ .004 & .803 $\pm$ .007 & .479 $\pm$ .052 & .242 $\pm$ .021 & .089 $\pm$ .012 & .086 $\pm$ .003 \\
 \cline{2-9} & \multirow[c]{9}{*}{\begin{rotate}{90}BiLSTM\end{rotate}}  &  IIBP-WR & \bfseries .731 $\pm$ .017 & .728 $\pm$ .018 & .727 $\pm$ .016 & .716 $\pm$ .014 & .631 $\pm$ .036 & .396 $\pm$ .040 \\
 &  & IIBP-FT & \bfseries .731 $\pm$ .017 & \bfseries .753 $\pm$ .018 & .751 $\pm$ .013 & .751 $\pm$ .015 & .742 $\pm$ .014 & \bfseries .693 $\pm$ .019 \\
 &  & IBP-AI & \bfseries .731 $\pm$ .017 & .729 $\pm$ .020 & .706 $\pm$ .019 & .616 $\pm$ .029 & .456 $\pm$ .036 & .224 $\pm$ .028 \\
 &  & IMP-WR & \bfseries .731 $\pm$ .017 & .726 $\pm$ .017 & .738 $\pm$ .015 & .745 $\pm$ .012 & .734 $\pm$ .011 & .481 $\pm$ .032 \\
 &  & IMP-FT & \bfseries .731 $\pm$ .017 & .751 $\pm$ .013 & .747 $\pm$ .014 & .745 $\pm$ .017 & \bfseries .746 $\pm$ .012 & .657 $\pm$ .028 \\
 &  & MP-AI & \bfseries .731 $\pm$ .017 & .740 $\pm$ .012 & .730 $\pm$ .014 & .705 $\pm$ .022 & .606 $\pm$ .026 & .393 $\pm$ .034 \\
 &  & IRP-FT & \bfseries .731 $\pm$ .017 & \bfseries .753 $\pm$ .015 & \bfseries .757 $\pm$ .015 & \bfseries .760 $\pm$ .014 & .743 $\pm$ .012 & .417 $\pm$ .042 \\
 &  & RP-AI & \bfseries .731 $\pm$ .017 & .731 $\pm$ .019 & .724 $\pm$ .015 & .661 $\pm$ .028 & .570 $\pm$ .030 & .377 $\pm$ .042 \\
\cline{1-9} \multirow[c]{16}{*}{\begin{rotate}{90}AAPD\end{rotate}} & \multirow[c]{9}{*}{\begin{rotate}{90}BERT\end{rotate}}  &  IIBP-WR & \bfseries .578 $\pm$ .007 & .547 $\pm$ .008 & .518 $\pm$ .009 & .482 $\pm$ .010 & .403 $\pm$ .018 & .179 $\pm$ .032 \\
 &  & IIBP-FT & \bfseries .578 $\pm$ .007 & .573 $\pm$ .009 & \bfseries .548 $\pm$ .009 & .462 $\pm$ .153 & \bfseries .476 $\pm$ .018 & \bfseries .316 $\pm$ .033 \\
 &  & IBP-AI & \bfseries .578 $\pm$ .007 & .539 $\pm$ .009 & .480 $\pm$ .015 & .398 $\pm$ .023 & .234 $\pm$ .042 & .091 $\pm$ .016 \\
 &  & IMP-WR & \bfseries .578 $\pm$ .007 & .567 $\pm$ .009 & .541 $\pm$ .007 & .483 $\pm$ .008 & .230 $\pm$ .029 & .080 $\pm$ .001 \\
 &  & IMP-FT & \bfseries .578 $\pm$ .007 & \bfseries .579 $\pm$ .009 & .546 $\pm$ .008 & \bfseries .521 $\pm$ .008 & .400 $\pm$ .019 & .080 $\pm$ .000 \\
 &  & MP-AI & \bfseries .578 $\pm$ .007 & .551 $\pm$ .008 & .508 $\pm$ .010 & .423 $\pm$ .014 & .145 $\pm$ .007 & .080 $\pm$ .000 \\
 &  & IRP-FT & \bfseries .578 $\pm$ .007 & .554 $\pm$ .009 & .312 $\pm$ .197 & .338 $\pm$ .133 & .082 $\pm$ .014 & .080 $\pm$ .000 \\
 &  & RP-AI & \bfseries .578 $\pm$ .007 & .524 $\pm$ .009 & .397 $\pm$ .015 & .261 $\pm$ .029 & .082 $\pm$ .007 & .080 $\pm$ .000 \\
 \cline{2-9} & \multirow[c]{9}{*}{\begin{rotate}{90}BiLSTM\end{rotate}}  &  IIBP-WR & \bfseries .468 $\pm$ .015 & .449 $\pm$ .022 & .441 $\pm$ .022 & .444 $\pm$ .020 & .346 $\pm$ .022 & .163 $\pm$ .028 \\
 &  & IIBP-FT & \bfseries .468 $\pm$ .015 & .429 $\pm$ .013 & .425 $\pm$ .015 & .436 $\pm$ .009 & \bfseries .486 $\pm$ .010 & \bfseries .396 $\pm$ .012 \\
 &  & IBP-AI & \bfseries .468 $\pm$ .015 & \bfseries .473 $\pm$ .014 & .459 $\pm$ .011 & .398 $\pm$ .029 & .190 $\pm$ .027 & .082 $\pm$ .004 \\
 &  & IMP-WR & \bfseries .468 $\pm$ .015 & .446 $\pm$ .018 & .454 $\pm$ .016 & .454 $\pm$ .013 & .406 $\pm$ .014 & .185 $\pm$ .019 \\
 &  & IMP-FT & \bfseries .468 $\pm$ .015 & .428 $\pm$ .014 & .421 $\pm$ .013 & .432 $\pm$ .015 & \bfseries .486 $\pm$ .009 & .330 $\pm$ .020 \\
 &  & MP-AI & \bfseries .468 $\pm$ .015 & \bfseries .473 $\pm$ .015 & \bfseries .473 $\pm$ .010 & .454 $\pm$ .011 & .385 $\pm$ .020 & .165 $\pm$ .025 \\
 &  & IRP-FT & \bfseries .468 $\pm$ .015 & .432 $\pm$ .012 & .451 $\pm$ .013 & \bfseries .486 $\pm$ .012 & .475 $\pm$ .010 & .167 $\pm$ .025 \\
 &  & RP-AI & \bfseries .468 $\pm$ .015 & .464 $\pm$ .014 & .453 $\pm$ .018 & .421 $\pm$ .023 & .356 $\pm$ .021 & .163 $\pm$ .025 \\ \hline
\end{tabular}
}
\caption{Average macro accuracy/F1 score and std over 30 model initializations. Pruning algo is the used pruning algorithm according to Table \ref{tab:pruningAlgoTable}.
The best results for each percentage of pruned parameters and combination of dataset and architecture are in bold.
}
\label{tab:eff_merged}
\end{table*}

%% file: Tables/PIEs/merged_original.tex
\begin{table*}
\centering

\resizebox{1.3\columnwidth}{!}{%
\begin{tabular}{cclll@{\hspace{0.5\tabcolsep}}lll@{\hspace{0.5\tabcolsep}}lll@{\hspace{0.5\tabcolsep}}lll@{\hspace{0.5\tabcolsep}}lll}

\multicolumn{17}{c}{\textbf{Single-label}} \\
 & \multicolumn{1}{l}{} & \multicolumn{1}{c}{\textbf{Pruner}} & \multicolumn{2}{c}{\textbf{20\%}} & \textbf{} & \multicolumn{2}{c}{\textbf{50\%}} & \textbf{} & \multicolumn{2}{c}{\textbf{70\%}} & \textbf{} & \multicolumn{2}{c}{\textbf{90\%}} & \textbf{} & \multicolumn{2}{c}{\textbf{99\%}} \\ \hline
\multicolumn{1}{c}{} &  & IIBP-WR & 0.245 & 0.755 &  & 0.200 & 0.800 &  & 0.191 & 0.809 &  & 0.188 & 0.812 &  & 0.182 & 0.818 \\
\multicolumn{1}{c}{} &  & IIBP-FT & 0.356 & 0.644 &  & 0.195 & 0.805 &  & 0.161 & 0.839 &  & 0.163 & 0.837 &  & 0.188 & 0.812 \\
\multicolumn{1}{c}{} &  & IBP-AI & 0.227 & 0.773 &  & 0.195 & 0.805 &  & 0.198 & 0.802 &  & 0.205 & 0.795 &  & 0.056 & \textbf{0.944} \\
\multicolumn{1}{c}{} &  & IMP-WR & 0.290 & 0.710 &  & 0.206 & 0.794 &  & 0.194 & 0.806 &  & 0.179 & 0.821 &  & 0.056 & \textbf{0.944} \\
\multicolumn{1}{c}{} &  & IMP-FT & 0.385 & 0.615 &  & 0.200 & 0.800 &  & 0.156 & 0.844 &  & 0.167 & 0.833 &  & 0.056 & \textbf{0.944} \\
\multicolumn{1}{c}{} &  & MP-AI & 0.262 & 0.738 &  & 0.197 & 0.803 &  & 0.192 & 0.808 &  & 0.180 & 0.820 &  & 0.056 & \textbf{0.944} \\
\multicolumn{1}{c}{} &  & IRP-FT & 0.220 & 0.780 &  & 0.161 & 0.839 &  & 0.172 & 0.828 &  & 0.056 & \textbf{0.944} &  & 0.056 & \textbf{0.944} \\
\multicolumn{1}{c}{} & \multirow{-8}{*}{\begin{rotate}{90}BERT\end{rotate}} & RP-AI & 0.198 & 0.802 &  & 0.199 & 0.801 &  & 0.198 & 0.802 &  & 0.056 & \textbf{0.944} &  & 0.056 & \textbf{0.944} \\ \cline{2-17} 
\multicolumn{1}{c}{} &  & IIBP-WR & 0.371 & 0.629 &  & 0.322 & 0.678 &  & 0.283 & 0.717 &  & 0.232 & 0.768 &  & 0.207 & 0.793 \\
\multicolumn{1}{c}{} &  & IIBP-FT & \cellcolor[HTML]{CBCEFB}0.604 & \cellcolor[HTML]{CBCEFB}0.396 &  & \cellcolor[HTML]{CBCEFB}0.616 & \cellcolor[HTML]{CBCEFB}0.384 &  & \cellcolor[HTML]{CBCEFB}0.598 & \cellcolor[HTML]{CBCEFB}0.402 &  & \cellcolor[HTML]{CBCEFB}0.555 & \cellcolor[HTML]{CBCEFB}0.445 &  & 0.471 & 0.529 \\
\multicolumn{1}{c}{} &  & IBP-AI & 0.382 & 0.618 &  & 0.253 & 0.747 &  & 0.209 & 0.791 &  & 0.206 & 0.794 &  & 0.168 & 0.832 \\
\multicolumn{1}{c}{} &  & IMP-WR & 0.471 & 0.529 &  & \cellcolor[HTML]{CBCEFB}0.542 & \cellcolor[HTML]{CBCEFB}0.458 &  & 0.480 & 0.520 &  & 0.470 & 0.530 &  & 0.218 & 0.782 \\
\multicolumn{1}{c}{} &  & IMP-FT & \cellcolor[HTML]{CBCEFB}0.644 & \cellcolor[HTML]{CBCEFB}0.356 &  & \cellcolor[HTML]{CBCEFB}0.658 & \cellcolor[HTML]{CBCEFB}0.342 &  & \cellcolor[HTML]{CBCEFB}0.584 & \cellcolor[HTML]{CBCEFB}0.416 &  & \cellcolor[HTML]{CBCEFB}0.613 & \cellcolor[HTML]{CBCEFB}0.387 &  & 0.395 & 0.605 \\
\multicolumn{1}{c}{} &  & MP-AI & 0.404 & 0.596 &  & 0.281 & 0.719 &  & 0.241 & 0.759 &  & 0.225 & 0.775 &  & 0.178 & 0.822 \\
\multicolumn{1}{c}{} &  & IRP-FT & \cellcolor[HTML]{CBCEFB}0.633 & \cellcolor[HTML]{CBCEFB}0.367 &  & \cellcolor[HTML]{CBCEFB}0.577 & \cellcolor[HTML]{CBCEFB}0.423 &  & \cellcolor[HTML]{CBCEFB}0.576 & \cellcolor[HTML]{CBCEFB}0.424 &  & 0.404 & 0.596 &  & 0.126 & 0.874 \\
\multicolumn{1}{c}{\multirow{-16}{*}{\begin{rotate}{90}IMDB\end{rotate}}} & \multirow{-8}{*}{\begin{rotate}{90}BiLSTM\end{rotate}} & RP-AI & 0.403 & 0.597 &  & 0.269 & 0.731 &  & 0.250 & 0.750 &  & 0.230 & 0.770 &  & 0.161 & 0.839 \\ \hline
\multicolumn{1}{c}{} &  & IIBP-WR & 0.250 & 0.688 &  & 0.177 & 0.753 &  & 0.155 & 0.782 &  & 0.091 & 0.855 &  & 0.074 & 0.878 \\
\multicolumn{1}{c}{} &  & IIBP-FT & 0.438 & 0.468 &  & 0.301 & 0.635 &  & 0.235 & 0.692 &  & 0.173 & 0.752 &  & 0.090 & 0.859 \\
\multicolumn{1}{c}{} &  & IBP-AI & 0.265 & 0.676 &  & 0.177 & 0.756 &  & 0.139 & 0.805 &  & 0.075 & 0.872 &  & 0.049 & 0.882 \\
\multicolumn{1}{c}{} &  & IMP-WR & 0.284 & 0.658 &  & 0.212 & 0.721 &  & 0.149 & 0.792 &  & 0.084 & 0.867 &  & 0.044 & \textbf{0.909} \\
\multicolumn{1}{c}{} &  & IMP-FT & 0.397 & 0.525 &  & 0.287 & 0.648 &  & 0.194 & 0.746 &  & 0.149 & 0.788 &  & 0.044 & \textbf{0.909} \\
\multicolumn{1}{c}{} &  & MP-AI & 0.275 & 0.656 &  & 0.175 & 0.748 &  & 0.083 & 0.853 &  & 0.069 & 0.873 &  & 0.044 & \textbf{0.909} \\
\multicolumn{1}{c}{} &  & IRP-FT & 0.347 & 0.582 &  & 0.192 & 0.738 &  & 0.136 & 0.803 &  & 0.044 & \textbf{0.909} &  & 0.044 & \textbf{0.909} \\
\multicolumn{1}{c}{} & \multirow{-8}{*}{\begin{rotate}{90}BERT\end{rotate}} & RP-AI & 0.224 & 0.709 &  & 0.087 & 0.855 &  & 0.074 & 0.873 &  & 0.044 & \textbf{0.909} &  & 0.044 & \textbf{0.909} \\ \cline{2-17} 
\multicolumn{1}{c}{} &  & IIBP-WR & 0.445 & 0.464 &  & 0.356 & 0.549 &  & 0.278 & 0.618 &  & 0.208 & 0.682 &  & 0.153 & 0.750 \\
\multicolumn{1}{c}{} &  & IIBP-FT & \cellcolor[HTML]{CBCEFB}0.467 & \cellcolor[HTML]{CBCEFB}0.413 &  & \cellcolor[HTML]{CBCEFB}0.529 & \cellcolor[HTML]{CBCEFB}0.366 &  & \cellcolor[HTML]{CBCEFB}0.517 & \cellcolor[HTML]{CBCEFB}0.368 &  & 0.391 & 0.493 &  & 0.184 & 0.719 \\
\multicolumn{1}{c}{} &  & IBP-AI & 0.382 & 0.518 &  & 0.298 & 0.582 &  & 0.258 & 0.626 &  & 0.177 & 0.722 &  & 0.124 & 0.760 \\
\multicolumn{1}{c}{} &  & IMP-WR & 0.434 & 0.454 &  & \cellcolor[HTML]{CBCEFB}0.461 & \cellcolor[HTML]{CBCEFB}0.429 &  & 0.447 & 0.451 &  & 0.225 & 0.670 &  & 0.068 & \textbf{0.824} \\
\multicolumn{1}{c}{} &  & IMP-FT & \cellcolor[HTML]{CBCEFB}0.495 & \cellcolor[HTML]{CBCEFB}0.381 &  & \cellcolor[HTML]{CBCEFB}0.496 & \cellcolor[HTML]{CBCEFB}0.379 &  & \cellcolor[HTML]{CBCEFB}0.522 & \cellcolor[HTML]{CBCEFB}0.361 &  & 0.375 & 0.522 &  & 0.141 & 0.758 \\
\multicolumn{1}{c}{} &  & MP-AI & 0.397 & 0.490 &  & 0.296 & 0.596 &  & 0.247 & 0.640 &  & 0.196 & 0.705 &  & 0.068 & \textbf{0.824} \\
\multicolumn{1}{c}{} &  & IRP-FT & \cellcolor[HTML]{CBCEFB}0.512 & \cellcolor[HTML]{CBCEFB}0.389 &  & \cellcolor[HTML]{CBCEFB}0.535 & \cellcolor[HTML]{CBCEFB}0.340 &  & \cellcolor[HTML]{CBCEFB}0.498 & \cellcolor[HTML]{CBCEFB}0.393 &  & 0.215 & 0.677 &  & 0.101 & \textbf{0.796} \\
\multicolumn{1}{c}{\multirow{-16}{*}{\begin{rotate}{90}SNLI\end{rotate}}} & \multirow{-8}{*}{\begin{rotate}{90}BiLSTM\end{rotate}} & RP-AI & 0.371 & 0.524 &  & 0.281 & 0.620 &  & 0.243 & 0.649 &  & 0.175 & 0.728 &  & 0.068 & \textbf{0.824} \\ \hline
\multicolumn{17}{c}{\textbf{Multi-label}} \\
 & \multicolumn{1}{l}{} & \multicolumn{1}{c}{\textbf{Pruner}} & \multicolumn{2}{c}{\textbf{20\%}} & \textbf{} & \multicolumn{2}{c}{\textbf{50\%}} & \textbf{} & \multicolumn{2}{c}{\textbf{70\%}} & \textbf{} & \multicolumn{2}{c}{\textbf{90\%}} & \textbf{} & \multicolumn{2}{c}{\textbf{99\%}} \\ \hline
 &  & IIBP-WR & 0.575 & 0.620 &  & 0.561 & 0.664 &  & 0.545 & 0.777 &  & 0.319 & 0.807 &  & 0.167 & \textbf{0.837} \\
 &  & IIBP-FT & \cellcolor[HTML]{CBCEFB}0.608 & \cellcolor[HTML]{CBCEFB}0.591 &  & \cellcolor[HTML]{CBCEFB}0.572 & \cellcolor[HTML]{CBCEFB}0.567 &  & 0.589 & 0.621 &  & 0.530 & 0.659 &  & 0.302 & 0.820 \\
 &  & IBP-AI & 0.572 & 0.656 &  & 0.506 & 0.780 &  & 0.276 & 0.825 &  & 0.182 & \textbf{0.838} &  & 0.096 & 0.836 \\
 &  & IMP-WR & 0.563 & 0.602 &  & 0.529 & 0.570 &  & 0.545 & 0.726 &  & 0.147 & \textbf{0.837} &  & 0.082 & 0.836 \\
 &  & IMP-FT & \cellcolor[HTML]{CBCEFB}0.619 & \cellcolor[HTML]{CBCEFB}0.602 &  & 0.555 & 0.596 &  & 0.590 & 0.627 &  & 0.393 & 0.794 &  & 0.085 & 0.836 \\
 &  & MP-AI & 0.555 & 0.610 &  & 0.555 & 0.743 &  & 0.359 & 0.819 &  & 0.127 & 0.836 &  & 0.086 & 0.836 \\
 &  & IRP-FT & 0.604 & 0.621 &  & 0.530 & 0.714 &  & 0.422 & 0.806 &  & 0.087 & 0.836 &  & 0.087 & 0.836 \\
 & \multirow{-8}{*}{\begin{rotate}{90}BERT\end{rotate}} & RP-AI & 0.560 & 0.666 &  & 0.428 & 0.815 &  & 0.196 & 0.825 &  & 0.089 & 0.836 &  & 0.086 & 0.836 \\ \cline{2-17} 
 &  & IIBP-WR & \cellcolor[HTML]{CBCEFB}0.466 & \cellcolor[HTML]{CBCEFB}0.462 &  & 0.498 & 0.500 &  & 0.483 & 0.509 &  & 0.509 & 0.620 &  & 0.362 & 0.701 \\
 &  & IIBP-FT & \cellcolor[HTML]{CBCEFB}0.476 & \cellcolor[HTML]{CBCEFB}0.423 &  & \cellcolor[HTML]{CBCEFB}0.490 & \cellcolor[HTML]{CBCEFB}0.440 &  & \cellcolor[HTML]{CBCEFB}0.511 & \cellcolor[HTML]{CBCEFB}0.442 &  & \cellcolor[HTML]{CBCEFB}0.508 & \cellcolor[HTML]{CBCEFB}0.432 &  & 0.496 & 0.509 \\
 &  & IBP-AI & 0.452 & 0.459 &  & 0.489 & 0.529 &  & 0.501 & 0.620 &  & 0.422 & 0.708 &  & 0.193 & 0.720 \\
 &  & IMP-WR & 0.464 & 0.470 &  & 0.445 & 0.448 &  & \cellcolor[HTML]{CBCEFB}0.483 & \cellcolor[HTML]{CBCEFB}0.451 &  & \cellcolor[HTML]{CBCEFB}0.521 & \cellcolor[HTML]{CBCEFB}0.485 &  & 0.435 & 0.696 \\
 &  & IMP-FT & \cellcolor[HTML]{CBCEFB}0.519 & \cellcolor[HTML]{CBCEFB}0.462 &  & \cellcolor[HTML]{CBCEFB}0.521 & \cellcolor[HTML]{CBCEFB}0.452 &  & \cellcolor[HTML]{CBCEFB}0.504 & \cellcolor[HTML]{CBCEFB}0.443 &  & \cellcolor[HTML]{CBCEFB}0.526 & \cellcolor[HTML]{CBCEFB}0.447 &  & 0.496 & 0.577 \\
 &  & MP-AI & \cellcolor[HTML]{CBCEFB}0.495 & \cellcolor[HTML]{CBCEFB}0.470 &  & 0.472 & 0.478 &  & 0.514 & 0.557 &  & 0.504 & 0.638 &  & 0.356 & 0.711 \\
 &  & IRP-FT & \cellcolor[HTML]{CBCEFB}0.500 & \cellcolor[HTML]{CBCEFB}0.423 &  & \cellcolor[HTML]{CBCEFB}0.480 & \cellcolor[HTML]{CBCEFB}0.399 &  & \cellcolor[HTML]{CBCEFB}0.488 & \cellcolor[HTML]{CBCEFB}0.416 &  & \cellcolor[HTML]{CBCEFB}0.512 & \cellcolor[HTML]{CBCEFB}0.440 &  & 0.375 & 0.704 \\
\multirow{-16}{*}{\begin{rotate}{90}Reuters\end{rotate}} & \multirow{-8}{*}{\begin{rotate}{90}BiLSTM\end{rotate}} & RP-AI & \cellcolor[HTML]{CBCEFB}0.453 & \cellcolor[HTML]{CBCEFB}0.446 &  & 0.510 & 0.517 &  & 0.510 & 0.593 &  & 0.507 & 0.676 &  & 0.346 & 0.710 \\ \hline
 &  & IIBP-WR & 0.471 & 0.511 &  & 0.453 & 0.529 &  & 0.432 & 0.553 &  & 0.367 & 0.563 &  & 0.175 & \textbf{0.580} \\
 &  & IIBP-FT & 0.498 & 0.506 &  & 0.476 & 0.515 &  & 0.417 & 0.542 &  & 0.418 & 0.556 &  & 0.292 & 0.576 \\
 &  & IBP-AI & 0.463 & 0.515 &  & 0.430 & 0.548 &  & 0.366 & 0.566 &  & 0.229 & \textbf{0.582} &  & 0.091 & 0.578 \\
 &  & IMP-WR & 0.492 & 0.507 &  & 0.475 & 0.525 &  & 0.428 & 0.552 &  & 0.225 & \textbf{0.582} &  & 0.080 & 0.578 \\
 &  & IMP-FT & 0.502 & 0.506 &  & 0.484 & 0.532 &  & 0.462 & 0.537 &  & 0.366 & 0.569 &  & 0.080 & 0.578 \\
 &  & MP-AI & 0.475 & 0.517 &  & 0.452 & 0.544 &  & 0.390 & 0.568 &  & 0.143 & \textbf{0.579} &  & 0.080 & 0.578 \\
 &  & IRP-FT & 0.483 & 0.519 &  & 0.295 & 0.560 &  & 0.311 & 0.565 &  & 0.082 & 0.578 &  & 0.080 & 0.578 \\
 & \multirow{-8}{*}{\begin{rotate}{90}BERT\end{rotate}} & RP-AI & 0.448 & 0.516 &  & 0.364 & 0.568 &  & 0.255 & \textbf{0.583} &  & 0.082 & 0.578 &  & 0.080 & 0.578 \\ \cline{2-17} 
 &  & IIBP-WR & 0.391 & 0.416 &  & 0.405 & 0.443 &  & 0.413 & 0.445 &  & 0.333 & 0.461 &  & 0.160 & \textbf{0.469} \\
 &  & IIBP-FT & 0.393 & 0.439 &  & 0.380 & 0.430 &  & 0.388 & 0.424 &  & \cellcolor[HTML]{CBCEFB}0.432 & \cellcolor[HTML]{CBCEFB}0.413 &  & 0.380 & 0.459 \\
 &  & IBP-AI & \cellcolor[HTML]{CBCEFB}0.402 & \cellcolor[HTML]{CBCEFB}0.392 &  & 0.410 & 0.422 &  & 0.378 & 0.454 &  & 0.184 & 0.468 &  & 0.082 & 0.468 \\
 &  & IMP-WR & 0.399 & 0.427 &  & 0.397 & 0.417 &  & 0.421 & 0.441 &  & 0.383 & 0.453 &  & 0.180 & \textbf{0.469} \\
 &  & IMP-FT & 0.389 & 0.435 &  & 0.375 & 0.432 &  & 0.386 & 0.432 &  & \cellcolor[HTML]{CBCEFB}0.442 & \cellcolor[HTML]{CBCEFB}0.425 &  & 0.318 & 0.462 \\
 &  & MP-AI & \cellcolor[HTML]{CBCEFB}0.393 & \cellcolor[HTML]{CBCEFB}0.385 &  & \cellcolor[HTML]{CBCEFB}0.434 & \cellcolor[HTML]{CBCEFB}0.430 &  & 0.421 & 0.439 &  & 0.365 & 0.457 &  & 0.162 & \textbf{0.469} \\
 &  & IRP-FT & 0.386 & 0.431 &  & 0.413 & 0.429 &  & \cellcolor[HTML]{CBCEFB}0.436 & \cellcolor[HTML]{CBCEFB}0.411 &  & \cellcolor[HTML]{CBCEFB}0.448 & \cellcolor[HTML]{CBCEFB}0.439 &  & 0.164 & 0.468 \\
\multirow{-16}{*}{\begin{rotate}{90}AAPD\end{rotate}} & \multirow{-8}{*}{\begin{rotate}{90}BiLSTM\end{rotate}} & RP-AI & 0.411 & 0.417 &  & 0.409 & 0.432 &  & 0.397 & 0.451 &  & 0.344 & 0.462 &  & 0.160 & \textbf{0.470} \\ \hline

\end{tabular}
}

\caption{
Average pruned and unpruned models' effectiveness on PIEs when pruning 20, 50, 70, 90, and 99\% of the parameters.
For each pruning percentage column, the first value refers to the effectiveness of the pruned models on PIEs, the second value represents the effectiveness of the unpruned models on the same set of PIEs.
We represent models' effectiveness through accuracy in Single-label and F1 macro in Multi-label settings.
The blue colour identifies cases where the pruned models have higher effectiveness on PIEs than the unpruned ones.
We represent in bold the cases where the effectiveness of the models on PIEs is higher than the effectiveness of the same models on the whole dataset instead.}
\label{tab:exp3_mc_app}

\end{table*}